\documentclass[10pt,reqno]{amsart}
\usepackage[sc]{mathpazo}

\usepackage[utf8]{inputenc}
\usepackage[T1]{fontenc}
\usepackage{amsmath,amssymb,amsfonts,amsthm,mathtools}
\usepackage{float}
\usepackage{wrapfig}
\usepackage{setspace}
\usepackage{enumitem}
\usepackage{booktabs}
\usepackage{nicefrac}
\usepackage{microtype}
\usepackage{comment}
\usepackage{cancel}
\usepackage{tikz-cd}
\usepackage{centernot}
\usepackage[many]{tcolorbox}
\usepackage{mathrsfs}
\usepackage{fullpage}
\usepackage{fourier}

\usepackage[authoryear,square]{natbib}

\usepackage[dvipsnames,x11names,svgnames,table]{xcolor}
\definecolor{purple2}{RGB}{153, 50, 204}
\definecolor{green}{RGB}{0, 128, 0}
\definecolor{myblue}{rgb}{0.21, 0.34, 0.74}

\usepackage{hyperref}

\emergencystretch=\maxdimen
\numberwithin{equation}{section}
\allowdisplaybreaks

\hypersetup{
    colorlinks=true,
    linkcolor=olive,
    filecolor=red,      
    urlcolor=olive,
    citecolor=olive
}

\makeatletter
\newcommand{\vast}{\bBigg@{3}}
\newcommand{\Vast}{\bBigg@{4}}
\makeatother

\newcommand{\EX}{\mathbb{E}}

\newtcolorbox{boxE}{
  enhanced, breakable,            
  colback=white,        
  colframe=BlueGreen,   
  boxrule=0.75pt,       
  arc=0pt,              
  left=6pt, right=6pt, top=6pt, bottom=6pt
}

\renewcommand{\text}[1]{\textnormal{#1}}

\title[Kähler landscapes for complex neural network descents]{Kähler landscapes for complex neural network descents and guarantees including a search and destroy of the Calabi-Yau manifold}

\author{Andrew Gracyk}
\address{Department of Mathematics\\
      Purdue University\\
      West Lafayette, IN 47907, USA}
\email{agracyk@purdue.edu}

\subjclass[2020]{53B35, 53B12, 53Z50, 90C25}
\keywords{information geometry, complex geometry, Kähler manifold, Kähler geometry, Calabi-Yau manifold, Calabi conjecture, Dolbeault, Monge-Amp\`ere, Chern curvature, strong convexity, Polyak-\L{}ojasiewicz, Dirichlet energy, negative curvature, Ricci curvature, canonical line bundle, sheaf of sections}

\begin{document}

\begin{abstract}
We study landscapes for complex-parameterized networks. Our approach is motivated with an information-theoretic manifold perspective of the parameter and via classical optimization guarantees although of complex geometric variety such as through Dolbeault asymptotics. The descent path admits a Kähler information metric under a cross-entropy via the Wirtinger Hessian on the log-likelihood potential. We descend according to a differentiated loss scaled by the inverse metric, so the descent path remains in the holomorphic tangent bundle. We emphasize information manifolds which profane theoretical guarantees via ill-curvature-conditioned, or lack thereof, landscapes. We focus on Calabi-Yau metrics specifically in a non-compact setting with a global potential, so defined geometrically rather than invoking the topological requirements of the Calabi conjecture. In non-compact settings, we can write the metric determinant with respect to a background in terms of a function, pluriharmonic or real. Using Yau's Liouville theorem, we get a small eigenvalue blow-up effect under nontrivial background interaction, geometric invariance of the endomorphism constant, and almost low rank assumptions. In an empirical setting, a Ricci-flat metric will not form, but large eigenvalues can be a local condition and hold empirically on open sets even in generalized Kähler cases due to lower bounds on the Ricci potential. Moreover, it has been discovered that negative curvature subverts the loss landscape, specifically sectional curvature, so we expand on this and draw interconnections to negative-definite Ricci curvature. Our arguments primarily exist via geometric modality, although we establish roots in deep learning theory such as through asymptotics at initialization and connections through failure modes of neural network guarantees under vanishing and negative Ricci curvature.
\end{abstract}

\maketitle

{
\hypersetup{linkcolor=olive}
\tableofcontents
}

\newpage

\section{Introduction}
\label{sec:introduction}

\noindent We attempt to lay foundations of optimization twofold: (1) for the complex-parametered neural network \cite{trabelsi2018deepcomplexnetworks}, and (2) under information manifolds \cite{lawson2023fishergeometrygeodesicsmultivariate}. Complex neural networks attempt to remedy a parameterization bottleneck, and can achieve various levels of performance gains and dominion over real counterparts \cite{abdalla2023complexvaluedneuralnetworks}. We possess roots in deep learning theory with relevant asymptotics, but our underlying mechanisms and strategies will be conducted via geometric analysis. Loss landscapes admit geometric structure via information manifolds \cite{10.1162/089976698300017746}, so our work is motivated by this perspective.

\vspace{2mm}

A primary focus of our work is in Dolbeault asymptotics. \cite{banerjee2023restricted}  establishes a modern Hessian bound, which we reinforce both theoretically and empirically in our work, in which the Hessian takes on desirable asymptotic scaling for sufficiently nice regions pertaining to space encompassed under spectral norms and cosine similarity. These qualities allow convexity arguments to take place. For us, this will take place in the form
\begin{align}
\nabla^2_\omega L(v,v) \geq C \|v\|_h^2 ,
\end{align}
which relates to convexity. Existing asymptotics exist for real-valued neural networks, and are for Euclidean gradient descent methods that are unaccommodating for manifold information-theoretic structure. Our work addresses these qualities through a Kähler geometric analysis.

\vspace{2mm}

Under a Boreal measure on the training data that corresponds to a sufficiently smooth Radon-Nikodym derivative that, disaffected from neural network parameter dependence, the parameter landscape admits a cross-entropy metric. This metric is reminscent of a Fisher metric, although not exactly because of the parameter invariance, that is moreover a mixed Wirtinger derivative of a potential since the derivatives commute with the iterated integral. The potential, for example in the quadratic case, is plurisubharmonic. Therefore, under this potential, the admitted information manifold is Kähler. Under further restriction of conditions, in various order of ascending strength, such as
\begin{align}
\text{Ric} \equiv 0 \quad \impliedby \quad i^{K^2} \Omega \wedge \overline{\Omega} - \text{constant} \cdot \frac{\omega^K}{K!} \equiv 0 ,
\end{align}
the geometry is Calabi-Yau for a specific Kähler class. We will attempt to investigate the pitfalls of Calabi-Yau metrics on information landscapes. We study these circumstances solely affected in the Calabi-Yau scenario via \ref{app:calabi_yau_oscillation}, \ref{app:KPL_failure}, but also draw connections in \ref{app:regret_bounds}, \ref{app:dirichlet_energy}, \ref{app:trajectory_bounds}, \ref{app:variance}, \ref{app:min_eigenvalue_bounds}, which are sections that discuss roles of curvature in general. In our analyses, the Ricci curvature, further tethered to metric eigenvalues that both diminish and blow-up due to a constant determinant condition\footnote{Here, we are using the Kähler version of Ricci curvature, since it is defined via a log determinant, and a determinant is a product of eigenvalues.}, have interconnections to our theory. As we will see, optimization arguments are not immune to structures profaned in curvature.

\vspace{2mm}

Pitfalls are not unique to Calabi-Yau metrics, and can be properties of curvature in general. In Appendices \ref{app:dirichlet_energy}, \ref{app:trajectory_bounds}, \ref{app:variance}, \ref{app:min_eigenvalue_bounds}, we demonstrate that Ricci curvature itself can mar the learning landscape. In fact, in some results, the severity of the effect of curvature can be in exact proportion to the magnitude of the curvature, so Calabi-Yau metrics are merely a threshold as to where adverse effects can start. Nonetheless, we also demonstrate pitfalls unique to Calabi-Yau metrics in \ref{app:calabi_yau_oscillation}, \ref{app:KPL_failure}.

\vspace{2mm}

\section{Related work}

\begin{figure}[t]
  \centering
  \includegraphics[width=0.55\textwidth]{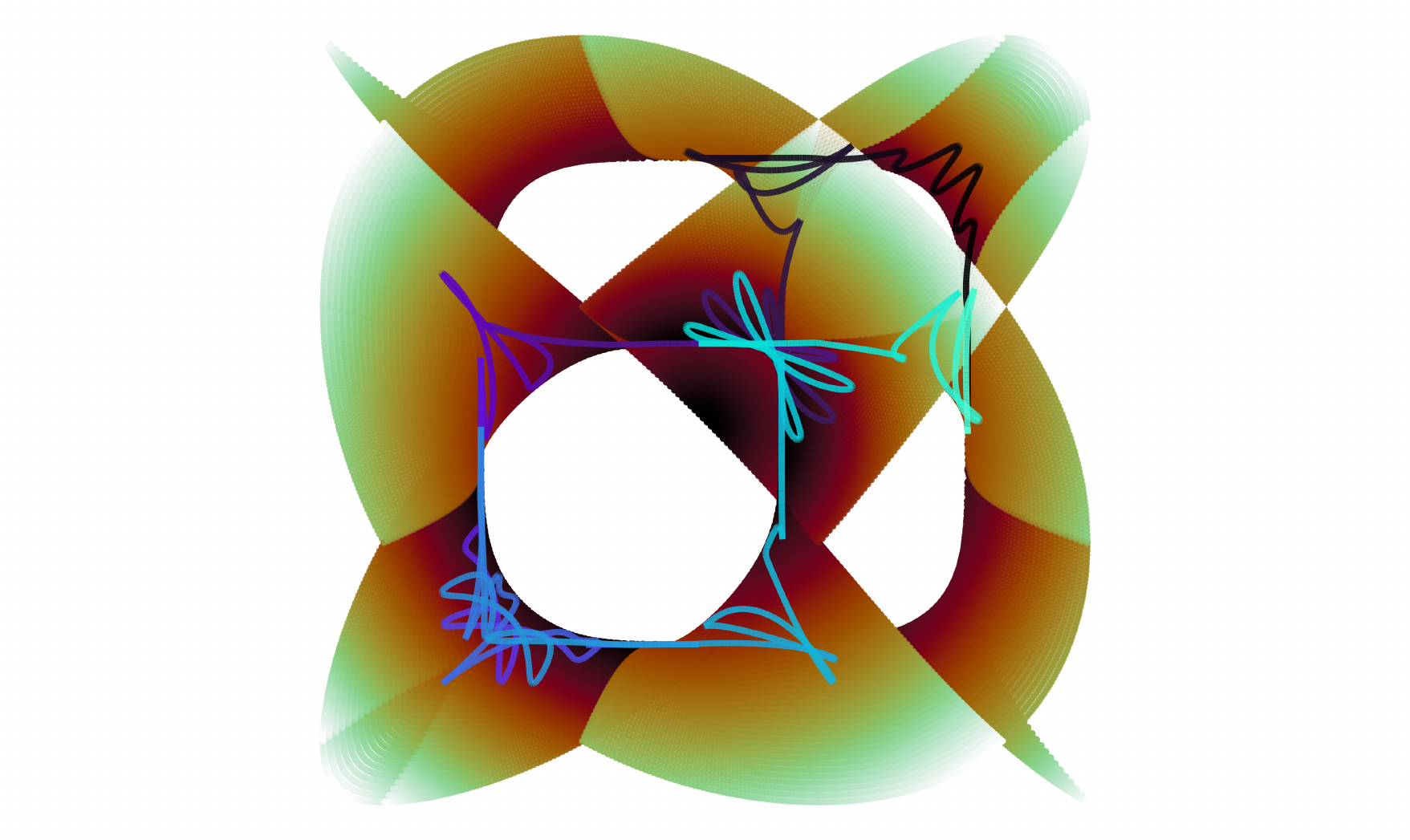}
  \vspace{4mm}
  \caption{We plot a cross-section and a curve corresponding to a descent path along a Calabi-Yau manifold. Our slice corresponds to the Fermat equation $Z_1^N + Z_2^N = 1$, $N=12$. The surface is projected onto the real plane by taking the real components $(\text{Re}(Z_1), \text{Re}(Z_2))$. This figure is somewhat toy because in a descending landscape scenario, the manifold is much more high-dimensional.}
  \label{fig:calabi_yau_manifold}
  \vspace{-0mm}
\end{figure}

\textbf{Connections to more traditional deep learning theory.} Our work possesses foundations in deep learning theory, namely at initialization. Literature that establish to varying levels neural network derivative bounds including use of Feynmann diagrams are \cite{banerjee2023restricted}  \cite{dyer2019asymptoticswidenetworksfeynman} \cite{zhu2022notelinearbottlenecknetworks} \cite{shemur2024weakcorrelationsunderlyingprinciple} \cite{aitken2020asymptoticswidenetworkspolynomial}. Bounding a Hessian norm can also be found in \cite{cisneros-velarde2025optimization}. Additional work relevant to deep learning theory and the roles of width in asymptotics include \cite{hanin2023randomfullyconnectedneural}
\cite{andreassen2020asymptoticswideconvolutionalneural} \cite{cirone2025genusexpansionnonlinearrandom} and especially pertaining to the neural tangent kernel \cite{guillen2026finitewidthneuraltangentkernels} \cite{huang2019dynamicsdeepneuralnetworks} \cite{liu2021linearitylargenonlinearmodels}. Much of our work also pertains to optimization through a geometric and analytic lens. Relevant work from a more real-analytic lens include \cite{javanmard2019analysistwolayerneuralnetwork} \cite{zhang2022meanfieldanalysistwolayerneural} with connections to Wasserstein geometry \cite{daneshmand2023efficientdisplacementconvexoptimization}. Optimization literature focusing specifically on (Riemannian) geometric analysis include \cite{amari2018statisticalneurodynamicsdeepnetworks} \cite{zavatoneveth2025doestrainingshaperiemannian} \cite{NIPS2016_14851003} \cite{Tron_2024} \cite{article}, so our work has high interconnections to these among those mentioned.

\vspace{2mm}

\noindent \textbf{Connections to negative sectional curvature.} The role of Ricci curvature in an optimization landscape has been studied \cite{Gigli_2017} \cite{lott2009ricci} \cite{Ambrosio_2015}, but these works are mostly separate from a deep learning theory perspective, i.e. neglect commentary on width, etc. Most of our curvature results comment on width in this context, such as \ref{app:dirichlet_energy}, \ref{app:trajectory_bounds}, \ref{app:gradient_flux_searches}, and \ref{app:some_bad_curvature_and_primitive}. \cite{Li2020} does study how positive Ricci curvature positively affects convergence, therefore Ricci flat and negatively-curved metrics lack this. On the other hand, \cite{pmlr-v178-criscitiello22a} shows that for many manifolds including Hadamard manifolds and hyperbolic spaces, i.e. with constant negative sectional curvature, gradient descent experiences pitfalls due to to the curvature. \cite{pmlr-v195-criscitiello23a} shows that, primarily under hyperbolic spaces, convexity results often worsen due to the curvature effects. Our work is reminiscent of these works, while our work emphasizes Ricci curvature and eigenvalues specifically. The lower-bounded determinant w.r.t. a background implies very large eigenvalues is unique to complex manifolds under Calabi-Yau metrics, therefore many of our results are lost in translation to Riemannian manifolds, since Ricci curvature being the log determinant of the metric is unique to Kähler manifolds. We build upon these works by examining the Ricci-flat scenario. In particular, our methods will also experience pitfalls under negative curvature as well, and so we also consider
\begin{align}
\text{Ric}(v,v) < 0 ,
\end{align}
and sometimes the Calabi-Yau manifold is simply the transition from the better to worse cases. In general, in sufficiently high dimensions, Ricci flatness is a weaker condition than a sectional curvature condition, since
\begin{align}
\text{Ric}(X, X) = \text{Tr}_h \text{Rm}(X, \cdot, X, \cdot) = \sum_i K(X, e_i) \equiv 0 \quad \centernot\implies \quad K(X, Y) \equiv 0 .
\end{align}
Since the eigenvalue explosion, on the other hand, is unique to our situation, our Calabi-Yau results particularly often go hand-in-hand with violating $\beta$-smoothness. 

\vspace{2mm}

\textbf{Optimization in complex geometry.} While some of our curvature-driven results are founded in width $m$, some are also driven by parameter dimension $K$; however, this is also the manifold dimension, therefore it would be non-rigorous to say results impacted by $K$ alone, independent of $m$, are tailored to deep learning theory. Therefore, we also argue our curvature-based results are primarily new in literature from the perspective that optimization in complex geometry is largely nonexistent rather than the Ricci curvature results are deeply interconnected with neural network width in every result.

\section{Neural network setup}
\label{sec:network_setup}

Consider training data $\{z_i, y_i\}_i, z_i \in \mathcal{Z} \subseteq \mathbb{C}^d, y_i \in \mathcal{Y} \subseteq \mathbb{R}$ (without loss of generality, we can restrict the imaginary component of $y_i$ to be $0$ if necessary, thus we maintain representations as generalized as possible), where $y \sim p(y | z, \theta) = \mathcal{N}(f(\theta; z), \sigma^2)$ follows a data distribution. Here, $\theta \in \Theta$ is the total collection of complex-valued weights. Consider a fully-connected complex neural network of the form \cite{banerjee2023restricted} 
\begin{align}
& \alpha^{(0)}(z) = z
\\
& h^{(l)}(z) = \frac{1}{\sqrt{m}} W^{(l)} \alpha^{(l-1)}(z)
\\
& \alpha^{(l)}(z) = \phi \left( h^{(l)}(z), \overline{h^{(l)}(z)} \right) , \quad l \in [L]
\\
& f(\theta; z, \overline{z}) = \alpha^{(L+1)}(z) = \frac{1} {\sqrt{m_L}} \text{Re} \left(  v^{\dagger} \alpha^{(L)}(z) \right) ,
\end{align}
where $W$ is a linear operator over the field of complex numbers $\mathbb{C}$, and $\phi : \mathbb{C} \rightarrow \mathbb{C}$ is a non-holomorphic activation. Here, $\theta \in \Theta \subseteq \mathbb{C}^{\sum_k m_k m_{k+1} + m_L} := \mathbb{C}^K$. For simplicity, assume $m_l = m$ for all $l$.

\section{Geometric setup}

\subsection{The Kähler descent landscape and its information geometry}
\label{sec:geometry}

Construct a probability measure such that $y_i \sim q(y|z)$. Notice $q$ does not depend on $\theta$. This is not unusual per se, although sometimes $q$ is parameterized. We can note a density of this form "violates injectivity" since the output of $y$ varies across single $z$. More specifically, $q$ is a Markov kernel $z \mapsto \mathbb{P}_{Y|Z=z}$ \cite{Fritz_2020}. The cross-entropy information metric is defined as
\begin{align}
\label{eqn:metric}
h_{i \overline{j}}(\theta) = \EX_{z \sim p_{\text{data}}} \left[ \EX_{y \sim q(y|z)} \left[  - \frac{\partial^2 \log p(y | z, \theta) }{\partial \theta^i \partial \overline{\theta}^j }  \right] \right] ,
\end{align}

which defines a Kähler manifold loss landscape $(M,\omega), \omega \in \Omega^{1,1}(M)$ under a preconditioned loss, or natural gradient descent where the descent update is scaled in accordance with the inverse information metric. In the above, $p$ is taken to be related to the loss. The above metric is Kähler since the Wirtinger derivatives commute with the integral under complex-valued Lebesgue dominated convergence \cite{Ziemer2017}, so 
\begin{align}
h_{i \overline{j}}(\theta) = \partial_i \partial_{\overline{j}} \Phi := \partial_i \partial_{\overline{j}} \ \text{potential}, \ \text{potential} \in \text{SPSH}(M) ,
\end{align}

by taking $\Phi = \EX_z \EX_q [ - \log p ]$. We can note the Wirtinger derivatives commute under $L^1$ decay via dominated convergence
\begin{align}
& \partial_{\theta^i} \partial_{\theta^{\overline{j}}} \underbrace{ \int_{\mathcal{Z}} \int_{\mathcal{Y}} -\log p(y | z,\theta) q(y | z) p_{\text{data}}(z)  dy  \frac{\omega_{\mathcal{Z}}^d}{d!}  }_{\displaystyle := \Phi}= \int_{\mathcal{Z}} \int_{\mathcal{Y}} -\partial_{\theta^i} \partial_{\theta^{\overline{j}}} \log p(y | z,\theta) q(y | z) p_{\text{data}}(z) dy  \frac{\omega_{\mathcal{Z}}^d}{d!} .
\end{align}
It is crucial to note we have removed the dependence on $\theta$ in the latter static $p$. Therefore, the above is more closely related to a cross-entropy Hessian rather than a true Fisher metric. When $p$ is an exponential likelihood, the negative log likelihood is quadratic, which is SPSH. In particular, a quadratic cost submits to a quadratic form $\Phi(\theta) = \frac{1}{2} \theta^\dagger A \theta + b^\dagger \theta + \theta^\dagger c + d$ without an expected value. As a last remark, we can note the expectation operator preserves a quadratic property since for a probability $(\Omega, \mathcal{F}, \mathbb{P})$ be a probability space with random vector $z \in L^2(\Omega; \mathbb{C}^n)$
\begin{align}
& \mathbb{E}[z^{\dagger} A z] = \mathbb{E}[\text{Tr}(z^{\dagger} A z)] = \mathbb{E}[\text{Tr}(A z z^{\dagger})]  = \text{Tr}(A  \mathbb{E}[z z^{\dagger}]) = \text{Tr}(A \Sigma) + \mu^{\dagger} A \mu .
\end{align}
The above line is more an illustrative example. In general, the expectation preserves an SPSH quality so it is indeed Kähler.

\vspace{2mm}

\noindent Denote $\mathcal{L}(\theta, \overline{\theta})$ the loss function under a complex parameter, so it dependent on the parameter itself and its complex conjugate. Under a traditional neural network trajectory, the gradient descent updates obey $\partial_t \theta^{\alpha}(t) = - \partial \mathcal{L} / \partial \overline{\theta}^{\alpha}$. For this work, we consider the geometrically-preconditioned backpropagated loss \cite{shrestha2023naturalgradientmethodsperspectives}
\begin{align}
\label{eqn:complex_parameter_trajectory}
\frac{d\theta^i(t)}{dt} = - h^{i \overline{j}} \frac{\partial \mathcal{L}}{\partial \overline{\theta}^j} ,
\end{align}
where $d\theta^i/dt$ lives in $T^{1,0}(M)$. Moreover, $h \in \Gamma(T^{*1,0}M \otimes T^{*0,1}M)$ is a section, so $h^{-1} \in \Gamma(T^{1,0}M \otimes T^{0,1}M)$ and the inverse information metric acts such that $\sharp: T^{*0,1}M \to T^{1,0}M$. Under this trajectory, the parameter obeys the information manifold Kähler structure, which serves as a loss landscape. We can note the descent trajectory is just one direction in the span of the holomorphic tangent bundle, and it is in the direction of steepest descent. In particular, $d\theta/dt \in T_{\theta(t)}^{1,0}M$ and we get for some unimportant orthogonal complement $\mathcal{W}$

\begin{align}
T_{\theta(t)}^{1,0}M = \text{span}_{\mathbb{C}} \left\{ \frac{d\theta}{dt} \right\} \oplus_h \mathcal{W} .
\end{align}

\noindent Let us verify our parameter update in the complex case indeed decreases the loss under \ref{eqn:complex_parameter_trajectory}. Consider the total loss differential $
d\mathcal{L} = \frac{\partial \mathcal{L}}{\partial \theta^i} d\theta^i + \frac{\partial \mathcal{L}}{\partial \overline{\theta}^j} d\overline{\theta}^j $.
Because the loss is real-valued, we have $
\overline{\left(\frac{\partial \mathcal{L}}{\partial \theta^i}\right)} = \frac{\partial \mathcal{L}}{\partial \overline{\theta}^i} $.
Using our preconditioned trajectory, the parameter differential and its conjugate are $d\theta^i = - \eta h^{i \overline{j}} \frac{\partial \mathcal{L}}{\partial \overline{\theta}^j}$ and $d\overline{\theta}^j = - \eta h^{i \overline{j}} \frac{\partial \mathcal{L}}{\partial \theta^i}$, where we use the fact that the metric is Hermitian. Into the total differential, we see
\begin{align}
d\mathcal{L} & = \frac{\partial \mathcal{L}}{\partial \theta^i} \left(- \eta h^{i \overline{j}} \frac{\partial \mathcal{L}}{\partial \overline{\theta}^j}\right) + \frac{\partial \mathcal{L}}{\partial \overline{\theta}^j} \left(- \eta h^{i \overline{j}} \frac{\partial \mathcal{L}}{\partial \theta^i}\right) = - 2\eta h^{i \overline{j}} \frac{\partial \mathcal{L}}{\partial \theta^i} \frac{\partial \mathcal{L}}{\partial \overline{\theta}^j} .
\end{align}
Because the information metric $h$ is positive definite, the quadratic form $h^{i \overline{j}} \frac{\partial \mathcal{L}}{\partial \theta^i} \frac{\partial \mathcal{L}}{\partial \overline{\theta}^j}$ is real and positive, guaranteeing $d\mathcal{L}$ is real and negative.

\vspace{2mm}

\noindent \textbf{Definition 1.} Let $(M, \omega)$ be a Kähler information manifold representing the parameter space, where $\omega$ is the Kähler form induced by the information potential. A smooth, real-valued loss function $L: M \to \mathbb{R}$ is said to satisfy $\nu$-(complex)  strong convexity if, for any $z' \in S$ and a reference $z \in M$, we have
\begin{align}
\label{eqn:strong_convexity}
L(\theta') \geq L(\theta) + 2\text{Re} \langle \partial L_\theta, \exp_\theta^{-1}(\theta')^{1,0} \rangle  + \frac{\nu}{2} d_\omega(\theta, \theta')^2 .
\end{align}
with $\nu > 0$. Here, $d_{\omega}$ is the geodesic distance. This condition is a bridge to the classical (Dolbeault) Hessian bound $i \partial \overline{\partial} L(z) \succeq \nu \omega$.

\vspace{2mm}

The above definition is useful (more so crucial) because we will draw bridges to it via Taylor expansions in some of our results.

\subsection{Calabi-Yau descents}
\label{sec:calabi_yau}

\textbf{Definition 2.} We say the Kähler manifold $M$ that admits metric $h$ is Calabi-Yau if its Ricci curvature is zero, i.e.
\begin{align}
\label{eqn:ricci_zero}
\text{Ric}_{i \overline{j}} & = - \partial_i \partial_{\overline{j}} \log \det (h)
\\
& = - \partial_i \partial_{\overline{j}} \log \det \left( \EX_{z \sim p_{\text{data}}} \left[ \EX_{y \sim q(y|z)} \left[  - \frac{\partial^2 \log p(y | z, \theta) }{\partial \theta^i \partial \overline{\theta}^j }  \right] \right] \right)  \equiv 0 .
\end{align}

\noindent This definition has topological and geometric nuance, so we elaborate. We adopt this geometric definition. Generally, to invoke the Calabi conjecture  \cite{yau1977calabi}, we require compactness of $M$, or at least of the submanifold in which the optimization trajectory exists. We remark this is slightly nonrigorous because the Calabi conjecture requires the manifold to be closed without boundary, and a compact submanifold would contain a boundary with Dirichlet or Neumann boundary conditions. This is problematic for us because our metric is defined via a global Kähler potential, and by the maximum principle, a globally defined strictly plurisubharmonic function on a compact manifold without boundary must be constant \cite{demailly_complex_analytic}. This completely destroys our defined metric, at least globally but not locally. Topologically, the manifold is governed by the vanishing of the first real Chern class, $c_1(M; \mathbb{R}) = 0$. While stronger algebraic definitions exist, such as requiring the canonical line bundle to be trivial, which equips $M$ with a nowhere-vanishing holomorphic $K$-form, the condition $c_1(M; \mathbb{R}) = 0$ is the, albeit weaker, requirement for Calabi-Yau manifolds. We bypass the Chern class requirement of Yau's theorem, and we define the Calabi-Yau manifold entirely geometrically rather than topologically by ensuring it is Ricci-flat. We remark the title of this work is also allusive to a search of the Ricci flat metric within a Kähler class, but since we relaxed compactness, this is not rigorous.

\begin{figure}[ht]
  \centering
  \includegraphics[width=0.55\textwidth]{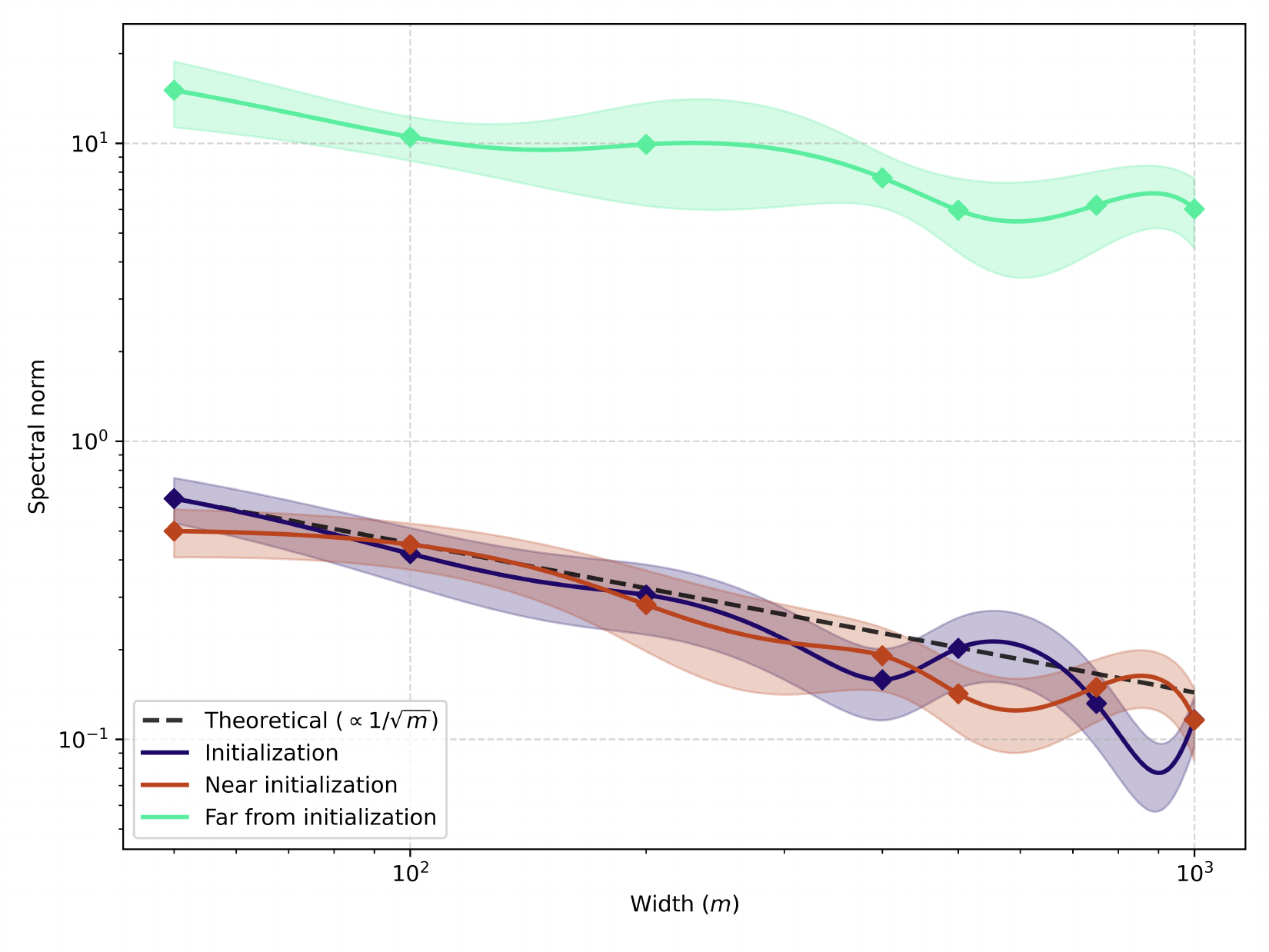}
  \caption{We plot $\| i \partial \overline{\partial} f \|_2$ asymptotics corresponding to \ref{app:dolb_hessian_bounds} on 20 instances, corresponding to exactly initialization, near initialization (perturbation $\epsilon=0.1$), and far from initialization ($\epsilon=2$).}
  \label{fig:spectral_hessian_scaling}
\end{figure}

\vspace{2mm}

\noindent We can note the global Ricci form $\rho = -i \partial \overline{\partial} \log \det(h)$ is, up to a constant, the curvature 2-form $F^{\nabla}$ of the induced Chern connection on the anti-canonical line bundle $K_M^{-1} = \Lambda^K T^{1,0}M$ over the parameter space, so we get the Calabi-Yau condition $\text{Ric} \equiv 0 \iff \rho \equiv 0 \iff F^{\nabla} \equiv 0$ and
\begin{align}
F^{\nabla} = \overline{\partial} \partial \log \left( \frac{\omega^K K!^{-1}}{ c_K \Omega \wedge \overline{\Omega}} \right) \equiv 0 
\end{align}

for a nowhere-vanishing holomorphic volume form $\Omega \in H^0(M, K_M)$, so the Dolbeault operators vanish the log of the geometric volume form and the algebraic volume form ratio. As a remark, note that
\begin{align}
i F^{\nabla} = \rho = i \text{Tr}(\Theta_h) .
\end{align}
Here, $\Theta_h$ is the Chern curvature form. A Calabi-Yau manifold will fail our convexity results because of the following scenarios:

\vspace{2mm}

\noindent We would consider the Calabi-Yau condition as a Monge–Ampère equation with constant determinant condition for constant $\kappa$ for manifold dimension $K$
\begin{align}
\label{eqn:constant_determinant_Phi}
(i \partial \overline{\partial} \Phi)^K = \kappa  dV_0\footnotemark 
\end{align}
\footnotetext{To be compatible with a complex manifold, specifically a Kähler manifold, $dV_0$  is the volume form generated by a fixed reference Kähler metric $\omega_0$ defined with $dV_0 = \frac{\omega_0^K}{K!}$. In particular, while standard, $dV_0$ is somewhat slight abuse of notation for a complex manifold, for example $\int_M dV_0$ over a Kähler manifold is somewhat informal notation.}up to a conventional factor of $K!$. Note that the above is not true in the compact case due to the argument following \ref{eqn:ricci_zero}, since a positive Kähler form on a compact manifold cannot be expressed globally as $\partial \overline{\partial} \Phi$. This is a constant determinant condition w.r.t. a flat background since $\omega = i \sum_{j,k=1}^K \left( \partial_j \partial_{\overline{k}} \Phi \right) d\theta^j \wedge d\overline{\theta}^k$, $dV_0 = \det(h_0) \left( \frac{i}{2} \right)^K d\theta^1 \wedge d\overline{\theta}^1 \wedge \dots \wedge d\theta^K \wedge d\overline{\theta}^K$, and substituting into \ref{eqn:constant_determinant_Phi}, with a background metric, being slightly nonrigorous with constants,
\begin{align}
& i^K K! \det \left( \partial_j \partial_{\overline{k}} \Phi \right) d\theta^1 \wedge d\overline{\theta}^1 \wedge \dots \wedge d\theta^K \wedge d\overline{\theta}^K  = \kappa \det(h_0) \left( \frac{i}{2} \right)^K d\theta^1 \wedge d\overline{\theta}^1 \wedge \dots \wedge d\theta^K \wedge d\overline{\theta}^K ,
\end{align}
which yields $\det \left( \partial_j \partial_{\overline{k}} \Phi \right) = \frac{\kappa \det(h_0)}{K!  2^K}$ solving for the determinant. A flat background, so $\det(h_0)=1$ simplifies this. Equivalently, \ref{eqn:constant_determinant_Phi} can be written if we choose it to be so
\begin{align}
 (i \partial \overline{\partial} \Phi)^K = (-1)^{\frac{K(K-1)}{2}} \left( \frac{i}{2} \right)^K \kappa  \Omega \wedge \overline{\Omega}  ,
\end{align}
for  nowhere-vanishing holomorphic $K$-form $\Omega$, again up to normalization. In the compact case, we suppose the Calabi–Yau metric is relative to a fixed background Hermitian metric $h_0$, which is Ricci-flat (and so is $h$), and we have that
\begin{align}
\label{eqn:det_const}
\det_{h_0}(h) = \det( h_0^{-1}h ) = \frac{\det(h)}{\det(h_0)}:= \det(H) \equiv \widetilde{\kappa} > 0 .
\end{align}
This follows from the maximum principle on a compact manifold. To be clear, in the Calabi conjecture, the background metric is not Ricci-flat. In \ref{eqn:det_const}, we are really just comparing two metrics which are already Ricci-flat. The above of \ref{eqn:det_const} is possible in a non-compact case as well, but not guaranteed (see following arguments). In general, the above is a global (or global guarantee), not a local, property. Now, let us discuss the non-compact case more closely.

\vspace{2mm}

More aptly for us, in the case of the non-compact Calabi-Yau manifold, we desire a blow-up condition of the form
\begin{align}
\lambda_{\max}(H) = \Omega\left(\frac{\kappa}{\mu^{\widetilde{K}-1}}\right) .
\end{align}
On both compact and non-compact Kähler manifolds,
\begin{align}
\operatorname{Ric}(h)-\operatorname{Ric}(h_0) = -i\partial\overline{\partial} \log\det(h_0^{-1}h) ,
\end{align}
which holds for all $h,h_0$, not necessarily Ricci-flat, and if only $h$ is  Ricci-flat,
\begin{align}
\partial\overline{\partial} \log\det(h_0^{-1}h)= - i \text{Ric}(h_0) . 
\end{align}
Instead of a constant determinant condition of \ref{eqn:det_const}, we get the condition, even when $h_0$ is not Ricci flat
\begin{align}
\label{eqn:not_ricci_flat}
\partial\overline{\partial} u = - i \text{Ric}(h_0) .
\end{align}
When $h_0$ is Ricci-flat, $u$ is real pluriharmonic. Note a pluriharmonic function need not be constant on a non-compact manifold. If $\text{Ric}(h_0)$ is not identically vanishing, corresponding to the above in \ref{eqn:not_ricci_flat}, we know $u$ is merely real-valued in addition to \ref{eqn:not_ricci_flat}. Moreover, it is also true that on a non-compact Kähler manifold
\begin{align}
\label{eqn:det_H}
\det(H) = \prod_{i \in \mathcal{I}_{\infty}} \lambda_{H,i}(\theta) \cdot \prod_{j \in \mathcal{I}_{0}} \lambda_{H,j}(\theta)  \cdot \prod_{k \in \mathcal{I}_{B}}  \lambda_{H,k}(\theta) = e^{u} ,
\end{align}
for suitable set of divergence of some eigenvalues $\mathcal{I}_{\infty}$, a collapsing set $\mathcal{I}_0$, and a sufficiently bounded, neutral set $\mathcal{I}_B$. Note that the above of \ref{eqn:det_H} is true for any pair of Kähler metrics, and is not specific to Calabi-Yau manifolds. Note that \ref{eqn:det_H} can simplify. 

\vspace{2mm}

We will impose in this work (in a theoretical, not empirical, setting)
\begin{align}
\text{Ric}(h) = \text{Ric}(h_0) = 0 .
\end{align}
Canonically choose a flat background metric if suitable, although this is not always ideal (see discussion below). Additionally, we assume (1) $h_0$ is complete (this is reasonable for us); (2) $u$ is bounded below (also reasonable). It follows from Yau's Liouville theorem \cite{Yau1975Harmonic} (\cite{SoupletZhang2006} is also slightly relevant), in this scenario, that we have
\begin{align}
\label{eqn:eu_kappa}
\det(H) = e^{u} = \kappa > 0 .
\end{align}
Note this holds even in the non-compact case, whereas previously in \ref{eqn:det_const}, a constant condition can also hold in the compact case but under different assumptions. Also, the above of \ref{eqn:eu_kappa} is a global, not local, condition. Suppose, for example, that $\widetilde{K}$ eigenvalues are small
\begin{align}
\label{eqn:mu}
& \lambda_{K-\widetilde{K}+1},\ldots,\lambda_K\le \mu, \qquad \mu\ll 1
\\
\implies & \kappa = \prod_{j=1}^{K-\widetilde{K}}\lambda_j \prod_{j=K-\widetilde{K}+1}^{K}\lambda_j \le \lambda_{\max}^{K-\widetilde{K}}\mu^{\widetilde{K}}  ,
\end{align}
which in turn therefore gives
\begin{align}
\label{eqn:blow-up}
\lambda_{\max}(H) \ge \left( \frac{\kappa} {\mu^{\widetilde{K}}} \right)^{\frac{1}{K-\widetilde{K}}} \implies \lambda_{\max}(H) = \Omega\!\left( \frac{\kappa}{\mu^{\widetilde{K}'-1}} \right) , 
\end{align}
for some $\widetilde{K}'$. We will not take $\mu$ to be some constant, and we will take $\widetilde{K}'$ to vary, thus we prevent ourselves from writing $\Omega(1)$, which is too weak. This condition is crucial to Lemma 6 in \ref{app:calabi_yau_oscillation}, which is one of our primary Calabi-Yau results. We will use this condition in \ref{sec:theory}. The above argument can be made a local property to hold pointwise; let us discuss further: 

\vspace{2mm}

Since \ref{eqn:eu_kappa} is a global condition, we can get a similar blow-up effect when curvature is locally zero but not globally by lower bounding $u$. Instead, we can consider a local condition, which holds in a general Kähler setting and not necessarily a Calabi-Yau setting,
\begin{align}
\label{eqn:eu_lower_bound}
\det(H(\theta)) = e^{u(\theta)} \geq e^C := \kappa_0 > 0 ,
\end{align}
The above is not guaranteed to hold over the entire manifold, so we assume it at least holds locally over our descent paths. A local blow-up condition is 
\begin{align}
\lambda_{\max}(H(\theta))
\geq
\left(\frac{\kappa_0}{\mu^{\widetilde K}}\right)^{\frac1{K-\widetilde K}}.
\end{align}
The above does not need global Ricci-flatness. This is our blow-up condition that holds reasonably in empirical settings. We can rewrite the eigenvalues as
\begin{align}
\prod_{j \in \mathcal{I}_0}\lambda_j = \frac{\kappa}{ \left(\prod_{i \in \mathcal{I}_\infty}\lambda_i\right) \left(\prod_{k \in \mathcal{I}_B}\lambda_k\right) }  .
\end{align}
Under the conditions on $u$ such as $0 < c_1 \le e^{u(\theta)} \le c_2 < \infty$, we get an inverse relationship
\begin{align}
\prod_{j \in \mathcal{I}_0} \lambda_{H,j}(\theta) \asymp \Biggl( \prod_{i \in \mathcal{I}_\infty} \lambda_{H,i}(\theta) \Biggr)^{\!-1} \xrightarrow[\theta \to \theta^*]{} 0 .
\end{align}

\vspace{2mm}

After examining equations \ref{eqn:eu_kappa} and \ref{eqn:eu_lower_bound}, a major part of our argument that Calabi-Yau manifolds specifically are adverse for optimization is now that the blow-up effect is more greatly characterized by the Calabi-Yau manifold than for generalized Kähler manifolds because the determinant remains fixed, while the lower bound of $e^{u(\theta}) = \kappa_0$ can drop well below $\kappa$. $\kappa$ larger is more conducive the the largest eigenvalue being large. The smaller the determinant is, then on average the lower the eigenvalues will be, but not necessarily the largest eigenvalue. If, on average, the eigenvalues are larger, then this is more conducive to the maximal eigenvalue being large, but this is not guaranteed. 

\vspace{2mm}

The above phenomenon is also affected by qualities such as how almost low rank the metric is. Generally, from \cite{Sun_2025}, the eigenvalues of $h$ are unstable in the sense that the rank varies. All we truly desire is that a instability of $H$ is possible via our results in Appendices \ref{app:calabi_yau_oscillation}, \ref{app:KPL_failure}. It does not imply trivial eigenvalue interaction when $h_0, h$ are multiplied. If the eigenvalues of $h_0, h$ are diverse, the eigenvalues of $H = h_0^{-1} h$ can be diverse.

\vspace{2mm}

\begin{figure}[ht]
  \centering
  \includegraphics[width=0.45\textwidth]{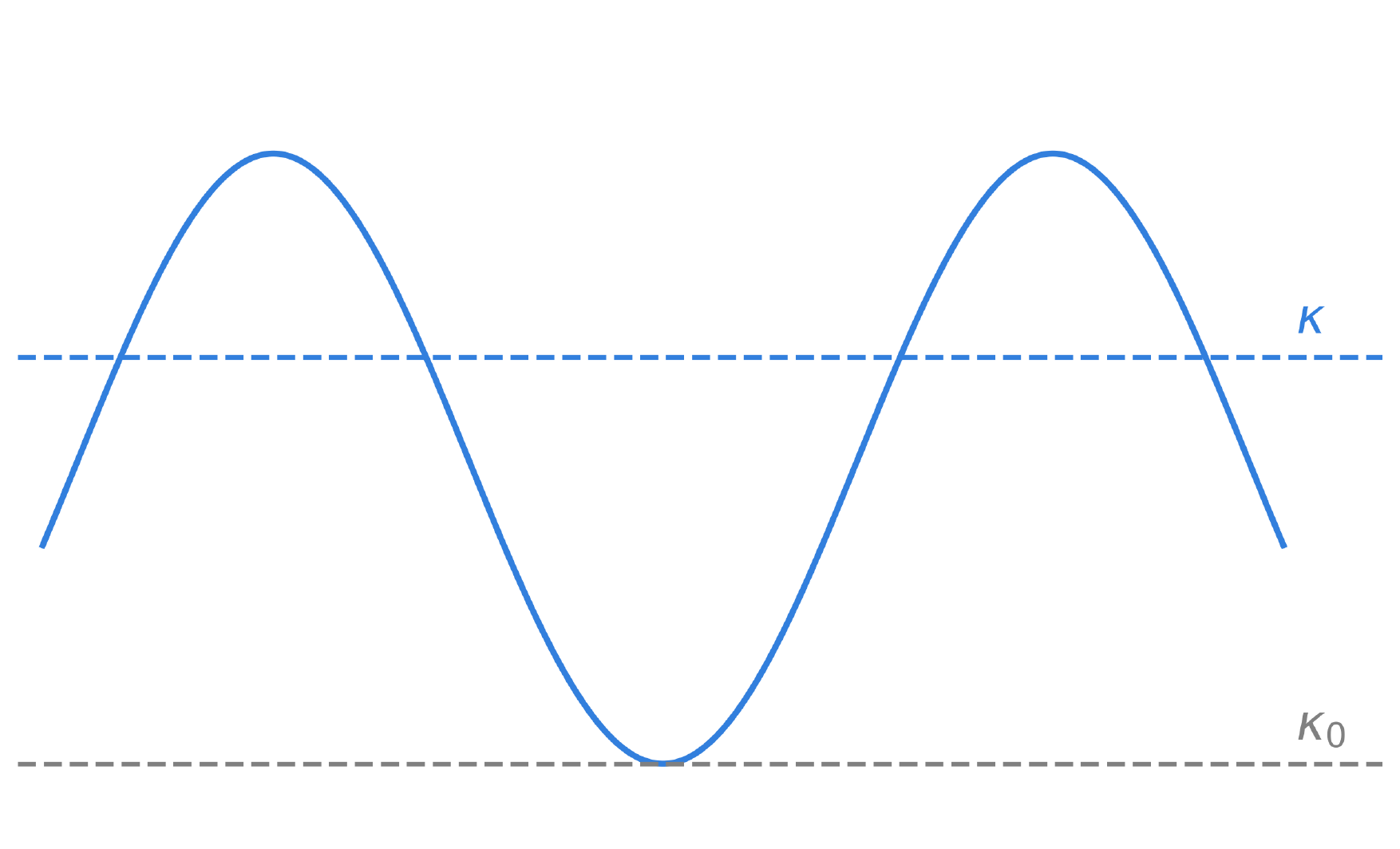}
  \caption{We illustrate a phenomenon as to why Calabi-Yau metrics may be more conducive to blow-up than general Kähler metrics. For a Kähler metric, the only eigenvalue guarantee is via lower bound $\kappa_0$, whereas Calabi-Yau $\kappa$ is constant and can be higher. It is true we can use the full Ricci potential term $e^{u}$ when we do an analysis for the general Kähler case, but this is nontrivial to analyze asymptotically. For example, it is not a guarantee. Also, it could experience non-uniform behavior (i.e. spikes). In this case, we must write $\lambda_{\max}(H(\theta)) = \Omega\left(\frac{e^{u(\theta)}}{\mu^{\widetilde{K}-1}}\right)$, and this is no longer a neural network guarantee but rather a fluctuating value. }
  \label{fig:kappa}
\end{figure}

We argue more thoroughly on a set of nonzero measure the constant $\kappa_0(\theta)$ is significantly smaller than $\kappa$. We will operate so that
\begin{align}
e^{u(\theta)} \geq \kappa_0 \ll \kappa ,
\end{align}
so that the Calabi-Yau blow-up effect is greater than that for generalized Kähler metrics. Indeed, since $u$ is nonconstant, it can "dip below to a sufficiently small" value, making $\kappa_0$ sufficiently small since $\kappa_0$ is only a lower bound. Since $\kappa_0$ is defined via a lower bound and not a global constant, we have less guarantee $\mu$ is as large in the Kähler case: again, this is because it is expected the eigenvalue guarantee is smaller in the generalized Kähler case. There is great discrepancy in this argument because the exact values of $\kappa,\kappa_0$ are to be determined and have nontrivial closed-forms in an analysis; however, under the assumption the function $e^{u}$ is "around" what would be $\kappa$ in the Calabi-Yau case, this is sufficient to argue $\kappa_0$ is well below $\kappa$. Hence, following this argument here, not only do we assume $\kappa_0 < \kappa$ but also $\mu_0 < \mu$ where $\mu_0$ is the upper bound on the first $\widetilde{K}$ eigenvalues in the Kähler case and $\mu$ is in the Calabi-Yau case.

\vspace{2mm}

Of course, just because the determinant is constant does not mean the eigenvalues themselves are constant. For example, consider the metric
\begin{align}
\label{eqn:h_example}
\begin{cases}
h=
\begin{pmatrix}
1+4|z_1|^2&2 z_1\\
2 \overline{z}_1&1
\end{pmatrix} 
\\
\det h=1 ,
\end{cases}
\end{align}
which is indeed both Ricci-flat and Kähler with potential $\Phi(z_1,z_2)=|z_1|^2+|z_2+z_1^2|^2$, but it is also low-dimensional, so it is a toy example. This is a contributing factor to a Calabi-Yau blow-up effect even more so because it will hold locally on open sets that a blow-up condition in some regions is more severe than other regions. Since $\kappa$ is constant, this will not change, but if the metric can enter "almost low rank" states at an even greater level at some regions in comparison to other regions, the larger eigenvalues will compensate to counteract the even more diminished eigenvalues. This effect may not exist as well in Kähler metrics alone because it is possible $u$ will get smaller in these "almost low rank" regions, but this cannot happen in our Calabi-Yau case.

\vspace{2mm}

Let us try to be convinced the blow-up of the maximum eigenvalue of $H$ is not circular reasoning. By definition, a determinant is a product of eigenvalues, so inherently all eigenvalues small will mean the determinant is small. In particular, let us try to be convinced it is not circular that at least one eigenvalue is sufficiently large because the determinant is sufficient large. Certainly, we already argued the determinant depends more on the average of the eigenvalues than the individual eigenvalues. We also argue $\kappa$ is a global constant, and is fixed a priori. Therefore, by the paragraph surrounding \ref{eqn:h_example}, we can in part justify the broadness of the eigenspectrum since it varies locally and will change across varying descent paths. $\kappa$ is fixed through, and the metric can become "more almost low rank" in some areas than others.

\vspace{2mm}

Notice we have restricted in a theoretical (not empirical setting) that $\text{Ric}(h_0)=0$ in order to apply Yau's Liouville theorem. We can canonically choose a flat metric if desirable, which is trivially Calabi-Yau. We have witnessed in our experiments that the interaction between $h_0,h$ when formulating $H$ is more conducive to a soft eigenvalue blow-up phenomenon than when examining $h$ individually. We refer to figures \ref{fig:kahler_blowup}, \ref{fig:kahler_blowup_greater}. We therefore find value when considering when $h_0$ is a more sophisticated Ricci-flat metric. Note that we can also canonically choose $h_0$ to correspond to $h_{t=0}$, which is Ricci-flat since $h$ is.

\vspace{2mm}

We return to the fact that we imposed $\text{Ric}(h_0)=0$. In the context of the Calabi conjecture, we remark this is not reasonable. As per the discussion in Appendix \ref{app:background}, this assumption is reasonable for us because we are in a deep learning setting. Certainly $h$ Ricci-flat is reasonable for us (in some of our theoretical arguments). In particular, we do not seek a background in a Kähler class as per the Calabi conjecture, but rather something canonical associated with $h$ to complete our deep learning arguments.

\vspace{2mm}

For additional discussion, we refer to \cite{karkada2024lazyntkrichmup}, which discusses the broad eigenspectrum range of an information metric. We also refer to Figure \ref{fig:kahler_blowup}.

\vspace{2mm}

\noindent Calabi-Yau manifolds can also offer advantages over negatively-curved spaces, but not as much as positively-curved spaces. We elaborate more on effects of curvature in general in \ref{sec:first_deriv_and_curv}. Under a continuous-time stochastic natural gradient descent, the trajectory variance under a stochastic Jacobi equation
\begin{align}
\frac{d}{dt} \mathbb{E} \left[ \| J_t \|_h^2 \right] = & - 2 \EX \left[ \nabla^{1,1}_{\omega} \mathcal{L} (J_t, \overline{J}_t) \right]    - \EX \left[ \text{Ric}(J_t, \overline{J}_t) \right] + \text{Tr}_h(\Sigma) .
\end{align}
The Ricci curvature vanishes under a Calabi-Yau manifold, but in a largely negatively curved space ($\text{Ric} \ll 0$), the Ricci curvature term $- \EX \left[ \text{Ric}(J_t, \overline{J}_t) \right]$ becomes positive and large, inducing high variance.

\subsection{Corrupted geometries under neural networks and eigenvalue collapse}
\label{sec:corrupted_geometries}

\noindent It is not immediately guaranteed the loss landscape encounters a split in eigenvalue behavior; however, it is a realistic neural network scenario given the following.

\vspace{2mm}

\noindent \textbf{Theorem 1.} \textit{Let there exist $K$ neural network parameters and real-data constraints $N \times k$. In the overparameterization regime, then there exist eigenvalues $\lambda_{\text{min}} = 0$.}

\label{theorem_1}

\vspace{2mm}

\noindent \textit{Remark.} It is not possible for a Calabi-Yau manifold to be low rank (except on a set of measure zero). By definition, a Kähler metric must be positive-definite. Not only this, but a non-invertible metric completely destroys our application of natural gradient descent.

\vspace{2mm}

\begin{figure}[ht]
  \centering
  \includegraphics[width=0.55\textwidth]{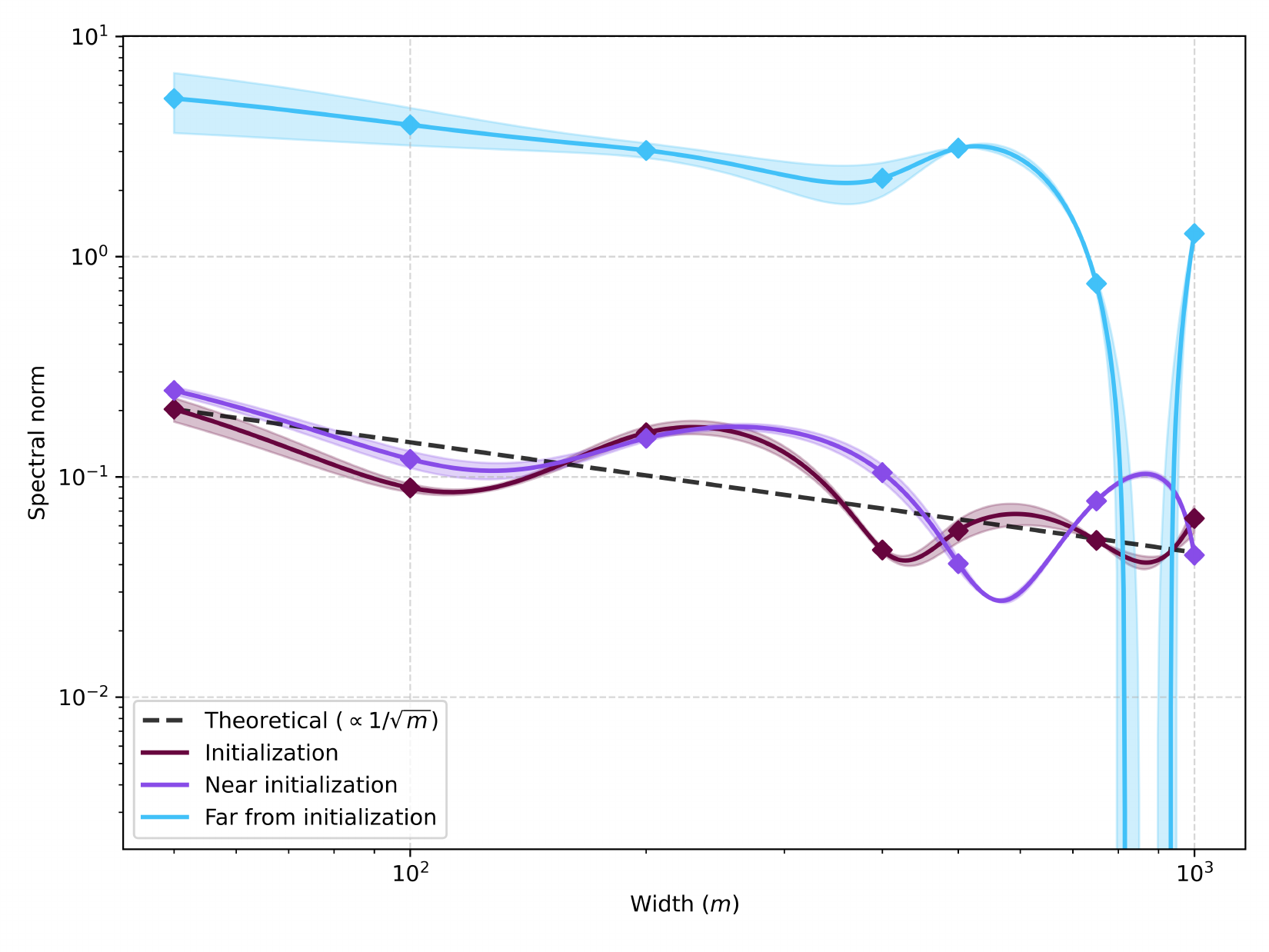}
  \caption{We plot asymptotics on the (2,0)-Hessian of $\| \mathcal{H}^{2,0}\|_2$ corresponding to \ref{app:2,0_hess_bounds} on 2 instances using an SVD power iteration approach, noting $H_{2,0} = \frac{1}{4}(H_{xx} - H_{yy}) + \frac{i}{4}(H_{xy} + H_{yx})$.}
  \label{fig:spectral_2,0hessian_scaling}
\end{figure}

\noindent \textit{Proof.} In the overparameterization regime, so $m$ is large, the metric $h$ is a $K \times K$ matrix effectively of rank at most $k$. The maximum possible rank of $h$ is $\min\{K,N \times k\}$, if $K$ is taken $K \gg N \times k$, for example $f(z; \theta) \in \mathbb{R}^k$ with a dataset of size $N$, $h$ becomes rank deficient \cite{Sun_2025} \cite{bakeer2026localinformationoperatorsspatial} \cite{dong2026quotientgeometryeffectivecurvature}. The kernel of $h$ has dimension at least $K - Nk > 0$. Therefore, there are zero eigenvalues, forcing $\lambda_{\text{min}} = 0$.  $\square$

\vspace{2mm}

\section{Regularization}
\label{sec:regularization}

\noindent It can be noted in the case of at least one degenerate eigenvalue for sufficiently large width threshold $m^*$, somewhat informally
\begin{gather}
\log \det(h) = -\infty \quad \forall m > m^*
\\
\text{Ric}_{i \overline{j}} =  - \partial_{i} \partial_{\overline{j}} (-\infty) = \text{DNE} .
\end{gather}
This will corrupt our information manifolds.  It can be noted \cite{banerjee2023restricted}  is not a work in information geometry and does not use a metric, therefore their work is allowed to focus on restricted strong convex regimes and this work largely ignores volume collapse. Since our approach is geometric, we cannot ignore corrupted geometries. To reconcile this inconsistency, we can assume our loss is regularized as
\begin{align}
\label{eqn:regularized_loss}
\widetilde{\mathcal{L}(\theta)} = \mathcal{L}(\theta) + 
\lambda \theta^{\dagger} \theta := - \log p( y | z, \theta) + \lambda \theta^{\dagger} \theta .
\end{align} 
Note this is also called Tikhonov/ridge regularization. This will in turn allow a bound on a minimum eigenvalue, since $h$ will take the form $\widetilde{h}_{i \overline{j}}(\theta) = h_{i \overline{j}}(\theta) + \lambda \delta_{i \overline{j}}$. It is crucial to note \ref{eqn:regularized_loss} provides the metric to be full rank in a theoretical setting, but empirical Fisher/cross-entropy calculations can still be low rank. We show this in Figure \ref{fig:eigenspectrum}.

\vspace{2mm}

In Appendix \ref{app:convexity_results}, we derive a result that is reminiscent of this although is not a global bound, but \ref{app:convexity_results} contradicts our geometric setup as discussed due to the eigenvalue singularity. Indeed, this proof is valid under the low-width regime, and is not immune to metric collapse. Therefore, in much of our work, we will typically work with the condition where a subset of eigenvalues of $h$ or $H$ or both are either really small or really large
\begin{align}
(M^*, \omega^*) \in \Bigg\{ (M,\omega) \Bigg| \text{Ric} \equiv 0 \iff (i \partial \overline{\partial} \Phi)^K - \kappa  dV_0 = 0, 0 \lesssim \lambda < \lambda_{\{h,H\}\text{min}} \ll 1, \lambda_{\{h,H\}\text{max}} \leq \lambda_{\gg 1}^*  < \infty \Bigg\} . 
\end{align}

\noindent \textbf{Eigenvalue lower bounds and convexity discussion.} From this theorem, it follows that the eigenvalue collapse is a consequence of overparameterization and not the Calabi-Yau artifact. The eigenvalue explosion is the consequence of the Calabi-Yau feature. Strong convexity necessitates $\nabla^2_\omega L = J^\dagger J + \mathcal{H}_{\text{net}} \succeq \nu I$ where $J^{\dagger} J$ is the contribution from the metric, and $\mathcal{H}_{\text{net}}$ is intrinsic contribution from the network. A nondegenerate metric is not sufficient to ensure strong convexity. Therefore, it is most reasonable for us to assume the regularized information metric (in all cases, not just the Calabi-Yau)
\begin{align}
h = \mathcal{F}_{\text{reg}} = \mathcal{F} + \lambda I, \quad \lambda_{\text{min}} ( \mathcal{F}_{\text{reg}}) \geq \lambda > 0
\end{align}
has eigenvalues bounded below by a positive value (see \cite{dufortlabbé2026navigatingpotholesgeometryawaresharpness} \cite{cayci2025riemannianoptimizationperspectivegaussnewton} for relevant literature), which derives from $\nabla^{1,1}_{\omega}\mathcal{L}(\theta) + \lambda \nabla^{1,1}_{\omega}\theta^{\dagger} \theta$. This equation is a direct consequence of \ref{eqn:regularized_loss}. We will employ this assumption in Appendix \ref{app:convexity_results} rather than assuming strong convexity. In general, $i \partial \overline{\partial} L \succeq \nu I$ is considered unreasonable in deep learning theory globally. \cite{banerjee2023restricted}  does not assume a static $i \partial \overline{\partial} \mathcal{L} \succeq \nu I$ result and this is a consequence of their restricted strong convexity. In Appendix \ref{app:convexity_results}, we will prove a convexity result.

\begin{figure}[t]
  \centering
  \includegraphics[width=0.65\textwidth]{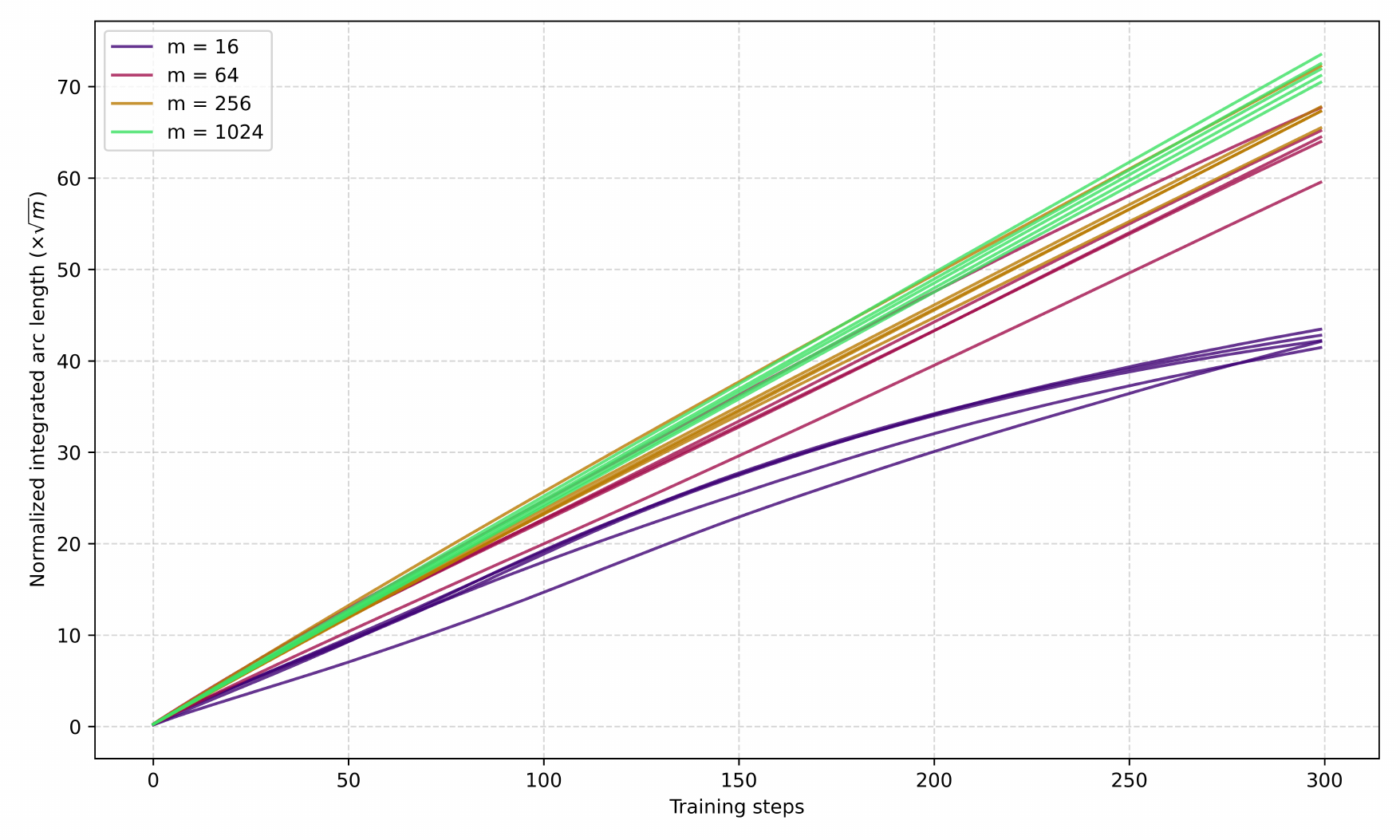}
  \caption{We plot the integrated arc length $\int  \| \dot{\theta}\|_h ds$ times $\sqrt{\text{width}}$ versus training steps across 5 trajectories per width with a high learning rate of $\gamma = 0.35$. We can note the \textcolor{Purple2}{narrowly}-parameterized networks dissipate in high training steps, while \textcolor{Green0}{wide} networks stay straight. This demonstrates an effect of feature learning: large $m$ is in the neural tangent kernel regime, or the "lazy" regime \cite{karkada2024lazyntkrichmup}, and the narrow networks are underparameterized and in the process of feature learning, or the "rich" regime. The narrow networks fall off because they adapt their features to finding a more efficient minima.}
  \label{fig:integrated_arc_length_falloff_effect}
\end{figure}

\vspace{2mm}

\noindent \textbf{Additional remarks.} Under an overparameterization collapse, the geodesic distance along a direction in the nullspace $\mathcal{N}_h = \{ v \in T_p M \mid h(v,v) = 0 \}$ will maintain identically vanishing geodesic distance, so a geodesic ball region collapses with respect to dimension. The inverse of the exponential map $\exp_p^{-1}$ possesses similar properties, and will collapse in norm, so the cosine similarity is ill-defined. The spectral norm $\| h \|_2$, on the other hand, is immune to rank deficiency and is immune to collapse.

\begin{tcolorbox}[colback=white, colframe=black,
  title=\textbf{A destroy via regularization of curvature}]
To regularize the loss landscape and support well-behaved curvature, we can penalize the loss further with
\begin{align}
\label{eqn:loss_curv_penalty}
\widehat{\mathcal{L}} = \mathcal{L}  + \lambda \theta^{\dagger} \theta - \alpha \log \det(h(\theta)) \quad \implies \quad
\alpha \rho \succeq - (i\partial\overline{\partial}\mathcal{L} + \lambda \omega_{\text{flat}})  ,
\end{align}
since the Hessian of the loss $\succeq 0$ at a local minima. Again, we set the loss to be the negative log-likelihood of $p$, so $- \log p$. Here, $\rho$ is the Ricci form. We can control the positive-definieness of the Ricci form based on $\alpha, \lambda$, and the loss, therefore regularizing the Ricci curvature of the manifold with the $\log \det$ penalty. We can note the negative sign in \ref{eqn:loss_curv_penalty} provides a lower bound as in \ref{eqn:loss_curv_penalty} and helps facilitate positive curvature, although it does not necessarily produce it.
\end{tcolorbox}

\vspace{2mm}

\section{Theoretical results}
\label{sec:theory}

In our theoretical results, we will typically make the following assumptions (unless stated otherwise).
\begin{enumerate}[label=\Roman*., nosep]
\item $(M,\omega)$ is the Kähler information manifold of \ref{eqn:metric} of dimension $K$ with metric $h$.
\item The parameter descent obeys natural gradient descent $\dot{\theta}^i = - h^{i \overline{j}} \partial_{\overline{j}} \mathcal{L}$.
\item $h$ is full rank by the regularized loss of \ref{eqn:regularized_loss}. Moreover, the minimum eigenvalue is bounded below $\lambda_{\text{min}} > \lambda$. In order for the nuclear norm to not pick up a factor of dimension, which we need in Theorem 2, we assume $\lambda = \mathcal{O}(1/K^2)$ when necessary. To enforce this, we can consider $\widetilde{\mathcal{L}(\theta)} = \mathcal{L}(\theta) + \frac{\lambda_0}{K^2} \theta^\dagger \theta$.
\item $\mathcal{S}$ is a (compact) geodesic ball on the Kähler manifold sufficiently close to initialization.
\end{enumerate}

\vspace{2mm}

We will moreover make the assumption specific to Calabi-Yau manifolds. To reiterate, we remark these results can be theoretical, and a Calabi-Yau loss manifold will not form in an actual training scenario. Instead, these results can still be reasonable empirically due to the lower bounds on the Ricci potential in the general Kähler case according to equation \ref{eqn:eu_lower_bound}. We list:
\begin{enumerate}[label=\Roman*., nosep]
\item The manifold $M$ is not compact ($\mathcal{S}$ still exists even if $M$ is not compact).
\item The "almost low rank" quality mentioned in \ref{sec:calabi_yau} holds for $H$.
\item The Calabi-Yau constant determinant condition $\det (H) \equiv \text{constant}$ is with respect to a constant sufficiently large, so not all of the eigenvalues via $\det (H) = \prod_i \lambda_i$ are very small. 
\item The minimum eigenvalue(s) of Calabi-Yau $H$ follows $\mu \geq \lambda_{\widetilde{K}} \geq \lambda_{\widetilde{K}-1} \geq \hdots \geq  \lambda_{\text{min}}(H) \geq \lambda(K) := \lambda$, where the definition is shorthand, so $h \succeq \lambda(K) h_0$, where $h_0$ is the background metric.
\item The maximum eigenvalue of Calabi-Yau $H$ follows $\lambda_{\max}(H) = \Omega\left(\frac{\kappa}{\mu^{\widetilde{K}-1}}\right)$, where $\kappa$ is a determinant constant and $0 \ll \widetilde{K} \leq K$ is some constant (this is our blow-up condition; if it does not hold globally for large $\widetilde{K}$, then assume it holds locally; note it is not the same as $\Omega(1)$ since $\mu,\widetilde{K}$ can vary).
\end{enumerate}

\begin{figure}[ht]
  \centering
  \includegraphics[width=0.7\textwidth]{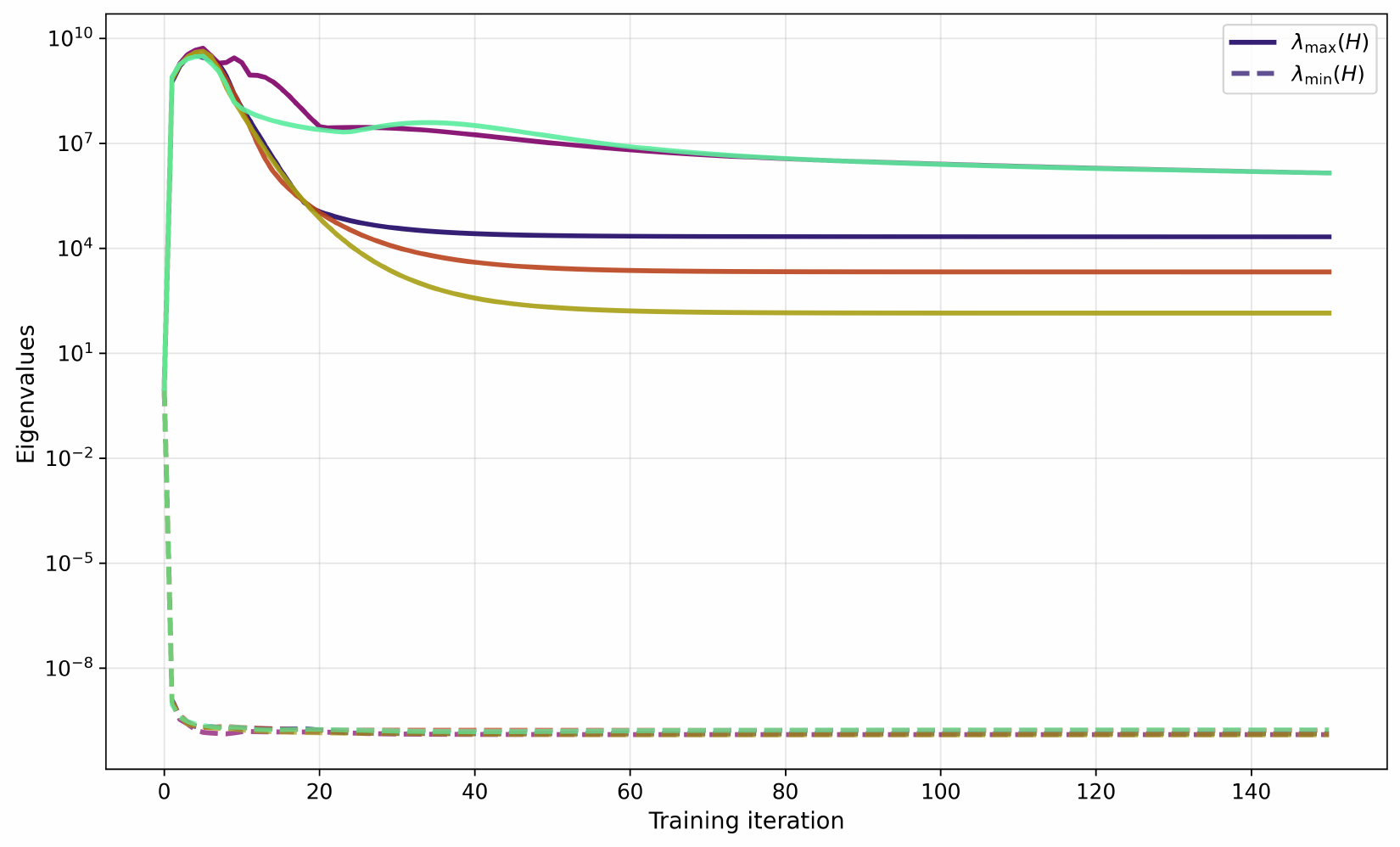}
  \caption{We plot a severe eigenvalue blow-up on $H$ that holds empirically based on the Kähler condition of \ref{eqn:eu_lower_bound} across five distinct training trajectories. Here, we use a low regularization floor $h_t=(F_t)_+ + \varepsilon I, \varepsilon=10^{-10}\lambda_{\max}((F_0)_+)$, where $F_t$ is the complex mixed Hessian of the binary cross-entropy loss. Here, we set a background metric $h_0$ to be the regularized metric $h_t$ at initialization $t=0$. We crucially remark most eigenvalues of $h_0$ are quite small because the metric is near degenerate at initialization, but also the metric is anisotropic. We also remark the exact regime of this figure is the overparameterization regime, which is more conducive to blow-up (largest eigenvalue is much larger than smallest eigenvalue in $H$). In particular, the interaction between $h_0$ and $h$ is highly nontrivial. Hence, this is not the same as just multiplying $h$ by a constant. Since $h_0$ is fixed in $\theta$, it is Ricci-flat in $\theta$, thus counts as Calabi-Yau by our terms.}
  \label{fig:kahler_blowup_severe}
\end{figure}

\begin{figure}[t]
  \centering
  \includegraphics[width=0.7\textwidth]{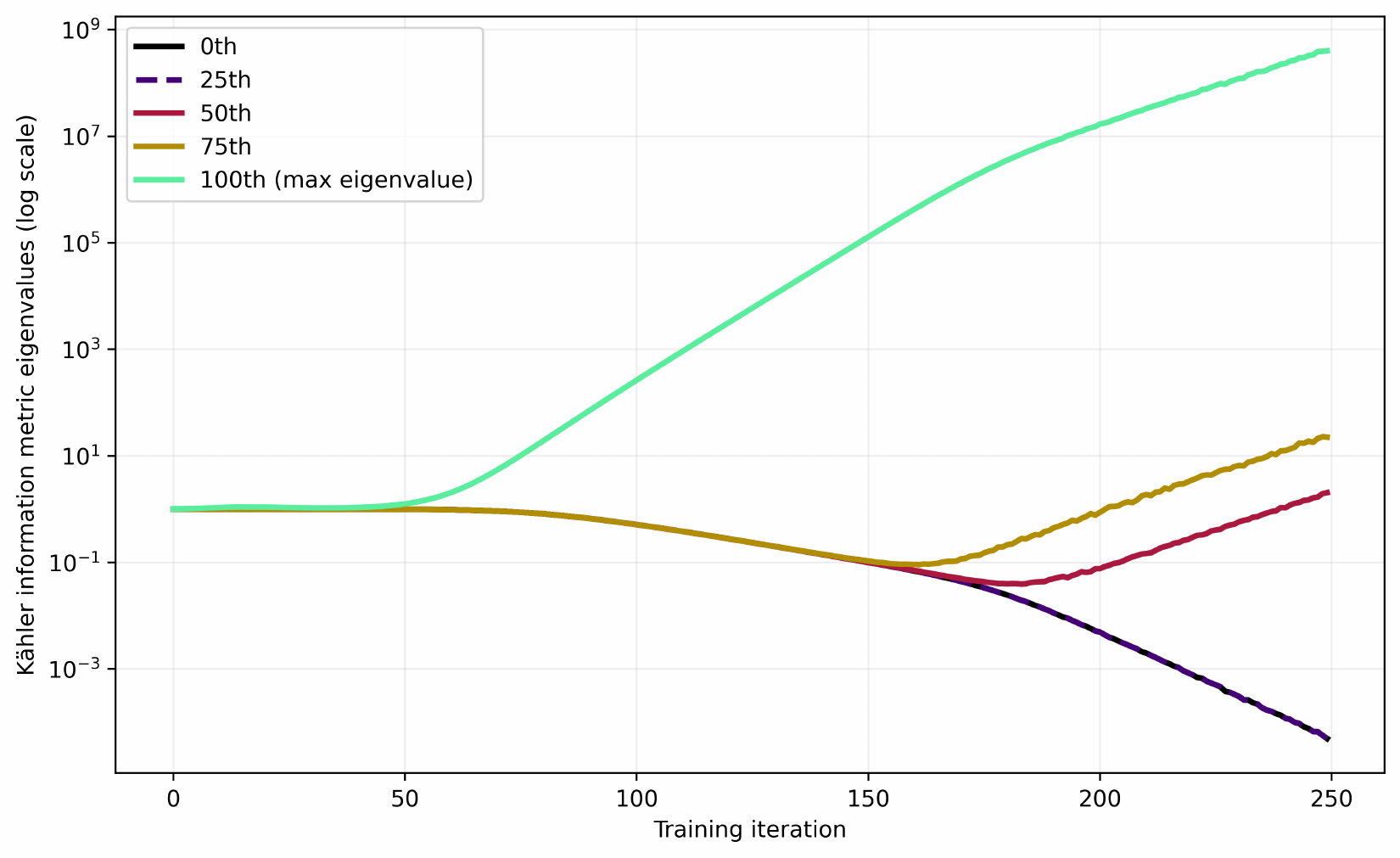}
  \caption{We illustrate a blow-up effect in a constant determinant empirical setting. We use a complex-parametered network with Gaussian data with a complex neural network. We use a cross-entropy information metric (not a Fisher metric, which is not Kähler). We force a constant determinant condition via multiplying the information metric by $\alpha = (\kappa / \det(h_{\text{reg}}))^{1/K}$, thus this yields a constant determinant condition. In particular, denote the empirical metric $h_{\text{reg}}$. We scale $H = \alpha h_{\text{reg}}$, thus the determinant evaluates to $\alpha^K \det(h_{\text{reg}}) = \kappa$. We regularize the metric so the lowest eigenvalue obeys $\lambda_t = (0.85)^{(\text{training iteration})}$, which is exponential (and indeed exhibits sufficient spectral decay that we need in our theory). This is not a Calabi-Yau case exactly (in reality, a Calabi-Yau manifold will never hold almost surely), but the metric is modified so that a Calabi-Yau scenario is reflected empirically.}
  \label{fig:empirical_blowup}
\end{figure}

\subsection{Second derivative results}
\label{sec:second_deriv_results}

\textbf{Theorem 2.} \textit{Let $(M,\omega)$ be a Kähler manifold with dimension $K$ and Kähler metric $\omega$ with $h$ full rank. Let $f(\theta; z) : M \rightarrow \mathbb{R}$ be a function such that the deformed metric obeys $\omega_f = \omega + i \partial \overline{\partial} f > 0$. Assume that the spectral norms $\| \partial^3 f \|_2$ and $\| i \partial \overline{\partial} f\|_2^2$ are $\mathcal{O}(1)$ or $\mathcal{O}(m^{-\alpha})$, $\alpha \geq 0$. Assume the hypotheses of Lemma 1 and Lemma 2 hold. Assume the nuclear norm $\| \cdot \|_1 \leq \text{constant} \cdot \| \cdot \|_2$ does not pick up a factor of dimension $K$ due to rapid eigenvalue decay. Suppose the hypotheses of the Mean Value Theorem with respect to $\mathcal{S}$ hold. Then the spectral norm of the Dolbeault Hessian on $\mathcal{S}$ is bounded by
\begin{align}
\sup_{\theta \in \mathcal{S}} \| i \partial \overline{\partial} f \|_2 = \mathcal{O}\left(\frac{1}{\sqrt{m}}\right) .
\end{align}
}

\vspace{2mm}

Theorem 2 is one of our main results, and is consistent with the scaling of \cite{banerjee2023restricted}. To the best of our knowledge, this bound is sharp, consistent with Figure \ref{fig:spectral_hessian_scaling}. This result is unique since it is specifically for Dolbeault asymptotics and complex networks. \textit{Sketch of proof.} The proof is an application of the Mean Value Theorem applied to a logarithmic volume ratio. The remainder of the proof primarily follows from applications of Lemma 1, Lemma 2, and a Taylor expansion. 

\vspace{2mm}

\noindent \textbf{Lemma 1.} \textit{Let $f(\theta;z) : M \to \mathbb{R}$ be as in \ref{app:dolb_hessian_bounds}. Assume the (2,0) Hessian obeys a bound with respect to the (1,1) Hessian $\| \partial^2_{ki} f \|_2 \leq C' \| i \partial \overline{\partial} f \|_2$. Under a Gaussian model, then the Dolbeault Hessian obeys the property
\begin{align}
\label{eqn:lemma_1}
\sup_{\|v\|_2=1} \| \nabla_v (i \partial \overline{\partial} f) \|_2 \leq \| \partial^3 f \|_2 + C \| i \partial \overline{\partial} f \|_2  \sup_{\theta \in \mathcal{S}} \| i \partial \overline{\partial} f \|_2 ,
\end{align}
where $\| \partial^3 f \|_2 := \sup_{\|v\|_2=1} \| v^k \partial_k \mathcal{H} \|_2$.}

\vspace{2mm}

\noindent \textbf{Lemma 2.} \textit{Assume the maximum eigenvalue of the Hessian $\mathcal{H}(\theta) = i\partial\overline{\partial} f(\theta)$ is finite over $\mathcal{S}$. Assume $\lambda_{\text{max}}(\mathcal{H}) > 0$, $- \lambda_{\text{min}}(\mathcal{H}) < \lambda_{\text{max}}(\mathcal{H})$. Suppose for test function $v(\theta_t, t) = \log(\lambda_{\max}(\mathcal{H}(\theta_t))) - \Psi(\theta_t) - \xi t$, the operator satisfies $(\partial_t - \Delta_{\widetilde{\omega}}) [ \log(\lambda_{\max}(\mathcal{H}(\theta_t))) - \Psi(\theta_t) ] \leq \xi$ on $\mathcal{S} \times [0,T]$ for some $\xi \geq 0$. Given the initialization bound $\| \mathcal{H}(\theta_0) \|_2 \leq \frac{C_{\text{init}}}{\sqrt{m}}$, there exists a geometric constant $C_{\text{geom}} > 0$ such that the spectral norm of the Hessian obeys
\begin{align}
\sup_{\theta \in \mathcal{S}} \| \mathcal{H} \|_2 \leq C_{\text{geom}} \| \mathcal{H}(\theta_0) \|_2 \exp\left( \sup_{\mathcal{S}} \Psi - \inf_{\mathcal{S}} \Psi \right) .
\end{align}
}

\vspace{2mm}

\noindent \textbf{Lemma 3.}  \textit{Assume the $(1,1)$ Hessian satisfies $\| \mathcal{H}^{1,1} \|_2 \leq \widetilde{C} \| \mathcal{H}^{2,0} \|_2$, and the $(2,0)$ Hessian obeys the covariant bound
\begin{align}
\| \nabla_\omega \mathcal{H}^{2,0} \|_2 \leq C_1 \| \partial^3 f \|_2 + C_2 \sup_{\theta \in \mathcal{S}} \| \mathcal{H}^{2,0} \|_2^2 .
\end{align}
If the spectral norm of the initial $(2,0)$ Hessian $\| \mathcal{H}^{2,0}(\theta_0) \|_2$ is $\mathcal{O}\left(\frac{1}{\sqrt{m}}\right)$ and the third derivative $A = \sup_{\theta \in \mathcal{S}} \| \partial^3 f \|_2$ is $\mathcal{O}(m^{-k})$ where $k \geq \frac{1}{2}$, then the spectral norm of the $(2,0)$ Hessian over the geodesic ball is bounded by
\begin{align}
\sup_{\theta \in \mathcal{S}} \| \mathcal{H}^{2,0} \|_2 = \mathcal{O}\left(\frac{1}{\sqrt{m}}\right) .
\end{align}}

\subsection{Initialization results}

\noindent \textbf{Theorem 3.} \textit{Let $f(\theta_0; z)$ be the output of the $L$-layer neural network of width $m$ as in \ref{sec:network_setup} exactly at initialization, parameterized by $\theta_0 = \{W^{(1)}, \dots, W^{(L)}, v\}$ where the weights are drawn i.i.d. from a standard complex Gaussian distribution $\mathcal{CN}(0,1)$. Assume $v=g/\|g\|_2$ with $g\sim\mathcal{CN}(0,I_m)$. Assume the base input vector satisfies $\|\alpha^{(0)}\|_2^2 = \mathcal{O}(m)$ and the activation function $\phi(h, \overline{h})$ has bounded first and second Wirtinger derivatives. Then, with high probability over the initialization, the spectral norm of the $(1,1)$ parameter Hessian $\mathcal{H}$ of $f(\theta_0)$ is bounded by
\begin{align}
\|\mathcal{H}(\theta_0)\|_2 = \mathcal{O}\left(\frac{1}{\sqrt{m}}\right).
\end{align}
}

\subsection{Convexity results}

Theorems 4,5 and Lemmas 4,5 are very close to results found in \cite{banerjee2023restricted} in the end result, but the proofs of each are different in spirit, since we are now relying on geometric structure and complex data to complete the proof. 
\vspace{2mm}

\noindent \textbf{Theorem 4 (convexity).} \textit{Let $(M, \omega)$ be Kähler. Consider the loss $L(\theta) = \frac{1}{n} \sum_{i=1}^n \ell_i(y_i, f_i(\theta))$ parameterized by $\theta \in M$. Assume the following regularity conditions locally: (i) the loss function satisfies $\ell_i'' \geq a$ for some constant $a > 0$; (ii) $\mathcal{F}_t(v) = \frac{2}{n} \sum_{i=1}^n \left( \text{Re} \left( \langle \nabla_\omega f_i(\widetilde{\theta}_t), v \rangle_h\right) \right)^2$ is bounded by $\nu \|v\|_h^2 \leq \mathcal{F}_t(v) \leq \rho \|v\|_h^2$ for some $\nu > 0$ and maximum eigenvalue bound $\rho$; (iii) the full Hessian norm is bounded by $C_{\mathcal{H}} \|v\|_h^2 / \sqrt{m}$, where $m$ is the network width parameter. Then we have
\begin{align}
\nabla^2_\omega L(\widetilde{\theta}_t)(v,v) \geq \Gamma\left(a, \nu, \rho, C_{\mathcal{H}}, \{\ell_i'(f_i(\widetilde{\theta}_t))\}_i, m,v\right) \|v\|_h^2 ,
\end{align}
where $\Gamma = \Gamma \left(a, \nu, \rho, C_{\mathcal{H}}, L(\widetilde{\theta}_t), m, v\right)$ is defined as
\begin{align}
\Gamma := a \nu  - \frac{ 2\sqrt{2}a \sqrt{\rho} C_{\mathcal{H}} \|v\|_h +  C_{\mathcal{H}} \sqrt{\frac{1}{n} \sum_{i=1}^n (\ell'_i(f_i(\widetilde{\theta}_t)))^2}}{\sqrt{m}}   .
\end{align}
}

\vspace{2mm}

\noindent \textbf{Theorem 5 ($\beta$-smoothness).} \textit{Let $(M, \omega)$ be Kähler. Let $L(\theta) = \frac{1}{n} \sum_{i=1}^n \ell_i(y_i, f_i(\theta))$ denote the empirical loss. Assume the regularity condition that the first-order Jacobian quadratic form is bounded by $\rho_J > 0$, such that $\frac{2}{n} \sum_{i=1}^n \ell''_i \left( \text{Re} \left( \langle\nabla_\omega f_i, v \rangle_h \right) \right)^2 \leq \rho_J \|v\|_h^2$. Assume the results of \ref{sec:second_deriv_results} hold. Then we have
\begin{align}
\frac{1}{2} \nabla^2_\omega L(\widetilde{\theta}_t)(v, v) \leq \left( \rho_J + \frac{C_{\mathcal{H}}}{\sqrt{m}} \sqrt{ \frac{1}{n} \sum_{i=1}^n (\ell'_i(f_i(\widetilde{\theta}_t)))^2 } \right) \|v\|_h^2 
\end{align}
for some constant $C_{\mathcal{H}}$.}

\vspace{2mm}

\noindent \textbf{Lemma 4 (dynamic Kähler Polyak-Łojasiewicz condition).} \textit{Let $(M, \omega)$ be a Kähler manifold and $L$ a loss function attaining its minimum at $\theta^*$. Define the dynamic strong convexity parameter along a geodesic $\gamma_t$ as
\begin{align}
\label{eqn:Gamma_t}
\Gamma_t := \inf_{s\in[0,1]} \frac{ \mathcal H_{i\overline j}\dot\gamma^i\overline{\dot\gamma^j} + \operatorname{Re}(\nabla_i\nabla_jL\dot\gamma^i\dot\gamma^j) }{ h_{i\overline j}\dot\gamma^i\overline{\dot\gamma^j} }  ,
\end{align}
where $w = \exp_{\theta_t}^{-1}(\theta^*)$, $\gamma_t(s) = \exp_{\theta_t}(s w)$, and $v = w^{1,0} = \dot{\gamma}_t(0) = \exp_{\theta_t}^{-1}(\theta^*)^{1,0}$ is an initial holomorphic tangent vector. Then $\Gamma_t$ is guaranteed to satisfy a strong convexity condition by the result of Appendix \ref{app:convexity_results}. Moreover, if $\Gamma_t > 0$, then $L$ satisfies the dynamic Kähler Polyak-Łojasiewicz condition
\begin{align}
\inf_{\theta \in U} L(\theta) \geq L(\theta_t) - \frac{1}{\Gamma_t} \vast\| \nabla_{\omega}^{1,0} L(\theta_t) \vast\|_h^2 .
\end{align}}

\vspace{2mm}

\textbf{Lemma 5 (geodesic descent under the Kähler-PL condition).} \textit{Let $(M, \omega)$ be a Kähler manifold and $L$ a loss function attaining its minimum at $\theta^*$. Consider the geodesic gradient descent update $ \theta_{t+1}=\exp_{\theta_t}(w_t), w_t=v_t+\overline{v_t}, v_t= \eta_t \nabla_\omega^{1,0} L(\theta_t),$ where $\eta_t>0$, and assume the geodesic $\sigma_t(s)=\exp_{\theta_t}(sw_t)$ exists for \(s\in[0,1]\). Define the dynamic $\beta$-smoothness parameter
\begin{align}
\beta_t=\max\left\{0, \sup_{\substack{s\in[0,1]\\
0\neq\xi\in T_{\sigma_t(s)}^{1,0}\mathcal S}} \frac{ \mathcal H_{i\overline j}\xi^i\overline{\xi^j} +\operatorname{Re}\left(\nabla_i\nabla_jL\xi^i\xi^j\right)}{h_{i\overline j}\xi^i\overline{\xi^j}}\right\}.
\end{align}
Suppose $\beta_t<\infty$ and $L$ satisfies the Kähler-PL condition at $\theta_t$ with parameter $\Gamma_t>0$. Then, for $0<\eta_t\leq 2/\beta_t$, the updated parameters satisfy
\begin{align} L(\theta_{t+1})-L(\theta^*) \leq
& \left[1-\Gamma_t\eta_t(2-\beta_t\eta_t)\right] \cdot \left(L(\theta_t)-L(\theta^*)\right).
\end{align}
When $\beta_t=0$, the above holds for any $\eta_t>0$.
}

\begin{figure}[ht]
  \centering
  \includegraphics[width=0.7\textwidth]{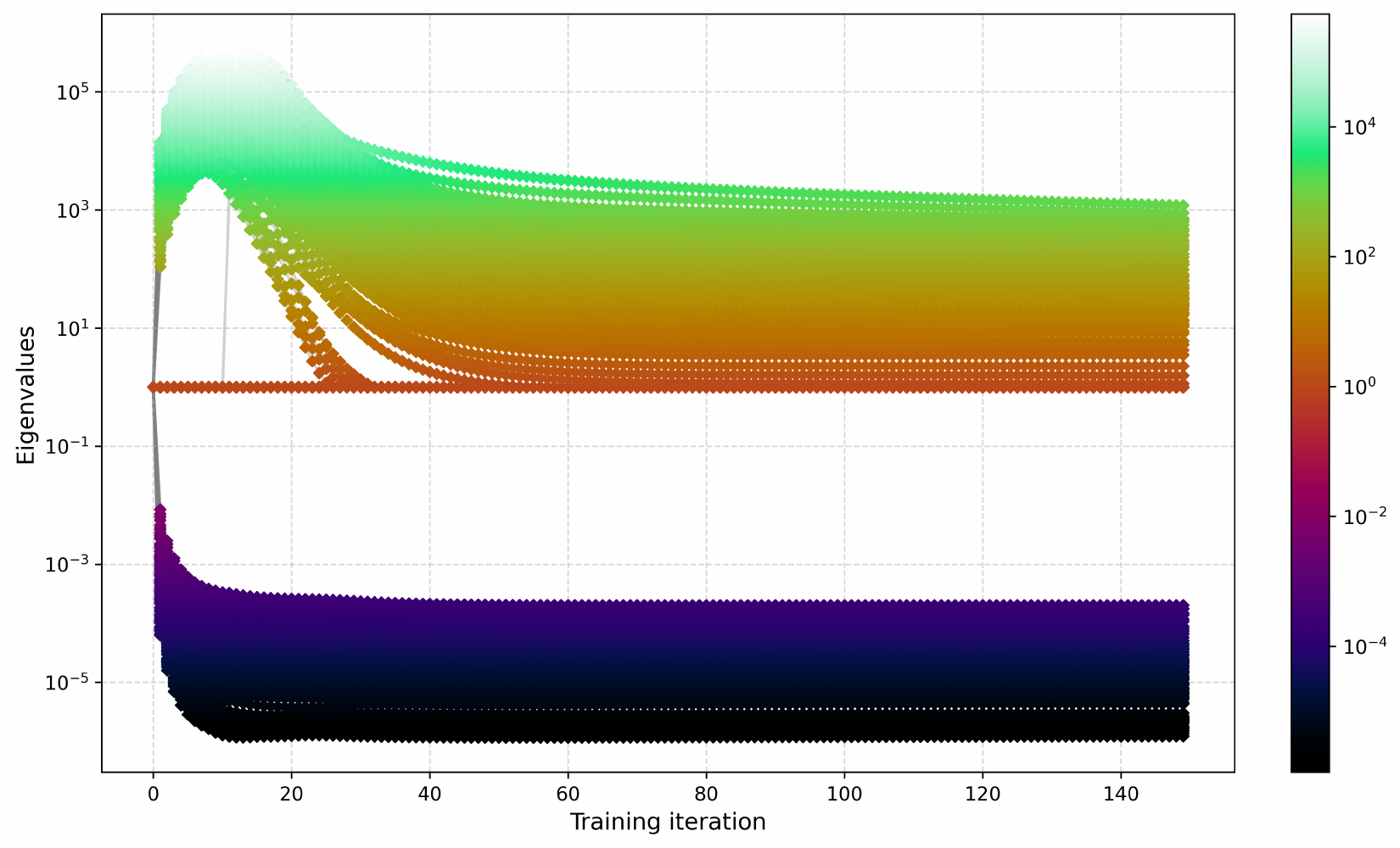}
  \caption{We plot all eigenvalues across one instance of training according to the same setup as Figure \ref{fig:kahler_blowup}, and this figure is also consistent with Figure \ref{fig:kahler_blowup_severe}. The purpose of this figure is to illustrate not all eigenvalues have blown-up due to the anisotropy of $h_0$.}
  \label{fig:kahler_all_eigenvalues}
\end{figure}

\vspace{2mm}

\textbf{Lemma 6 (learning rate deterioration under Calabi-Yau (and sometimes Kähler) blow-up under nondegeneracy).} \textit{Let $(M,\omega)$ be a Kähler information manifold equipped with the regularized loss of \ref{eqn:regularized_loss}, background metric $h_0$, and relative metric endomorphism $H = h_0^{-1}h$. Assume the near-collapse regime $\lambda_{\min}(H) \geq \lambda$, the Calabi-Yau constraint $\det(H) \geq \kappa > 0$, and the Hessian bound $\|\partial^2f\|_2 = \Theta(\frac{1}{\sqrt{m}})$. Assume the eigenvalue blow-up $
\lambda_{\max}(H) = \Omega\left( \frac{\kappa}{\mu^{\widetilde{K}-1}} \right)$, where the first $\widetilde{K}$ eigenvalues of $H$ satisfy $\lambda_i \leq \mu$. Under the nondegeneracy assumptions of Appendix \ref{app:calabi_yau_oscillation}, the covariant Hessian constant and $\beta$-smoothness parameter satisfy $C_{\mathcal H} = \Omega (\lambda^{-2} \kappa\mu^{-(\widetilde{K}-1)} ), \beta = \Omega ( \frac{\kappa}{\lambda^2 \sqrt m} \mu^{-(\widetilde{K}-1)} \sqrt{L(\widetilde\theta_t)} ).$
Consequently, the learning rate is restricted to $
\eta_t = \mathcal O( \frac{\lambda^2 \sqrt m}{\kappa\sqrt{L(\widetilde\theta_t)}} \mu^{\widetilde{K}-1} ).$ If $\beta\eta_t \geq 2$, the monotone descent guarantee of Appendix \ref{app:convergence} is lost.}

\vspace{2mm}

Lemma 6 also relies on a reasonable nondegeneracy assumption, which has strongest justification when: (1) the geometry of $H$ is coupled with $f$ (since $h$ is constructed using $f$, this is true); (2) there is no serious cancellation among the appropriate tensor contractions; (3) the constant $c_0$ is uniform as $\mu$ changes/becomes zero, but $\mu$ is generally bounded away from zero, so this is reasonable as well.

\vspace{2mm}

\noindent \textbf{Lemma 7 (degeneration of Kähler-PL parameter under eigenvalue blow-up).} \textit{Let $(M, \omega)$ be a Kähler information manifold with background metric $h_0$. Define the relative metric endomorphism $H := h_0^{-1}h$. Assume the Calabi-Yau constraint $\det(H) \equiv \kappa > 0$, the eigenvalue blow-up $\lambda_{\max}(H) \geq \Omega ( \frac{\kappa}{\mu^{\widetilde{K}-1}} )$ for some $0 \ll \widetilde{K} \leq K$. Let $\Gamma_t$ be as in \ref{eqn:Gamma_t}. Suppose we have the following assumptions: $
\langle H\dot\gamma,\dot\gamma\rangle_{h_0} \geq c_2\lambda_{\max}(H)\|\dot\gamma\|_{h_0}^2$, and assume suitable smoothness when necessary. Then $\Gamma_t$ is bounded above by $
\Gamma_t \leq \mathcal{O}\left(\frac{\mu^{\widetilde{K}-1}}{\kappa}\right) $. In particular, $\Gamma_t \rightarrow 0$ as $\widetilde{K} \rightarrow \infty$.}

\vspace{2mm}

\noindent \textit{Remark.} Lemmas 6, 7 are our primary results that are most applicable to Calabi-Yau manifolds alone. As we saw in section \ref{sec:calabi_yau}, some of these results are also applicable for general Kähler manifolds but the affects are more severe for Calabi-Yau manifolds with constant determinant. The results in \ref{sec:first_deriv_and_curv} also applicable to Calabi-Yau manifolds, but are more so for manifolds of a range of Ricci curvature. Lemma 11 in \ref{app:gradient_flux_searches} is also a failure mode of Calabi-Yau manifolds specifically.

\subsection{First derivative results and results relating to negative Ricci curvature}
\label{sec:first_deriv_and_curv}

\textbf{Lemma 8 (regret bounds).} \textit{Let $\theta^*$ denote the optimal parameter. Assume $\theta_t$ evolves according to stochastic natural gradient descent $d\theta_t=-\nabla_h\mathcal L(\theta_t)dt+\sqrt{\eta}h(\theta_t)^{-1}\Sigma^{1/2}dW_t$. Define the expected regret along $\theta_t$ over interval $[0,T]$ as
\begin{align}
\label{eqn:regret}
\EX[\mathcal{R}(T)] & = - \EX_{\theta_0} \EX_W \left[ \int_0^T \text{Re} \langle \nabla_h \mathcal{L}(\theta_t), \exp_{\theta_t}^{-1}(\theta^*)^{1,0} \rangle_h dt \right]  .
\end{align}
Then the regret obeys the lower bound $
\EX[\mathcal{R}(T)] \leq \EX[ \delta(\theta_0)] - \EX[\delta(\theta_T)] + \eta K T + \EX[\mathcal{E}_{\text{Ric}}] + \EX[ \mathcal{E}_{\text{diff}} ]$,
where the term $\mathcal{E}_{\text{Ric}}$ is modulated by the Ricci curvature. Because $\text{Ric} \equiv 0$ on the Ricci-flat/Calabi-Yau manifold, the $\mathcal{E}_{\text{Ric}}$ vanishes, term provides no curvature contribution.}

\vspace{2mm}

\noindent \textbf{Lemma 9 (Dirichlet energy).}  \textit{Suppose the regularized loss of \ref{eqn:regularized_loss} holds. Then, the total Dirichlet energy of the network over the dataset, defined for each data point as $E(f_\alpha) = \frac{1}{\text{Vol}_{\omega}(S)} \int_S i \partial f_\alpha \wedge \overline{\partial} f_\alpha \wedge \frac{\omega^{K-1}}{(K-1)!}$, is bounded above by
\begin{align}
0 \leq \sum_{\alpha=1}^N E(f_\alpha) \leq \mathcal{O}(NK + \frac{NK}{\sqrt{m}}) ,
\end{align}
where $N$ is the number of data points in the loss. Note that the lower bound is trivial.
}

\vspace{2mm}

\noindent \textbf{Lemma 10 (gradient bounds).} \textit{Let $(M, \omega)$ be Kähler with metric $h$ full rank. Assume a negative log-likelihood loss. Consider the natural gradient flow $\dot{\theta}^i = -h^{i\overline{j}} \partial_{\overline{j}} \mathcal{L}$. Let $V(t) = \|\nabla_h \mathcal{L}\|_h^2$ denote the gradient norm. Define the uniform Hessian bounds $H(t) = \sup_{\theta \in \Omega} \|\nabla^{2,0}_{\omega} \mathcal{L}\|_h$ and $\mu(t) = \inf_{\theta \in \Omega} \lambda_{\text{min}}(\nabla^{1,1}_{\omega} \mathcal{L})$. Assume the Hessians are locally $L$-Lipschitz with respect to the Kähler metric connection, and holomorphic bisectional curvature is bounded below by $\kappa > 0$. We have
\begin{align}
V(t) \leq V(0) \exp\left( 2t \cdot \mathcal{O}(\frac{1}{\sqrt{m}}) \right) ;
\end{align}
the material derivative of the minimum eigenvalue of the $(1,1)$ Hessian obeys
\begin{align}
\frac{d}{dt} \mu(t) \geq \mu(t)^2 + \|N(X, \cdot)\|_h^2 + \kappa V(t) - X^i X^{\overline{j}} \nabla_{i\overline{j}} (\|\nabla \mathcal{L}\|_h^2) ,
\end{align}
denoting $N_{ij} = \nabla_{ij} \mathcal{L}$ for short.}

\vspace{2mm}

\noindent \textbf{Lemma 11 (integrated gradient flux).} \textit{Let $(M,\omega)$ be Kähler. Denote $\mathcal{W}(\epsilon)$ the quantity as in \ref{eqn:oint_flux_noW}. Then $\mathcal{W}(\epsilon)$ obeys with some shorthand
\begin{align}
\mathcal{W}(\epsilon) = \mathcal{O} \left( \Delta_0 \mathcal{L} \epsilon^{2K+1} + \big[ \langle \text{Ric}, \mathcal{H}_{\mathcal{L}} \rangle - R \Delta_0 \mathcal{L} \big] \epsilon^{2K+3} + \epsilon^{2K+5} \right) .
\end{align}
When $(M,\omega)$ is Calabi-Yau, then the second term vanishes. Moreover, $\mathcal{W}(\epsilon) > \mathcal{W}_{CY}(\epsilon)$ when the manifold is negatively-curved, where $\mathcal{W}_{CY}(\epsilon)$ corresponds to $\mathcal{W}$ in the Calabi-Yau case, meaning the nonzero curvature case has greater escape at initialization than that of vanishing curvature.
}

\vspace{2mm}

\noindent \textit{Remark.} In $\mathcal{W}(\epsilon)$, we are interested in the quantity of the flux across radii $
\label{eqn:oint_flux_noW}
\int_0^\epsilon \left( \oint_{\partial B_r(\theta_0)} d^c \mathcal{L} \wedge \omega^{K-1} \right) dr $, as described in Appendix \ref{app:gradient_flux_searches}. This formulation has connections to the flux under the Riemannian divergence theorem
\begin{align}
\oint_{\partial B_r} \langle \overline{\nabla}\mathcal{L}, n \rangle_h dA \propto \int_{B_r} \overline{\text{div}}(\overline{\nabla}\mathcal{L}) \frac{\omega^K}{K!} \propto \int_{B_r} (\Delta_{\overline{\partial}}\mathcal{L}) \frac{\omega^K}{K!}.
\end{align}
The above says that the flux of the gradient across the boundary is governed by the Dolbeault Laplacian, up to a convention of constants.

\vspace{2mm}

\textit{Remark.} Lemma 11 supports positive Ricci curvature at initialization is desirable.

\vspace{2mm}

\noindent \textbf{Lemma 12 (variance of the parameter).} \textit{Let $U \subseteq M$ be an open subset. Let $\mathcal{L}(\theta): U \rightarrow \mathbb{R}$ be a smooth loss (not necessarily quadratic) with a local minimum at the critical point $\theta^*$. Assume a stochastic natural gradient descent update on the parameter with learning rate $\eta > 0$ over $U$. Furthermore, assume the system reaches a steady-state probability measure (via Fokker-Planck; steady with respect to the gradient descent) given by
\begin{align}
\rho_\infty(\theta) = \frac{1}{\mathcal{Z}} e^{-\frac{2}{\eta} \mathcal{L}(\theta)} ,
\end{align}
where $\mathcal{Z}$ is the normalization constant over the volume form $\frac{\omega^K}{K!}$. Then, the asymptotic variance $V_{\infty} = \mathbb{E} \left[ \|\theta - \theta^*\|_h^2 \right]$ in a local neighborhood of the critical point $\theta^*$ is given by
\begin{align}
\label{eqn:v_infty}
V_\infty = \frac{\eta}{2} \text{Tr}_h \left( \left[ \nabla^{1,1}_{\omega} \mathcal{L} + \frac{\eta}{2} \text{Ric} - \overline{\nabla_{\omega}^{2,0}} \mathcal{L} \left( \overline{\nabla^{1,1}_{\omega}} \mathcal{L} + \frac{\eta}{2} \overline{\text{Ric}} \right)^{-1} \nabla_{\omega}^{2,0} \mathcal{L} \right]^{-1} \right) \Bigg|_{\theta^*} + \mathcal{O}(\eta^2) .
\end{align}
}

\vspace{2mm}

\noindent \textit{Remark.} We can note the linear algebra fact
\begin{align}
\begin{cases}
A + B - \overline{C}(\overline{A} + \overline{B})^{-1} C \succeq A - \overline{C}\overline{A}^{-1} C
\\
B \succeq 0 .
\end{cases}
\end{align}
In this theorem, from the interior tensor in \ref{eqn:v_infty}, we have positive Ricci curvature acts a restoring force to help the optimization with lower variance. Global positivity on the Ricci curvature is nontrivially restrictive. Instead, we can formulate this condition via a "weak restoring force" by evaluating the positivity of the holomorphic tangent bundle. We say the landscape provides a sufficient restoring force at $\theta \in M$ if \cite{popovici2026mpositivestabilityholomorphicvector}
\begin{align}
\label{eqn:semi_positivity}
\Bigg\{(i\Theta_h(T^{1,0}M) \wedge \omega^{q-1} \wedge \Omega)u, u \Bigg\}_h \geq 0 ,
\end{align}
for $u \in T^{1,0}M_\theta$, integer $1 \leq q \leq K$, and a nowhere-vanishing form $\Omega \in \Lambda^{K-q, K-q}T_\theta^* M$ with $\Omega > 0$ (metrically weakly). Here, $i\Theta_h(T^{1,0}M)$, an $\text{End}(T^{1,0}M)$-valued $(1,1)$-form, represents the Chern curvature form of the tangent bundle. Because the Ricci form is the trace of this Chern curvature, an $\omega$-$q$-semi-positive parameter landscape guarantees that the metric geometry prevents high variance in \ref{eqn:v_infty} in high negative curvature regimes, even if the manifold is locally Calabi-Yau or exhibits minor eigenvalue collapse. In particular, the landscape is allowed to have areas and directions of bad curvature that adversely affect \ref{eqn:v_infty}, but the net effective curvature, when weighted against the specific metric structure of $\Omega$ in the relevant dimensions $q$, remains non-negative. This discussion ties into Lemma 14.

\vspace{2mm}

\noindent \textbf{Lemma 13 (minimum eigenvalue of the Witten Laplacian).} \textit{Let $U \subseteq M$ be an open subset of a Kähler manifold and $K \subseteq U$ a compact subset. Consider the deformed Laplacian $\Delta_\eta = \overline{\partial}_\eta \overline{\partial}_\eta^\dagger + \overline{\partial}_\eta^\dagger \overline{\partial}_\eta$, where $\overline{\partial}_\eta = \overline{\partial} + \frac{1}{\eta} \overline{\partial}\mathcal{L} \wedge$. Assume the Ricci curvature is bounded above such that $\text{Ric}(h) \leq - \kappa \omega$ for some constant $\kappa > 0$. Define $\beta_{1,1} = \sup_{\theta \in U} \|\nabla^{1,1}_{\omega}\mathcal{L}\|_2, M_\eta \leq \frac{C_0}{\eta\text{Tr}\big(g\cdot\mathrm{Hess}\mathcal{L}(\theta_0)\big)}$, where $g$ is the realification of $h$, up to constants, and $C_0$ is some constant. Denote $\theta_0$ a unique minimum of $U$. Then, the minimal eigenvalue $\lambda_1$ of $\Delta_\eta$ is bounded by
\begin{align}
\lambda_1 \leq \frac{(2+K)\beta_{1,1}}{\eta} - \kappa + M_\eta .
\end{align}
}

\vspace{2mm}

\noindent \textit{Remark.}  The above dividies by $\eta$ and picks up a factor of $K$ due to a trace, which are not ideal. Moreover, it is possible to redo the proof of Lemma 13 requiring a division on a gradient term, but this definitely diverges with an extrema in the set of interest. We find the above to be a better result. 

\vspace{2mm}

\noindent \textit{Remark.} The choice of the Witten Laplacian is meaningful because it incorporates the loss $\mathcal{L}$ itself into the geometric object. The Witten Laplacian \cite{Michel2019AboutSmallEigenvalues} was first interested by Witten \cite{158d02b66ce6495492b1b30126556c95} to study Morse inequalities, but it applies to our context here too. We examine the minimum eigenvalue of the Witten Laplacian. Our Witten Laplacian is $\Delta_\eta = \overline{\partial}_\eta \overline{\partial}_\eta^\dagger + \overline{\partial}_\eta^\dagger \overline{\partial}_\eta$, where $\overline{\partial}_\eta = \overline{\partial} + \frac{1}{\eta} \overline{\partial}\mathcal{L} \wedge$. The eigenvalues of this operator determine a rate of convergence. To understand this, we must turn to the kernel of the differential operator, since a steady state corresponds to both an optimal state and the kernel of the operator. 

\vspace{2mm}

\noindent \textbf{Lemma 14 (divergence and diffusion bounds with Chern curvature)}. \textit{Let $(M,\omega)$ be a Kähler manifold of complex dimension $K$ with Kähler form $\omega$. Let $\mathcal{L}$ be with the natural gradient descent vector field given by $V = -\nabla^{1,0}_h \mathcal{L}$, and let $\Theta = \text{div}_h(V)$ denote its divergence. Suppose $M$ satisfies the $\omega-q$-semi-positivity hypothesis with respect to a positive $(p,p)$-form $\Omega$ such that
\begin{align}
\Bigg\{(i\Theta_h(T^{1,0}M) \wedge \omega^{q-1} \wedge \Omega) V, V \Bigg\}_h \geq 0 .
\end{align}
Then, the material derivative of the divergence, bounded by diffusion and expansion, satisfy the inequalities
\begin{align}
-\mathcal{O}\left(\frac{1}{m}\right) + \kappa_{\max} \dot{\mathcal{V}} \leq \dot{\Theta} + \frac{1}{2} \Delta_h \dot{\mathcal{V}} \leq - \|\nabla^{1,1}_{\omega} \mathcal{L}\|_h^2 - \|\nabla^{2,0}_{\omega} \mathcal{L}\|_h^2 + \frac{K}{M_{q,\Omega}} \mathcal{P}_{q,\Omega}(V) ,
\end{align}
where $\dot{\mathcal{V}} = -\|V\|_h^2, M_{q,\Omega} = \star(\omega^q \wedge \Omega), \mathcal{P}_{q,\Omega}(V) = \star(\Theta_{V, \text{prim}} \wedge \omega^{q-1} \wedge \Omega)$ is the primitive curvature term. Furthermore, as $K \to \infty$, this upper bound diverges to $+\infty$ since the primitive curvature term scales linearly in $K$, $\Omega(1)$. Consequently, in high-dimensional spaces, the parameter flow has potential to "disperse," as the guarantee of convergence is lost.
}

\section{Conclusions} We studied information cross-entropy manifolds in a complex geometric lens for optimization. We focused on results based in \cite{banerjee2023restricted} and adapted and significantly changed these results for arguments of a greater geometric flavor, which use a deep learning theory base via the result of \ref{app:initialization}. We have expanded upon these results and considered diverse setups, namely through a dynamic Kähler Polyak-Łojasiewicz condition. We discussed failure modes of Calabi-Yau manifolds and modes pertaining to small and Ricci negative curvature in general. These results are notable because standard literature results are for sectional curvature \cite{pmlr-v178-criscitiello22a}, not Ricci curvature, at least in a deep learning context: we remark Ricci curvature involvement in optimization exists in literature in general \cite{lott2009ricci}. One limitation of our methods is that in practice it is nontrivial to compute our geometric structures such as local neighborhoods of the Kähler manifold corresponding to the cross-entropy metric of \ref{eqn:metric}. Moreover, in order to perform natural gradient descent, the metric must be computed, but the expectations can be approximated so this is more of an inconvenience and not a limitation. In general, natural gradient descent is standard \cite{shrestha2023naturalgradientmethodsperspectives}. Our work is mostly theoretical, yet confirmable with some experiments.

\bibliographystyle{plainnat}
\bibliography{bibliography}

@inproceedings{
banerjee2023restricted,
title={Restricted Strong Convexity of Deep Learning Models with Smooth Activations},
author={Arindam Banerjee and Pedro Cisneros-Velarde and Libin Zhu and Misha Belkin},
booktitle={The Eleventh International Conference on Learning Representations },
year={2023},
url={https://openreview.net/forum?id=PINRbk7h01}
}

@inproceedings{
dyer2019asymptoticswidenetworksfeynman,
title={Asymptotics of Wide Networks from Feynman Diagrams},
author={Ethan Dyer and Guy Gur-Ari},
booktitle={International Conference on Learning Representations},
year={2020},
url={https://openreview.net/forum?id=S1gFvANKDS}
}

@misc{zhu2022notelinearbottlenecknetworks,
      title={A note on Linear Bottleneck networks and their Transition to Multilinearity}, 
      author={Libin Zhu and Parthe Pandit and Mikhail Belkin},
      year={2022},
      eprint={2206.15058},
      archivePrefix={arXiv},
      primaryClass={cs.LG},
      url={https://arxiv.org/abs/2206.15058}, 
}

@inproceedings{
shemur2024weakcorrelationsunderlyingprinciple,
title={Weak Correlations as the Underlying Principle for Linearization of Gradient-Based Learning Systems},
author={Ori Shem-Ur and Khen Cohen and Yaron Oz},
booktitle={The Fourteenth International Conference on Learning Representations},
year={2026},
url={https://openreview.net/forum?id=vokk8t1gnp}
}

@misc{aitken2020asymptoticswidenetworkspolynomial,
      title={On the asymptotics of wide networks with polynomial activations}, 
      author={Kyle Aitken and Guy Gur-Ari},
      year={2020},
      eprint={2006.06687},
      archivePrefix={arXiv},
      primaryClass={cs.LG},
      url={https://arxiv.org/abs/2006.06687}, 
}

@misc{yaida2022metaprincipledfamilyhyperparameterscaling,
      title={Meta-Principled Family of Hyperparameter Scaling Strategies}, 
      author={Sho Yaida},
      year={2022},
      eprint={2210.04909},
      archivePrefix={arXiv},
      primaryClass={cs.LG},
      url={https://arxiv.org/abs/2210.04909}, 
}

@inproceedings{
guillen2026finitewidthneuraltangentkernels,
title={Finite-Width Neural Tangent Kernels from Feynman Diagrams},
author={Max Guillen and Philipp Misof and Jan E Gerken},
booktitle={Forty-third International Conference on Machine Learning},
year={2026},
url={https://openreview.net/forum?id=SOlPHMdSY3}
}

@inproceedings{Herzlich2000,
  author    = {Marc Herzlich},
  title     = {Refined Kato inequalities in Riemannian Geometry},
  booktitle = {Journées Équations aux dérivées partielles},
  year      = {2000},
  pages     = {1--11},
  publisher = {Journées Équations aux dérivées partielles},
  url       = {https://www.numdam.org/item/10.5802/jedp.570.pdf}
}

@article{hanin2023randomfullyconnectedneural,
  author  = {Boris Hanin},
  title   = {Random Fully Connected Neural Networks as Perturbatively Solvable Hierarchies},
  journal = {Journal of Machine Learning Research},
  year    = {2024},
  volume  = {25},
  number  = {267},
  pages   = {1--58},
  url     = {http://jmlr.org/papers/v25/23-0643.html}
}

@InProceedings{huang2019dynamicsdeepneuralnetworks,
  title = 	 {Dynamics of Deep Neural Networks and Neural Tangent Hierarchy},
  author =       {Huang, Jiaoyang and Yau, Horng-Tzer},
  booktitle = 	 {Proceedings of the 37th International Conference on Machine Learning},
  pages = 	 {4542--4551},
  year = 	 {2020},
  editor = 	 {III, Hal Daumé and Singh, Aarti},
  volume = 	 {119},
  series = 	 {Proceedings of Machine Learning Research},
  month = 	 {13--18 Jul},
  publisher =    {PMLR},
  url = 	 {https://proceedings.mlr.press/v119/huang20l.html}
}

@misc{andreassen2020asymptoticswideconvolutionalneural,
      title={Asymptotics of Wide Convolutional Neural Networks}, 
      author={Anders Andreassen and Ethan Dyer},
      year={2020},
      eprint={2008.08675},
      archivePrefix={arXiv},
      primaryClass={cs.LG},
      url={https://arxiv.org/abs/2008.08675}, 
}

@misc{cirone2025genusexpansionnonlinearrandom,
      title={Genus expansion for non-linear random matrix ensembles with applications to neural networks}, 
      author={Nicola Muca Cirone and Jad Hamdan and Cristopher Salvi},
      year={2025},
      eprint={2407.08459},
      archivePrefix={arXiv},
      primaryClass={math.PR},
      url={https://arxiv.org/abs/2407.08459}, 
}

@article{liu2021losslandscapesoptimizationoverparameterized,
title = {Loss landscapes and optimization in over-parameterized non-linear systems and neural networks},
journal = {Applied and Computational Harmonic Analysis},
volume = {59},
pages = {85-116},
year = {2022},
note = {Special Issue on Harmonic Analysis and Machine Learning},
issn = {1063-5203},
doi = {https://doi.org/10.1016/j.acha.2021.12.009},
url = {https://www.sciencedirect.com/science/article/pii/S106352032100110X},
author = {Chaoyue Liu and Libin Zhu and Mikhail Belkin}
}

@article{
cisneros-velarde2025optimization,
title={Optimization and Generalization Guarantees for Weight Normalization},
author={Pedro Cisneros-Velarde and Zhijie Chen and Sanmi Koyejo and Arindam Banerjee},
journal={Transactions on Machine Learning Research},
issn={2835-8856},
year={2025},
url={https://openreview.net/forum?id=gpHOtQQPJG},
note={}
}

@inproceedings{
taheri2024sharperguaranteeslearningneural,
title={Sharper Guarantees for Learning Neural Network Classifiers with Gradient Methods},
author={Hossein Taheri and Christos Thrampoulidis and Arya Mazumdar},
booktitle={The Thirteenth International Conference on Learning Representations},
year={2025},
url={https://openreview.net/forum?id=h7GAgbLSmC}
}

@inproceedings{liu2021linearitylargenonlinearmodels,
author = {Liu, Chaoyue and Zhu, Libin and Belkin, Mikhail},
title = {On the linearity of large non-linear models: when and why the tangent kernel is constant},
year = {2020},
isbn = {9781713829546},
publisher = {Curran Associates Inc.},
address = {Red Hook, NY, USA},
booktitle = {Proceedings of the 34th International Conference on Neural Information Processing Systems},
articleno = {1338},
numpages = {11},
location = {Vancouver, BC, Canada},
series = {NIPS '20}
}

@article{riis2018geometricintegrationapproachnonsmooth,
  author   = {Riis, Erlend S. and Ehrhardt, Matthias J. and Quispel, G. R. W. and Sch\"{o}nlieb, Carola-Bibiane},
  title    = {A Geometric Integration Approach to Nonsmooth, Nonconvex Optimisation},
  journal  = {Foundations of Computational Mathematics},
  volume   = {22},
  number   = {5},
  pages    = {1351--1394},
  year     = {2022},
  month    = {10},
  issn     = {1615-3383},
  doi      = {10.1007/s10208-020-09489-2},
  url      = {https://doi.org/10.1007/s10208-020-09489-2}
}

@article{Ehrhardt_2024,
   title={A geometric integration approach to smooth optimization: foundations of the discrete gradient method},
   volume={45},
   ISSN={1464-3642},
   url={http://dx.doi.org/10.1093/imanum/drae037},
   DOI={10.1093/imanum/drae037},
   number={3},
   journal={IMA Journal of Numerical Analysis},
   publisher={Oxford University Press (OUP)},
   author={Ehrhardt, Matthias J and Riis, Erlend S and Ringholm, Torbjørn and Schönlieb, Carola-Bibiane},
   year={2024},
   month=July, pages={1269–1299} }

@article{ringholm2018variationalimageregularizationeulers,
author = {Ringholm, Torbj{\o}rn and Lazi{\'c}, Jasmina and Sch{\"o}nlieb, Carola-Bibiane},
title = {Variational Image Regularization with Euler's Elastica Using a Discrete Gradient Scheme},
journal = {SIAM Journal on Imaging Sciences},
volume = {11},
number = {4},
pages = {2665-2691},
year = {2018},
doi = {10.1137/17M1162354},

URL = { 
    
        https://doi.org/10.1137/17M1162354
},
eprint = { 
    
        https://doi.org/10.1137/17M1162354
}
}

@misc{niu2026continuoustimedynamicsdifferenceofconvexalgorithm,
      title={Continuous-Time Dynamics of the Difference-of-Convex Algorithm}, 
      author={Yi-Shuai Niu},
      year={2026},
      eprint={2604.06926},
      archivePrefix={arXiv},
      primaryClass={math.OC},
      url={https://arxiv.org/abs/2604.06926}, 
}

@misc{niu2022convergenceanalysisdca,
      title={On the convergence analysis of DCA}, 
      author={Yi-Shuai Niu},
      year={2022},
      eprint={2211.10942},
      archivePrefix={arXiv},
      primaryClass={math.OC},
      url={https://arxiv.org/abs/2211.10942}, 
}

@article{HauerMazon2019,
  author   = {Hauer, Daniel and Maz\'{o}n, Jos\'{e} M.},
  title    = {Kurdyka–Łojasiewicz–Simon inequality for gradient flows in metric spaces},
  journal  = {Trans. Amer. Math. Soc.},
  volume   = {372},
  year     = {2019},
  pages    = {4917--4976},
  doi      = {10.1090/tran/7801},
  mrnumber = {4009443},
  url={https://doi.org/10.1090/tran/7801}
}

@article{Dello_Schiavo_2024,
   title={Local conditions for global convergence of gradient flows and proximal point sequences in metric spaces},
   ISSN={1088-6850},
   url={http://dx.doi.org/10.1090/tran/9156},
   DOI={10.1090/tran/9156},
   journal={Transactions of the American Mathematical Society},
   publisher={American Mathematical Society (AMS)},
   author={Dello Schiavo, Lorenzo and Maas, Jan and Pedrotti, Francesco},
   year={2024},
   month=Apr }

@article{Sun_2025,
   title={A geometric modeling of Occam’s razor in deep learning},
   volume={8},
   ISSN={2511-249X},
   url={http://dx.doi.org/10.1007/s41884-025-00167-2},
   DOI={10.1007/s41884-025-00167-2},
   number={S1},
   journal={Information Geometry},
   publisher={Springer Science and Business Media LLC},
   author={Sun, Ke and Nielsen, Frank},
   year={2025},
   month=June, pages={233–273} }

@misc{bakeer2026localinformationoperatorsspatial,
      title={Local Information Operators for Spatial Identifiability in Distributed-Parameter Inverse Problems in Computational Mechanics}, 
      author={Tammam Bakeer},
      year={2026},
      eprint={2605.28601},
      archivePrefix={arXiv},
      primaryClass={cs.CE},
      url={https://arxiv.org/abs/2605.28601}, 
}

@misc{dong2026quotientgeometryeffectivecurvature,
      title={Quotient Geometry, Effective Curvature, and Implicit Bias in Simple Shallow Neural Networks}, 
      author={Hang-Cheng Dong and Pengcheng Cheng},
      year={2026},
      eprint={2603.21502},
      archivePrefix={arXiv},
      primaryClass={cs.LG},
      url={https://arxiv.org/abs/2603.21502}, 
}

@inproceedings{schwachhöfer2017congruentfamiliesinvarianttensors,
  author    = {Schwachh{\"o}fer, Lorenz and Ay, Nihat and Jost, J{\"u}rgen and V{\^a}n L{\^e}, H{\^o}ng},
  editor    = {Ay, Nihat and Gibilisco, Paolo and Mat{\'u}{\v{s}}, Franti{\v{s}}ek},
  title     = {Congruent Families and Invariant Tensors},
  booktitle = {Information Geometry and Its Applications},
  year      = {2018},
  publisher = {Springer International Publishing},
  address   = {Cham},
  pages     = {157--187},
  isbn      = {978-3-319-97798-0},
  doi       = {10.1007/978-3-319-97798-0_6}
}

@misc{gnandi2026constructionexponentialfamiliesstatistical,
      title={Construction of Exponential Families from Statistical Manifolds}, 
      author={Emmanuel Gnandi},
      year={2026},
      eprint={2511.23444},
      archivePrefix={arXiv},
      primaryClass={math.DG},
      url={https://arxiv.org/abs/2511.23444}, 
}

@article{Hezari2016,
  author   = {Hezari, Hamid and Kelleher, Casey and Seto, Shoo and Xu, Hang},
  title    = {Asymptotic Expansion of the Bergman Kernel via Perturbation of the Bargmann--Fock Model},
  journal  = {The Journal of Geometric Analysis},
  year     = {2016},
  volume   = {26},
  number   = {4},
  pages    = {2602--2638},
  month    = {10},
  issn     = {1559-002X},
  doi      = {10.1007/s12220-015-9641-3},
  url      = {https://doi.org/10.1007/s12220-015-9641-3}
}

@article{ruan1996canonicalcoordinatesbergmanmetrics,
      title={Canonical coordinates and Bergman metrics}, 
      author={Wei-Dong Ruan},
      journal = {Communications in Analysis and Geometry},
      year    = {1998},
      volume  = {6},
      number  = {3},
      url     = {https://intlpress.com/site/pub/files/_fulltext/journals/cag/1998/0006/0003/CAG-1998-0006-0003-a005.pdf}
}

@inproceedings{
murray2023characterizingspectrumntkpower,
title={Characterizing the spectrum of the {NTK} via a power series expansion},
author={Michael Murray and Hui Jin and Benjamin Bowman and Guido Montufar},
booktitle={The Eleventh International Conference on Learning Representations },
year={2023},
url={https://openreview.net/forum?id=Tvms8xrZHyR}
}

@article{10.1162/089976698300017746,
    author = {Amari, Shun-ichi},
    title = {Natural Gradient Works Efficiently in Learning},
    journal = {Neural Computation},
    volume = {10},
    number = {2},
    pages = {251-276},
    year = {1998},
    month = {02},
    issn = {0899-7667},
    doi = {10.1162/089976698300017746},
    url = {https://doi.org/10.1162/089976698300017746},
    eprint = {https://direct.mit.edu/neco/article-pdf/10/2/251/813415/089976698300017746.pdf},
}

@article{mishra2025hermitianyangmillsconnectionsgeneral,
  author   = {Mishra, Challenger and Tan, Justin},
  title    = {Hermitian Yang-Mills connections on general vector bundles: geometry and physical Yukawa couplings},
  journal  = {Journal of High Energy Physics},
  volume   = {2026},
  number   = {6},
  pages    = {93},
  year     = {2026},
  month    = {06},
  issn     = {1029-8479},
  doi      = {10.1007/JHEP06(2026)093},
  url      = {https://doi.org/10.1007/JHEP06(2026)093}
}

@misc{karkada2024lazyntkrichmup,
      title={The lazy (NTK) and rich ($\mu$P) regimes: a gentle tutorial}, 
      author={Dhruva Karkada},
      year={2024},
      eprint={2404.19719},
      archivePrefix={arXiv},
      primaryClass={cs.LG},
      url={https://arxiv.org/abs/2404.19719}, 
}

@misc{dufortlabbé2026navigatingpotholesgeometryawaresharpness,
      title={Navigating Potholes with Geometry-Aware Sharpness Minimization}, 
      author={Simon Dufort-Labbé and Mehrab Hamidi and Razvan Pascanu and Ioannis Mitliagkas and Damien Scieur and Aristide Baratin},
      year={2026},
      eprint={2605.16134},
      archivePrefix={arXiv},
      primaryClass={cs.LG},
      url={https://arxiv.org/abs/2605.16134}, 
}

@misc{cayci2025riemannianoptimizationperspectivegaussnewton,
      title={A Riemannian Optimization Perspective of the Gauss-Newton Method for Feedforward Neural Networks}, 
      author={Semih Cayci},
      year={2025},
      eprint={2412.14031},
      archivePrefix={arXiv},
      primaryClass={math.OC},
      url={https://arxiv.org/abs/2412.14031}, 
}

@misc{fang2025laplaciancomparisontheoremscomplete,
      title={Laplacian comparison theorems on complete K\"ahler manifolds and applications}, 
      author={Jiaxuan Fang and Zhiyao Xiong and Xiaokui Yang},
      year={2025},
      eprint={2510.01548},
      archivePrefix={arXiv},
      primaryClass={math.DG},
      url={https://arxiv.org/abs/2510.01548}, 
}

@article{Tam2012,
  author   = {Tam, Luen-Fai and Yu, Chengjie},
  title    = {Some comparison theorems for K\"{a}hler manifolds},
  journal  = {Manuscripta Mathematica},
  year     = {2012},
  volume   = {137},
  number   = {3},
  pages    = {483--495},
  month    = {03},
  issn     = {1432-1785},
  doi      = {10.1007/s00229-011-0477-2},
  url      = {https://doi.org/10.1007/s00229-011-0477-2}
}

@misc{wang2024indexform,
  author       = {Zuoqin Wang},
  title        = {Lecture 20: The Index Form},
  year         = {2024},
  howpublished = {Lecture notes for Riemannian Geometry, University of Science and Technology of China},
  url          = {http://staff.ustc.edu.cn/~wangzuoq/Courses/24S-RiemGeom/Notes/Lec20.pdf}
}

@misc{abdalla2023complexvaluedneuralnetworks,
      title={Complex-valued Neural Networks -- Theory and Analysis}, 
      author={Rayyan Abdalla},
      year={2023},
      eprint={2312.06087},
      archivePrefix={arXiv},
      primaryClass={cs.LG},
      url={https://arxiv.org/abs/2312.06087}, 
}

@inproceedings{trabelsi2018deepcomplexnetworks,
  title={Deep Complex Networks},
  author={Chiheb Trabelsi and Olexa Bilaniuk and Ying Zhang and Dmitriy Serdyuk and Sandeep Subramanian and Jo{\~a}o Felipe Santos and Soroush Mehri and Negar Rostamzadeh and Yoshua Bengio and Christopher J Pal},
  booktitle={International Conference on Learning Representations},
  year={2018},
  url={https://openreview.net/forum?id=H1T2hmZAb}
}

@misc{lawson2023fishergeometrygeodesicsmultivariate,
      title={The Fisher Geometry and Geodesics of the Multivariate Normals, without Differential Geometry}, 
      author={Brodie A. J. Lawson and Kevin Burrage and Kerrie Mengersen and Rodrigo Weber dos Santos},
      year={2023},
      eprint={2306.01278},
      archivePrefix={arXiv},
      primaryClass={math.ST},
      url={https://arxiv.org/abs/2306.01278}, 
}

@article{javanmard2019analysistwolayerneuralnetwork,
  author  = {Javanmard, Adel and Mondelli, Marco and Montanari, Andrea},
  title   = {Analysis of a two-layer neural network via displacement convexity},
  journal = {The Annals of Statistics},
  volume  = {48},
  number  = {6},
  pages   = {3619--3642},
  year    = {2020},
  month   = {12},
  doi     = {10.1214/20-AOS1945},
  url     = {https://projecteuclid.org/journals/annals-of-statistics/volume-48/issue-6/Analysis-of-a-two-layer-neural-network-via-displacement-convexity/10.1214/20-AOS1945.full}
}

@InProceedings{daneshmand2023efficientdisplacementconvexoptimization,
  title = 	 {Efficient displacement convex optimization with particle gradient descent},
  author =       {Daneshmand, Hadi and Lee, Jason D. and Jin, Chi},
  booktitle = 	 {Proceedings of the 40th International Conference on Machine Learning},
  pages = 	 {6836--6854},
  year = 	 {2023},
  editor = 	 {Krause, Andreas and Brunskill, Emma and Cho, Kyunghyun and Engelhardt, Barbara and Sabato, Sivan and Scarlett, Jonathan},
  volume = 	 {202},
  series = 	 {Proceedings of Machine Learning Research},
  month = 	 {23--29 Jul},
  publisher =    {PMLR},
  url = 	 {https://proceedings.mlr.press/v202/daneshmand23a.html}
}

@misc{zhang2022meanfieldanalysistwolayerneural,
      title={Mean-Field Analysis of Two-Layer Neural Networks: Global Optimality with Linear Convergence Rates}, 
      author={Jingwei Zhang and Xunpeng Huang and Jincheng Yu},
      year={2022},
      eprint={2205.09860},
      archivePrefix={arXiv},
      primaryClass={cs.LG},
      url={https://arxiv.org/abs/2205.09860}, 
}

@article{amari2018statisticalneurodynamicsdeepnetworks,
  author  = {Amari, Shun-ichi and Karakida, Ryo and Oizumi, Masafumi},
  title   = {Statistical neurodynamics of deep networks: Geometry of signal spaces},
  journal = {Nonlinear Theory and Its Applications, IEICE},
  year    = {2019},
  volume  = {10},
  number  = {4},
  pages   = {322--336}
}

@InProceedings{zavatoneveth2025doestrainingshaperiemannian,
  title = 	 {How does training shape the Riemannian geometry of neural network representations?},
  author =       {Zavatone-Veth, Jacob A and Yang, Sheng and Rubinfien, Julian Alex and Pehlevan, Cengiz},
  booktitle = 	 {Proceedings of the 4th (2025) and 3rd (2024) NeurIPS Workshops on Symmetry and Geometry in Neural Representations},
  pages = 	 {792--883},
  year = 	 {2026},
  editor = 	 {Acosta, Francisco and Azeglio, Simone and Tolooshams, Bahareh and van de Geijn, Chase and Shewmake, Christian and Sanborn, Sophia and Miolane, Nina},
  volume = 	 {282},
  series = 	 {Proceedings of Machine Learning Research},
  month = 	 {14 Dec 2024--07 Dec 2025},
  publisher =    {PMLR},
  url = 	 {https://proceedings.mlr.press/v282/zavatone-veth26a.html}
}

@inproceedings{NIPS2016_14851003,
 author = {Poole, Ben and Lahiri, Subhaneil and Raghu, Maithra and Sohl-Dickstein, Jascha and Ganguli, Surya},
 booktitle = {Advances in Neural Information Processing Systems},
 editor = {D. Lee and M. Sugiyama and U. Luxburg and I. Guyon and R. Garnett},
 pages = {},
 publisher = {Curran Associates, Inc.},
 title = {Exponential expressivity in deep neural networks through transient chaos},
 url = {https://proceedings.neurips.cc/paper_files/paper/2016/file/148510031349642de5ca0c544f31b2ef-Paper.pdf},
 volume = {29},
 year = {2016}
}

@article{Tron_2024,
   title={Cartan moving frames and the data manifolds},
   volume={7},
   ISSN={2511-249X},
   url={http://dx.doi.org/10.1007/s41884-024-00159-8},
   DOI={10.1007/s41884-024-00159-8},
   number={S2},
   journal={Information Geometry},
   publisher={Springer Science and Business Media LLC},
   author={Tron, Eliot and Fioresi, Rita and Couëllan, Nicolas and Puechmorel, Stéphane},
   year={2024},
   month=Nov, pages={883–912} }

@article{article,
author = {Kaul, Piyush and Lall, Brejesh},
year = {2019},
month = {06},
pages = {1410-1416},
title = {Riemannian Curvature of Deep Neural Networks},
volume = {31},
journal = {IEEE Transactions on Neural Networks and Learning Systems},
doi = {10.1109/TNNLS.2019.2919705}
}

@article{Gigli_2017,
   title={Nonsmooth differential geometry– An approach tailored for spaces with Ricci curvature bounded from below},
   volume={251},
   ISSN={0065-9266},
   url={http://dx.doi.org/10.1090/memo/1196},
   DOI={10.1090/memo/1196},
   number={1196},
   journal={Memoirs of the American Mathematical Society},
   publisher={American Mathematical Society (AMS)},
   author={Gigli, Nicola},
   year={2017},
   month=Nov }

@article{lott2009ricci,
  title={Ricci curvature for metric-measure spaces via optimal transport},
  author={Lott, John and Villani, C{\'e}dric},
  journal={Annals of Mathematics},
  volume={169},
  number={3},
  pages={903--991},
  year={2009},
  doi={10.4007/annals.2009.169.903},
  mrnumber={2480619},
  zmnumber={1178.53038},
  url={https://doi.org/10.4007/annals.2009.169.903}
}

@article{Ambrosio_2015,
   title={Bakry–Émery curvature-dimension condition and Riemannian Ricci curvature bounds},
   volume={43},
   ISSN={0091-1798},
   url={http://dx.doi.org/10.1214/14-AOP907},
   DOI={10.1214/14-aop907},
   number={1},
   journal={The Annals of Probability},
   publisher={Institute of Mathematical Statistics},
   author={Ambrosio, Luigi and Gigli, Nicola and Savaré, Giuseppe},
   year={2015},
   month=Feb }

@InProceedings{pmlr-v178-criscitiello22a,
  title = 	 {Negative curvature obstructs acceleration for strongly geodesically convex optimization, even with exact first-order oracles},
  author =       {Criscitiello, Christopher and Boumal, Nicolas},
  booktitle = 	 {Proceedings of Thirty Fifth Conference on Learning Theory},
  pages = 	 {496--542},
  year = 	 {2022},
  editor = 	 {Loh, Po-Ling and Raginsky, Maxim},
  volume = 	 {178},
  series = 	 {Proceedings of Machine Learning Research},
  month = 	 {02--05 Jul},
  publisher =    {PMLR},
  url = 	 {https://proceedings.mlr.press/v178/criscitiello22a.html}
}

@InProceedings{pmlr-v195-criscitiello23a,
  title = 	 {Curvature and complexity: Better lower bounds for geodesically convex optimization},
  author =       {Criscitiello, Christopher and Boumal, Nicolas},
  booktitle = 	 {Proceedings of Thirty Sixth Conference on Learning Theory},
  pages = 	 {2969--3013},
  year = 	 {2023},
  editor = 	 {Neu, Gergely and Rosasco, Lorenzo},
  volume = 	 {195},
  series = 	 {Proceedings of Machine Learning Research},
  month = 	 {12--15 Jul},
  publisher =    {PMLR},
  url = 	 {https://proceedings.mlr.press/v195/criscitiello23a.html}
}

@misc{shrestha2023naturalgradientmethodsperspectives,
      title={Natural Gradient Methods: Perspectives, Efficient-Scalable Approximations, and Analysis}, 
      author={Rajesh Shrestha},
      year={2023},
      eprint={2303.05473},
      archivePrefix={arXiv},
      primaryClass={cs.LG},
      url={https://arxiv.org/abs/2303.05473}, 
}

@inproceedings{stoica2020chiralasymmetryweakinteraction,
  title={Some applications of Clifford Algebras to the Standard Model},
  author={Stoica, Ovidiu Cristinel},
  booktitle={Proceedings of the 10th Mathematical Physics Meeting: School and Conference on Modern Mathematical Physics},
  editor={Dragovich, Branko and Salom, Igor and Vojinovic, Marko},
  year={2020},
  publisher={Institute of Physics Belgrade},
  address={Belgrade, Serbia},
  url={http://www.mphys10.ipb.ac.rs/proceedings10/Stoica.pdf}
}

@misc{gilgarcía2026torsionparallelpurespinors,
      title={Torsion parallel pure spinors on neutral manifolds}, 
      author={Alejandro Gil-García},
      year={2026},
      eprint={2607.06358},
      archivePrefix={arXiv},
      primaryClass={math.DG},
      url={https://arxiv.org/abs/2607.06358}, 
}

@article{McNeal2015,
  author   = {McNeal, Jeffery D. and Varolin, Dror},
  title    = {{$L^2$} estimates for the {$\bar{\partial }$} operator},
  journal  = {Bulletin of Mathematical Sciences},
  year     = {2015},
  volume   = {5},
  number   = {2},
  pages    = {179--249},
  month    = {07},
  issn     = {1664-3615},
  doi      = {10.1007/s13373-015-0068-8},
  url      = {https://doi.org/10.1007/s13373-015-0068-8}
}

@article{Fritz_2020,
   title={A synthetic approach to Markov kernels, conditional independence and theorems on sufficient statistics},
   volume={370},
   ISSN={0001-8708},
   url={http://dx.doi.org/10.1016/j.aim.2020.107239},
   DOI={10.1016/j.aim.2020.107239},
   journal={Advances in Mathematics},
   publisher={Elsevier BV},
   author={Fritz, Tobias},
   year={2020},
   month=Aug, pages={107239} }

@book{Ziemer2017,
  author    = {William P. Ziemer},
  title     = {Modern Real Analysis},
  edition   = {2},
  series    = {Graduate Texts in Mathematics},
  publisher = {Springer International Publishing},
  address   = {Cham},
  year      = {2017},
  isbn      = {978-3-319-64628-2},
  doi       = {10.1007/978-3-319-64629-9},
  url       = {https://link.springer.com/book/10.1007/978-3-319-64629-9}
}

@misc{popovici2026mpositivestabilityholomorphicvector,
      title={$m$-Positive Stability of Holomorphic Vector Bundles and Moduli Spaces}, 
      author={Dan Popovici},
      year={2026},
      eprint={2607.17203},
      archivePrefix={arXiv},
      primaryClass={math.DG},
      url={https://arxiv.org/abs/2607.17203}, 
}

@book{wells1980differential,
  title={Differential Analysis on Complex Manifolds},
  author={Wells, R. O.},
  series={Graduate Texts in Mathematics},
  edition={2},
  year={1980},
  publisher={Springer New York},
  doi={10.1007/978-1-4757-3946-6},
  isbn={978-1-4757-3946-6},
  url={https://link.springer.com/book/10.1007/978-1-4757-3946-6}
}

@inproceedings{Griffiths1978PrinciplesOA,
  title={Principles of Algebraic Geometry},
  author={Phillip A. Griffiths and Joe W. Harris},
  year={1978},
  url={https://api.semanticscholar.org/CorpusID:118963833}
}

@article{Michel2019AboutSmallEigenvalues,
  author    = {Laurent Michel},
  title     = {About small eigenvalues of the Witten Laplacian},
  journal   = {Pure and Applied Analysis},
  volume    = {1},
  number    = {2},
  pages     = {149--204},
  year      = {2019},
  publisher = {Mathematical Sciences Publishers},
  doi       = {10.2140/paa.2019.1.149},
  url = {https://msp.org/paa/2019/1-2/paa-v1-n2-p01-p.pdf}
}

@article{158d02b66ce6495492b1b30126556c95,
title = "Supersymmetry and Morse theory",
author = "Edward Witten",
year = "1982",
month = dec,
doi = "10.4310/jdg/1214437492",
language = "English (US)",
volume = "17",
pages = "661--692",
journal = "Journal of Differential Geometry",
issn = "0022-040X",
publisher = "International Press, Inc.",
number = "4",
}

@misc{petrov2024essencerhamcohomology,
      title={The Essence of de Rham Cohomology}, 
      author={Alice Petrov},
      year={2024},
      eprint={2411.06296},
      archivePrefix={arXiv},
      primaryClass={math.AT},
      url={https://arxiv.org/abs/2411.06296}, 
}

@misc{anghel2026akszdescentmanifoldsordinary,
      title={AKSZ Descent on Manifolds with Ordinary Corners}, 
      author={Cristian Anghel},
      year={2026},
      eprint={2608.02928},
      archivePrefix={arXiv},
      primaryClass={math.SG},
      url={https://arxiv.org/abs/2608.02928}, 
}

@article{Li2020,
  author = {Li, Wuchen and Montúfar, Guido},
  title = {Ricci curvature for parametric statistics via optimal transport},
  journal = {Information Geometry},
  volume = {3},
  number = {1},
  pages = {89--117},
  year = {2020},
  doi = {10.1007/s41884-020-00026-2},
  url = {https://doi.org/10.1007/s41884-020-00026-2},
  issn = {2511-249X}
}

@article{yau1977calabi,
  author  = {Yau, Shing-Tung},
  title   = {Calabi's conjecture and some new results in algebraic geometry},
  journal = {Proceedings of the National Academy of Sciences},
  volume  = {74},
  number  = {5},
  pages   = {1798--1799},
  year    = {1977},
  doi     = {10.1073/pnas.74.5.1798},
  url     = {https://doi.org/10.1073/pnas.74.5.1798}
}

@misc{bennequin2020extrafinesheavesinteractiondecompositions,
      title={Extra-fine sheaves and interaction decompositions}, 
      author={Daniel Bennequin and Olivier Peltre and Grégoire Sergeant-Perthuis and Juan Pablo Vigneaux},
      year={2020},
      eprint={2009.12646},
      archivePrefix={arXiv},
      primaryClass={math.AT},
      url={https://arxiv.org/abs/2009.12646}, 
}

@unpublished{demailly_complex_analytic,
  author = {Demailly, Jean-Pierre},
  title  = {Complex Analytic and Differential Geometry},
  year   = {2012},
  note   = {Manuscript, Université de Grenoble},
  url    = {https://people.math.harvard.edu/~demarco/Math274/Demailly_ComplexAnalyticDiffGeom.pdf}
}

@misc{meng2025combinatorialcourantfischerweylminimaxprinciple,
      title={Combinatorial Courant-Fischer-Weyl Minimax Principle on Cheeger $k$-constants of Weighted Forests}, 
      author={Zijun Meng and Dong Zhang},
      year={2025},
      eprint={2510.06301},
      archivePrefix={arXiv},
      primaryClass={math.CO},
      url={https://arxiv.org/abs/2510.06301}, 
}

@misc{zein2014sheafcohomologyalgebraicrham,
      title={From Sheaf Cohomology to the Algebraic de Rham Theorem}, 
      author={Fouad El Zein and Loring W. Tu},
      year={2014},
      eprint={1302.5834},
      archivePrefix={arXiv},
      primaryClass={math.AG},
      url={https://arxiv.org/abs/1302.5834}, 
}

@article{Deligne1975,
author = {Deligne P., Griffiths P. and Morgan, J.},
journal = {Inventiones mathematicae},
pages = {245-274},
title = {Real Homotopy Theory of Kähler Manifolds.},
url = {http://eudml.org/doc/142341},
volume = {29},
year = {1975},
}

@article{L_ger_2023,
   title={A geometric Laplace method},
   volume={5},
   ISSN={2578-5893},
   url={http://dx.doi.org/10.2140/paa.2023.5.1041},
   DOI={10.2140/paa.2023.5.1041},
   number={4},
   journal={Pure and Applied Analysis},
   publisher={Mathematical Sciences Publishers},
   author={Léger, Flavien and Vialard, François-Xavier},
   year={2023},
   month=Dec, pages={1041–1080} }

@article{SoupletZhang2006,
  author  = {Souplet, Philippe and Zhang, Qi S.},
  title   = {Sharp Gradient Estimate and {Yau}'s Liouville Theorem for the Heat Equation on Noncompact Manifolds},
  journal = {Bulletin of the London Mathematical Society},
  volume  = {38},
  number  = {6},
  pages   = {1045--1053},
  year    = {2006},
  doi     = {10.1112/S0024609306018947}
}

@article{Yau1975Harmonic,
  author    = {Shing-Tung Yau},
  title     = {Harmonic functions on complete Riemannian manifolds},
  journal   = {Communications on Pure and Applied Mathematics},
  volume    = {28},
  number    = {2},
  pages     = {201--228},
  year      = {1975},
  publisher = {Wiley Online Library},
  doi       = {10.1002/cpa.3160280203},
  url = {https://onlinelibrary.wiley.com/doi/abs/10.1002/cpa.3160280203}
}

@misc{freire2024yau,
  author       = {Freire, Alexandre},
  title        = {Yau’s theorem on positive harmonic functions in nonnegative Ricci
curvature},
  howpublished = {Lecture notes, University of Tennessee, Knoxville},
  url          = {https://web.math.utk.edu/~afreire/teaching/m661f24/Yau_theorem_positive_harmonic_nonnegative_Ricci.pdf}
}

\newpage

\section{Main notations}
\begin{table}[H]
\centering
\renewcommand{\arraystretch}{1.0}
\small
\begin{tabular}{@{} l p{11cm} @{}}
\toprule
\textbf{Symbol} & \textbf{Description} \\ \midrule
\rowcolor{NavajoWhite!15} $\Phi$ & Kähler potential \\
\addlinespace[6pt] \rowcolor{ProcessBlue!15} $h$ & Kähler information metric \\
\addlinespace[6pt] \rowcolor{NavajoWhite!15} $H$ & Metric with respect to background metric \\
\addlinespace[6pt] \rowcolor{ProcessBlue!15} $\text{Ric}_{i \overline{j}}$ & Complex Ricci curvature \\
\addlinespace[6pt] \rowcolor{NavajoWhite!15} $\Omega$ & Nowhere-vanishing holomorphic form / test form \\
\addlinespace[6pt] \rowcolor{ProcessBlue!15} $\partial, \overline{\partial}$ & Dolbeault operators \\
\addlinespace[6pt] \rowcolor{NavajoWhite!15} $\mathcal{H}$ & Dolbeault Hessian ($i\partial\overline{\partial}f$) \\
\addlinespace[6pt] \rowcolor{ProcessBlue!15} $z$ & Neural network input \\
\addlinespace[6pt] \rowcolor{NavajoWhite!15} $f$ & Neural network output \\
\addlinespace[6pt] \rowcolor{ProcessBlue!15} $\frac{\omega^K}{K!}$ & Volume form \\
\addlinespace[6pt] \rowcolor{NavajoWhite!15} $\theta$ & Complex neural network parameters \\
\addlinespace[6pt] \rowcolor{ProcessBlue!15} $\mathcal{L}$ & Loss function \\
\addlinespace[6pt] \rowcolor{NavajoWhite!15} $m$ & Network width \\
\addlinespace[6pt] \rowcolor{ProcessBlue!15} $K$ & Parameter dimension \\
\addlinespace[6pt] \rowcolor{NavajoWhite!15} $\Delta_{\overline{\partial}}$ & Dolbeault Laplacian \\
\addlinespace[6pt] \rowcolor{ProcessBlue!15} $\Theta_h$ & Chern curvature form \\
\addlinespace[6pt] \rowcolor{NavajoWhite!15} $\nabla_{\omega}^{1,1}$ & $(1,1)$-Hessian \\
\addlinespace[6pt] \rowcolor{ProcessBlue!15} $\nabla_{\omega}^{2,0}$ & $(2,0)$-Hessian \\
\addlinespace[6pt] \rowcolor{NavajoWhite!15} $\nabla_{\omega}^{0,2}$ & $(0,2)$-Hessian \\
\addlinespace[6pt] \rowcolor{ProcessBlue!15} $\| \cdot \|_2$ & Spectral norm \\
\addlinespace[6pt] \rowcolor{NavajoWhite!15} $\nabla_{k}$ & Covariant derivative with respect to $k$ \\
\addlinespace[6pt] \rowcolor{ProcessBlue!15} $\lambda$ ($\lambda(K)$) & Loss regularization coefficient (often a function of $K$) \\
\addlinespace[6pt] \rowcolor{NavajoWhite!15} $\mu$ & Upper bound on the first $\widetilde{K}$ eigenvalues of $H$ \\
\addlinespace[6pt] \rowcolor{ProcessBlue!15} $\kappa$ & Determinant constant of $H$ \\
\bottomrule
\end{tabular}
\end{table}

\section{Additional figures}

\begin{figure}[ht]
  \centering
  \includegraphics[width=0.7\textwidth]{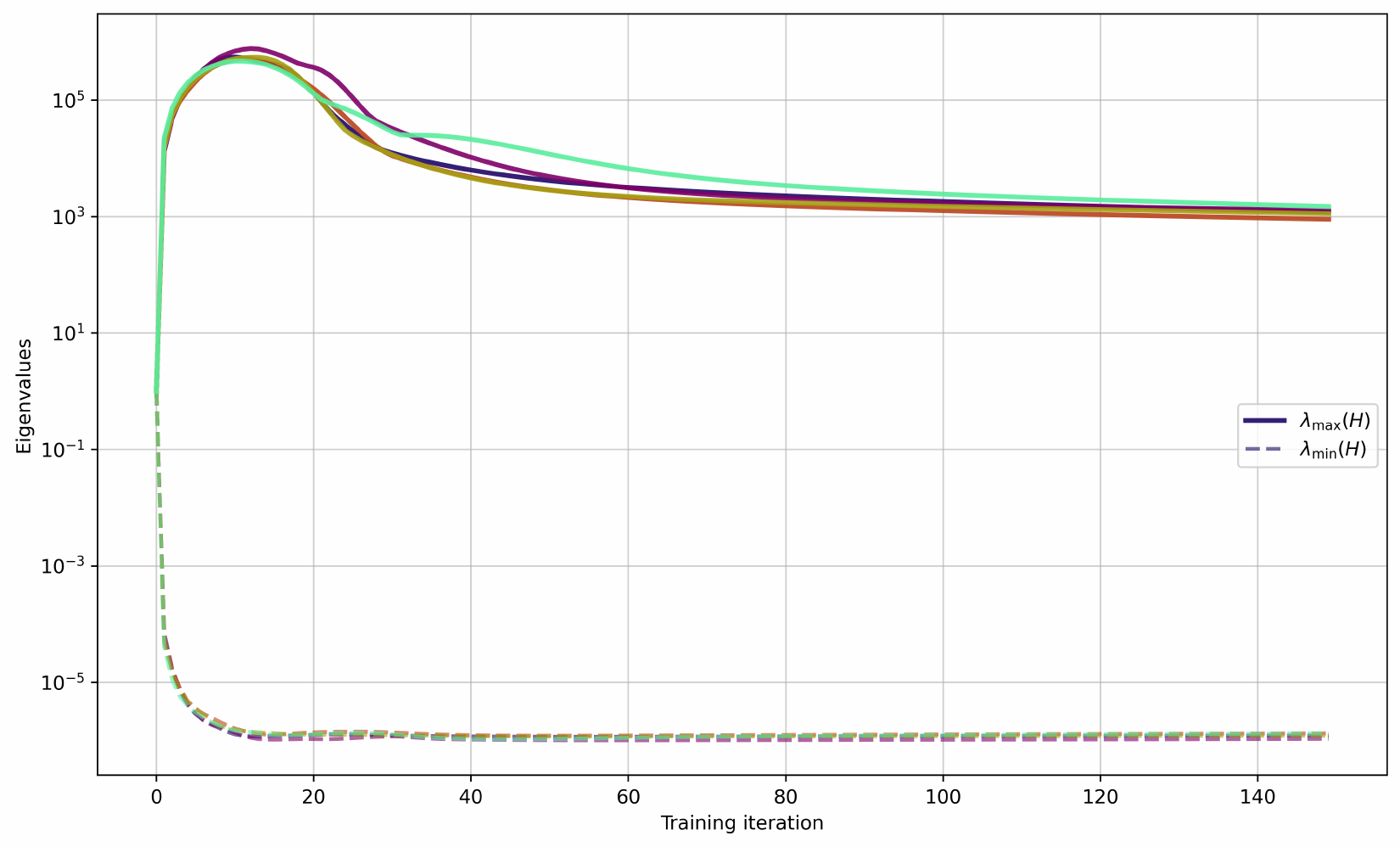}
  \caption{We plot a softer eigenvalue blow-up on $H$ that holds reasonably well in an empirical setting based on the Kähler condition of \ref{eqn:eu_lower_bound} across five distinct training trajectories. We crucially remark we indeed use a Kähler cross-entropy metric for this figure. This figure is consistent with \ref{fig:kahler_blowup_severe}. We set background metric $h_0$ to $h_{t=0}$ and fix it across training, which is a canonical choice and not Ricci-flat. As a remark, in order to apply Yau's Liouville theorem, we need both $h_0,h$ Ricci-flat, and here the background is not Ricci-flat.}
  \label{fig:kahler_blowup}
\end{figure}

\begin{figure}[ht]
  \centering
  \includegraphics[width=0.7\textwidth]{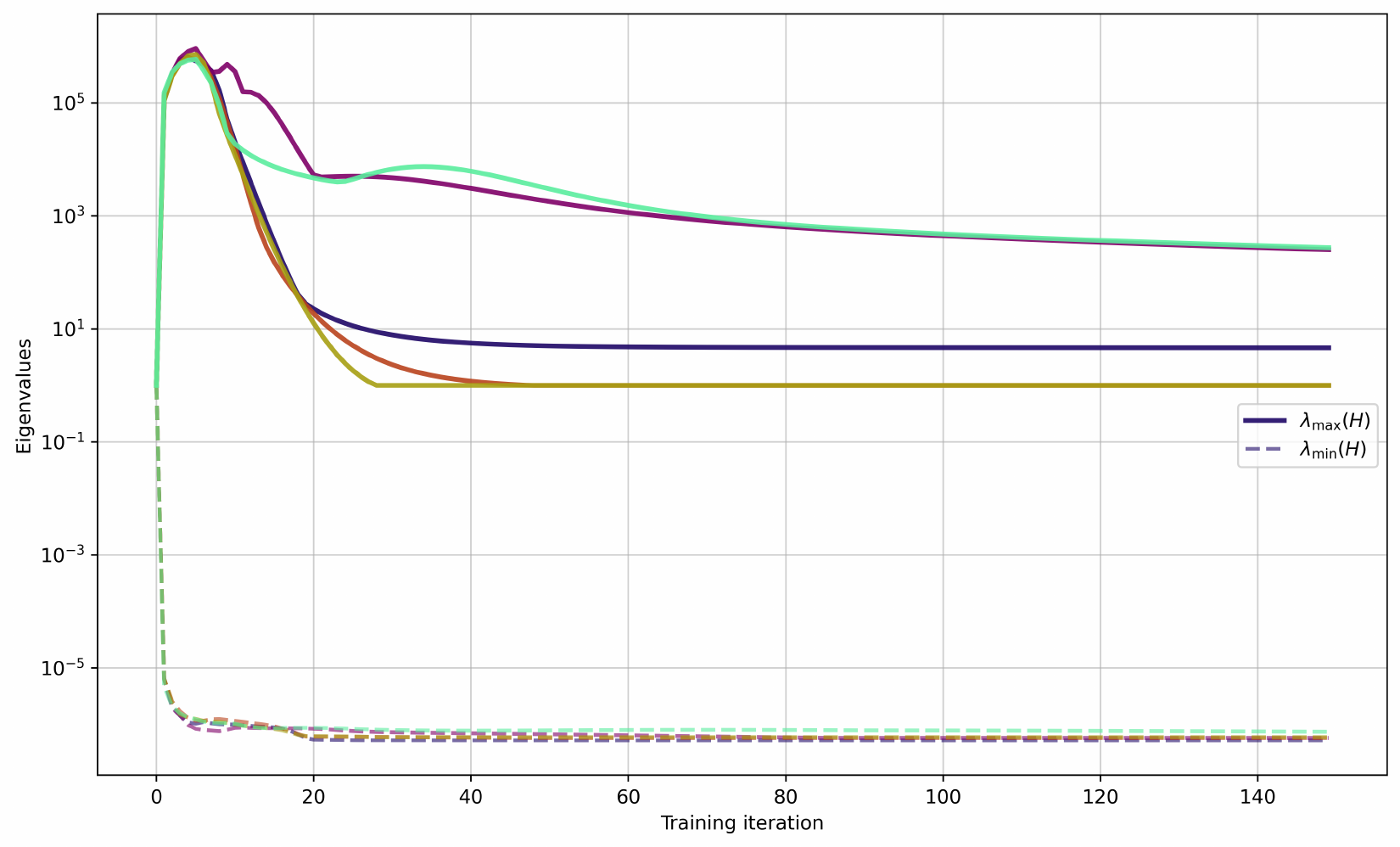}
  \caption{We plot the same figure as \ref{fig:kahler_blowup} but we increased the learning rate from $\gamma=0.05$ to $\gamma=0.15$ (the minimum eigenvalue plateaus above this because the plot is with respect to $H$ not $h$).}
  \label{fig:kahler_blowup_greater}
\end{figure}

\begin{figure}[ht]
  \centering
  \includegraphics[width=0.7\textwidth]{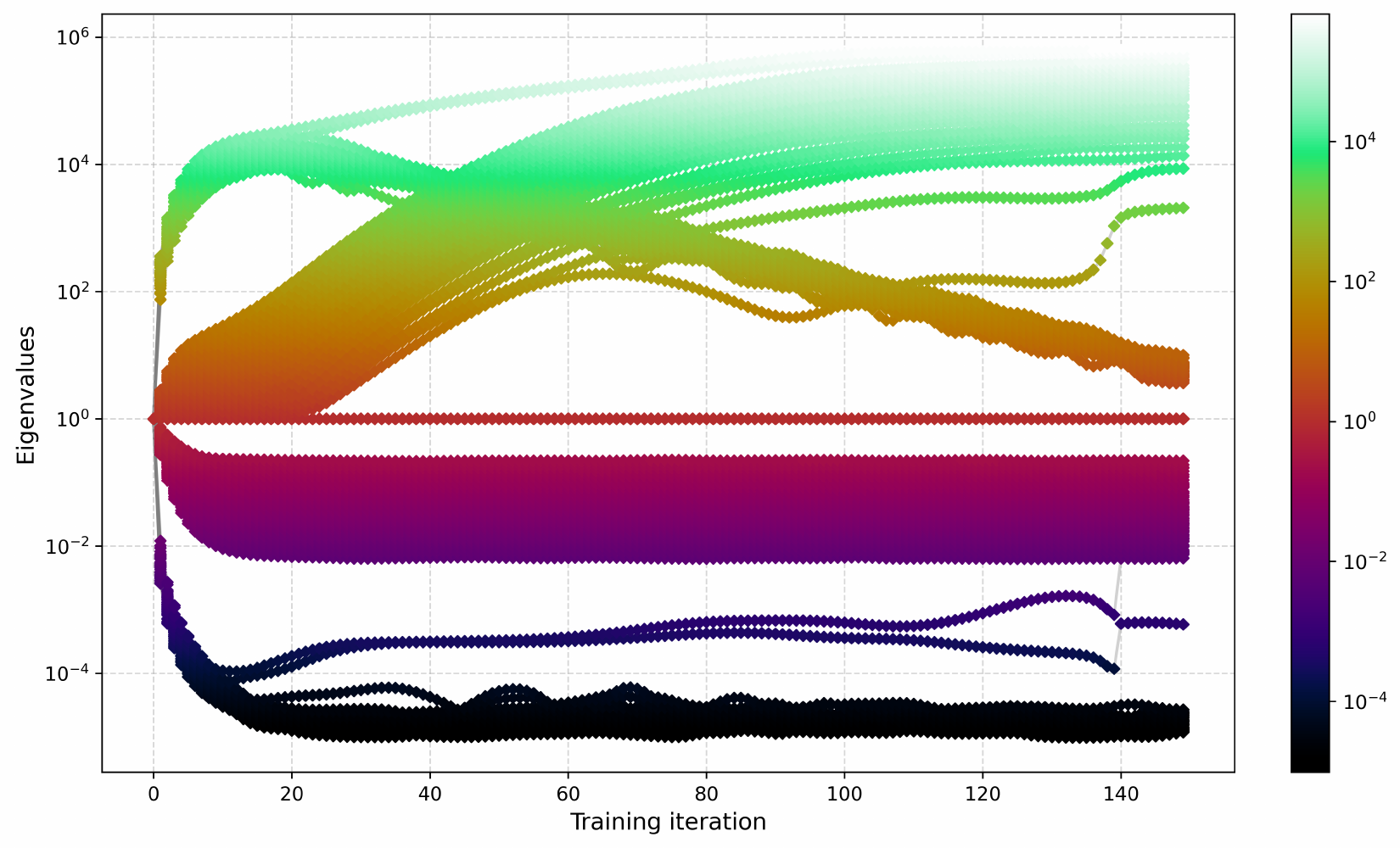}
  \caption{We plot a figure of all eigenvalues of $H$ across one instance of training, similar to \ref{fig:kahler_all_eigenvalues}, and the scenario almost corresponds to that of Figure \ref{fig:kahler_blowup}. Here, we have slightly modified hyperparameters to appear visually distinct from \ref{fig:kahler_all_eigenvalues}. In particular, we have doubled width, lowered the learning rate, added weight decay, and changed initialization spread.}
  \label{fig:kahler_all_eigenvalues_appendix}
\end{figure}

\begin{figure}[ht]
  \centering
  \includegraphics[width=0.55\textwidth]{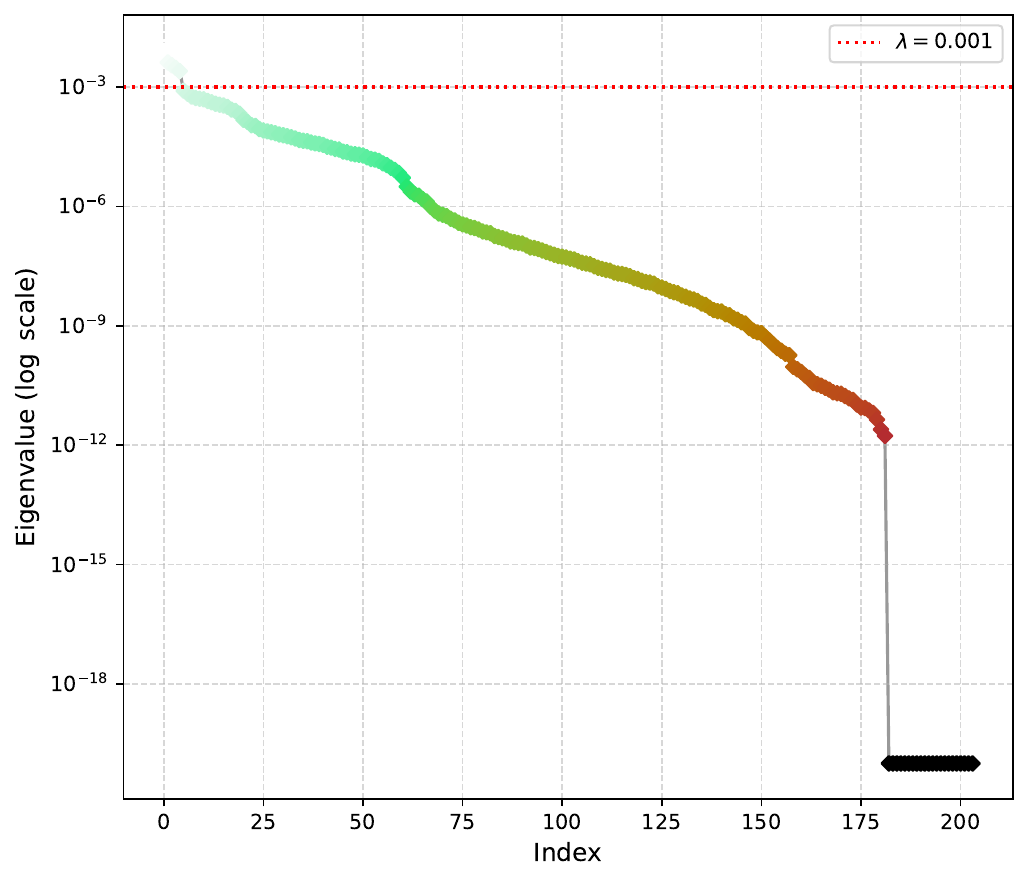}
  \caption{We plot the top all 204 eigenvalues corresponding to a regularized metric with the loss of \ref{eqn:regularized_loss} in a real scenario with Tikhonov regularization with $\lambda=1\mathrm{e}{-3}$. We can note the metric is actually low rank here, although it is close to the "almost low rank" quality as mentioned in \ref{sec:calabi_yau}.}
  \label{fig:eigenspectrum}
\end{figure}

\begin{figure}[ht]
  \centering
  \includegraphics[width=0.55\textwidth]{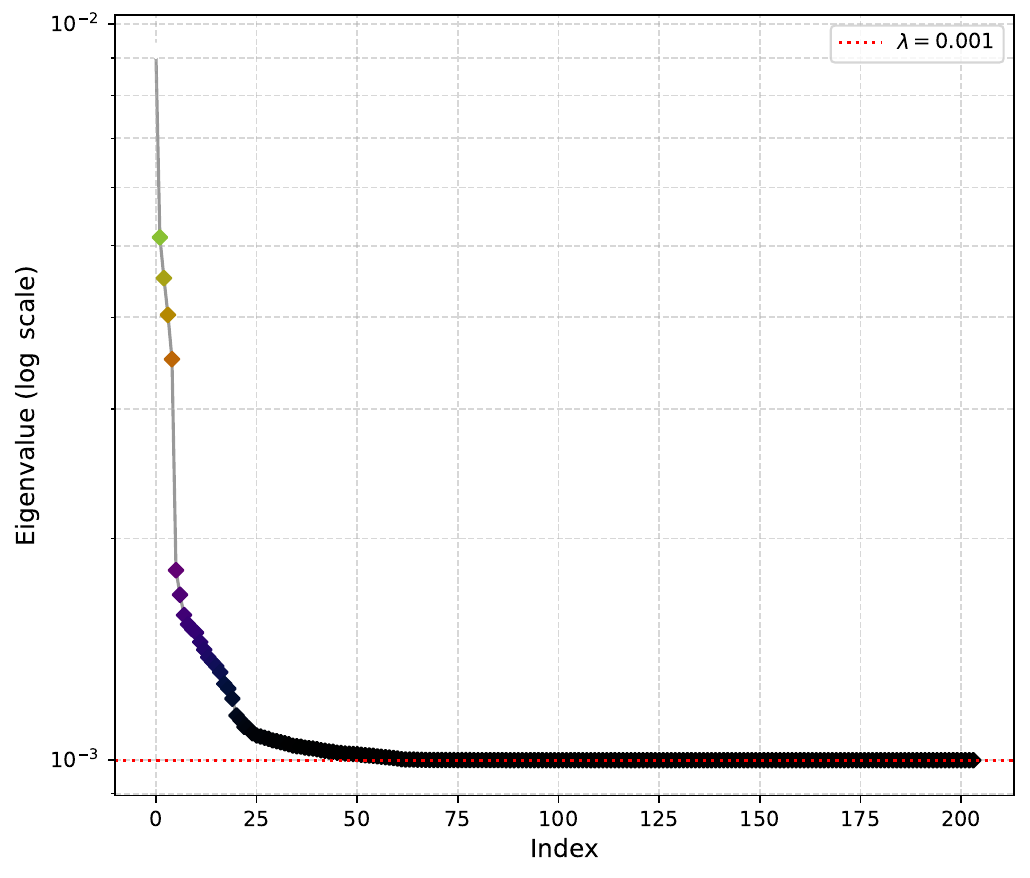}
  \caption{We plot the same as \ref{fig:eigenspectrum} but we use the strategy of Jacobian augmentation here. Notice the metric is full rank here, and simultaneously "almost low rank."}
  \label{fig:eigenspectrum_jacobian}
\end{figure}

We refer to additional figures \ref{fig:kahler_blowup}, \ref{fig:kahler_blowup_greater}, \ref{fig:kahler_all_eigenvalues_appendix}, \ref{fig:eigenspectrum}, \ref{fig:eigenspectrum_jacobian}, which are empirically reasonable.

\section{Contributions}
\label{sec:contributions}

Due to lack of space in section \ref{sec:introduction}, we outline our contributions here.
\begin{enumerate}[label=\Roman*., nosep]
\item First, we provide additional discussion relevant to section \ref{sec:calabi_yau}. Let us provide the following theorem:

\vspace{2mm}

\noindent \textbf{Theorem (Yau)}. Let $(M,g)$ be a connected, complete Riemannian manifold without boundary satisfying
\begin{align}
\operatorname{Ric}(g)\succeq 0.
\end{align}
Then every positive harmonic function $f:M\to\mathbb R$ is constant
\begin{align}
\Delta_g f=0,\qquad f>0 \quad\Longrightarrow\quad f\equiv C.
\end{align}

\vspace{2mm}

\noindent This is based in \cite{Yau1975Harmonic}, and is found essentially verbatim in \cite{freire2024yau}. To apply this to our Calabi-Yau setting, we take $\operatorname{Ric}(h)=\operatorname{Ric}(h_0)=0$. From the relative Ricci identity,
$\partial\overline\partial u=0, u=\log\det(h_0^{-1}h)$.
Therefore, $u$ is pluriharmonic. In particular,
$\Delta_{h_0}u = 2h_0^{i\overline j}\partial_i\partial_{\overline j}u =0$. Now, we assume $(M,h_0)$ is complete and connected without boundary, and that $u$ is globally bounded below. Let us define $f=u-\inf_Mu+1$. Then $f>0, \Delta_{h_0}f=0$. Since $\operatorname{Ric}(h_0)=0$, Yau's theorem applies, so $f\equiv C$. Consequently,
\begin{align}
\label{eqn:u_const}
u\equiv\log\kappa \quad\Longrightarrow\quad \det(H)=e^u=\kappa>0.
\end{align}
Note that this result also applies to Kähler manifolds since a Kähler manifold is a specific type of Riemannian manifold. We will use \ref{eqn:u_const} in many of our Calabi-Yau results (see section \ref{sec:calabi_yau} for lengthier discussion).

\item \textbf{Dolbeault Hessian scaling.} In Appendix \ref{app:dolb_hessian_bounds}, we establish the result the (1,1)-Hessian scales as
\begin{align}
\text{(1,1)-Hessian} \sim \mathcal{O} \left( \frac{1}{\sqrt{m}} \right) .
\end{align}
This is arguably our most valuable result, not because it is particularly meaningful standalone, but it complements many of our alternative results, including the convexity arguments of Appendix \ref{app:convexity_results}. Our proof relies on an application of the Mean Value Theorem (MVT) around a ball on the manifold. Generally, the MVT requires a sufficiently nice domain so that an intermediary point does not "lift off" the domain, i.e. convexity of the domain. While a large ball can fail, a sufficiently small ball is fine for our purposes, so we will allow the MVT to hold (we refer to Whitehead's theorem on convex neighborhoods). It is crucial to remark we require a nuclear norm inequality to hold without picking up a factor of dimension, therefore regularization techniques compatible with \ref{sec:theory} are mostly vital. In this result, we also rely on the supplemental results of Appendices \ref{app:dolb_cov_bound}, \ref{app:yau_estimate}. We also remark we use the norm
\begin{align}
\| i \partial \overline{\partial} f \|_2 = \sup_{v \neq 0} \frac{\left| \sum_{j,k=1}^K \frac{\partial^2 f}{\partial \theta^j \partial \overline{\theta}^k} v^j \overline{v}^k \right|}{\sum_{j,k=1}^K h_{j\overline{k}} v^j \overline{v}^k}  = \max_{1 \leq m \leq K} |\lambda_m|
.
\end{align}
where $v \in T_\theta^{(1,0)} M$ is a non-zero tangent vector, and the eigenvalues $\lambda_m$ are the solutions to the characteristic equation
\begin{align}
\det \left( \frac{\partial^2 f}{\partial \theta^j \partial \overline{\theta}^k} - \lambda h_{j\overline{k}} \right) = 0 .
\end{align}
\item \textbf{Covariant Hessian bounds.} In Appendix \ref{app:dolb_cov_bound}, we establish the result of a bound on the covariant of the (1,1)-Hessian, so $\| \nabla_{\omega} \mathcal{H} \|_2 \leq \text{upper bound}$, which is non-essential result and is mostly used to complement the result in Appendix \ref{app:dolb_hessian_bounds}. This proof is one of our most information-theoretic results.
\item \textbf{Bounded oscillation.} In Appendix \ref{app:yau_estimate}, we establish a PDE-type result that allows us to achieve an oscillatory result on the quantity we call $\Psi := \log \det ( I + \omega^{-1} i \partial \overline{\partial} f )$, which is important for our $\mathcal{O}\left( \frac{1}{\sqrt{m}} \right)$ result.
\item \textbf{A Kato inequality result.} In Appendix \ref{app:2,0_hess_bounds}, we prove a (2,0)-Hessian scaling law, and the (0,2)-law follows similarly in a corollary via complex conjugate arguments and via Spectral theorem. The main lemma follows by the Kato inequality \cite{Herzlich2000}
\begin{align}
\label{eqn:kato}
\left| \frac{d}{dt} \| \mathcal{H}^{2,0}(\gamma(t)) \|_2 \right|
&\leq \| \nabla_{\dot{\gamma}(t)} \mathcal{H}^{2,0} \|_2
\end{align}
and via standard ordinary differential equation techniques by dropping the absolute value on the left-hand side of \ref{eqn:kato}.
\item \textbf{Morphisms, sheaves, Jacobians, and eigenvalues.} In Appendix \ref{app:cohomology}, we discuss how the Jacobian of the network (with a Dolbeault, $\partial f$) is a surjective morphism of sheaves (but not injective since it maps from high to low dimensional spaces). Moreover, the metric under a closed-form log-likelihood can be represented first-order instead of second-order. Now, $df$ is a smooth bundle map, mapping the entire real tangent bundle $T_{\mathbb{R}}M$ to the trivial vector bundle $M \times \mathbb{R}^{Nk}$, also a map from high-to-low dimension. In particular, the connection between this smooth bundle map and $\partial f$ is given by
\begin{align}
\label{eqn:df}
df = \partial f + \overline{\partial} f ,
\end{align}
so they are related high-to-low maps. $\partial f$ is a morphism of sheaves, and so is $df$ via $\mathcal{E}_M(T_{\mathbb{R}}M) \ \xrightarrow{\ df\ } \ \mathcal{E}_M(\underline{\mathbb{R}}^{Nk})$, via definitions
\begin{align}
& \mathcal{E}_M(U) = \{ f: U \to \mathbb{R}^{Nk} \mid f \text{ is smooth} \} .
\end{align}
We can note $\mathcal{E}_M(\underline{\mathbb{R}}^{Nk}) \longrightarrow 0$ since the sequence is exact. Also, as in sheaf cohomology, $H^0(M; \mathcal{F})$ represents the global sections of the sheaf $\mathcal{F}$ over the entire manifold $M$. The 0-th order chain follows similarly from the exact sequence with $df$, and higher orders vanish.

\vspace{2mm}

\noindent Subsequently, we discuss eigenvalue gaps. Previously, in section \ref{sec:corrupted_geometries}, we discussed regularized eigenvalues which provide a lower bound on eigenvalues of the Kähler metric. We also discuss an upper bound for these minimal eigenvalues in terms of the norm of the regularization. In particular, regularization provides both a lower and upper bound on the eigenvalues via $h \succeq \lambda \delta$ and $\lambda_{\text{min}} \leq \lambda \| \delta \|_2$, where $\lambda$ is the regularization tuning coefficient. If $\delta$ is the identity times a coefficient, then since the spectral norm of the identity is the value of each entry, we get an eigenvalue bound
\begin{align}
\lambda \leq \lambda_{\text{min}} \leq \lambda ,
\end{align}
which is totally possible.
\item \textbf{Deep learning theory: asymptotics as initialization.} In Appendix \ref{app:initialization}, we provide a scaling law $\mathcal{O}(\frac{1}{\sqrt{m}})$ exactly at initialization. We consider (1) forward passes; (2) backward passes with error signals; and (3) and via a unit-norm perturbation analysis. This is our main deep learning theoretic result.
\item \textbf{Convexity result.} In Appendix \ref{app:convexity_results}, we provide convexity results, similarly to a restricted strong convexity analysis as in \cite{banerjee2023restricted}, although our analysis is more geometric in argument rather than deep learning theoretic. Our convexity results for both a lower bound and upper bound, or via $\beta$-smoothness, rely on bound
\begin{align}
L(\theta_0) = L(\theta_t) + 2\text{Re}\langle \nabla_\omega L(\theta_t), \exp_{\theta_t}^{-1}(\theta_0) \rangle_h + \frac{1}{2} \nabla^2_\omega L(\widetilde{\theta}_t)(v, v) .
\end{align}
The lower bound relies on the term
\begin{align}
\Gamma\left(a, \nu, \rho, C_{\mathcal{H}}, \{\ell_i'(f_i(\widetilde{\theta}_t))\}_i, m, v\right)  := 2 \left( a \nu  - \frac{ 2\sqrt{2}a \sqrt{\rho} C_{\mathcal{H}} \|v\|_h +  C_{\mathcal{H}} \sqrt{\frac{1}{n} \sum_{i=1}^n (\ell'_i(f_i(\widetilde{\theta}_t)))^2}}{\sqrt{m}}   \right) ,
\end{align}
which almost vanishes in the wide-width limit $m \rightarrow \infty$ except for the constant $a \nu$ term, and the upper bound term
\begin{align}
\rho_J + \frac{C_{\mathcal{H}} \sqrt{ \frac{1}{n} \sum_{i=1}^n (\ell'_i(f_i(\widetilde{\theta}_t)))^2 } }{\sqrt{m}} ,
\end{align}
which also almost vanishes in the limit except for the $\rho_J$ term. Via these results, we can establish a Kähler Polyak-Łojasiewicz and a loss decent update rule. 

\vspace{2mm}

\noindent In particular, large eigenvalues are mostly adverse regarding $\beta$-smoothness result. $\beta$-smoothness is motivated by upper bounds on the second order Taylor expansion, and Hessian eigenvalue explosion will violate a low bound. This is our overarching idea as to why Calabi-Yau manifolds specifically are catastrophic in an optimization context, since this property is largely specific to Calabi-Yau manifolds. 
\item \textbf{Kähler Polyak-Łojasiewicz condition.} With respect to the above discussion, we establish theoretical results that Calabi-Yau manifolds, under a light blow-up assumption $ \lambda_{\max}(H) = \Omega( \frac{\kappa}{\mu^{\widetilde K-1}} )$, fail to satisfy the Kähler Polyak-Łojasiewicz and the loss decent update rule.
\end{enumerate}

\vspace{2mm}

We proceed onto our more curvature-driven results.
\begin{enumerate}[label=\Roman*., nosep]
\item \textbf{Regret bounds.} In Appendix \ref{app:regret_bounds}, we establish an upper bound on regret. Generally, in a regret analysis, it is of interest to show success of an algorithmic process by establishing an upper bound, therefore we will be showing success of stochastic natural gradient descent. Our proof for this result relies on a Ricci curvature claim following the index lemma. Since we examine a geodesic ball, this is along the manifold, thus depends on local curvature information. In particular, we will be integrating curvature, thus it is not curvature at a single point.
\item \textbf{Dirichlet energy.} We examine a Dirichlet energy result in \ref{app:dirichlet_energy} around a geodesic ball $\mathcal{S}$ on the manifold. We can show at around initialization, the normalized Dirichlet energy follows $\mathcal{O}(K  + \frac{K}{\sqrt{m}})$, which is linear in highest order with respect to the parameter. Dirichlet energy is useful because it contributes to how fast the parameter descends, therefore we find an upper bound on this "descent speed."
\item \textbf{Expected paths and minimum eigenvalues.} In Appendix \ref{app:trajectory_bounds}, we first establish the claim the integrated norm of the parameter path differentiated obeys $\int_0^t \|\dot{\theta}_s\|_h ds = \mathcal{O}(\frac{1}{\sqrt{m}})$. We assume natural gradient descent, so the parameter update is non-stochastic, and we develop: (1) an upper bound on the gradient norm at time $t$ with respect to the starting gradient norm and an exponential term that uses the $\frac{1}{\sqrt{m}}$ scaling; we establish a similar upper bound in terms of the minimum eigenvalue, which is closely related to a stability result.
\item \textbf{Flux of the gradient.} In Appendix \ref{app:gradient_flux_searches}, we establish a value that quantities "how well we can escape close to initialization" via the flux
\begin{align}
\int_0^\epsilon \left( \oint_{\partial B_r(\theta_0)} d^c \mathcal{L} \wedge \omega^{K-1} \right) dr ,
\end{align}
and generally a higher negative value is desirable. In particular, the Calabi-Yau case is a failure mode since a curvature term vanishes here. In fact, positive curvature leads to greater escape here, so this result is consistent with the abstract of this work.
\item \textbf{Parameter variance.} In Appendix \ref{app:variance}, we establish a result that tells us the variance of the traversed parameter under a stochastic natural gradient descent update rule (which is reasonable in a true empirical setting). We develop the variance in terms of Ricci curvature at the local neighborhood, or target parameter. This result favors positive curvature, and negative Ricci curvature at the (target) parameter is less desirable. This result relies on an analysis of the matrix system based on 
\begin{align}
A + B - \overline{C}(\overline{A} + \overline{B})^{-1} C  .
\end{align}
\item \textbf{Witten Laplacian eigenvalue bounds.} In Appendix \ref{app:min_eigenvalue_bounds}, we examine the spectral gap via the Witten Laplacian, which establishes a "speed limit" for convergence. The choice of the Witten Laplacian is meaningful and say not the Hessian of the loss because this only measures the curvature a specific point, and the eigenvalues can be negative or positive. The Witten Laplacian consists of wide-neighborhood to even the point of considering global information, and moreover its eigenvalues are nonnegative. As we see in \ref{eqn:modified_Bochner}, the modified Bochner equation gives us the upper bound, and it is defined via integration, hence global information.
\item \textbf{Chern curvature bounds.} In Appendix \ref{app:some_bad_curvature_and_primitive}, we examine with respect to a positive $(p,p)$-form $\Omega$ such that
\begin{align}
\Bigg\{(i\Theta_h(T^{1,0}M) \wedge \omega^{q-1} \wedge \Omega) V, V \Bigg\}_h \geq 0 .
\end{align}
This is a semi-positivity condition that acts similary to a "restoring force." We have (1) $i\Theta_h(T^{1,0}M)$ as a curvature operator; (2) $\omega^{q-1}$ as a background Kähler scaling; (3) and $\Omega$, a geometric weight. We add $(1,1) + (q-1, q-1) + (p,p)$ forms, so this combined operator is a large differential form that eats the test field $V$, applies the curvature transformation, scales it by the metric, and weighs it by $\Omega$. It guarantees that the geometry of the parameter space does not have any runaway or hyperbolic directions under the weight of $\Omega$. Every direction $V$ yields that $\geq 0$ restoring force.
\end{enumerate}

\section{Additional background}
\label{app:background}

Let us provide some background on background metrics.

\vspace{2mm}

Let $\omega_0$ be a background form corresponding to a background metric $h_0$, which are both representations of the same geometric object: $h_0$ is a Hermitian metric on the holomorphic tangent bundle $T^{1,0}M$. In local holomorphic coordinates $\theta^1, \dots, \theta^K$, $h_0$ is represented by a positive-definite, Hermitian matrix function $(h_{0, j\overline{k}})$, meaning $h_{0, j\overline{k}} = \overline{h_{0, k\overline{j}}}$. As a tensor, it is written as
\begin{align}
h_0 = \sum_{j,k=1}^K h_{0, j\overline{k}} d\theta^j \otimes d\overline{\theta}^k .
\end{align}
Correspondingly, the differential form is a smooth section of the cotangent bundle, written as
\begin{align}
\omega_0 = i \sum_{j,k=1}^K h_{0, j\overline{k}} d\theta^j \wedge d\overline{\theta}^k  .
\end{align}
Both of these definitions correspond to a background geometric object. By defining $\omega_0$ as a closed, strictly positive $(1,1)$-form, we establish a well-defined de Rham cohomology class $[\omega_0] \in H^2(M; \mathbb{R})$. When we consider $\text{Ric}(\omega) = 0$, we can find a further Calabi-Yau metric given the cohomology condition
\begin{align}
[\omega] = [\omega_0] ,
\end{align}
The $\partial \overline{\partial}$-lemma then dictates that any other Kähler metric $\omega$ in this exact same class must take the form $\omega = \omega_0 + i\partial\overline{\partial}\varphi$.

\vspace{2mm}

With respect to the Calabi conjecture, we seek a metric $h$ in the same Kähler class as a starting metric $h_0$, i.e. via
\begin{align}
h_{i\bar{j}} = (h_0)_{i\bar{j}} + \partial_i \bar{\partial}_j \varphi .
\end{align}
In our contexts, this does not apply. Our metric is defined via the neural network losses/descents. A Kähler class is largely a topological/cohomological concept, and recall from \ref{sec:calabi_yau} our arguments exist via geometry. In this work, we have some freedom in choosing a background metric.

\vspace{2mm}

The Calabi conjecture is intended for compact Kähler manifolds. Therefore, the above only loosely applies to our contexts, since our manifolds are more unrestricted and have greater association to non-compact scenarios (see section \ref{sec:calabi_yau}). Even in the non-compact case, we can write
\begin{align}
\omega = \omega_0 + i\partial\overline{\partial}\varphi  .
\end{align}
$\varphi$ is a valid potential function that typically exhibits some sort of decay to maintain sufficiently nice properties. Therefore, our Kähler metric $h$ is associated to the (1,1)-form $\omega$. Since our metric is defined globally (which is valid in a non-compact case), we can write
\begin{align}
\omega = i \sum_{j,k=1}^K \left( \partial_j \partial_{\overline{k}} \Phi \right) d\theta^j \wedge d\overline{\theta}^k .
\end{align}
The relationship on potentials is given by $\Phi = \Phi_0 + \varphi$. Therefore, our true Kähler metric, which we take in this work to be
\begin{align}
\partial_{\theta^i} \partial_{\theta^{\overline{j}}} \int_{\mathcal{Z}} \int_{\mathcal{Y}} -\log p(y | z,\theta) q(y | z) p_{\text{data}}(z)  dy  \frac{\omega_{\mathcal{Z}}^d}{d!}  ,
\end{align}
can be, roughly speaking, decomposed into two potentials.

\section{Second derivative results}

\subsection{Complex Hessian background}

In this section, we discuss our strategies for connecting the types of complex Hessians.

\vspace{2mm} 

\noindent We first note the complexified cotangent bundle splits $T^* M \otimes \mathbb{C} = T^{1,0 *} M \oplus T^{0,1 *} M$, and the second derivative splits
\begin{align}
\nabla_{\omega}^2 f = \nabla^{2,0}_{\omega}f + \nabla^{1,1}_{\omega}f + \nabla^{0,2}_{\omega}f .
\end{align}
In local holomorphic coordinates, the holomorphic tangent bundle itself decomposes
\begin{align}
T^{1,0}M = \bigoplus_{i=1}^K \mathbb{C} \frac{\partial}{\partial \theta_i} .
\end{align}

\vspace{2mm}

\noindent We will attempt to bound various versions of the complex Hessian. For example, we will attempt to bound the section of $T^* M \otimes T^* M$ in local holomorphic coordinates
\begin{align}
\nabla^{1,1}_{\omega}f = \frac{\partial^2 f }{ \partial \theta^i \partial \overline{\theta}^j } d\theta^i \otimes d\overline{\theta}^j .
\end{align}
This is most closely related to the (1,1)-form
\begin{align}
i \partial \overline{\partial} f = i \sum_{ij} \frac{\partial^2 f }{ \partial \theta^i \partial \overline{\theta}^j } d\theta^i \wedge d\overline{\theta}^j .
\end{align} 
The (2,0) part has the closed form 
\begin{align}
\nabla^{2,0}_{\omega}f = \sum_{ij} \left( \frac{\partial^2 f }{ \partial \theta^i \partial \theta^j } - \sum_k {\Gamma_{ij}}^k \frac{\partial f}{\partial \theta^k} \right) d\theta^i \otimes d\theta^j ,
\end{align}
therefore this requires a different bound. We can note the difference in the wedge and tensor products. One approach could be to bound two terms simultaneously, although we find the approach to bound the terms individually more straightforward in practice. If we take the Kähler metric as $\omega = i h_{i\overline{j}} d\theta^i \wedge d\overline{\theta}^j$, we can take the trace (the dual Lefschetz operator $\Lambda$) of the $(1,1)$-form $i \partial \overline{\partial} f = i \frac{\partial^2 f}{\partial \theta^i \partial \overline{\theta}^j} d\theta^i \wedge d\overline{\theta}^j$, giving
\begin{align}
\star \left( i \partial \overline{\partial} f \wedge \frac{\omega^{K-1}}{(K-1)!} \right) = \Lambda(i \partial \overline{\partial} f) = h^{i \overline{j}} \frac{\partial^2 f}{\partial \theta^i \partial \overline{\theta}^j} = \Delta_{\overline{\partial}} f ,
\end{align}
which we will use in \ref{app:dolb_hessian_bounds}.

\vspace{2mm}

\noindent We can note the decomposition in the bilinear form case is
\begin{align}
\frac{1}{2} \nabla^2 L(v,v) = \text{Re}(\nabla^{2,0}_{\omega}L(v,v) ) + \nabla^{1,1}_{\omega}L(v, \overline{v} ) ,
\end{align}
which is the holomorphic part and its conjugate, and the Hermitian part.

\subsection{Dolbeault Hessian bounds}
\label{app:dolb_hessian_bounds}

In this section, we will provide a bound on $i \partial \overline{\partial} f$, which is connected to the (1,1) Hessian $\nabla_{\omega}^{1,1} f$. We will draw a connection between these two terms, which will help us in the following sections.

\vspace{2mm}

\noindent \textit{Proof of Theorem 2.} Let $f(\theta ; z) : M \rightarrow \mathbb{R}$, where $\theta \in M$ is a parameter along the Kähler manifold $M$ with Kähler metric $\omega$. Define the deformed form
\begin{align}
\omega_f = \omega + i \partial \overline{\partial} f ,
\end{align}
where $f$ is not necessarily SPSH but we have $\omega + i \partial \overline{\partial} f > 0$.

\vspace{2mm}

\noindent The spectral norm of the Hessian in \cite{banerjee2023restricted}  is analogous to the maximum eigenvalue of $i \partial \overline{\partial} f$ for us, so we will attempt to bound
\begin{align}
\| i \partial \overline{\partial} f \|_2 ,
\end{align}
where $\| \cdot \|_2$ is the spectral norm. Let us define $\Psi : M \rightarrow \mathbb{R}$ as the logarithmic volume ratio
\begin{align}
\Psi := \log \left( \frac{(\omega + i \partial \overline{\partial} f)^K}{\omega^K} \right) ,
\end{align}
(observe it exists) which immediately implies the complex Monge-Ampère equation
\begin{align}
(\omega + i \partial \overline{\partial} f)^K = e^\Psi \omega^K ,
\end{align}
where $K$ is the integer such that $\Theta \subseteq \mathbb{C}^{\sum_k m_k m_{k+1} + m_L} := \mathbb{C}^K$, and we use notation $\bigwedge_{k=1}^K \omega = \omega^K$. We will canonically choose $\omega$ flat.

\vspace{2mm}

\noindent Our proof strategy for the next step will be to apply the Mean Value Theorem and then bound the gradient of $\Psi$ term. It is nontrivial to bound $\| i \partial \overline{\partial} f \|_2$ directly, but instead we will show a relation to a bound $\| \nabla_{\omega} i \partial \overline{\partial} f \|_2$.  Let us examine the oscillation, or supremum and infimum difference, and examine the geodesic ball diameter. By the Mean Value Theorem\footnote{Here, we assume the MVT is applicable, which it not always is on a manifold. MVT over a more Euclidean domain would need it so the evaluated point lies within the region, for example requiring convexity. The issue along a manifold is not necessarily the ball is too large, but that the region may fail to be geodesically convex. We mention this briefly in section \ref{sec:contributions} via Whitehead's theorem. Without loss of generality, we can work in a sufficiently small ball if needed, which resolves issues, thus we can mostly relax concerns about this geodesic convexity.}
\begin{align}
\sup_{\mathcal{S}} \Psi - \inf_{\mathcal{S}} \Psi \leq \left( \sup_{\theta \in \mathcal{S}} \| \nabla_{\omega} \Psi \|_h \right) \times \text{diam}(\mathcal{S})  ,
\end{align}
where $\mathcal{S}$ is the geodesic ball $\mathcal{S} = B_{\omega}(\theta_0, R) = \{ \theta \in M | d_\omega(\theta_0, \theta) \leq R \}$. $R$ is fixed, so $\text{diam}(\mathcal{S})$ is $\mathcal{O}(1)$. The above norm is the $\omega$-dual norm.

\vspace{2mm}

\noindent Let us examine
\begin{align}
\det(I + \omega^{-1} i \partial \overline{\partial} f) = e^\Psi .
\end{align}
This is true since $(A \omega) \wedge (A \omega) \wedge \dots \wedge (A \omega) = \det(A) (\omega \wedge \omega \wedge \dots \wedge \omega)$, and
\begin{align}
(\omega + i \partial \overline{\partial} f)^K = \det(I + \omega^{-1} i \partial \overline{\partial} f) \omega^K .
\end{align}
In particular, via eigenvalues
\begin{align}
\omega = i \sum_{j=1}^K d\theta^j \wedge d\overline{\theta}^j, \quad i \partial \overline{\partial} f = i \sum_{j=1}^K \lambda_j d\theta^j \wedge d\overline{\theta}^j .
\end{align}
The volume element obeys 
\begin{align}
\omega^K = K!  i^K d\theta^1 \wedge d\overline{\theta}^1 \wedge \dots \wedge d\theta^K \wedge d\overline{\theta}^K ,
\end{align}
so the deformed metric follows
\begin{align}
\omega + i \partial \overline{\partial} f = i \sum_{j=1}^K (1 + \lambda_j) d\theta^j \wedge d\overline{\theta}^j .
\end{align}
Therefore, taking the $K$-th wedge power, the combinatorial properties follow and
\begin{align}
(\omega + i \partial \overline{\partial} f)^K = K! \left( \prod_{j=1}^K (1 + \lambda_j) \right) i^K d\theta^1 \wedge d\overline{\theta}^1 \wedge \dots \wedge d\theta^K \wedge d\overline{\theta}^K .
\end{align}
By definition, the product of eigenvalues is the determinant, so we can match the above product term to the determinant and 
\begin{align}
(\omega + i \partial \overline{\partial} f)^K = \det(I + \omega^{-1} i \partial \overline{\partial} f) \omega^K .
\end{align}
Since $\Psi$ is the ratio, we get
\begin{align}
e^\Psi = \det(I + \omega^{-1} i \partial \overline{\partial} f) \implies
\Psi = \log \det(I + \omega^{-1} i \partial \overline{\partial} f) .
\end{align}
Now, 
\begin{align}
\nabla_k \Psi = \nabla_k \log \det(I + \omega^{-1} i \partial \overline{\partial} f) = \frac{1}{\det(I + \omega^{-1} i \partial \overline{\partial} f)} \nabla_k \det(I + \omega^{-1} i \partial \overline{\partial} f) .
\end{align}
Using Jacobi’s formula, 
\begin{align}
\nabla_k \det(I + \omega^{-1} i \partial \overline{\partial} f) = \det(I + \omega^{-1} i \partial \overline{\partial} f) \text{Tr} \Big( ( I + \omega^{-1} i \partial \overline{\partial} f)^{-1} \nabla_k ( I + \omega^{-1} i \partial \overline{\partial} f) \Big) .
\end{align}
Therefore,
\begin{align}
\label{eqn:Psi_deriv}
\nabla_k \Psi & = \cancel{\frac{\det(I + \omega^{-1} i \partial \overline{\partial} f)}{\det(I + \omega^{-1} i \partial \overline{\partial} f)}} \text{Tr}((I + \omega^{-1} i \partial \overline{\partial} f)^{-1} \omega^{-1} \nabla_k (  i \partial \overline{\partial} f) )\footnotemark.
\end{align}
\footnotetext{We have used the fact that $\nabla_k(\omega^{-1} i \partial \overline{\partial} f) = \underbrace{(\nabla_k \omega^{-1})}_{=0} (i \partial \overline{\partial} f) + \omega^{-1} \nabla_k (i \partial \overline{\partial} f)$, which follows from metric compatibility.} We have set $A = I + \omega^{-1} i \partial \overline{\partial} f$. The trace is not necessarily $\mathcal{O}(K)$, it is instead a sum upon $K$ terms. Therefore, with suitable geometric assumptions such as decay, the factor of $K$ can be prevented.

\vspace{2mm}

\noindent We note nuclear norm inequality $\| M \|_1 \leq  C \| M \|_2$ with the absence of $K$. $K$ is a worst-case bound, and is omitted due to spectral decay (see section \ref{sec:theory}). We get
\begin{align}
\|\nabla_\omega \Psi\|_h \leq \|(I + \omega^{-1} i \partial \overline{\partial} f)^{-1} \omega^{-1}\|_2 \|\nabla_v (i \partial \overline{\partial} f)\|_1\footnotemark.
\end{align}
\footnotetext{Note that we can also apply the 1-norm on the other term in the product, but this still requires an application of a nuclear norm inequality with rapid eigenvalue decay.}We have noted the full gradient norm in the Kähler metric, $\|\nabla_{\omega} \Psi\|_h$, is the supremum of $|\nabla_v \Psi|$ over all tangent vectors with $\|v\|_h \leq 1$, but $\Psi \neq i \partial \overline{\partial f}$. We have noted independence of direction in the above. By nuclear norm with decay, so no extra factor of dimension, 
\begin{align}
\|\nabla_\omega \Psi\|_h \leq C \|A^{-1}\|_2 \|h^{-1}\|_2 \|\nabla_v (i \partial \overline{\partial} f)\|_2  .
\end{align}
Because $\|A^{-1}\|_2$ and $\|h^{-1}\|_2$\footnote{This is $\mathcal{O}(1)$ under a regularized loss; in general, it is not true that it is $\mathcal{O}(1)$ in unrestricted settings.} are $\mathcal{O}(1)$, this yields
\begin{align}
\|\nabla_\omega \Psi\|_h \leq \widetilde{C} \|\nabla_v (i \partial \overline{\partial} f)\|_2 .
\end{align}

\vspace{2mm}

\noindent Now, our result in \ref{app:dolb_cov_bound} is similar to the recursion rules as in \cite{yaida2022metaprincipledfamilyhyperparameterscaling} \cite{guillen2026finitewidthneuraltangentkernels}. Hence, we use
\begin{align}
\sup_{\|v\|_2=1} \| \nabla_v (i \partial \overline{\partial} f) \|_2 \leq \| \partial^3 f \|_2 + C_2 \| i \partial \overline{\partial} f \|_2 \sup_{\theta \in \mathcal{S}} \| i \partial \overline{\partial} f \|_2 .
\end{align}
Substituting the recursive bound of Lemma 2, and by our norm relation,
\begin{align}
\| \nabla_{\omega} \Psi \|_h \leq  \widetilde{C}  \left( \| \partial^3 f \|_2 + C \sup_{\theta \in \mathcal{S}} \| i \partial \overline{\partial} f \|_2^2 \right) .
\end{align}
Returning to the Mean Value Theorem setup,
\begin{align} 
\sup_{\mathcal{S}} \Psi - \inf_{\mathcal{S}} \Psi \leq \widetilde{C} \left(  \sup_{\theta \in \mathcal{S}} (  \| \partial^3 f \|_2 + C \| i \partial \overline{\partial} f \|_2^2 ) \right) \text{diam}_\omega(\mathcal{S}) .
\end{align}
The maximum principle of Appendix \ref{app:dolb_cov_bound} ensures the spectral norm of the complex Hessian is bounded by the oscillation
\begin{align}
\sup_{\theta \in \mathcal{S}} \| i \partial \overline{\partial} f \|_2 & \leq \frac{C_0}{\sqrt{m}} \exp\left( \sup_{\mathcal{S}} \Psi - \inf_{\mathcal{S}} \Psi \right) 
\\
& \leq \frac{\widetilde{C}}{\sqrt{m}} \exp\left( D \sup_{\theta \in \mathcal{S}} (  \| \partial^3 f \|_2 + C \| i \partial \overline{\partial} f \|_2^2 )  \right)
\\
& = \frac{\widetilde{C}}{\sqrt{m}} \left\{ 1 +  \sum_{j=1}^{\infty} \frac{1}{j!} \left(  D  \sup_{\theta \in \mathcal{S}}  (  \| \partial^3 f \|_2 + C \| i \partial \overline{\partial} f \|_2^2 )\right)^j \right\}
\\
& \leq \frac{\widetilde{C}}{\sqrt{m}} \left( 1 +  \mathcal{O}((g(m))_{\text{degree}(g) \leq 0}) + \mathcal{O}((g(m))_{\text{degree}(g) \leq 0})   \right) 
\\
& = \mathcal{O}(\frac{1}{\sqrt{m}} ).
\end{align}
We use the estimate in \ref{app:yau_estimate} to obtain a transcendental inequality, and we then use a Taylor expansion. We operate under assumption $\| \partial^3 f \|_2, \| i \partial \overline{\partial} f\|_2^2$ are $\mathcal{O}(m^{-\alpha})$, $\alpha \geq 0$. We remark this argument is an a priori estimate since we only need $\mathcal{O}(1)$ in the above, which is reasonable and less restrictive than our result. We refer to the end of Appendix \ref{app:dolb_cov_bound} for references that the above is reasonable.

\noindent $\square$

\subsection{Dolbeault covariant bound}
\label{app:dolb_cov_bound}

\begin{figure}[t]
  \centering
  \includegraphics[width=0.45\textwidth]{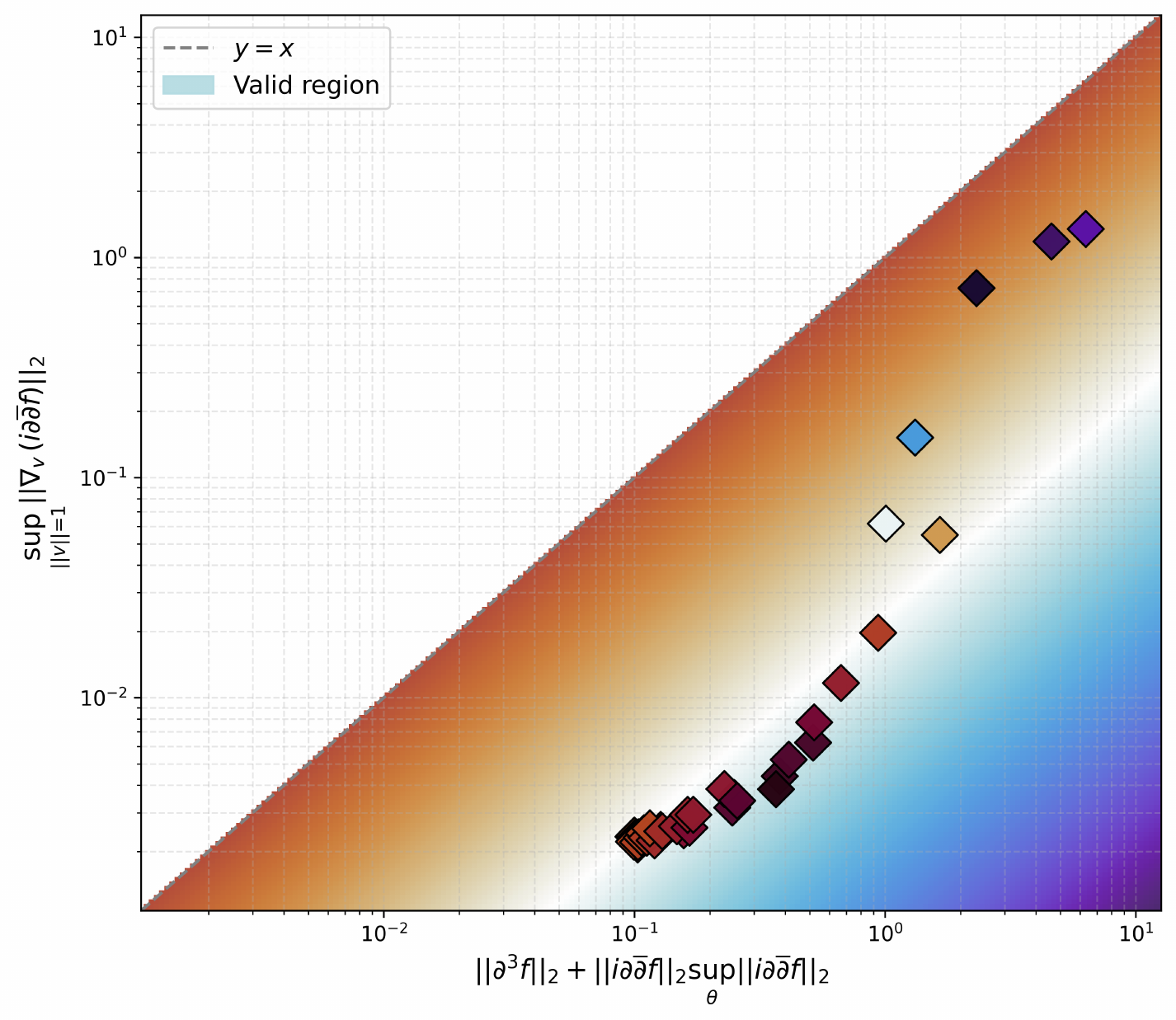}
  \vspace{4mm}
  \caption{We illustrate that Lemma 1 holds, where the shaded region represents a valid upper bound, the background color representing the bound gap. Here, we choose constant $C=1$. The x-axis is the right-hand side of \ref{eqn:lemma_1} and the y-axis the left-hand side.}
  \label{fig:covariant_bound}
  \vspace{-0mm}
\end{figure}

\noindent \textit{Proof of Lemma 1.} We apply the covariant derivative to the Hessian compatible with the Kähler metric. Because the manifold is Kähler, the Levi-Civita connection is compatible with the complex structure, so the mixed symbols such as ${\Gamma_{k \overline{i}}}^l$ vanish. The Levi-Civita connection on the Hessian yields 
\begin{align}
\nabla_k \partial_{i} \partial_{\overline{j}} f = \partial_k \partial_i \partial_{\overline{j}} f - {\Gamma_{ki}}^l \partial_l \partial_{\overline{j}} f
\end{align}
and
\begin{align}
\nabla_{\overline{k}} \partial_{i} \partial_{\overline{j}} f = \partial_{\overline{k}} \partial_i \partial_{\overline{j}} f - \overline{{\Gamma_{kj}}^l} \partial_i \partial_{\overline{l}} f .
\end{align}
By the triangle inequality on the operator norm, and denoting $\mathcal{H}_{i \overline{j}} = \partial_i \partial_{\overline{j}} f$,
\begin{align}
\| \nabla_{\omega} \mathcal{H} \|_2 \leq \| \partial^3 f \|_2 + \| \Gamma \cdot \mathcal{H} \|_2 .
\end{align}
We can note the Christoffel symbols are governed by ${\Gamma_{ki}}^l = h^{l \overline{m}} \partial_k h_{i \overline{m}}$. Our proof strategy will be to show the third derivative term is at least of equivalent order as the covariant derivative term, and the Christoffel symbol term is $\mathcal{O}(\frac{1}{\sqrt{m}})$, therefore we have the recursive relationship.

\vspace{2mm}

\noindent The probabilistic model $p$ obeys $p(y|z, \theta) = \mathcal{CN}(f(\theta; z), \sigma^2)$. We notice the log likelihood obeys $  \log p = -\frac{1}{\sigma^2} |y - f(\theta; z)|^2 + C = -\frac{1}{\sigma^2} (y - f)(\overline{y} - \overline{f}) + C$. Denote the loss residual $\epsilon = y - f$. The derivative of the log likelihood w.r.t. $\theta$ is
\begin{align}
\partial_i \log p = \frac{1}{\sigma^2} \left( \overline{\epsilon} \partial_i f + \epsilon \partial_i \overline{f} \right) . 
\end{align}

\vspace{2mm}

\noindent In information geometry, the derivative of the Fisher metric follows the Amari-Chentsov tensor \cite{schwachhöfer2017congruentfamiliesinvarianttensors} \cite{gnandi2026constructionexponentialfamiliesstatistical}, although recall from section \ref{sec:geometry} our metric is more closely a cross-entropy one. Therefore, we actually have
\begin{align}
\label{eqn:metric_deriv}
\partial_k h_{i\overline{m}} = \EX_q \left[ \frac{\partial^2 \log p}{\partial \theta^k \partial \theta^i} \frac{\partial \log p}{\partial \overline{\theta}^m} \right] + \EX_q \left[ \frac{\partial \log p}{\partial \theta^i} \frac{\partial^2 \log p}{\partial \theta^k \partial \overline{\theta}^m} \right] - \EX_q \left[ \partial_k \left( \frac{ \partial_i \partial_{\overline{m}} p}{p} \right) \right].
\end{align}
This partially follows from
\begin{align}
-\partial_i \partial_{\overline{m}} \log p = \partial_i \log p \partial_{\overline{m}} \log p - \frac{\partial_i \partial_{\overline{m}} p}{p} 
\end{align}
and differentiating. Let us examine the first term. Differentiating $\log p$ again,
\begin{align}
\partial^2_{ki} \log p = \frac{1}{\sigma^2} \left( -\partial_k \overline{f} \partial_i f + \overline{\epsilon} \partial^2_{ki} f - \partial_k f \partial_i \overline{f} + \epsilon \partial^2_{ki} \overline{f} \right) . 
\end{align}
Evaluating the first term of $\partial_k h_{i \overline{m}}$,
\begin{align}
\label{eqn:ex_logp}
\EX_q \left[ \partial^2_{ki} \log p \partial_{\overline{m}} \log p \right] = \frac{1}{\sigma^4} \EX_q \left[ \left( \overline{\epsilon} \partial^2_{ki} f + \epsilon \partial^2_{ki} \overline{f} - \partial_k \overline{f} \partial_i f - \partial_k f \partial_i \overline{f} \right) \left( \overline{\epsilon} \partial_{\overline{m}} f + \epsilon \partial_{\overline{m}} \overline{f} \right) \right] . 
\end{align}
By Wick's theorem, isolated terms in $\epsilon, \overline{\epsilon}$ vanish and what remains is $\EX_q [\epsilon \overline{\epsilon}] = \sigma^2$. The surviving components are
\begin{align}
\EX_q \Bigg[ \partial^2_{ki} \log p  \partial_{\overline{m}} \log p \Bigg] = \frac{1}{\sigma^2} \EX_q \Bigg[ \partial^2_{ki} f \partial_{\overline{m}} \overline{f} + \partial^2_{ki} \overline{f} \partial_{\overline{m}} f \Bigg] . 
\end{align}
A similar argument for the second term of $\partial_k h_{i \overline{m}}$ yields
\begin{align}
\EX_q \Bigg[ \partial_i \log p  \partial^2_{k\overline{m}} \log p \Bigg] = \frac{1}{\sigma^2} \EX_q \Bigg[ \partial_i f \partial^2_{k\overline{m}} \overline{f} + \partial_i \overline{f} \partial^2_{k\overline{m}} f \Bigg] . 
\end{align}
Evaluating the third term of \ref{eqn:metric_deriv},
\begin{align}
-\EX_q \left[ \partial_k \left( \frac{ \partial_i \partial_{\overline{m}} p}{p} \right) \right] = \frac{1}{\sigma^2} \EX_q \Bigg[ \partial_k \overline{f} \partial^2_{i\overline{m}} f + \partial_k f \partial^2_{i\overline{m}} \overline{f} \Bigg] .
\end{align}
Consolidating the terms, 
\begin{align}
\partial_k h_{i\overline{m}} = \frac{1}{\sigma^2} \EX_q \Bigg[ \mathcal{H}_{ki} \partial_{\overline{m}} \overline{f} + \partial^2_{ki} \overline{f} \partial_{\overline{m}} f + \partial_i f \mathcal{H}_{k\overline{m}}^{\dagger} + \partial_i \overline{f} \mathcal{H}_{k\overline{m}} + \partial_k \overline{f} \mathcal{H}_{i\overline{m}} + \partial_k f \mathcal{H}_{i\overline{m}}^{\dagger} \Bigg] .
\end{align}
Taking the norm, by Jensen's inequality, and taking the supremum it follows
\begin{align}
\label{eqn:partial_h}
\| \partial h \|_2 \leq \frac{1}{\sigma^2} \EX_q \Bigg[ 2 \sup_{\theta \in \mathcal{S}}\| \partial^{2,0} f \|_2 \| \partial f \|_2 + 4 \sup_{\theta \in \mathcal{S}} \| \partial f \|_2 \| \mathcal{H} \|_2 \Bigg]  . 
\end{align}
By definition, the Dolbeault Hessian norm is exactly $\| \mathcal{H} \|_2 = \| i \partial \overline{\partial} f \|_2 = \| \partial^2_{k\overline{m}} f \|_2$. We will allow the spectral norm of the (2,0) derivative to be bounded proportionally by the Dolbeault Hessian, meaning there exists a constant $C' > 0$ such that $\| \partial^2_{ki} f \|_2 \leq C' \| \mathcal{H} \|_2$. This is an assumption, which is reasonable. Taking the norm and a supremum to account for the expectation, for \ref{eqn:partial_h} we get
\begin{align}
\| \partial_k h_{i\overline{m}} \|_2 \leq  C \sup_{\theta \in \mathcal{S}} \| \mathcal{H} \|_2 \| \partial f \|_2 .
\end{align}
Recall the Christoffel symbols are defined via ${\Gamma_{ki}}^l = h^{l\overline{m}} \partial_k h_{i\overline{m}}$. As a remark, we can note Christoffel symbols are invariant to constant rescalings of the metric, so $(c^{-1} h^{l\overline{m}}) \partial_k (c \cdot h_{i\overline{m}}) = h^{l\overline{m}} \partial_k h_{i\overline{m}}$.  The operator norm of the inverse metric is bounded $\mathcal{O}(K^{-1})$ normally. but recall we have instilled rapid spectral decay via the hypotheses, so it is $\mathcal{O}(1)$. Therefore,
\begin{align}
\sup_{\|v\|_2=1} \| \Gamma(v, \cdot) \|_2 \leq  \| h^{-1} \|_2 \sup_{\theta \in \mathcal{S}} \| \partial_k h_{i\overline{m}} \|_2 \leq  C \| h^{-1} \|_2 \sup_{\theta \in \mathcal{S}} \| \partial f \|_2  \| \mathcal{H} \|_2 .
\end{align}
It immediately follows from the neural network definition the Jacobian of $f$ extracts an $\mathcal{O}(1)$ scaling factor, thus we conclude the bound on the Christoffel symbols
\begin{align}
\sup_{\|v\|_2=1} \| \Gamma(v, \cdot) \|_2  \leq C_1  \sup_{\theta \in \mathcal{S}} \| \mathcal{H} \|_2 .
\end{align}
Returning to the triangle inequality, we have
\begin{align}
\| \Gamma \cdot \mathcal{H} \|_2 \leq \sup_{\|v\|_2=1} \| \Gamma(v, \cdot) \|_2 \| \mathcal{H} \|_2 \leq C_1 \| \mathcal{H} \|_2 \sup_{\theta \in \mathcal{S}} \| \mathcal{H} \|_2 .
\end{align}

\vspace{2mm}

\noindent Next, we address the third derivative term $\| \partial^3 f \|_2$. It suffices to show $\| \partial^3 f \|$ is lower than $\mathcal{O}(1)$. By \cite{banerjee2023restricted}  we have the real Hessian is $\mathcal{O}(\frac{1}{\sqrt{m}})$ and a neural network will not increase the order to $\mathcal{O}(\sqrt{m})$. This phenomenon is known to be true such as in \cite{aitken2020asymptoticswidenetworkspolynomial} \cite{hanin2023randomfullyconnectedneural} \cite{dyer2019asymptoticswidenetworksfeynman} \cite{hanin2023randomfullyconnectedneural} \cite{guillen2026finitewidthneuraltangentkernels} \cite{huang2019dynamicsdeepneuralnetworks} \cite{andreassen2020asymptoticswideconvolutionalneural} \cite{cirone2025genusexpansionnonlinearrandom}.

\noindent $ \square $

\begin{figure}[t]
  \centering
  \includegraphics[width=0.55\textwidth]{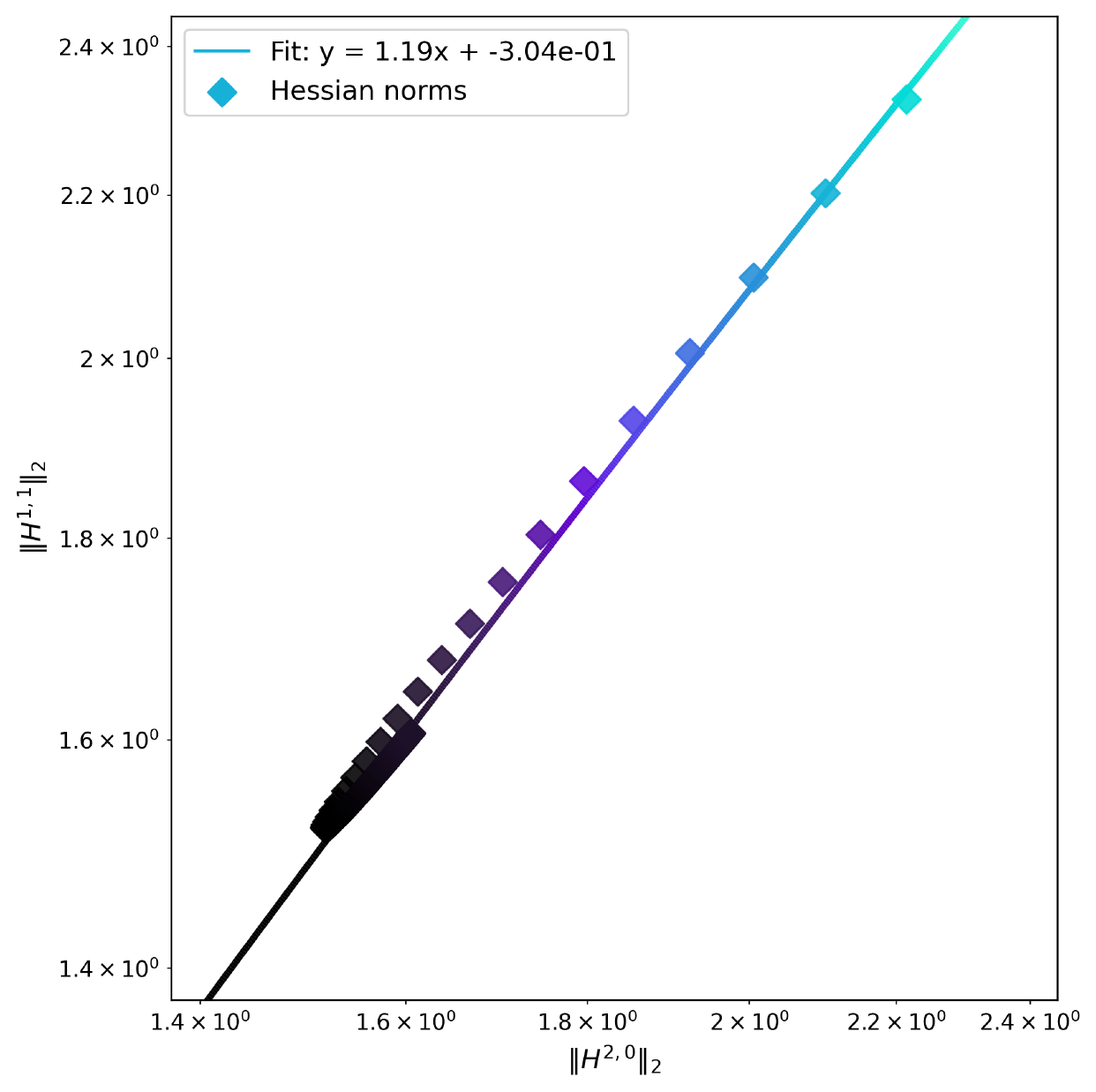}
  \caption{We illustrate a proportion relationship between the (1,1) and (2,0)-Hessians. We plot $\| \mathcal{H}^{1,1}\|_2, \|\mathcal{H}^{2,0}\|_2$ exactly without the constant. The line is the best fit line. We evaluate at 80 distinct parameter points.}
  \label{fig:Hessian_1,1_2,0_proportions}
\end{figure}

\vspace{2mm}

\textit{Remark.} If we consider the deformed metric $\omega_f = \omega + i\partial\overline{\partial}f$, construct a tensor representing the difference between the Chern connections of the deformed metric and the base metric. We get a difference formula
\begin{align}
\Psi^k_{ij} = \Psi^k_{ij} (\omega_f, \omega) := {\Gamma(\omega_f)_{ij}}^k - {\Gamma(\omega)_{ij}}^k = (\omega_f)^{k\overline{l}} \nabla^\omega_i (\omega_f)_{j\overline{l}} .
\end{align}
Because the base metric is covariantly constant with respect to its own connection, we can write
\begin{align}
\nabla^\omega_i (\omega_f)_{j\overline{l}} = \nabla^\omega_i (\omega_{j\overline{l}} + \mathcal{H}_{j\overline{l}}) = \nabla^\omega_i \mathcal{H}_{j\overline{l}} .
\end{align}
Therefore, we can see
\begin{align}
\Psi^k_{ij} = (\omega + \mathcal{H})^{k\overline{l}} \nabla^\omega_i \mathcal{H}_{j\overline{l}} .
\end{align}
Taking the norm, and using \ref{eqn:lemma_1} from the lemma
\begin{align}
\|\Psi\|_2 & \leq \|(\omega + \mathcal{H})^{-1}\|_2 \cdot \|\nabla^\omega \mathcal{H}\|_2  \\
& \leq \|(\omega + \mathcal{H})^{-1}\|_2 \left( \|\partial^3 f\|_2 + C\|\mathcal{H}\|_2 \sup_{\theta \in \mathcal{S}} \|\mathcal{H}\|_2 \right) .
\end{align}

\subsection{Spectral $C^2$ estimate}
\label{app:yau_estimate}

\noindent Let us use the fact that the scaling limit of $\mathcal{H}(\theta_0) = i\partial\overline{\partial} f(\theta_0)$ obeys at initialization a pointwise bound (as aside, see \cite{banerjee2023restricted} \cite{liu2021losslandscapesoptimizationoverparameterized} \cite{taheri2024sharperguaranteeslearningneural} \cite{liu2021linearitylargenonlinearmodels} for other relevant literature)
\begin{align}
\| \mathcal{H}(\theta_0) \|_2 \leq \frac{C_{\text{init}}}{\sqrt{m}} .
\end{align}
This is a localized fact, and the burden of the proof is shifted to showing that the Monge-Ampère dynamics prevent the Hessian from escaping this scale globally as $\theta$ moves across $\mathcal{S}$.

\vspace{2mm}

\noindent We avoid use of the trace functional since trace is affected by $K$, thus our final bound would involve $K$, which is undesirable. We need only work with the spectral norm. Therefore, we find a maximum principle applicable for us. 

\vspace{2mm}

\noindent \textit{Proof of Lemma 2.} Define the test function
\begin{align}
\label{eqn:v}
v(\theta,t) = \log(\lambda_{\max}(\mathcal{H}(\theta_t))) - \Psi(\theta_t) - \xi t.
\end{align}
Applying the parabolic operator to the definition of $v$ yields
\begin{align}
(\partial_t - \Delta_{\widetilde{\omega}}) v = (\partial_t - \Delta_{\widetilde{\omega}}) [\log ( \lambda_{\text{max}} (\mathcal{H}(\theta_t))) - \Psi(\theta_t)] - \xi \leq \xi - \xi = 0.
\end{align}
Because $(\partial_t - \Delta_{\widetilde{\omega}}) v \leq 0$ on $\mathcal{S} \times [0, T]$, $v$ is a subsolution. By the parabolic maximum principle, $v$ must attain its supremum on the parabolic boundary $t=0$. Therefore, for any $t \in [0, T]$, we have $v(\theta_t,t) \leq \sup_{\theta_0 \in \mathcal{S}} v(\theta_0,0)$.
\vspace{2mm}

\noindent Hence
\begin{align}
\log(\lambda_{\max}(\mathcal{H}(\theta_t))) - \Psi(\theta_t) - \xi t \leq \sup_{\theta_0 \in \mathcal{S}} \left( \log(\lambda_{\max}(\mathcal{H}(\theta_0))) - \Psi(\theta_0) \right) .
\end{align}
We can rewrite
\begin{align}
\log(\lambda_{\max}(\mathcal{H}(\theta_t))) \leq \sup_{\theta_0 \in \mathcal{S}} \log(\lambda_{\max}(\mathcal{H}(\theta_0))) + \Psi(\theta_t) - \inf_{\theta_0 \in \mathcal{S}} \Psi(\theta_0) + \xi t .
\end{align}
By definition of the spectral norm and exponentiating,
\begin{align}
\label{eqn:v_exp}
\|\mathcal{H}(\theta_t)\|_2 \leq \left( \sup_{\theta_0 \in \mathcal{S}} \|\mathcal{H}(\theta_0)\|_2 \right) \exp \left( \xi t \right) \exp\left( \Psi(\theta_t) - \inf_{\mathcal{S}} \Psi \right) .
\end{align}
Rewrite the first exponential in \ref{eqn:v_exp} via an upper bound since $t \leq T$, and call it $C_{\text{geom}}$. Taking the supremum over all $\theta_t \in \mathcal{S}$ on both sides yields as desired
\begin{align}
\sup_{\theta \in \mathcal{S}} \|\mathcal{H}\|_2 \leq C_{\text{geom}} \sup_{\theta_0 \in \mathcal{S}} \|\mathcal{H}(\theta_0)\|_2 \exp\left( \sup_{\mathcal{S}} \Psi - \inf_{\mathcal{S}} \Psi \right) .
\end{align}

\noindent $ \square $

\vspace{2mm}

\textit{Remark.} The inclusion of $\xi$ is useful to obtain a constant $C_{\text{geom}}$ in front. The constant is necessary to keep the assumptions reasonable and it does not affect the proof in \ref{app:dolb_hessian_bounds}, which is a crucial result that the result here supplements.

\subsection{$\nabla_{\omega}^{2,0} f$ bounds}
\label{app:2,0_hess_bounds}

\noindent \textit{Proof of Lemma 3.} As before, let us begin with the property
\begin{align}
\| \nabla_\omega \mathcal{H}^{2,0} \|_2 \leq C_1 \| \partial^3 f \|_2 + C_2 \sup_{\theta \in \mathcal{S}} \| \mathcal{H}^{2,0} \|_2^2 .
\end{align}
This bound follows similarly to \ref{app:dolb_cov_bound}, but we can note 
\begin{align}
\partial_k h_{i\overline{m}} = \frac{1}{\sigma^2} \EX_q \Bigg[ \mathcal{H}_{ki} \partial_{\overline{m}} \overline{f} + \partial^2_{ki} \overline{f} \partial_{\overline{m}} f + \partial_i f \mathcal{H}_{k\overline{m}}^{\dagger} + \partial_i \overline{f} \mathcal{H}_{k\overline{m}} + \partial_k \overline{f} \mathcal{H}_{i\overline{m}} + \partial_k f \mathcal{H}_{i\overline{m}}^{\dagger} \Bigg] .
\end{align}
and
\begin{align}
\| \partial_k h_{i\overline{m}} \|_2 \leq \frac{2}{\sigma^2} \left[ \| \mathcal{H}^{2,0} \|_2 \| \partial f \|_2 + 2 \| \partial f \|_2 \| \mathcal{H}^{1,1} \|_2 \right] .
\end{align}
Moreover, we assume $\| \mathcal{H}^{1,1} \|_2 \asymp  \| \mathcal{H}^{2,0} \|_2$ (or equivalently $c \| \mathcal{H}^{2,0} \|_2 \leq \| \mathcal{H}^{1,1} \|_2 \leq C \| \mathcal{H}^{2,0} \|_2$), and the remainder of the proof is identical, so we omit the details.

\vspace{2mm}

\noindent Let us consider a geodesic along the Kähler manifold within the geodesic ball $\mathcal{S} = B_\omega(\theta_0, R)$. Let us denote the diameter $D = 2R$. We can note the absolute derivative is equal to the covariant derivative along the tangent vector $\frac{D}{dt} \mathcal{H}^{2,0} = \nabla_{\dot{\gamma}(t)} \mathcal{H}^{2,0}$. Now, we can note
\begin{align}
\left| \frac{d}{dt} \| \mathcal{H}^{2,0}(\gamma(t)) \|_2 \right|
&\leq \| \nabla_{\dot{\gamma}(t)} \mathcal{H}^{2,0} \|_2 \\
&\leq \| \nabla_\omega \mathcal{H}^{2,0} \|_2 \| \dot{\gamma}(t) \|_h \\
&= \| \nabla_\omega \mathcal{H}^{2,0} \|_2 \\
&\leq C_1  \| \partial^3 f \|_2 + C_2 \sup_{\theta \in \mathcal{S}} \| \mathcal{H}^{2,0} \|_2^2 .
\end{align}
The first inequality is by a geometric Kato inequality \cite{Herzlich2000}, the second inequality follows from properties of 2-norms on the covariant derivative, and the equality follows from the fact that the curve is unit speed. Let us denote $A = \sup_{\theta \in \mathcal{S}} \| \partial^3 f \|_2, S = \sup_{\theta \in \mathcal{S}} \| \mathcal{H}^{2,0} \|_2$ for short.

\vspace{2mm}

\noindent The $(2,0)$ Hessian at initialization obeys $\mathcal{O}(\frac{1}{\sqrt{m}})$ scaling, and the third derivative obeys at least as high scaling. We integrate the ordinary differential equation 
\begin{align}
\frac{d}{dt} \| \mathcal{H}^{2,0}(\gamma(t)) \|_2 \leq C_1 A + C_2 S^2 . 
\end{align}
Integrating this from $t = 0$ to $t \leq D$ gives the bound
\begin{align}
\| \mathcal{H}^{2,0}(\gamma(t)) \|_2 \leq \| \mathcal{H}^{2,0}(\theta_0) \|_2 + D (C_1 A + C_2 S^2) .
\end{align}
Therefore, we can note
\begin{align}
S \leq \| \mathcal{H}^{2,0}(\theta_0) \|_2 + D C_1 A + D C_2 S^2 . 
\end{align}
Rearranging,
\begin{align}
(D C_2) S^2 - S + \left( \| \mathcal{H}^{2,0}(\theta_0) \|_2 + D C_1 A \right) \geq 0 . 
\end{align}
For the inequality to hold, $S$ must lie outside the roots of the corresponding parabola. Let $K = \| \mathcal{H}^{2,0}(\theta_0) \| + D C_1 A$. By assumption, both the initial Hessian and $A$ scale as $\mathcal{O}(\frac{1}{\sqrt{m}})$, therefore $K = \mathcal{O}(\frac{1}{\sqrt{m}})$. The roots of the quadratic $D C_2 S^2 - S + K = 0$ are given by
\begin{align}
S_{\pm} = \frac{1 \pm \sqrt{1 - 4 D C_2 K}}{2 D C_2} .
\end{align}
By continuity of the Hessian with respect to the initial scaling, $S$ must remain on the lower branch $S \leq S_-$. Applying the identity $1 - \sqrt{1-x} = \frac{x}{1+\sqrt{1-x}}$, we find
\begin{align}
S_- & = \frac{1 - \sqrt{1 - 4 D C_2 K}}{2 D C_2}
\\
& = \frac{4 D C_2 K}{2 D C_2 (1 + \sqrt{1 - 4 D C_2 K})}
\\
& = \frac{2 K}{1 + \sqrt{1 - 4 D C_2 K}}  .
\end{align}
Since $\sqrt{1 - 4 D C_2 K} \geq 0$, the denominator is bounded below by $1$, which gives the upper bound $S_- \leq 2 K$. We arrive at
\begin{align}
\label{eqn:2,0_asymptotics}
\sup_{\theta \in \mathcal{S}} \| \mathcal{H}^{2,0} \|_2 = \mathcal{O}\left(\frac{1}{\sqrt{m}}\right) .
\end{align}

\noindent $\square$

\vspace{2mm}

\noindent \textit{Remark.} The application of the geometric Kato inequality follows since
\begin{align}
\frac{d}{dt} \| \mathcal{H}^{2,0} \|_2^2 = \frac{d}{dt} \langle \mathcal{H}^{2,0}, \mathcal{H}^{2,0} \rangle = 2 \langle \nabla_{\dot{\gamma}(t)} \mathcal{H}^{2,0}, \mathcal{H}^{2,0} \rangle .
\end{align}
It also follows from the chain rule
\begin{align}
\frac{d}{dt} \| \mathcal{H}^{2,0} \|_2^2 = 2 \| \mathcal{H}^{2,0} \|_2 \frac{d}{dt} \| \mathcal{H}^{2,0} \|_2  .
\end{align}
Equating the two,
\begin{align}
\| \mathcal{H}^{2,0} \|_2 \frac{d}{dt} \| \mathcal{H}^{2,0} \|_2 = \langle \nabla_{\dot{\gamma}(t)} \mathcal{H}^{2,0}, \mathcal{H}^{2,0} \rangle  .
\end{align}
Applying the absolute value and the Cauchy-Schwarz inequality,
\begin{align}
\| \mathcal{H}^{2,0} \|_2 \left| \frac{d}{dt} \| \mathcal{H}^{2,0} \|_2 \right| = \left| \langle \nabla_{\dot{\gamma}(t)} \mathcal{H}^{2,0}, \mathcal{H}^{2,0} \rangle \right| \leq \| \nabla_{\dot{\gamma}(t)} \mathcal{H}^{2,0} \|_2 \| \mathcal{H}^{2,0} \|_2 .
\end{align}
Dividing by $\| \mathcal{H}^{2,0} \|_2$ recovers what we desire.

\vspace{2mm}

\textit{\textbf{Corollary.} Given the bound $\sup_{\theta \in \mathcal{S}} \|\mathcal{H}^{2,0}\|_2 = \mathcal{O}(\frac{1}{\sqrt{m}})$, the spectral norm of the $(0,2)$ Hessian over the geodesic ball is similarly bounded
\begin{align}
\sup_{\theta \in \mathcal{S}} \|\mathcal{H}^{0,2}\|_2 = \mathcal{O}\left(\frac{1}{\sqrt{m}}\right) .
\end{align}}

\vspace{2mm}

\noindent \textit{Proof.} Because the $f$ output is real-valued, i.e. $f : M \rightarrow \mathbb{R}$, the function is equal to its complex conjugate, $f = \overline{f}$. On a Kähler manifold, the Levi-Civita connection is compatible with the complex structure, meaning the only non-vanishing Christoffel symbols are of type ${\Gamma_{ij}}^k$ and their conjugates ${\Gamma_{\overline{i}\overline{j}}}^{\overline{k}} = \overline{{\Gamma_{ij}}^k}$. The $(2,0)$ covariant Hessian in local coordinates is given by
\begin{align}
\nabla_i \nabla_j f = \partial_i \partial_j f - {\Gamma_{ij}}^k \partial_k f .
\end{align}
Taking the complex conjugate of this expression and applying the reality of $f$, we obtain
\begin{align}
\overline{\nabla_i \nabla_j f} = \partial_{\overline{i}} \partial_{\overline{j}} \overline{f} - \overline{{\Gamma_{ij}}^k} \partial_{\overline{k}} \overline{f} = \partial_{\overline{i}} \partial_{\overline{j}} f - {\Gamma_{\overline{i}\overline{j}}}^{\overline{k}} \partial_{\overline{k}} f = \nabla_{\overline{i}} \nabla_{\overline{j}} f .
\end{align}
This demonstrates that the $(0,2)$ Hessian is exactly the complex conjugate of the $(2,0)$ Hessian, $\mathcal{H}^{0,2} = \overline{\mathcal{H}^{2,0}}$. For any linear operator represented in coordinates, the spectral norm induced by the localized metric $h$ is invariant under complex conjugation. The nonzero eigenvalues of $A^\dagger A$ are real by the Spectral theorem since $(A^\dagger A)^\dagger = A^\dagger (A^\dagger)^\dagger = A^\dagger A$, and thus $\lambda_{\max}(A^\dagger A) = \lambda_{\max}(\overline{A}^\dagger \overline{A})$. Therefore, the operator norms coincide 
\begin{align}
\|\mathcal{H}^{0,2}\|_2 = \|\overline{\mathcal{H}^{2,0}}\|_2 = \|\mathcal{H}^{2,0}\|_2 .
\end{align}
Applying the supremum over the geodesic ball $\mathcal{S}$ and substituting the result of \ref{eqn:2,0_asymptotics}, we conclude
\begin{align}
\sup_{\theta \in \mathcal{S}} \|\mathcal{H}^{0,2}\|_2 = \sup_{\theta \in \mathcal{S}} \|\mathcal{H}^{2,0}\|_2 = \mathcal{O}\left(\frac{1}{\sqrt{m}}\right) .
\end{align}

\noindent $\square$

\begin{figure}[htbp]
  \centering
  \includegraphics[width=0.55\textwidth]{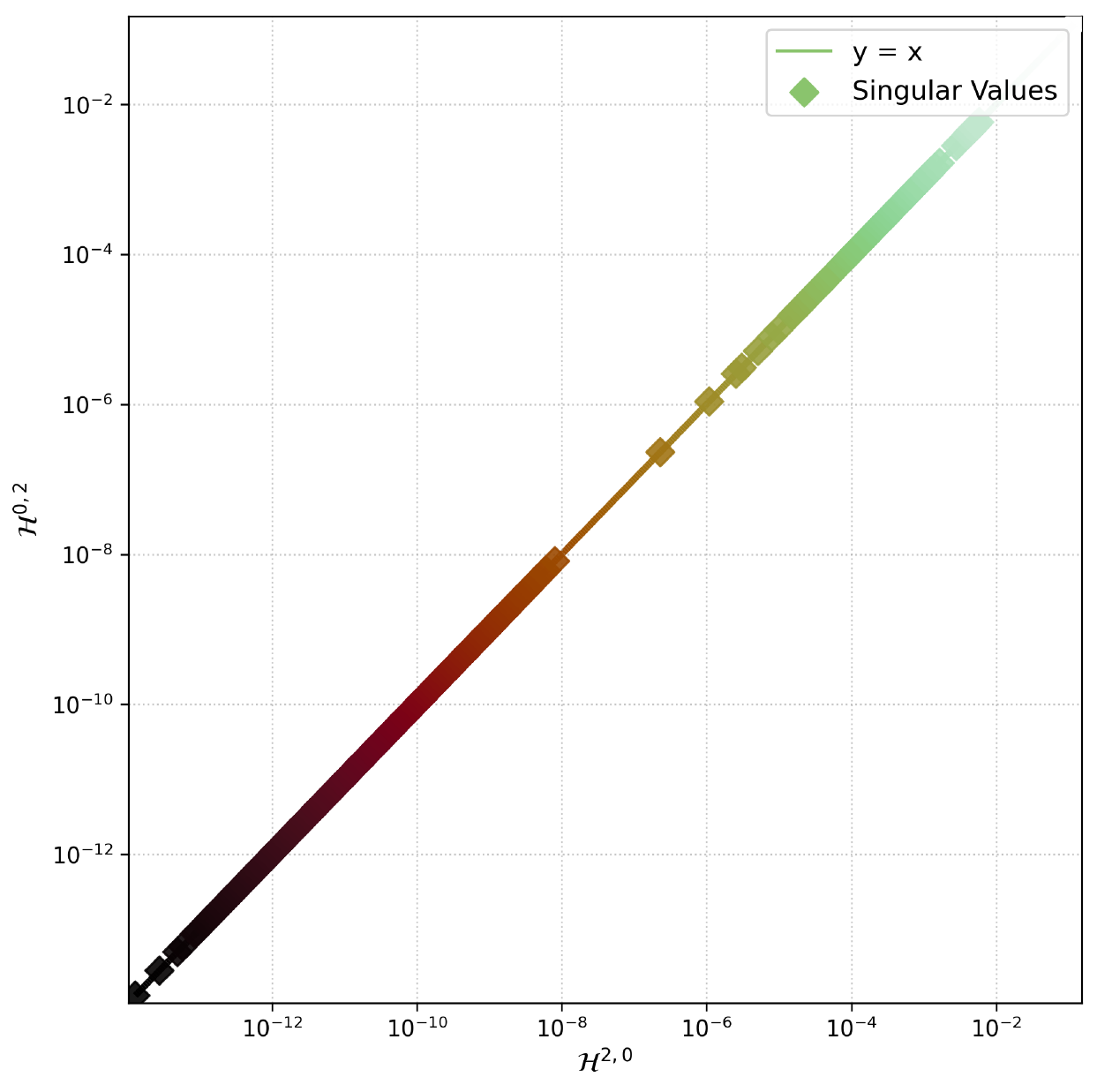}
  \caption{We plot the corollary of Lemma 2, and that the singular values of $\mathcal{H}^{2,0}$ and $\mathcal{H}^{0,2}$ match. We plot all 2,750 singular values of Hessian matrices of the same dimension. We choose network width $m=50$ and input dimension $d=4$, where input $z$ is sampled randomly. Here, color indicates value.}
  \label{fig:2,0_0,2_corollary}
\end{figure}

\section{Initialization results}
\label{app:initialization}

\noindent \textit{Proof of Theorem 3.} Assume the depth $L$, and assume the first and second Wirtinger derivatives of the activation $\phi$ are uniformly bounded. At initialization, let the entries of each hidden weight matrix $W^{(l)}$ be drawn independently from $\mathcal{CN}(0,1)$, and let the output vector $v$ be initialized  with $ \|v\|_2 = 1$.
For example, one may take $v=g/\|g\|_2$ with $g\sim\mathcal{CN}(0,I_m)$. We first examine the forward pass. Assume for the sake of induction that 
\begin{align} 
\|\alpha^{(l-1)}\|_2^2 = \mathcal{O}(m). 
\end{align} 
Given $\alpha^{(l-1)}$, the variance of a pre-activation is 
\begin{align}
\operatorname{Var}\left(h_i^{(l)}\right) = \operatorname{Var}\left( \frac{1}{\sqrt m} \sum_{j=1}^{m} W_{ij}^{(l)} \alpha_j^{(l-1)} \right) = \frac{1}{m} \sum_{j=1}^{m} \left|\alpha_j^{(l-1)}\right|^2 = \mathcal{O}(1).
\end{align} 
Under regularity of $\phi$, the above then implies with high probability, 
\begin{align} 
\label{eqn:alpha_Om} 
\|\alpha^{(l)}\|_2^2 = \mathcal{O}(m), 
\end{align} 
and taking the square root gives $
\|\alpha^{(l)}\|_2 = \mathcal{O}(\sqrt m)$
at every fixed layer, which completes the inductive argument. Moreover, analysis of complex Gaussian matrices gives $
\left\| \frac{1}{\sqrt m}W^{(l)} \right\|_2 = \mathcal{O}(1)$
with high probability, similar to \cite{banerjee2023restricted}.

\vspace{2mm}

\noindent Now, we  examine the backward pass. Define the Wirtinger error signals 
\begin{align}
\label{eqn:ed}
\epsilon^{(l)} & := \nabla_{h^{(l)}}f(\theta_0) \quad \text{(holomorphic gradient)} 
\\
\delta^{(l)} & := \nabla_{\overline h^{(l)}}f(\theta_0) \quad \text{(antiholomorphic gradient)},
\end{align} 
and concatenate them into
\begin{align} 
\label{eqn:bl}
b^{(l)} := \begin{pmatrix} \epsilon^{(l)} \\ \delta^{(l)} \end{pmatrix}. 
\end{align}
For the final hidden layer, $f=\frac{1}{2\sqrt m}\left(v^\dagger\alpha^{(L)} + \alpha^{(L)\dagger}v\right)$,
so the chain rule on \ref{eqn:ed} gives 
\begin{align} 
\begin{cases}
\epsilon_i^{(L)} &=\frac{1}{2\sqrt m}\left[\overline v_i\partial_h\phi_i + v_i\partial_h\overline{\phi_i}\right],
\\ \delta_i^{(L)} &=\frac{1}{2\sqrt m}\left[\overline v_i\partial_{\overline h}\phi_i + v_i\partial_{\overline h}\overline{\phi_i}\right] .
\end{cases}
\end{align} 
If $ 
|\partial_h\phi| + |\partial_{\overline h}\phi| \leq C_\phi$,
then it follows
\begin{align} 
\label{eqn:bL}
\|b^{(L)}\|_2^2 \leq \frac{C_\phi^2}{m} \sum_{i=1}^m |v_i|^2 = \frac{C_\phi^2}{m}\|v\|_2^2 = \mathcal{O}\left(\frac{1}{m}\right). 
\end{align}
We have used the properties of $v$ previously mentioned in the proof. Consequently, by taking the square root of the above of \ref{eqn:bL}
\begin{align} 
\label{eqn:bL} 
\|b^{(L)}\|_2 = \mathcal{O}\left(\frac{1}{\sqrt m}\right).
\end{align}
Let us consider a backward direction, and define the augmented Wirtinger Jacobian 
\begin{align} 
\mathcal J^{(l)} & := \frac{\partial \left(h^{(l)},\overline h^{(l)}\right)} {\partial \left(h^{(l-1)},\overline h^{(l-1)}\right)} 
\\[1em]
& :=   \frac{1}{\sqrt m}
\begin{pmatrix}
W^{(l)}D_h^{(l-1)}
&
W^{(l)}D_{\overline h}^{(l-1)}
\\
\overline{W^{(l)}}\overline{D_{\overline h}^{(l-1)}}
&
\overline{W^{(l)}}\overline{D_h^{(l-1)}}
\end{pmatrix}
\\[1em]
& :=   \frac{1}{\sqrt m}
\begin{pmatrix}
W^{(l)}\operatorname{diag}\left( \partial_h\phi\left(
h_i^{(l-1)}, \overline{h_i^{(l-1)}} \right) \right)_{i=1}^m
&
W^{(l)} \operatorname{diag}\left( \partial_{\overline h}\phi\left( h_i^{(l-1)}, \overline{h_i^{(l-1)}} \right) \right)_{i=1}^m
\\
\overline{W^{(l)}}\overline{\operatorname{diag}\left( \partial_{\overline h}\phi\left( h_i^{(l-1)}, \overline{h_i^{(l-1)}} \right) \right)_{i=1}^m}
&
\overline{W^{(l)}}\overline{\operatorname{diag}\left( \partial_h\phi\left( h_i^{(l-1)}, \overline{h_i^{(l-1)}} \right) \right)_{i=1}^m}
\end{pmatrix} .
\end{align} 
We can note $\mathcal J^{(l)}$ contains both holomorphic and antiholomorphic components generated by 
\begin{align} 
h^{(l)} = \frac{1}{\sqrt m} W^{(l)} \phi\left( h^{(l-1)},\overline h^{(l-1)} \right). 
\end{align} 
Since the normalized matrices $\frac{1}{\sqrt{m}}W^{(l)}$ have bounded spectral norm and the first Wirtinger derivatives of $\phi$ are uniformly bounded, there exists a constant $\gamma=\mathcal{O}(1)$, independent of $m$, such that $
\|\mathcal J^{(l)}\|_2 \leq \gamma$
with high probability. The multivariable Wirtinger chain rule via \ref{eqn:bl} with iterating backwards on layers yields
\begin{align} 
\label{eqn:chain_rule}
b^{(l)} & = \left(\mathcal J^{(l+1)}\right)^\dagger b^{(l+1)}
\\
\implies \|b^{(l)}\|_2 & \leq \left( \prod_{r=l+1}^L \|\mathcal J^{(r)}\|_2 \right) \|b^{(L)}\|_2 
\\
& \leq \gamma^{L-l} \mathcal{O}\left(\frac{1}{\sqrt m}\right) 
\\
& = \mathcal{O}\left(\frac{1}{\sqrt m}\right), 
\end{align}
where we have used that $L$ is fixed. In particular, 
\begin{align} 
\label{eqn:delta_infinity} 
\|\delta^{(l)}\|_\infty \leq \|\delta^{(l)}\|_2 \leq \|b^{(l)}\|_2 = \mathcal{O}\left(\frac{1}{\sqrt m}\right). 
\end{align} 
We next establish bounds with respect to a parameter perturbation. Let $X$ be a perturbation of the weight matrix $W^{(l)}$ with $ 
\|X\|_F = 1$. At layer $l$, $ 
\Delta_X h^{(l)} = \frac{1}{\sqrt m} X\alpha^{(l-1)}$.
Here, we use directional derivative notation
\begin{align}
\Delta_X Q := \lim_{t \to 0} \frac{Q(W^{(l)} + tX) - Q(W^{(l)})}{t} .
\end{align}
Hence, by \ref{eqn:alpha_Om}, 
\begin{align} 
\label{eqn:Delta_h}
\|\Delta_X h^{(l)}\|_2 \leq \frac{1}{\sqrt m} \|X\|_F \|\alpha^{(l-1)}\|_2 = \mathcal{O}(1). 
\end{align} 
Moreover, recalling the definition of $\alpha$ as in \ref{sec:network_setup}
\begin{align} 
\label{eqn:forward_sensitivity} 
\|\Delta_X h^{(r)}\|_2 + \|\Delta_X \alpha^{(r)}\|_2 = \mathcal{O}(1), \qquad r\geq l ,
\end{align}
since by combining $
\|\mathcal J^{(l)}\|_2 \leq \gamma$ and \ref{eqn:Delta_h}
\begin{align}
\left\|
\begin{pmatrix}
\Delta_X h^{(r)} \\
\Delta_X \overline{h}^{(r)}
\end{pmatrix}
\right\|_2
&\leq\left(\prod_{s=l+1}^{r}\left\|\mathcal{J}^{(s)}\right\|_2\right)\left\|
\begin{pmatrix}
\Delta_X h^{(l)} \\
\Delta_X \overline{h}^{(l)}
\end{pmatrix}
\right\|_2 =
\mathcal{O}(1).
\end{align}

\vspace{2mm}

\noindent We now establish a backwards perturbation argument. We will establish an argument based on a directional derivative supremum on $E^{(k)}$, which corresponds to the second derivative. This will coincide with the spectral norm on the Hessian, giving us the result. Via the chain rule of \ref{eqn:chain_rule}, consider
\begin{align} 
b^{(r)} & = \left(\mathcal J^{(r+1)}\right)^\dagger b^{(r+1)} 
\\
\implies \Delta_X b^{(r)} & = \left( \Delta_X\mathcal J^{(r+1)} \right)^\dagger b^{(r+1)} + \left( \mathcal J^{(r+1)} \right)^\dagger \Delta_X b^{(r+1)}
\end{align} 
in the direction $X$, which is yielded by product rule. Because the second Wirtinger derivatives $
\partial_h^2\phi,  \partial_h\partial_{\overline h}\phi, \partial_{\overline h}^2\phi$ 
are uniformly bounded, the forward perturbation estimate \ref{eqn:forward_sensitivity}, together with $
\left\| \frac{1}{\sqrt m} W^{(r)} \right\|_2 = \mathcal{O}(1),$ 
implies $
\|\Delta_X\mathcal J^{(r)}\|_2 = \mathcal{O}(1)$.
At the final layer, differentiation of the explicit formulas for $\epsilon^{(L)}$ and $\delta^{(L)}$ gives
\begin{align} 
\|\Delta_X b^{(L)}\|_2 \leq \frac{C}{\sqrt m} \|v\|_2 \|\Delta_X h^{(L)}\|_2 = \mathcal{O}\left(\frac{1}{\sqrt m}\right), 
\end{align} 
where we used $\|v\|_2=1$, bounded second Wirtinger derivatives, and \ref{eqn:forward_sensitivity}. Using $
\|b^{(r+1)}\|_2 = \mathcal{O}\left(\frac{1}{\sqrt m}\right)$,
we can follow through \ref{eqn:bl} to give
\begin{align} 
\label{eqn:backward_sensitivity} 
\|\Delta_X b^{(r)}\|_2 = \mathcal{O}\left(\frac{1}{\sqrt m}\right). 
\end{align} 
In particular, via \ref{eqn:ed} $
\|\Delta_X\delta^{(r)}\|_2 = \mathcal{O}\left(\frac{1}{\sqrt m}\right)$.
We now examine the $(1,1)$ parameter Hessian. For the parameter block $W^{(k)}$, the Wirtinger gradient is 
\begin{align} 
E^{(k)} := \nabla_{\overline W^{(k)}}f(\theta_0) = \frac{1}{\sqrt m} \delta^{(k)} \left( \alpha^{(k-1)} \right)^\dagger. 
\end{align} 
Now, let $X$ be an arbitrary unit perturbation of $W^{(l)}$. Differentiating $E^{(k)}$ in the direction $X$ gives 
\begin{align} 
\label{eqn:Delta_E}
\Delta_X E^{(k)} & = \frac{1}{\sqrt m} \left( \left( \Delta_X\delta^{(k)} \right) \left( \alpha^{(k-1)} \right)^\dagger + \delta^{(k)} \Delta_X \left( \alpha^{(k-1)\dagger} \right) \right)
\end{align}
and then taking the norm
\begin{align}
\|\Delta_X E^{(k)}\|_F &  \leq \frac{1}{\sqrt m} \left( \left\| \Delta_X\delta^{(k)} \right\|_2 \left\| \alpha^{(k-1)} \right\|_2 + \left\| \delta^{(k)} \right\|_2 \left\| \Delta_X \alpha^{(k-1)} \right\|_2 \right) 
\\[0.5em]
& \leq \frac{1}{\sqrt m} \left( \mathcal{O}\left(\frac{1}{\sqrt m}\right) \mathcal{O}(\sqrt m) + \mathcal{O}\left(\frac{1}{\sqrt m}\right) \mathcal{O}(1) \right)  
\\[0.5em]
& = \mathcal{O}\left(\frac{1}{\sqrt m}\right) . 
\end{align} 
If $l\geq k$, then $\alpha^{(k-1)}$ is independent of $W^{(l)}$, so the second term vanishes. If $l<k$, both terms are present, and we bound them. We have used the rank-one identity $
\|ab^\dagger\|_F = \|a\|_2\|b\|_2$, 
together with
\begin{align} 
\begin{cases}
\|\Delta_X\delta^{(k)}\|_2 &= \mathcal{O}\left(\frac{1}{\sqrt m}\right) \\
\|\alpha^{(k-1)}\|_2 &= \mathcal{O}(\sqrt m) \\
\|\delta^{(k)}\|_2 &= \mathcal{O}\left(\frac{1}{\sqrt m}\right) \\
\|\Delta_X\alpha^{(k-1)}\|_2 &= \mathcal{O}(1) .
\end{cases}
\end{align} 
Taking the supremum over $\|X\|_F=1$ therefore gives from \ref{eqn:Delta_E}
\begin{align} 
\label{eqn:H_weight_weight}
\sup_{\| X \|_F = 1} \left\| \Delta_X E^{(k)} \right\|_F = \left\| \mathcal H_{W^{(l)}\overline W^{(k)}} \right\|_2 = \mathcal{O}\left(\frac{1}{\sqrt m}\right)  ,
\end{align} 
for every pair of hidden-layer parameter blocks $l,k$. It remains to examine the blocks involving the output parameter $v$. Since $ \nabla_{\overline v}f =\frac1{2\sqrt m}\alpha^{(L)}$,
we have 
\begin{align} 
\left\| \mathcal H_{W^{(l)}\overline v} \right\|_2 = \frac{1}{2 \sqrt m} \sup_{\|X\|_F=1} \|\Delta_X\alpha^{(L)}\|_2 = \mathcal{O}\left(\frac{1}{\sqrt m}\right).
\end{align} 
Moreover, $ 
\mathcal H_{v\overline v} = 0$,
because the network output is linear in $\overline v$. Thus every block of the $(1,1)$ parameter Hessian is bounded by $
\mathcal{O}\left(\frac{1}{\sqrt m}\right)$. 
Putting everything together, since $L$ is fixed, the spectral norm of the full $(1,1)$ parameter Hessian satisfies 
\begin{align} 
\left\| \nabla_{\theta} \nabla_{\overline\theta} f(\theta_0) \right\|_2 = \Vast\| \begin{pmatrix}
H_{11} & \cdots & H_{1L} & H_{1v} \\
\vdots & \ddots & \vdots & \vdots \\
H_{L1} & \cdots & H_{LL} & H_{Lv} \\
H_{v1} & \cdots & H_{vL} & H_{vv}
\end{pmatrix} \Vast\|_2 = \mathcal{O}\left(\frac{1}{\sqrt m}\right) ,
\end{align} 
where $H_{kl} = \mathcal H_{W^{(l)}\overline{W}^{(k)}}$. This completes the proof.

\noindent $\square$

\section{Convexity results}
\label{app:convexity_results}

\noindent \textit{Proof of Theorem 4.} Denote $L(\theta) = \frac{1}{n} \sum_{i=1}^n \ell_i(y_i,f(\theta; z_i))$. Consider the second-order Riemannian Taylor expansion along the geodesic $\gamma(s)$ connecting $\theta_t$ to $\theta_0$
\begin{align}
L(\theta_0) = L(\theta_t) + 2\text{Re}\langle \nabla_\omega L(\theta_t), \exp_{\theta_t}^{-1}(\theta_0) \rangle_h + \frac{1}{2} \nabla^2_\omega L(\widetilde{\theta}_t)(v, v) ,
\end{align}
where $\widetilde{\theta}_t$ is an intermediary point on the geodesic, $v = \exp_{\theta_t}^{-1}(\theta_0)$, and $\nabla^2_\omega$ denotes the covariant Riemannian Hessian with respect to the Kähler metric $\omega$.

\vspace{2mm}

\noindent Let us bound the Hessian terms. On a Kähler manifold, the covariant Hessian decomposes. The mixed Christoffel symbols vanish. We have the decomposition
\begin{align}
\frac{1}{2} \nabla^2_\omega L(\widetilde{\theta}_t)(v, v) = \underbrace{ \text{Re}\left( v^T \left( \nabla^{2,0}_\omega L(\widetilde{\theta}_t) \right) v \right) }_{= A_1} + \underbrace{ v^\dagger \left( \nabla^{1,1}_\omega L(\widetilde{\theta}_t) \right) v }_{= A_2} .
\end{align}
Applying the covariant chain rule to the composite loss, we observe
\begin{align}
\label{eqn:A_1}
A_1 = \text{Re} \left\{ \frac{1}{n} \sum_{i=1}^n \left[ \underbrace{ \ell''_i(f_i(\widetilde{\theta}_t)) \left( \langle \nabla_\omega f_i(\widetilde{\theta}_t), v \rangle_h \right)^2 }_{= B_1} + \underbrace{ \ell'_i(f_i(\widetilde{\theta}_t)) \left( v^T \nabla^{2,0}_\omega f_i(\widetilde{\theta}_t) v \right) }_{= B_2} \right] \right\} ,
\end{align}
and
\begin{align}
A_2 = \frac{1}{n} \sum_{i=1}^n \left[ \ell''_i(f_i(\widetilde{\theta}_t)) \left| \langle \nabla_\omega f_i(\widetilde{\theta}_t), v \rangle_h \right|^2 + \ell'_i(f_i(\widetilde{\theta}_t)) \left( v^\dagger \nabla^{1,1}_\omega f_i(\widetilde{\theta}_t) v \right) \right] .
\end{align}
Let us examine the term $B_1$ of \ref{eqn:A_1} and the Jacobian term of $A_2$
\begin{align}
\text{Re}(B_1) + A_{2, \text{Jac}} = \frac{2}{n} \sum_{i=1}^n \ell''_i(f_i(\widetilde{\theta}_t)) \left( \text{Re} \left( \langle \nabla_\omega f_i(\widetilde{\theta}_t; z_i), v \rangle_h \right) \right)^2 .
\end{align}
Using a lower bound $\ell_i'' \geq a$ on the loss, we perform the Taylor expansion along the geodesic $\gamma(s)$ parameterized from $s=0$ to $s=1$ connecting $\theta_t$ to $\widetilde{\theta}_t$ for the intermediary $\widetilde{\theta}_t$
\begin{align}
\label{eqn:reB1_A1}
\text{Re}(B_1) + A_{2, \text{Jac}} & \geq \frac{2a}{n} \sum_{i=1}^n \left( \text{Re} \left( \langle \nabla_\omega f_i(\widetilde{\theta}_t), v \rangle_h \right) + \text{Re} \left( \int_{0}^{1} \nabla^2_\omega f_i(\gamma(s))(v, v) ds \right)  \right)^2 
\\
& \geq \frac{2a}{n} \sum_{i=1}^n \left( \text{Re} \left( \langle \nabla_\omega f_i(\theta_t), v \rangle_h\right) \right)^2 - \frac{4a}{n} \sum_{i=1}^n \left| \text{Re} \left( \langle \nabla_\omega f_i(\theta_t), v \rangle_h\right) \right| \vast| \text{Re} \left( \int_{0}^{1} \nabla^2_\omega f_i(\gamma(s))(v, v) ds \right) \vast| .
\end{align}
We have used the fact that $(X + Y)^2 \geq X^2 - 2|X\|Y|$ and that the covariant Hessian is a bilinear form. Let $\mathcal{F}_t(v) = \frac{2}{n} \sum_{i=1}^n \left( \text{Re} \left( \langle \nabla_\omega f_i(\widetilde{\theta}_t), v \rangle_h\right) \right)^2$ be the quadratic form of the empirical Fisher information. Applying the Cauchy-Schwarz inequality to \ref{eqn:reB1_A1}
\begin{align}
\text{Re}(B_1) + A_{2, \text{Jac}} \geq a \mathcal{F}_t(v) - 2 \sqrt{2} a \sqrt{ \mathcal{F}_t(v) } \sqrt{ \frac{1}{n} \sum_{i=1}^n \left( \int_{0}^{1} \nabla^2_\omega f_i(\gamma(s))(v, v) ds \right)^2 } .
\end{align}
Assume that the loss landscape satisfies strong convexity in the local neighborhood $\mathcal{S}$ with parameter $\nu > 0$, such that the Fisher quadratic form is bounded below by $\mathcal{F}_t(v) \geq \nu \|v\|_h^2$. Let the constant $C_{\mathcal{H}}$ be such that the full covariant Hessian norm is bounded by $C_{\mathcal{H}} \|v\|_h^2 / \sqrt{m}$, since the Hessian bound picks up a factor of $m^{-\frac{1}{2}}$. Further, letting $\rho$ bound the maximum eigenvalue of the Fisher matrix such that $\mathcal{F}_t(v) \leq \rho \|v\|_h^2$, we arrive at the bound
\begin{align} 
\text{Re}(B_1) + A_{2, \text{Jac}} \geq \left( a \nu - \frac{2\sqrt{2}a \sqrt{\rho} C_{\mathcal{H}} \|v\|_h }{\sqrt{m}} \right) \|v\|_h^2 . \end{align}
Now, let us consider the $B_2$ term with the Hessian term of $A_2$
\begin{align}
\text{Re}(B_2) + A_{2, \text{Hes}} & = \frac{1}{n} \sum_{i=1}^n \ell'_i(f_i(\widetilde{\theta}_t)) \left( \text{Re} \left[ v^T \nabla^{2,0}_\omega f_i(\widetilde{\theta}_t) v \right] + v^\dagger \nabla^{1,1}_\omega f_i(\widetilde{\theta}_t) v \right) 
\\
\stackrel{\text{Cauchy Schwarz}}{\geq} & - \sqrt{\frac{1}{n} \sum_{i=1}^n (\ell'_i(f_i(\widetilde{\theta}_t)))^2} \sqrt{ \frac{1}{n} \sum_{i=1}^n \left| \text{Re} \left[ v^T \nabla^{2,0}_\omega f_i(\widetilde{\theta}_t) v \right] + v^\dagger \nabla^{1,1}_\omega f_i(\widetilde{\theta}_t) v \right|^2 } 
\\
&  \geq - \sqrt{\frac{1}{n} \sum_{i=1}^n (\ell'_i(f_i(\widetilde{\theta}_t)))^2} \frac{C_{\mathcal{H}}}{\sqrt{m}} \|v\|_h^2 .
\end{align}
Recall we proved in Appendix \ref{app:dolb_hessian_bounds} and Appendix \ref{app:2,0_hess_bounds} the Hessians scale $\mathcal{O}(\frac{1}{\sqrt{m}})$. Combining the bounds
\begin{align}
\nabla^2_\omega L(\widetilde{\theta}_t)(v,v) \geq \underbrace{ 2\left( a \nu  - \frac{ 2\sqrt{2}a \sqrt{\rho} C_{\mathcal{H}} \|v\|_h +  C_{\mathcal{H}} \sqrt{\frac{1}{n} \sum_{i=1}^n (\ell'_i(f_i(\widetilde{\theta}_t)))^2}}{\sqrt{m}} \right)  }_{\displaystyle :=\Gamma\left(a, \nu, \rho, C_{\mathcal{H}}, \{\ell_i'(f_i(\widetilde{\theta}_t))\}_i, m, v\right)  } \|v\|_h^2 .
\end{align}
We also remark that $\theta_0$ lies within a local geodesic ball of radius $D$ centered at $\theta_t$, meaning $\theta_0 \in B_\omega^D(\theta_t)$. Therefore, the geodesic distance is bounded $\|v\|_h = \|\exp_{\theta_t}^{-1}(\theta_0)\|_h \leq D$, although this is not crucial to the above. 

\noindent $ \square $

\vspace{2mm}

\noindent Under a metric collapse, since the collapse implies $\lambda_{\text{min}} = 0$, we get
\begin{align}
\nabla^2_\omega L(\widetilde{\theta}_t) (v,v) \geq \liminf_{\nu \to 0}    \liminf_{\substack{\rho \to \infty \\ C_{\mathcal{H}} \to \infty}}  \Gamma\left(a, \nu, \rho, C_{\mathcal{H}}, \{\ell_i'(f_i(\widetilde{\theta}_t))\}_i, m,v\right)  \|v\|_h^2 = - \infty ,
\end{align}
which destroys the possibility of a finite lower bound. The divergence of constants is because the covariant Hessian relies on the Levi-Civita connection (under choice of connection) and the inverse metric, and a singular metric implies its inverse diverges (under real analytic conventions so that it exists, for example take the limit if appropriate). With a regularized metric, this is not as feasible. In the scenario of a Calabi-Yau manifold, and when $m \rightarrow \infty$, we get
\begin{align}
\nabla^2_\omega L(\widetilde{\theta}_t) (v,v) \geq \liminf_{m \rightarrow \infty} 
\liminf_{\nu \rightarrow 0}
\liminf_{\substack{ \rho \rightarrow r < \infty \\ C_{\mathcal{H}} \to c < \infty }}
\Gamma\left(a, \nu, \rho, C_{\mathcal{H}}, \{\ell_i'(f_i(\widetilde{\theta}_t))\}_i, m,v\right) \|v\|_h^2 > - \infty .
\end{align}
We are slightly informal in our use of the $\liminf$, since we are not particularly examining if the limit exists.

\subsection{$\beta$-smoothness}
\label{app:beta_smoothness}

\textit{Proof of Theorem 5.} By the second-order Riemannian Taylor expansion about $\theta_t$ along the geodesic $\gamma(s)$, we have
\begin{align}
	L(\theta_0) = L(\theta_t) + 2\text{Re} \langle \nabla_\omega^{1,0} L(\theta_t), v \rangle_h + \frac{1}{2} \nabla^2_\omega L(\widetilde{\theta}_t)(v, v),
\end{align}
where $v = \exp_{\theta_t}^{-1}(\theta_0)^{1,0}$. Let us bound the Hessian term. On a Kähler manifold, the covariant Hessian decomposes. The mixed Christoffel symbols vanish. We have the decomposition
\begin{align}
	\frac{1}{2} \nabla^2_\omega L(\widetilde{\theta}_t)(v, v)
	= \underbrace{ \text{Re}\left( v^T \left( \nabla^{2,0}_\omega L(\widetilde{\theta}_t) \right) v \right) }_{= A_1}
	+ \underbrace{ v^\dagger \left( \nabla^{1,1}_\omega L(\widetilde{\theta}_t) \right) v }_{= A_2}.
\end{align}
As before, decompose
\begin{align}
	A_1 = \text{Re} \left\{ \frac{1}{n} \sum_{i=1}^n \left[ \underbrace{ \ell''_i(f_i(\widetilde{\theta}_t)) \left( \langle \nabla_\omega f_i(\widetilde{\theta}_t), v \rangle_h \right)^2 }_{= B_1} + \underbrace{ \ell'_i(f_i(\widetilde{\theta}_t)) \left( v^T \nabla^{2,0}_\omega f_i(\widetilde{\theta}_t) v \right) }_{= B_2} \right] \right\},
\end{align}
and
\begin{align}
	A_2 = \frac{1}{n} \sum_{i=1}^n \left[ \ell''_i(f_i(\widetilde{\theta}_t)) \left| \langle \nabla_\omega f_i(\widetilde{\theta}_t), v \rangle_h \right|^2 + \ell'_i(f_i(\widetilde{\theta}_t)) \left( v^\dagger \nabla^{1,1}_\omega f_i(\widetilde{\theta}_t) v \right) \right].
\end{align}
Let us regroup these terms into first-order and second-order terms. Let us examine the $B_1$ term and the Jacobian term of $A_2$
\begin{align}
	\text{Re}(B_1) + A_{2, \text{Jac}} = \frac{2}{n} \sum_{i=1}^n \ell''_i(f_i(\widetilde{\theta}_t)) \left( \text{Re} \left( \langle \nabla_\omega f_i(\widetilde{\theta}_t), v \rangle_h \right) \right)^2.
\end{align}
For the square loss, $\ell''_i = 2$. The gradient term is reminiscent of a quadratic form via the empirical Fisher information. Assuming a sufficiently nice bound on the gradient over $\mathcal{S}$, and since it is quadratic in $v$, we assume there exists $\rho_J$ so that
\begin{align}
	\text{Re}(B_1) + A_{2, \text{Jac}} \leq \rho_J \|v\|_h^2.
\end{align}
Now, let us consider the $B_2$ term with the Hessian term of $A_2$:
\begin{align}
	\text{Re}(B_2) + A_{2, \text{Hes}} & = \frac{1}{n} \sum_{i=1}^n \ell'_i(f_i(\widetilde{\theta}_t)) \left( \text{Re} \left[ v^T \nabla^{2,0}_\omega f_i(\widetilde{\theta}_t) v \right] + v^\dagger \nabla^{1,1}_\omega f_i(\widetilde{\theta}_t) v \right)
    \\
    & \leq \sqrt{ \frac{1}{n} \sum_{i=1}^n (\ell'_i(f_i(\widetilde{\theta}_t)))^2 } \sqrt{ \frac{1}{n} \sum_{i=1}^n \left| \text{Re} \left[ v^T \nabla^{2,0}_\omega f_i(\widetilde{\theta}_t) v \right] + v^\dagger \nabla^{1,1}_\omega f_i(\widetilde{\theta}_t) v \right|^2 } 
    \\
    & \leq \frac{C_{\mathcal{H}}}{\sqrt{m}} \sqrt{ \frac{1}{n} \sum_{i=1}^n (\ell'_i(f_i(\widetilde{\theta}_t)))^2 }  \|v\|_h^2 .
\end{align}
The first inequality is by Cauchy-Schwarz. Again we invoke the established asymptotic bound on the (1,1)-Hessian in Appendix \ref{app:dolb_hessian_bounds} and (2,0)-Hessian in Appendix \ref{app:2,0_hess_bounds}. Due to the squaring and the bilinear nature, the above is quadratic in $v$. Combining what we have,
\begin{align}
\label{eqn:beta}
\frac{1}{2} \nabla^2_\omega L(\widetilde{\theta}_t)(v, v) \leq \left( \rho_J + \frac{C_{\mathcal{H}} \sqrt{ \frac{1}{n} \sum_{i=1}^n (\ell'_i(f_i(\widetilde{\theta}_t)))^2 } }{\sqrt{m}} \right) \|v\|_h^2.
\end{align}
Since $d_\omega(\theta_0, \theta_t)^2 = 2 \|v\|_h^2$, we can also rewrite the above with this. This completes the proof.

\noindent $\square$

\begin{figure}[!b]
  \centering
  \includegraphics[width=0.45\textwidth]{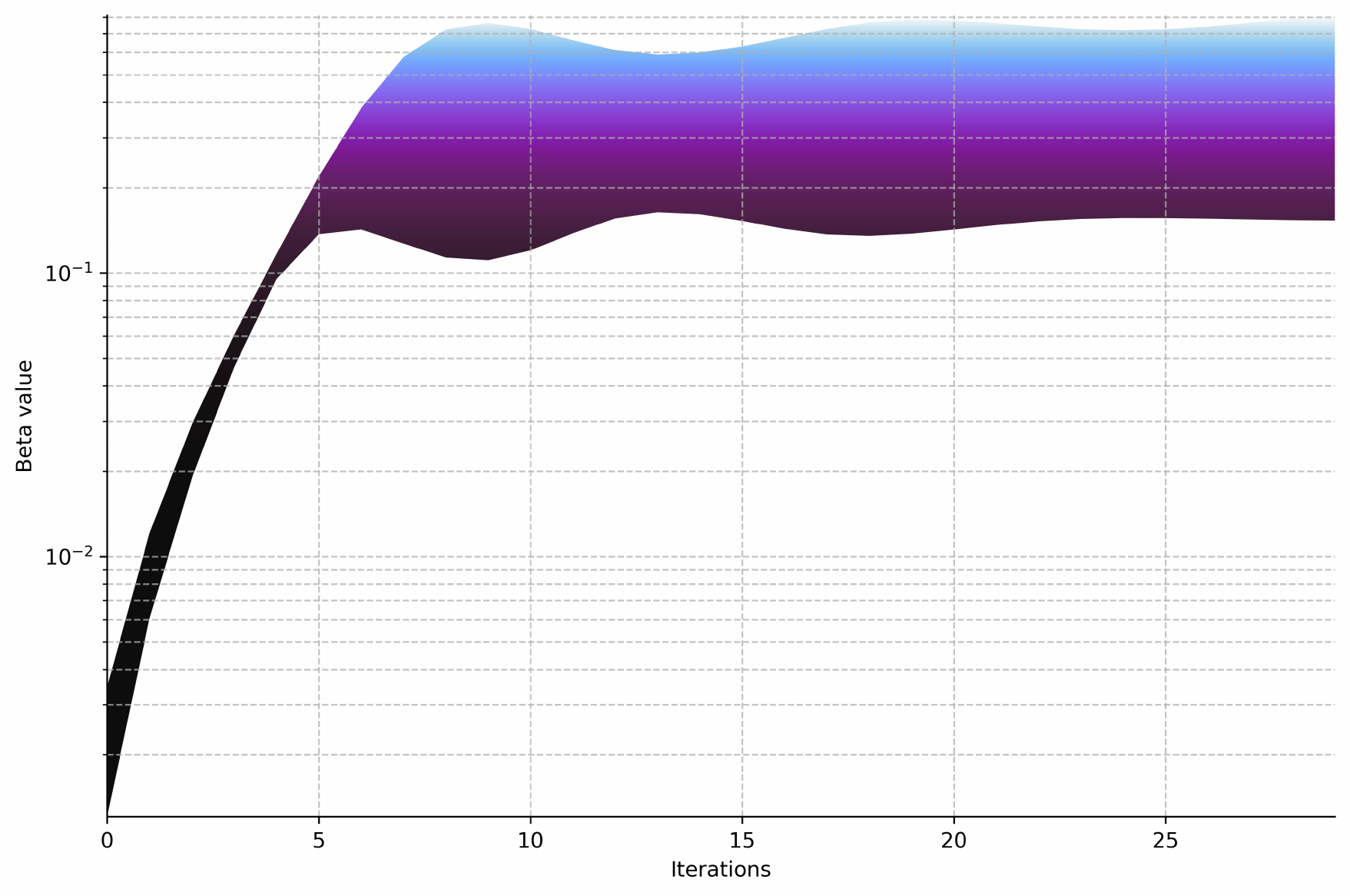}
  \vspace{4mm}
  \caption{We plot $\beta$-smoothness coefficient, $\left( \rho_J + \frac{C_{\mathcal{H}} \sqrt{ \frac{1}{n} \sum_{i=1}^n (\ell'_i(f_i(\widetilde{\theta}_t)))^2 } }{\sqrt{m}} \right)$, across a spectrum of paths ranging from $m=50$ (the upper bound) and $m=800$ (the lower bound).}
  \label{fig:beta_smoothness}
  \vspace{-0mm}
\end{figure}

\vspace{2mm}

In the Calabi-Yau scenario, assuming the minimum eigenvalues have no strict lower bound, we can note the following. Consider the constants in the bound: $\rho_J$, which bounds the first-order Jacobian quadratic form; and $C_{\mathcal{H}}$, which bounds the covariant Hessian norm. In a Calabi-Yau constant determinant condition, both of these will diverge, or at least become very large. Assuming overparameterization and insufficient regularization, we get the divergence case
\begin{align}
\limsup_{ \lambda_{\text{max}} \to \infty}
\beta = \limsup_{\rho_J \to \infty}
\limsup_{ C_{\mathcal{H}} \to \infty}
\left( \rho_J + \frac{C_{\mathcal{H}} \sqrt{ \frac{1}{n} \sum_{i=1}^n (\ell'_i(f_i(\widetilde{\theta}_t)))^2 } }{\sqrt{m}} \right) = \infty 
\end{align}
due to the eigenvalue explosion. If we cohere more closely to the sufficiently regularized loss of \ref{eqn:regularized_loss}, we will assume a lower bound on the minimum eigenvalues, although very small. In this case,
\begin{align}
\limsup_{\lambda_{\text{min}} \to \lambda_{\ll 1} > 0}
\limsup_{ \lambda_{\text{max}} \to \lambda^* \gg 1}
\beta = \limsup_{\rho_J \to \text{very large}}
\limsup_{ C_{\mathcal{H}} \to \text{very large}}
\left( \rho_J + \frac{C_{\mathcal{H}} \sqrt{ \frac{1}{n} \sum_{i=1}^n (\ell'_i(f_i(\widetilde{\theta}_t)))^2 } }{\sqrt{m}} \right) \gg 1  .
\end{align}
Therefore, we get
\begin{align}
\frac{1}{2} \nabla^2_\omega L(\widetilde{\theta}_t)(v, v) \leq \frac{1}{2} \left( \text{explosive value} \right) d_\omega(\theta_0, \theta_t)^2 ,
\end{align}
and the $\beta$-smoothness result will be destroyed.

\subsection{Dynamic Kähler Polyak-Łojasiewicz condition}
\label{app:kahler_PL}

\noindent \textit{Proof of Lemma 4.} Let $U \subseteq \mathcal{S}$ be a local coordinate chart equipped with the Kähler metric $h_{i\overline{j}}$. Let $\theta^*$, be a minima, and let $\gamma(s) = \exp_{\theta_t}(s w)$, $w = \exp_{\theta_t}^{-1}(\theta^*)$ for $s \in [0,1]$ be the geodesic connecting $\theta_t$ to $\theta^*$, with initial holomorphic tangent vector $v  = \exp_{\theta_t}^{-1}(\theta^*)^{1,0} = w^{1,0}$. The second-order Taylor expansion of the loss $L$ along this geodesic is
\begin{align}
\label{eqn:mu_hessian} 
L(\theta^*) &= L(\theta_t) + 2\operatorname{Re}\langle\nabla_\omega^{1,0}L,v\rangle_h \\ &\quad + 2\int_0^1(1-s) \Big[ \mathcal H_{i\overline j}\dot\gamma^i\overline{\dot\gamma^j} + \operatorname{Re}\left( \nabla_i\nabla_jL\dot\gamma^i\dot\gamma^j \right) \Big]ds   .
\end{align}
Define a dynamic $\Gamma_t$ that behaves similarly to the $\Gamma$ of \ref{app:convexity_results} corresponding to the mixed Hessian and the (2,0)-Hessian so that
\begin{align}
\label{eqn:Gamma_t}
\Gamma_t := \inf_{s\in[0,1]} \frac{ \mathcal H_{i\overline j}\dot\gamma^i\overline{\dot\gamma^j} + \operatorname{Re}(\nabla_i\nabla_jL\dot\gamma^i\dot\gamma^j) }{ h_{i\overline j}\dot\gamma^i\overline{\dot\gamma^j} } .
\end{align}
We know $\Gamma_t$ is guaranteed to satisfy a strong convexity result by the result of Appendix \ref{app:convexity_results}, and since we can always at least take
\begin{align}
\Gamma_t \geq \Gamma ,
\end{align}
and attain $\Gamma$ to guarantee it holds. $\gamma$ is contained in $U$, and the result holds for all points (up to smoothness) in $U$, so $\Gamma$ is a worst-case bound. It is not clear that the definition of $\Gamma_t$ can be connected to the result of \ref{app:convexity_results} from definition alone, since the definition looks dissimilar to the definition of $\Gamma$ we had in \ref{app:convexity_results}. Instead, we can notice both  are subsidiary of the fact that we examine $L(\theta^*) \geq L(\theta_t) + 2\text{Re}\langle \nabla_{\omega}^{1,0} L(\theta_t), v \rangle_h + \text{term to be bounded}$. In both cases, we are finding the a sufficient bound on the remaining term, therefore $\Gamma_t, \Gamma$ have the same role in a strong convexity argument. Therefore, we can bound
\begin{align}
2\int_0^1(1-s) \Big[ \mathcal H_{i\overline j}\dot\gamma^i\overline{\dot\gamma^j} &+ \operatorname{Re}( \nabla_i\nabla_jL\dot\gamma^i\dot\gamma^j) \Big]ds \\ &\geq 2\Gamma_t\int_0^1(1-s) \|\dot\gamma(s)^{1,0}\|_h^2 ds \\ &= \Gamma_t\|v\|_h^2   ,
\end{align}
since by definition a geodesic has constant speed $\|\dot{\gamma}(s)^{1,0}\|_h^2 = \|\dot{\gamma}(0)^{1,0}\|_h^2 = \| v \|_h^2$. Substituting back into \ref{eqn:mu_hessian},
\begin{align}
L(\theta^*) & \geq L(\theta_t)
+2\operatorname{Re}\langle\nabla_\omega^{1,0}L,v\rangle_h
+\Gamma_t\|v\|_h^2
\\
\implies L(\theta^*) & \geq 
L(\theta_t)
-\frac{1}{\Gamma_t}\|\nabla_\omega^{1,0}L\|_h^2 +\Gamma_t\left\|
v+\frac{\nabla_\omega^{1,0}L}{\Gamma_t}
\right\|_h^2
\\
& \geq  L(\theta_t)
-\frac{1}{\Gamma_t}\|\nabla_\omega^{1,0}L\|_h^2 .
\end{align}
The second inequality follows by completing the square. We establish the dynamic Kähler-PL condition
\begin{align}
\inf_{\theta \in U} L(\theta) \geq L(\theta_t) - \frac{1}{\Gamma_t} \left\| \nabla_{\omega}^{1,0} L(\theta_t) \right\|_h^2 .
\end{align}

\noindent $ \square $

\vspace{2mm}

\textit{Remark.} We can achieve a similar effect using a Kähler Polyak-Łojasiewicz condition but with lesser restriction, drawing connections to \ref{eqn:semi_positivity}. The definition of $\Gamma_t$ relies on the eigenvalue of the Hessian being positive, but since $\mathcal{H}$ is with respect to $f$, this is not the same as strong convexity. Instead, we examine
\begin{align}
\Gamma_t := \inf_{s \in [0,1]} \frac{ i\Theta_{e^{-L}}(\dot{\gamma}_t(s), J\dot{\gamma}_t(s)) }{ \|\dot{\gamma}_t(s)\|_h^2 } ,
\end{align}
where the metric $e^{-L}$ is a Hermitian metric on the fibers of a trivial complex line bundle $\mathcal{L} = M \times \mathbb{C}$ constructed over $M$ and the Chern curvature form $i\Theta_{e^{-L}}(\mathcal{L}) = i\partial\overline{\partial}L$. We can note
\begin{align}
i\Theta = - i \partial\overline{\partial}\log H  \quad \implies \quad i\Theta_{e^{-L}}(\mathcal{L}) = - i \partial\overline{\partial}\log(e^{-L}) = i\partial\overline{\partial}L .
\end{align}
From this definition of $\Gamma_t$, we have that for  $\mathcal{T} = \text{span}_{\mathbb{C}}\{\dot{\gamma}_t\}$ defining the rank-1 descent subsheaf,  $\Gamma_t \gg 0$ sufficiently positive along $\mathcal{T}$ will satisfy a PL condition. It must be sufficiently positive because of
\begin{align}
\nabla^2_{\text{real}} L = \nabla^{1,1} L + \operatorname{Re}(\nabla^{2,0} L) 
\end{align}
and $\Gamma_t$ must be sufficiently positive to dominate negative curvature from the $(2,0)$ part. We can relax the condition for $\Gamma_t$ to exist, and as long as curvature is sufficiently nice along the geodesic path, we get the same result. We can note the definition of $\Gamma_t^{(m)}$ has connections to $\omega$-$K$-semi-positivity since this is by definition
\begin{align}
i\Theta \wedge \omega^{K-1} \wedge \Omega \geq 0.
\end{align}
We are interested in net curvature being positive along the geodesic.

\subsection{Convergence}
\label{app:convergence}

\noindent \textit{Proof of Lemma 5.} We now analyze the forward step on the loss to guarantee it descends. Let $g_t=\nabla_\omega^{1,0}L(\theta_t)$ be a first derivative and $v_t = -\eta_t \nabla_{\omega}^{1,0} L(\theta_t)$ be a first derivative step with adjustable learning rate $\eta_t$, moving along the geodesic $\sigma(s) =\exp_{\theta_t}(sw_t), w_t=v_t+\overline{v_t}$ to the next parameter state $\theta_{t+1}$. Moreover, note $\theta_{t+1} = \sigma(1)$ and the Taylor expansion of the loss at the updated parameters is
\begin{align}
\label{eqn:beta_smoothness_variable_loss}
L(\theta_{t+1}) = &L(\theta_t) +2\operatorname{Re}\langle g_t,v_t\rangle_h\\ &+2\int_0^1(1-s) \Big[ \mathcal H_{i\overline j}\dot\sigma^i\overline{\dot\sigma^j}\\ &\qquad+\operatorname{Re}\left( \nabla_i\nabla_jL\dot\sigma^i\dot\sigma^j \right) \Big]ds.
\end{align} 
Similarly to $\Gamma_t$, define continually-updated $\beta$-smoothness parameter via 
\begin{align}
\beta_t= \max\left\{ 0,\ \sup_{\substack{s\in[0,1]\\ \xi\in T_{\sigma(s)}^{1,0}\mathcal S,\ \xi\neq0}} \frac{ \mathcal H_{i\overline j}\xi^i\overline\xi^j +\operatorname{Re}(\nabla_i\nabla_jL\xi^i\xi^j) }{ h_{i\overline j}\xi^i\overline\xi^j } \right\}. 
\end{align}
Again, $\beta_t$ is guaranteed to satisfy a $\beta$-smoothness condition because we can always take a worst-case bound
\begin{align}
\beta_t \leq \beta ,
\end{align}
and attain $\beta$ to guarantee the condition, where $\beta$ is as in \ref{app:beta_smoothness}. We can note the geodesic $\sigma$ exists in the $U$ in which $\beta$ is defined, and $\beta$ must hold (almost) everywhere in this region. Bounding the integral term of \ref{eqn:beta_smoothness_variable_loss},
\begin{align}
&2\int_0^1(1-s) \Big[ \mathcal H_{i\overline j}\dot\sigma^i\overline{\dot\sigma^j} +\operatorname{Re}( \nabla_i\nabla_jL\dot\sigma^i\dot\sigma^j) \Big]ds
\\
&\qquad\leq 2\beta_t\int_0^1(1-s) \|\dot\sigma(s)^{1,0}\|_h^2 ds
\\
&\qquad= \beta_t\|v_t\|_h^2
\\
&\qquad= \beta_t\eta_t^2\|g_t\|_h^2.
\end{align}
For the linear term, we use the definition of $v$
\begin{align}
2\operatorname{Re}\langle g_t,v_t\rangle_h =-2\eta_t\|g_t\|_h^2 .
\end{align}
Substituting everything back into \ref{eqn:beta_smoothness_variable_loss},
\begin{align}
L(\theta_{t+1}) \leq L(\theta_t) -\eta_t(2  - \beta_t\eta_t)\|g_t\|_h^2 .
\end{align}
With the Kähler-PL condition of \ref{app:kahler_PL}, we note $\|g_t\|_h^2 \geq \Gamma_t\big(L(\theta_t)-L(\theta^*)\big)$ and we get
\begin{align}
L(\theta_{t+1})-L(\theta^*) \leq{}& \left[ 1-\Gamma_t\eta_t(2-\beta_t\eta_t) \right] \cdot\big(L(\theta_t)-L(\theta^*)\big) .
\end{align}

\noindent $\square$

\subsection{Learning rate deterioration under Calabi-Yau (and sometimes Kähler) blow-up under nondegeneracy}
\label{app:calabi_yau_oscillation}

\textit{Proof of Lemma 6.} Let $(M,\omega)$ be a Kähler information manifold under the regularized loss, and fix a background Kähler metric $h_0$. Define the relative metric endomorphism
$H := h_0^{-1}h.$ Let $\lambda_{\min}(H) = \Theta(\lambda),$
and consider the assumed Calabi-Yau eigenvalue blow-up scaling $ \lambda_{\max}(H) = \Omega\left( \frac{\kappa}{\mu^{\widetilde K-1}} \right),$ where $\det H \geq \kappa > 0$ and $0 \ll \widetilde K \leq K$.

\vspace{2mm}

Fix a point in this blow-up region and choose $h_0$-normal holomorphic coordinates at that point. Thus, $
h_0 = I,\Gamma(h_0) = 0$, and
\begin{align}
A^m_{ik}=(H^{-1})^m_{\ \ell}\partial_iH^\ell_{\ k} .
\end{align}
Generally our results in Appendix \ref{app:dolb_cov_bound} are with respect to $h$, but here we use $H$, thus we need to adjust. Hence, for a scalar network output $f$, the $(2,0)$-covariant Hessian of Appendix \ref{app:2,0_hess_bounds} satisfies
\begin{align}
\nabla_{\omega}^{2,0}f(u,u) = \partial^2f(u,u) - u^iu^k(H^{-1})^m_{\ \ell} (\partial_iH^\ell_{\ k})\partial_mf.
\end{align}
Applying the reverse triangle inequality gives
\begin{align}
\label{eqn:reverse_triangle}
\left|\nabla_{\omega}^{2,0}f(u,u)\right| \geq \left| u^iu^k(H^{-1})^m_{\ \ell} (\partial_iH^\ell_{\ k})\partial_mf \right| - |\partial^2f(u,u)|.
\end{align}
Choose an $h$-unit vector $u$ in a minimum-eigenvalue direction of $H$. Since $\lambda_{\min}(H)=\Theta(\lambda)$,
\begin{align}
\label{eqn:H_lmin}
\|u\|_{h_0}^2=\Theta(\frac{1}{\lambda}),\qquad\|H^{-1}\|_2=\Theta(\frac{1}{\lambda}).
\end{align}
Note as in \ref{app:cohomology} $
h \approx \mathbb E\left[(\partial f)^\dagger\partial f\right] + \lambda h_0$ and also $\|\partial f\|_2^2 = \Omega\left(\lambda_{\max}(H)\right) = \Omega\left( \kappa\mu^{-(\widetilde K-1)} \right)$. Borrowing from \ref{app:2,0_hess_bounds}, \ref{app:dolb_hessian_bounds}, we make the assumption, also consistent with Figures \ref{fig:spectral_2,0hessian_scaling}, \ref{fig:spectral_hessian_scaling}
\begin{align}
\|\partial^2f\|_2 = \Theta(\frac{1}{\sqrt{m}}) .
\end{align}
We differentiate $h$. Via product rule, omitting expectation $
\partial h \sim (\partial^2f)^\dagger\partial f + (\partial f)^\dagger\partial^2f$. We obtain a lower bound. We invoke the nondegeneracy assumption. Thus, for some constant $c_0 > 0$, and substituting in \ref{eqn:H_lmin} and via assumption
\begin{align}
\left| u^iu^k(H^{-1})^m_{\ \ell} (\partial_iH^\ell_{\ k})\partial_mf \right| & \geq c_0 \|u\|_{h_0}^2 \|H^{-1}\|_2 \|\partial^2f\|_2 \|\partial f\|_2^2
\\
\label{eqn:christoffel_cont}
&= \Omega\left( \lambda^{-1}\lambda^{-1}\frac{1}{\sqrt{m}} \kappa\mu^{-(\widetilde K-1)} \right) \\ 
&= \Omega\left( \frac{\kappa}{\lambda^2 \sqrt m}  \mu^{-(\widetilde K-1)} \right).
\end{align}
We have used the nondegeneracy assumption in the first inequality. We next compare this with the ordinary second-derivative contribution. Notice
\begin{align}
|\partial^2f(u,u)| &\leq \|u\|_{h_0}^2\|\partial^2f\|_2 = \mathcal O\left( \frac{1}{\lambda \sqrt m} \right).
\end{align}
Thus the term \ref{eqn:christoffel_cont} has $\lambda^{-2} \mu^{-(\widetilde K-1)}$ scaling, whereas the ordinary Hessian contribution scales only as $\lambda^{-1}$. Therefore, \ref{eqn:christoffel_cont} dominates, and we obtain
\begin{align}
\sup_{\|u\|_h=1} \left| \nabla_{\omega}^{2,0}f(u,u) \right| & = \Omega\left( \frac{\kappa}{\lambda^2 \sqrt m} \mu^{-(\widetilde K-1)} \right) \leq \frac{C_{\mathcal{H}}}{\sqrt m}.
\end{align}
Consequently,
\begin{align}
C_{\mathcal{H}} = \Omega\left( \frac{\kappa}{\lambda^2} \mu^{-(\widetilde K-1)} \right).
\end{align}
For the square loss, the previously established $\beta$-smoothness scalar has the form \ref{eqn:beta}
\begin{align}
\beta & = \rho_J + \frac{2C_{\mathcal{H}}}{\sqrt m} \sqrt{L(\widetilde\theta_t)}.
\\
& = \Omega\left( \frac{\kappa}{\lambda^2 \sqrt m} \mu^{-(\widetilde K-1)} \sqrt{L(\widetilde\theta_t)} \right).
\end{align}
Finally, from Appendix \ref{app:convergence}, the one-step descent estimate is
\begin{align}
L(\theta_{t+1}) - L(\theta^*) \leq \left( 1 - 2\Gamma_t\eta_t + \Gamma_t\beta\eta_t^2 \right) \left( L(\theta_t) - L(\theta^*) \right).
\end{align}
The contraction coefficient factors as
\begin{align}
1 - 2\Gamma_t\eta_t + \Gamma_t\beta\eta_t^2 = 1 - \Gamma_t\eta_t(2 - \beta\eta_t).
\end{align}
Therefore, for a positive learning rate, a sufficient condition for the contraction coefficient to remain strictly below $1$ is
\begin{align}
0 < \eta_t < \frac{2}{\beta}.
\end{align}
Using the preceding lower-bound scaling of $\beta$, the admissible learning rate must satisfy
\begin{align}
\eta_t = \mathcal O\left( \frac{\lambda^2 \sqrt m} {\kappa\sqrt{L(\widetilde\theta_t)}} \mu^{\widetilde K-1} \right).
\end{align}
Thus, for fixed $0 < \mu < 1$, the admissible learning rate scale deteriorates rapidly as $\widetilde K$ increases. Although the regularization prevents exact metric collapse, the Calabi-Yau determinant constraint together with the assumed eigenvalue blow-up regime can drive the $\beta$-smoothness parameter to a large value and correspondingly shrink the learning-rate.

$ \square $

\begin{figure}[htbp]
  \centering
  \includegraphics[width=0.85\textwidth]{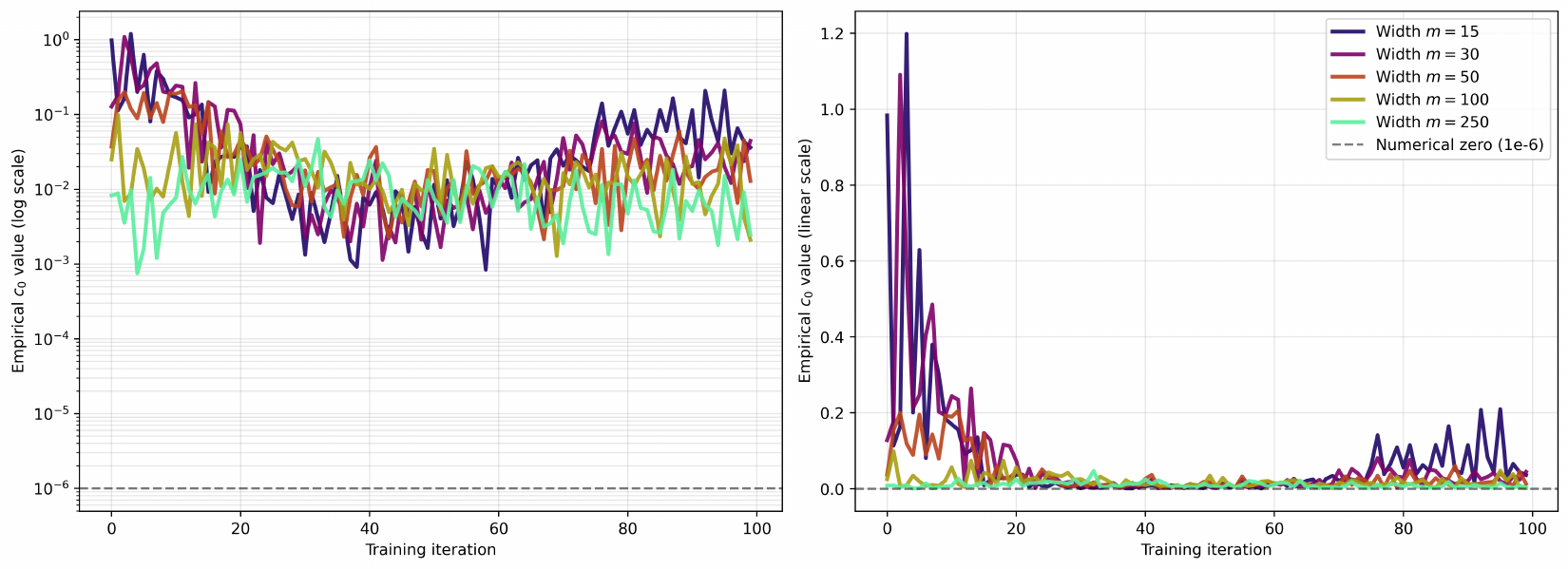}
  \vspace{4mm}
  \caption{We plot $c_0$ coefficient $c_0 = \frac{\left| u^i u^k (H^{-1})^m_{\ \ell} (\partial_i H^\ell_{\ k}) \partial_m f \right|}{\|u\|_{h_0}^2 \|H^{-1}\|_2 \|\partial^2 f\|_2 \|\partial f\|_2^2}$ in an empirical setting of varying width. We plot on a log scale (left) and a linear baseline scale (right). If $c_0 = 0$, then the bound would not hold. Alternatively, $c_0$ sufficiently nonzero means the bound holds. Note $10^{-6}$ is the zero threshold for floating point calculation. We remark the code for this figure took a long time to run (almost an hour).}
  \label{fig:c0_nondegeneracy}
  \vspace{-0mm}
\end{figure}

\subsection{Failure of the Kähler Polyak-Łojasiewicz condition under Calabi-Yau and lower-bounded Ricci potential metrics}
\label{app:KPL_failure}

In this section, along with \ref{app:calabi_yau_oscillation}, we also demonstrate the failure of the results of \ref{app:kahler_PL} via an eigenvalue blow-up effect of Calabi-Yau manifolds. This result is unique to eigenvalue blow-up, hence Calabi-Yau metrics, and not inherent to negative curvature.

\vspace{2mm}

\noindent \textit{Proof of Lemma 7.} Recall the definition of the Kähler Polyak–Łojasiewicz parameter
\begin{align}
\Gamma_t :=\inf_{s\in[0,1]} \frac{ \mathcal H_{i\overline j}\dot\gamma^i\overline{\dot\gamma^j} +\operatorname{Re}(\nabla_i\nabla_jL\dot\gamma^i\dot\gamma^j)
}{
h_{i\overline j}\dot\gamma^i\overline{\dot\gamma^j}}.
\end{align}
Define the directional Hessian quadratic form
\begin{align}
\mathcal Q_L(\xi)= \mathcal H_{i\overline j}\xi^i\overline{\xi^j} +\operatorname{Re}(\nabla_i\nabla_jL\xi^i\xi^j).
\end{align}
Rewriting this using our background metric $H = h_0^{-1}h$, we can write
\begin{align}
\Gamma_t= \inf_{s\in[0,1]} \frac{ \mathcal Q_L(\dot\gamma(s)) }{ \langle H\dot\gamma(s),\dot\gamma(s)\rangle_{h_0} }.
\end{align}
We invoke the assumption on the relation between the quadratic form and maximum eigenvalue: at some $s_0\in[0,1]$ within the eigenvalue blow-up region, there exists a constant $c_0>0$ such that
\begin{align}
\langle H\dot\gamma,\dot\gamma\rangle_{h_0}
\geq c_0\lambda_{\max}(H)\|\dot\gamma\|_{h_0}^2 .
\end{align}
From our previous asymptotic scaling results near initialization, we can note the (1,1), (2,0)-Hessians as in \ref{app:dolb_hessian_bounds}, \ref{app:2,0_hess_bounds}  scale as $\frac{1}{\sqrt{m}}$, but we do not have this use exactly here. We use by assumption\footnote{Note that we have defined $\mathcal{Q}_L$ with only the numerator of $\Gamma_t$, so we can ignore asymptotics on $h$ in this assumption. We assume a relation between $f$ and $L$ to hold so the asymptotic relation holds. Note that $L$ is defined with $f$. For quadratic loss, an $\mathcal{O}(\frac{1}{\sqrt{m}})$ assumption is not reasonable. In particular, a quadratic loss leads to the term $\nabla f \otimes \nabla f$.}
\begin{align}
\mathcal Q_L(\dot\gamma(s_0))
\leq C\|\dot\gamma(s_0)\|_{h_0}^2 
\end{align}
near initialization. Consequently,
\begin{align}
\Gamma_t &\leq \frac{\mathcal Q_L(\dot\gamma(s_0))} {\langle H\dot\gamma(s_0),\dot\gamma(s_0)\rangle_{h_0}}
\\
&\leq \frac{C}{c_0 \underbrace{ \lambda_{\max}(H(s_0)) }_{=\Omega\left(\kappa\mu^{-(\widetilde K-1)}\right)}}
\\
& \leq\frac{\mathcal{O}(1)}{\Omega(\kappa \mu^{-(\widetilde K - 1)})} 
\\
& = \mathcal{O}\left( \frac{\mu^{\widetilde K - 1}}{\kappa } \right) . 
\end{align}
By the eigenvalue bound of \ref{eqn:mu}, $0 < \mu \ll 1$ is small but $\mu \nrightarrow 0$ and $\widetilde{K}$ is large, especially in the overparameterization regime, so $\Gamma_t$ is very small. In fact, $\Gamma_t$ is converging to $0$ as $\widetilde{K}$ grows but not because $\mu$ goes to $0$. Therefore, when examining
\begin{align}
L(\theta_{t+1}) - L(\theta^*) \leq (1 - \Gamma_t \eta_t(2 - \beta_t \eta_t))(L(\theta_t) - L(\theta^*)) ,
\end{align}
the constant on the right-hand side is $\sim 0.9999999 \hdots$ for large $K$, and so the convergence guarantee is diminished in effect.

\noindent $\square$

\section{Additional discussion with cohomologies, eigenvalues, and Jacobians}
\label{app:cohomology}

\vspace{2mm}

\noindent \textbf{Jacobians, rank deficiencies and sufficiencies in cohomology, and relations to Theorem 1.} Let us consider the overparameterization regime with a full rank metric, and moreover consider the network's prediction as a morphism of sheaves. We will consider sheaves of sections but also vector bundles because we will assume the Jacobian is full rank. We will rewrite the metric using Jacobians. Recall the metric is full rank by \ref{sec:regularization}. Also note we will take an expected value in \ref{eqn:jacobian}, but expected values of full rank matrices are not necessarily full rank (almost surely); however, our matrix is positive semi-definite, and so this matters\footnote{The expected value of a low-rank positive semi-definite matrix can actually be full rank via examining $\ker(\mathbb{E}[A]) = \bigcap_{\omega \in \text{supp}(A)} \ker(A(\omega))$.}. Since $J$ is full rank, the kernel $\mathcal{V}$ is a valid vector bundle; however, we will do cohomological algebra on these modules, and note that vector bundles are not an abelian category. Also note by rank-nullity,
\begin{align}
\dim(\ker(J_\theta)) = K - \text{rank}(J_\theta) .
\end{align}

\vspace{2mm}

Considering sheaves, let $\mathcal{E}_M^{\oplus K}$ be the sheaf of smooth complex-valued sections\footnote{We consider the complexified tangent sheaf because $f$ is real-valued (hence non-holomorphic). We want to use Wirtinger derivatives on a non-holomorphic function, so we will consider $\partial f$ and $\overline{\partial} f$ (see \ref{eqn:df}).} of the trivial vector bundle of all parameters; and $\mathcal{E}_M^{\oplus Nk}$ be the sheaf of sections of the trivial bundle of network outputs over the dataset. Note considering the trivial bundle has nothing to do with the rank of the Jacobian.

\vspace{2mm}

The complexified network Jacobian $J = \partial f(z; \theta)$ is a morphism of sheaves
\begin{align}
J: \mathcal{E}_M^{\oplus K} \longrightarrow \mathcal{E}_M^{\oplus Nk} ,
\end{align}
since the Jacobian admits a linear/matrix representation mapping from $\mathcal{E}_M^{\oplus K}$ to $\mathcal{E}_M^{\oplus Nk}$. We note on the zero-loss manifold, $J$ is surjective because of overparameterization. Note this is actually nuanced because overparameterization is necessary but not sufficient to be surjective, for example because it can be low rank. Without loss of generality, assume it is surjective.

\vspace{2mm}

We have a short exact sequence of sheaves $
0 \longrightarrow \mathcal{V} \xrightarrow{\iota} \mathcal{E}_M^{\oplus K} \xrightarrow{J} \mathcal{E}_M^{\oplus Nk} \longrightarrow 0 $. We use $\mathcal{E}_M^{\oplus Nk}$ and not $\text{Im}(J) \subseteq \mathcal{E}_M^{\oplus Nk}$ because any map is surjective on its own image, so that would be tautological. Here, $\mathcal{V} = \ker(J)$ is the subsheaf that defines the flat minima. Note that the fibers of the associated vector bundle represent the flat minima, being the directions in parameter space where the information metric $h = \EX [ J^{\dagger} J ]$ has zero eigenvalues. Geometrically, this is the tangent space of the zero-loss manifold. In particular, a fiber is the vector space
\begin{align}
\label{eqn:fiber}
V_\theta = \{ \delta\theta \in \mathbb{C}^K \ | \ \EX \ \|J_\theta \delta\theta\|^2 = 0 \}   .
\end{align}
$h$ takes this form since $-\log p(y|z, \theta) = \frac{1}{2} \sum_k (f_k(z;\theta) - y_k)\overline{(f_k(z;\theta) - y_k)} + C$. Expanding the second derivative yields 
\begin{align}
\partial_i \partial_{\overline{j}} \left[ (f-y)\overline{(f-y)} \right] = (\partial_i f)(\partial_{\overline{j}} \overline{f}) + (\partial_{\overline{j}} f)(\partial_i \overline{f}) + (f-y) \partial_i \partial_{\overline{j}} \overline{f} + \overline{(f-y)} \partial_i \partial_{\overline{j}} f .
\end{align}
When taking the expectation over $y \sim q(y | z)$ as in \ref{sec:geometry}, the residual $(f-y)$ is mean-zero, causing the second-order derivative terms to vanish. Therefore, 
\begin{align}
\label{eqn:jacobian}
h_{i\overline{j}}(\theta) = \mathbb{E}_{z \sim p_{\text{data}}} \left[ \mathbb{E}_{y \sim q(y|z)} \left[ (J^\dagger J)_{i\overline{j}} \right] \right].
\end{align}
The fiber as in \ref{eqn:fiber} corresponds to minima since for a specific eigenvector $v \in V_\theta$, $h v = \mathbb{E}[J^\dagger J] v = \mathbb{E}[J^\dagger (J v)] = \mathbb{E}[0] = 0$. In other words, since the prediction does not change, it corresponds to flat directions tangent to the minimized-loss manifold.

\vspace{2mm}

Consider neural network prediction $f = (f_1, \dots, f_{Nk}) : M \to \mathbb{R}^{Nk}$ is a smooth, real-valued map as in \ref{sec:network_setup}, and $f$ is not holomorphic. Therefore, the real differential acts as a bundle map on the real tangent bundle $T_{\mathbb{R}}M$, which has rank $2K$
\begin{align}
\label{eqn:df}
df: T_{\mathbb{R}}M \longrightarrow \underline{\mathbb{R}}^{Nk}, \qquad df_\theta(v) = \big((df_1)_\theta(v),\dots,(df_{Nk})_\theta(v)\big) .
\end{align}
Assuming $df$ maintains constant rank $r = Nk$ across the zero-loss locus due to overparameterization, the flat directions form a smooth real vector subbundle $\mathcal{V} := \ker_{\mathbb{R}}(df) \subseteq T_{\mathbb{R}}M$. Let us take the sheaf of smooth sections $\mathcal{E}_M$, and so we obtain a short exact sequence of sheaves of smooth real vector bundles using \ref{eqn:df}
\begin{align}
0 \longrightarrow \mathcal{E}_M(\mathcal{V}) \longrightarrow \mathcal{E}_M(T_{\mathbb{R}}M) \ \xrightarrow{\ df\ } \ \mathcal{E}_M(\underline{\mathbb{R}}^{Nk}) \longrightarrow 0 .
\end{align}
The sheaf of smooth sections of any real vector bundle admits partitions of unity. We refer to sheaf theory, and so this property makes $\mathcal{E}_M$ a fine sheaf \cite{bennequin2020extrafinesheavesinteractiondecompositions}, and fine sheaves are acyclic \cite{zein2014sheafcohomologyalgebraicrham}, meaning $H^i(M; \mathcal{E}_M(\cdot)) = 0$ for all $i > 0$. Because $H^1(M; \mathcal{E}_M(\mathcal{V})) = 0$ automatically, the long exact sequence in cohomology splits trivially. This yields a surjective map on the global sections with no (non-vanishing) connecting homomorphism or higher cohomological obstructions \cite{Deligne1975} \cite{mishra2025hermitianyangmillsconnectionsgeneral} 
\begin{center}
\label{eqn:cohomology}
\makebox[\linewidth][c]{
\begin{tikzcd}[ampersand replacement=\&, column sep=small, row sep=small]
0 \arrow[r]
\& H^0(M; \mathcal{E}_M(\mathcal{V})) \arrow[r]
\& H^0(M; \mathcal{E}_M(T_{\mathbb{R}}M)) \arrow[r, "df_*"]
\& H^0(M; \mathcal{E}_M(\underline{\mathbb{R}}^{Nk}))
  \arrow[dll, blue, "!\delta" description, out=-10, in=170, looseness=1.5] \\
\& H^{i \geq 1}(M; \mathcal{E}_M(\cdot)) = 0 \arrow[r]
\& \dots 
\end{tikzcd}
}
\end{center}

\vspace{2mm}

\noindent We previously discussed how overparameterization can cause degeneracy, i.e. in \ref{sec:corrupted_geometries}. Note the subsheaf $\mathcal{V} = \ker(J)$ captures (at least) two distinct sources of degeneracy: from overparameterization (recall the Jacobian is still full rank; necessary but not sufficient), but also from symmetries belonging to neural network architectures (see de Rham cohomology section below). As an aside example, modReLU networks with activations $f(z) = \sigma(|z|)\footnote{$\sigma$ is a nonlinear activation.} e^{i \arg(z)}$ (ReLU has no canonical existence in the complex numbers, although complex ReLU can be defined) have an equivariance quality
\begin{align}
\label{eqn:relu_symmetry}
f(cz) = f(e^{i\phi}z) = \sigma(|e^{i\phi}z|) e^{i \arg(e^{i\phi}z)} = \sigma(|z|) e^{i (\arg(z) + \phi)} = e^{i\phi} \big( \sigma(|z|) e^{i \arg(z)} \big) = c f(z)   .
\end{align}
We remark the above is done with $z$ and not $\theta$. It is a bit of an aside to our de Rham cohomology arguments, but a similar analogy holds via \ref{eqn:widetilde_output}, so it is still relevant. In particular, we will elaborate more on why a symmetry property is useful in the context of de Rham cohomology. Moreover, the existence of the nontrivial subsheaf $\mathcal{V}$ provides the basis for Theorem 1. The unregularized information metric is constructed via $h = \EX J^\dagger J$ and is degenerate along the fibers associated to $\mathcal{V}$. This degeneracy prevents $h$ from being positive-definite, and recall from the remark of Theorem 1 it therefore cannot be Calabi-Yau.

\vspace{2mm}

\noindent \textbf{Near-zero spectral gap under regularization.} In this section, we will establish an upper bound on the eiegnvalues. This is notable because we previously worked with lower bounds throughout this work. Some of our results are driven by upper bounds on eigenvalues too (see Appendix \ref{app:min_eigenvalue_bounds}). We can note the cohomological vanishing in \ref{eqn:cohomology} holds because $\mathcal{E}_M(\mathcal{V})$ is a fine sheaf, independent of any metric. Let us consider the regularized loss of \ref{eqn:regularized_loss}, and this loss does not modify this property: for every $\lambda > 0$, $H^1(M; \mathcal{E}_M(\mathcal{V})) = 0$ still holds by the same argument. Regularization changes is not the cohomology but the eigenvalues of the information metric. We saw this in \ref{sec:regularization}, and we elaborate further on this here. Note $M$ is non-compact, as we outlined in the beginning of section \ref{sec:calabi_yau}. 

\vspace{2mm}

Denote $\delta$ on $\mathcal{E}_M^{\oplus K}$ the nondegenerate baseline, i.e. oftentimes the flat Euclidean metric under the loss of \ref{eqn:regularized_loss}, so
\begin{align}
h_\lambda(\theta) := h(\theta) + \lambda \delta(\theta), \quad \lambda > 0 .
\end{align}
Since $\delta$ is positive-definite and $h$ corresponds to that unregularized and is positive semi-definite, $h_\lambda$ is positive-definite for every $\lambda > 0$, and $h_\lambda \to h$ as $\lambda \to 0$. Since $V_\theta = \ker h(\theta)$ and $h(\theta)$ is Hermitian positive semi-definite, the orthogonal projector $\Pi_\theta$ onto $V_\theta$ satisfies $h(\theta)\Pi_\theta = \Pi_\theta h(\theta) = 0$, i.e. $h(\theta) = \Pi_\theta^\perp h(\theta) \Pi_\theta^\perp$ where $\Pi_\theta^\perp = I - \Pi_\theta$. Writing $h_\lambda(\theta)$ in block form with respect to the orthogonal decomposition $\mathbb{C}^K = V_\theta \oplus V_\theta^\perp$,
\begin{align}
h_\lambda(\theta) =
\begin{pmatrix}
\lambda\Pi_\theta \delta(\theta) \Pi_\theta & \lambda\Pi_\theta \delta(\theta) \Pi_\theta^\perp \\[4pt]
\lambda\Pi_\theta^\perp \delta(\theta) \Pi_\theta &  h(\theta)\big|_{V_\theta^\perp} + \lambda\Pi_\theta^\perp \delta(\theta) \Pi_\theta^\perp
\end{pmatrix}_{V_\theta \oplus V_\theta^\perp} .
\end{align}
At each $\theta$ on the zero-loss manifold, both are Hermitian forms on the same finite-dimensional fiber $\mathbb{C}^K$, so we may compare their eigenvalues directly via the Courant-Fischer minimax characterization \cite{meng2025combinatorialcourantfischerweylminimaxprinciple}. Writing $\lambda_1(\theta) \leq \cdots \leq \lambda_K(\theta)$ for the eigenvalues of $h(\theta)$ and $\lambda_1^\lambda(\theta) \leq \cdots \leq \lambda_K^\lambda(\theta)$ for those of $h_\lambda(\theta)$, Weyl's inequality gives
\begin{align}
\lambda_i(\theta) \leq \lambda_i^\lambda(\theta) \leq \lambda_i(\theta) + \lambda \|\delta(\theta)\|_{2} .
\end{align}
Recall from \ref{eqn:fiber} that $V_\theta = \ker h(\theta)$ has dimension $d(\theta) := \dim_{\mathbb{C}} V_\theta$, so $\lambda_1(\theta) = \cdots = \lambda_{d(\theta)}(\theta) = 0$. The inequality above then forces
\begin{align}
0  < \lambda_i^\lambda(\theta) \leq \lambda \|\delta(\theta)\|_{2} \quad \text{for } i = 1, \hdots, d(\theta) ,
\end{align}
so the bottom $d(\theta)$ eigenvalues of $h_\lambda$ vanish linearly in $\lambda$, uniformly on any compact subset of $M$. 

\vspace{2mm}

\noindent \textbf{Parameter holes and connections to de Rham cohomology.} As illustrated in \ref{eqn:relu_symmetry}, neural networks can possess intrinsic qualities and symmetries. Specifically, incoming and outgoing weights can be rotated by a phase without altering the network's output or the loss. Let us consider a $U(1)$ transformation on the parameter landscape, so we define the map $(w_1, w_2) \mapsto (e^{i\phi} w_1, e^{-i\phi} w_2)$ where $w_1,w_2$ are parameter vectors. Notice $z$ transforms as $z \mapsto e^{i\phi}z$. The neuron output is
\begin{align}
\label{eqn:widetilde_output}
\widetilde{\text{output}} = (e^{-i\phi} w_2) \sigma(e^{i\phi} z) \in \mathbb{C}^m ,
\end{align}
and the exponentials cancel when $\sigma(e^{i\phi} z) = e^{i\phi} \sigma(z)$, which is reasonable for certain complex neural network activations. Induced due to the symmetry is the group $U(1) \cong S^1$. If we consider the parameter plane of a complex weight, the origin where the phase becomes undefined is a puncture. De Rham cohomology is a tool to detect holes in the space \cite{petrov2024essencerhamcohomology}, since integrating an exact form around a closed path is zero. Let $\varphi$ represent the angular coordinate wrapping around this puncture. We have a nontrivial cohomology class $[d\varphi] \in H_{\text{dR}}^1(M; \mathbb{R})$, and so we develop
\begin{center}
\label{eqn:derham_mayer_vietoris}
\makebox[\linewidth][c]{
\begin{tikzcd}[ampersand replacement=\&, column sep=small, row sep=small]
0 \arrow[r]
\& H^0_{\text{dR}}(M; \mathbb{R}) \arrow[r]
\& H^0_{\text{dR}}(U_1; \mathbb{R}) \oplus H^0_{\text{dR}}(U_2; \mathbb{R}) \arrow[r, "\text{res}"]
\& H^0_{\text{dR}}(U_1 \cap U_2; \mathbb{R})
  \arrow[dll, blue, "d^*" description, out=-10, in=170, looseness=1.5] \\
\& H^1_{\text{dR}}(M; \mathbb{R}) \arrow[r]
\& H^1_{\text{dR}}(U_1; \mathbb{R}) \oplus H^1_{\text{dR}}(U_2; \mathbb{R}) = 0 \arrow[r]
\& \dots 
\end{tikzcd}
}
\end{center}
for two overlapping, simply connected open sets $U_1$ and $U_2$, $U_1 \cap U_2$ disconnected. By contrast, applying the Hodge star maps the angular form to the exact radial form $\star d\varphi = -d(\log r)$. Therefore, integrating over a loop $\gamma$ and a closed contour $\Sigma$ enclosing the singularity yields
\begin{align}
\oint_{\gamma} d\varphi = 2\pi \neq 0, \quad \oint_\Sigma \star d\varphi = \oint_\Sigma -d(\log r) = 0 .
\end{align}

\section{First derivative norm bounds and roles of negative curvature}

\subsection{Regret bounds}
\label{app:regret_bounds}

\textit{Proof of Lemma 8.} We can note $\Big\|\exp_{\theta_t}^{-1}(\theta^*)^{1,0}\Big\|_h$ is the geodesic distance $d_\omega(\theta_t, \theta^*)$. Let us denote
\begin{align}
\delta(\theta) := \frac{1}{2} d_\omega(\theta, \theta^*)^2 .
\end{align}
The Riemannian gradient can be computed as
\begin{align}
\label{eqn:s_deriv}
\nabla_h \delta(\theta) = -\exp_\theta^{-1}(\theta^*)^{1,0} .
\end{align}
Under stochastic natural gradient descent, the parameter updates $d\theta_t=-\nabla_h\mathcal L(\theta_t)dt +\sqrt{\eta}h(\theta_t)^{-1}\Sigma^{1/2}dW_t$. By Itô's lemma, the differential of $\delta$ can be written as
\begin{align}
d\delta(\theta_t)=-\text{Re}\langle\nabla_h\delta(\theta_t),\nabla_h\mathcal L(\theta_t)\rangle_h dt+dM_t+\mathscr D_\Sigma\delta(\theta_t)dt ,
\end{align}
where $M_t$ is the corresponding Itô martingale and where $\mathscr D_\Sigma$ denotes the second-order diffusion part. Substituting in \ref{eqn:s_deriv} and the parameter update rule,
\begin{align}
d\delta(\theta_t)=\text{Re}\langle\nabla_h\mathcal L(\theta_t),\exp_{\theta_t}^{-1}(\theta^*)^{1,0}\rangle_h dt+dM_t+\mathscr D_\Sigma\delta(\theta_t)dt .
\end{align}
The first term here is found in the regret term of \ref{eqn:regret}. Integrating,
\begin{align}
\label{eqn:s_difference}
\delta(\theta_T) - \delta(\theta_0) & = \underbrace{ \int_0^T \text{Re} \langle \nabla_h \mathcal{L}(\theta_t), \exp_{\theta_t}^{-1}(\theta^*)^{1,0} \rangle_h dt }_{= -\mathcal{R}(T)} 
\\
& \quad \quad \quad \quad - \underbrace{ \int_0^T dM_t }_{  \EX[M_T] =  0 } + \int_0^T \mathscr D_\Sigma\delta(\theta_t)dt  .
\end{align}
Because an Itô integral with respect to Brownian motion is a martingale, we get a vanishing effect under integrability conditions. Simplifying and rearranging, we get a decomposition
\begin{align}
\mathbb{E}[\mathcal R(T)]=\mathbb{E}[\delta(\theta_0)]-\mathbb{E}[\delta(\theta_T)]+\mathbb{E}\left[\int_0^T\mathscr D_\Sigma\delta(\theta_t)dt\right].
\end{align}
We can now insert zero
\begin{align}
\mathscr D_\Sigma\delta=\eta\Delta_{\overline\partial}\delta+\left(\mathscr D_\Sigma-\eta\Delta_{\overline\partial}\right)\delta . 
\end{align}
Define the diffusion correction by $\mathbb{E}[\mathcal E_{\mathrm{diff}}]:=\mathbb{E}\left[\int_0^T\left(\mathscr D_\Sigma-\eta\Delta_{\overline\partial}\right)\delta(\theta_t)dt\right]$. Therefore,
\begin{align}
\mathbb{E}[\mathcal R(T)]=\mathbb{E}[\delta(\theta_0)]-\mathbb{E}[\delta(\theta_T)]+\eta\mathbb{E}\left[\int_0^T\Delta_{\overline\partial}\delta(\theta_t)dt\right]+\mathbb{E}[\mathcal E_{\mathrm{diff}}] .
\end{align}
Now, by the Laplacian comparison theorem and the claim (see \cite{fang2025laplaciancomparisontheoremscomplete} \cite{Tam2012} for relevant literature, although this exact equation is not given), for a Kähler manifold of complex dimension $K$, the Laplacian of the squared distance function is bounded along the minimizing geodesic $\gamma(s)$ parameterized by arc length $s \in [0, d_\omega(\theta_t, \theta^*)]$ via:

\vspace{2mm}

\textbf{Claim.} We have
\begin{align}
\label{eqn:laplacian_comparison}
\Delta_{\overline{\partial}} \delta(\theta_t) \leq  K - \frac{1}{2 d_\omega(\theta_t,\theta^*)}\int_0^{d_\omega(\theta_t,\theta^*)} s^2 \text{Ric}\big(\dot\gamma(s),\dot\gamma(s)\big) ds .
\end{align}

\vspace{2mm}

Now, define the defect term
\begin{align}
\mathbb{E}[\mathcal E_{\mathrm{Ric}}]:=-\frac{\eta}{2}\mathbb{E}\left[\int_0^T\frac{1}{r_t}\int_0^{r_t}s^2\mathrm{Ric}\big(\dot\gamma(s),\dot\gamma(s)\big)dsdt\right] .
\end{align}
Into our regret bound,
\begin{align}
\mathbb{E}[\mathcal R(T)]\leq\mathbb{E}[\delta(\theta_0)]-\mathbb{E}[\delta(\theta_T)]+\eta KT+\mathbb{E}[\mathcal E_{\mathrm{Ric}}]+\mathbb{E}[\mathcal E_{\mathrm{diff}}]  .
\end{align}
On a Ricci-flat Kähler manifold, $\mathrm{Ric}\equiv0$, implying $\mathbb{E}[\mathcal E_{\mathrm{Ric}}]=0$. Consequently,
\begin{align}
\mathbb{E}[\mathcal R(T)]\leq\mathbb{E}[\delta(\theta_0)]-\mathbb{E}[\delta(\theta_T)]+\eta KT+\mathbb{E}[\mathcal E_{\text{diff}}].
\end{align}
In the additional case of intrinsic isotropic diffusion, $\mathcal E_{\text{diff}}=0$.

\noindent $\square$

\vspace{2mm}

\textit{Proof of claim.} We prove equation \ref{eqn:laplacian_comparison}, which is nontrivial and challenging to find in literature. Let $(M, \omega)$ be a Kähler manifold of complex dimension $K$. For any $\theta$ where $\delta$ is smooth, let $\gamma(s)$ be the unit-speed minimizing geodesic from $\theta^*$ to $\theta$ parameterized by arc length $s \in [0, r]$, where $r = d_\omega(\theta, \theta^*)$. Now, we first use the real Riemannian Laplacian $\Delta_d$. The Laplacian of the distance function $r(\theta)$ can be bounded using the index form of the second variation of arc length. Let $E_1(s), \dots, E_{2K-1}(s)$ be an orthonormal frame of parallel vector fields along $\gamma$ that are orthogonal to $\dot\gamma$. Construct the Jacobi test fields $Y_i(s) = \frac{s}{r} E_i(s)$. The index lemma \cite{wang2024indexform} provides the upper bound 
\begin{align}
\Delta_d r(\theta) \leq \sum_{i=1}^{2K-1} I(Y_i, Y_i) = \int_0^r \sum_{i=1}^{2K-1} \left( \| \nabla_{\dot\gamma} Y_i\|_h^2 - \langle R(Y_i, \dot\gamma)\dot\gamma, Y_i \rangle_h \right) ds .
\end{align}
Because the frame $E_i$ is parallel, the covariant derivative simplifies to $\nabla_{\dot\gamma} Y_i = \frac{1}{r} E_i$, meaning the first term sums to $\frac{2K-1}{r^2}$. For the curvature term, substituting $Y_i(s)$ pulls out a factor of $\frac{s^2}{r^2}$. Summing the Riemann tensor over the orthonormal frame recovers Ricci curvature
\begin{align}
\sum_{i=1}^{2K-1} \langle R(E_i, \dot\gamma)\dot\gamma, E_i \rangle_h = \text{Ric}(\dot\gamma, \dot\gamma) .
\end{align}
Integrating over $[0, r]$ gives
\begin{align}
\Delta_d r(\theta) \leq \frac{2K-1}{r} - \frac{1}{r^2} \int_0^r s^2 \text{Ric}(\dot\gamma(s), \dot\gamma(s)) ds .
\end{align}We transition to the squared distance $\delta = \frac{1}{2}r^2$. By the chain rule, $\Delta_d \delta = r \Delta_d r + \|\nabla r\|^2$. Since $\|\nabla r\|^2 = 1$, we multiply our bound by $r$ and add $1$
\begin{align}
\Delta_d \delta(\theta) \leq 2K - \frac{1}{r} \int_0^r s^2 \text{Ric}(\dot\gamma(s), \dot\gamma(s)) ds .
\end{align}
Finally, on a Kähler manifold, the real Laplacian and the complex Dolbeault Laplacian acting on functions are related by $\Delta_d = 2\Delta_{\overline{\partial}}$. Dividing by 2 completes the proof.

\noindent $ \square $

\subsection{Dirichlet energy bounds}
\label{app:dirichlet_energy}

In this section, we examine a Dirichlet asymptotic scaling for sufficient set $S$. Dirichlet bounds are relevant in optimization literature
\cite{riis2018geometricintegrationapproachnonsmooth} \cite{Ehrhardt_2024} \cite{ringholm2018variationalimageregularizationeulers} \cite{niu2026continuoustimedynamicsdifferenceofconvexalgorithm} \cite{niu2022convergenceanalysisdca} \cite{HauerMazon2019} \cite{Dello_Schiavo_2024} for measuring sensitivity of the neural network with respect to its weights. This has connections to how fast the parameter descends since the squared norm of parameter gradient is the trace of the neural tangent kernel (NTK), and recall a trace is the sum of eigenvalues. The eigenvalues of the NTK are often intertwined with the learning process such as through convergence speed and spectral bias \cite{murray2023characterizingspectrumntkpower}.

\vspace{2mm}

\noindent \textit{Proof of Lemma 9.} Since $f$ is real-valued, we get a split $df = \partial f + \overline{\partial} f$. Let us begin with the (form variety) of the Dirichlet energy
\begin{align}
\label{eqn:dirichlet_energy}
E(f) = \frac{1}{ \text{Vol}_{\omega}(\mathcal{S})} \int_\mathcal{S} i \partial f \wedge \overline{\partial} f \wedge \frac{\omega^{K-1}}{(K-1)!} .
\end{align}
The goal is to shift the Dolbeault operator off of $\partial f$ to establish a Hessian formulation. We use the exterior derivative $d = \partial + \overline{\partial}$ and apply the Leibniz rule to the $(2K-1)$-form $f \overline{\partial} f \wedge \omega^{K-1}$, which gives
\begin{align}
d\left(f \overline{\partial} f \wedge \omega^{K-1}\right) = df \wedge \overline{\partial} f \wedge \omega^{K-1} + f d\left(\overline{\partial} f\right) \wedge \omega^{K-1} + f \overline{\partial} f \wedge d\left(\omega^{K-1}\right) .
\end{align}
$d \omega =0$ since the manifold is Kähler. Expanding $df = \partial f + \overline{\partial} f$, we note that $\overline{\partial} f \wedge \overline{\partial} f = 0$, leaving only $\partial f \wedge \overline{\partial} f$.
Furthermore, $d(\overline{\partial} f) = (\partial + \overline{\partial})\overline{\partial} f = \partial \overline{\partial} f$. Substituting in, and scaling by a constant,
\begin{align}
d \left( i f \overline{\partial} f \wedge \frac{\omega^{K-1}}{(K-1)!} \right) = \underbrace{ i \partial f \wedge \overline{\partial} f \wedge \frac{\omega^{K-1}}{(K-1)!} }_{\text{Dirichlet integrand}} + f \left( i \partial \overline{\partial} f \wedge \frac{\omega^{K-1}}{(K-1)!} \right) .
\end{align}
Notice the middle term is the integrand of \ref{eqn:dirichlet_energy}. Integrating, and by Stokes' theorem,
\begin{align}
\frac{1}{ \text{Vol}_{\omega}(\mathcal{S})} \oint_{\partial \mathcal{S}} i f \overline{\partial} f \wedge \frac{\omega^{K-1}}{(K-1)!} & = E(f) + \frac{1}{ \text{Vol}_{\omega}(\mathcal{S})} \int_\mathcal{S} f \underbrace{ \left( i \partial \overline{\partial} f \wedge \frac{\omega^{K-1}}{(K-1)!} \right) }_{\displaystyle  = (\Delta_{\overline{\partial}} f) \frac{\omega^K}{K!}} 
\\
& = E(f) + \frac{1}{ \text{Vol}_{\omega}(\mathcal{S})} \int_\mathcal{S} f (\Delta_{\overline{\partial}} f) \frac{\omega^K}{K!}.
\end{align}
Hence, the Dirichlet energy can be written as
\begin{align}
\label{eqn:dirichlet_energy_decomposed}
E(f) & = - \frac{1}{ \text{Vol}_{\omega}(\mathcal{S})} \int_\mathcal{S} f (\Delta_{\overline{\partial}} f) \frac{\omega^K}{K!} + \frac{1}{ \text{Vol}_{\omega}(\mathcal{S})} \oint_{\partial \mathcal{S}} i f \overline{\partial} f \wedge \frac{\omega^{K-1}}{(K-1)!}  
\\
& \leq \|f\|_{L^2} \|\Delta_{\overline{\partial}} f\|_{L^2} + |\mathcal{B}_{\partial \mathcal{S}}(f) | .
\end{align}
This follows from Hölder's and the second term is by definition. We can note $\| f\|_{L^2}$ is $\mathcal{O}(1)$ since
\begin{align}
\|f\|_{L^2(\mathcal{S})} = \left( \frac{1}{ \text{Vol}_{\omega}(\mathcal{S})} \int_\mathcal{S} |f(z)|^2 \frac{\omega^K}{K!} \right)^{1/2} \leq \mathcal{O}(1) \cancel{ \frac{ \sqrt{\text{Vol}_\omega(\mathcal{S})} }{\sqrt{\text{Vol}_{\omega}(\mathcal{S})}} } = \mathcal{O}(1)  .
\end{align}
We previously established
\begin{align}
\sup_{\theta \in \mathcal{S}} \| i \partial \overline{\partial} f \|_2 = \mathcal{O}(\frac{1}{\sqrt{m}} )
\end{align}
for sufficient ball $\mathcal{S}$, but the norms here differ, and so the above has greater connections to the Laplacian term of \ref{eqn:dirichlet_energy_decomposed}. Note the identity $\Delta_{\overline{\partial}} f = \text{Tr}_\omega(i \partial \overline{\partial} f)$. Therefore, it follows
\begin{align}
\|\Delta_{\overline{\partial}} f\|_{L^2(\mathcal{S})} \leq \left( \frac{1}{ \text{Vol}_{\omega}(\mathcal{S})}  \int_\mathcal{S} \mathcal{O}\left(\frac{K^2}{m}\right) \frac{\omega^K}{K!} \right)^{1/2} = \mathcal{O}\left(\frac{K}{\sqrt{m}}\right)   . 
\end{align}
In particular, the norm is (averaged) $L^2$ so we get a squaring $\frac{K}{\sqrt{m}} \rightarrow \frac{K^2}{m}$ since the trace uses a $K$ bound. Turning to the boundary term, we can restrict the $L^2$ norm over a subset of the domain. Moreover, we can bound
\begin{align}
|\mathcal{B}_{\partial \mathcal{S}}(f)| \leq \underbrace{ \frac{ \text{Vol}_\omega(\partial \mathcal{S}) }{ \text{Vol}_{\omega}(\mathcal{S})} }_{= \mathcal{O}(K)} \underbrace{ \left( \sup_{\theta \in \partial \mathcal{S}} |f(\theta)| \right) }_{=\mathcal{O}(1)} \underbrace{ \left( \sup_{\theta \in \partial \mathcal{S}} \|\overline{\partial} f(\theta)\|_h \right) }_{=\mathcal{O}(1)} = \mathcal{O}(K)\footnote .
\end{align}
\footnotetext{This assumes the radius of the geodesic ball is $\mathcal{O}(1)$. This is consistent with what we assumed in \ref{app:dolb_hessian_bounds}.}We have used the $\omega$ norm $\| \overline{\partial} f \|_h^2 = h^{j\overline{k}} (\overline{\partial} f)_{\overline{k}} \overline{(\overline{\partial} f)_{\overline{j}}}$. Therefore, around initialization,
\begin{align}
E(f) \leq \mathcal{O}(1) \times \underbrace{ \mathcal{O}(\frac{K}{\sqrt{m}}) }_{\| \Delta_{\overline{\partial}} f \|_{L^2(\mathcal{S})}}  + \mathcal{O}(K)  = \mathcal{O}(K + \frac{K}{\sqrt{m}}) .
\end{align}
The Dirichlet energy is across a single data point, therefore summing across $N$ data points
\begin{align}
\sum_{\alpha=1}^N E(f_\alpha) \leq \sum_{\alpha=1}^N \left( \mathcal{O}(1) \times \underbrace{ \mathcal{O}\left(\frac{K}{\sqrt{m}}\right) }_{\| \Delta_{\overline{\partial}} f_\alpha \|_{L^2(\mathcal{S})}}  + \mathcal{O}(K) \right) = \mathcal{O}(NK + \frac{NK}{\sqrt{m}}) . 
\end{align}

\vspace{2mm}

\noindent We will verify the lower bound, which is trivially nonnegative. Define the unregularized (empirical) Fisher metric $\mathcal{F} = \sum_{\alpha=1}^N \partial f_\alpha \wedge \overline{\partial} f_\alpha$. We instilled a (Fisher, or recall we used cross-entropy) eigenvalue condition $\mathcal{F}_{\text{reg}} = \mathcal{F} + \lambda I,  \lambda_{\text{min}} ( \mathcal{F}_{\text{reg}}) \geq \lambda > 0$. Now, the trace of the unregularized information metric with respect to $\omega$ is the sum of the squared gradient norms, which coincides with the trace of the Neural Tangent Kernel (NTK)
\begin{align}
\text{Tr}(\mathcal{F}) = \sum_{\alpha=1}^N \|\partial f(z_\alpha)\|_h^2 = \text{Tr}(K_{\text{NTK}}) .
\end{align}
Moreover, let us note
\begin{align}
E(f(z_\alpha)) & = \frac{1}{ \text{Vol}_{\omega}(\mathcal{S})} \int_\mathcal{S} \|\partial f(z_\alpha)\|_h^2 \frac{\omega^K}{K!}
\\
\implies \sum_{\alpha=1}^N E(f(z_\alpha)) & = \sum_{\alpha=1}^N \left( \frac{1}{ \text{Vol}_{\omega}(\mathcal{S})} \int_\mathcal{S} \|\partial f(z_\alpha)\|_h^2 \frac{\omega^K}{K!} \right)
\\
& \stackrel{\substack{\text{the sum is finite} \\ \text{the functions are nonnegative}}}{=} \frac{1}{ \text{Vol}_{\omega}(\mathcal{S})} \int_\mathcal{S} \left( \sum_{\alpha=1}^N \|\partial f(z_\alpha)\|_h^2 \right) \frac{\omega^K}{K!}
\\
& = \frac{1}{ \text{Vol}_{\omega}(\mathcal{S})} \int_\mathcal{S} \text{Tr}(K_{\text{NTK}}) \frac{\omega^K}{K!} .
\end{align}
By taking the trace of our regularized Fisher metric, we get
\begin{align}
\label{eqn:tr_ntk}
\text{Tr}(\mathcal{F}_{\text{reg}}) = \text{Tr}(\mathcal{F}) + \text{Tr}(\lambda I) = \text{Tr}(K_{\text{NTK}}) + \lambda K .
\end{align}
Because we assumed $\lambda_{\text{min}} ( \mathcal{F}_{\text{reg}}) \geq \lambda$, the trace is bounded below by the sum of its minimal eigenvalues across all $K$ dimensions
\begin{align}
\text{Tr}(\mathcal{F}_{\text{reg}}) \geq \lambda K .
\end{align}
Isolating the trace of the NTK, using \ref{eqn:tr_ntk}, we get
\begin{align}
\text{Tr}(K_{\text{NTK}}) = \sum_{\alpha=1}^N \|\partial f(z_\alpha)\|_h^2 \geq 0 .
\end{align}
Our total Dirichlet energy of the network over the dataset is the integral of this NTK trace over the set $\mathcal{S}$
\begin{align}
\sum_{\alpha=1}^N E(f(z_\alpha)) = \frac{1}{ \text{Vol}_{\omega}(\mathcal{S})}  \int_\mathcal{S} \text{Tr}(K_{\text{NTK}}) \frac{\omega^K}{K!} \geq \frac{1}{ \text{Vol}_{\omega}(\mathcal{S})}  \int_\mathcal{S} 0 \frac{\omega^K}{K!}  = 0.
\end{align}

\noindent $\square$

\subsection{Trajectory bounds}
\label{app:trajectory_bounds}

\noindent In this section, we analyze the asymptotic regime of the descent path of the parameter. Let us begin by first proving a claim.

\vspace{2mm}

\noindent \textit{\textbf{Claim.} We have $\int_0^t \|\dot{\theta}_s\|_h ds = \mathcal{O}(\frac{1}{\sqrt{m}})$ near initialization.}

\vspace{2mm}

\noindent \textit{Proof.} Under natural gradient descent, we have $\dot{\theta}_s = -h^{-1} \nabla \mathcal{L}(\theta_s)$. From discussion in earlier sections such as \ref{sec:network_setup} and \ref{app:initialization}, we define $f(\theta) = \frac{1}{\sqrt{m}} v^\dagger \alpha^{(L)}$ and we note $\|\delta^{(l)}\|_\infty = \mathcal{O}(\frac{1}{\sqrt{m}})$.  We have $\|\nabla f(\theta)\|_2 = \mathcal{O}(\frac{1}{\sqrt{m}})$. Since under negative log-likelihood loss, $\nabla \mathcal{L}(\theta) = (p(y|x) - y) \nabla f(\theta)$, we get $\|\nabla \mathcal{L}(\theta_0)\|_2 = \mathcal{O}(\frac{1}{\sqrt{m}})$\footnote{We note $(p-y)$ is $\mathcal{O}(1)$.}. Now, we can note the norm on $\dot{\theta}_s$ obeys
\begin{align}
\|\dot{\theta}_s\|_h = \sqrt{(\nabla \mathcal{L}(\theta_s))^\dagger h^{-1} (\nabla \mathcal{L}(\theta_s))} .
\end{align}
Under the regularized loss assumption of \ref{eqn:regularized_loss} $h_{\text{reg}} = h + \lambda I$, the minimum eigenvalue is bounded away from zero, $\lambda_{\text{min}}(h) \ge \lambda > 0$. Therefore, the spectral norm of the inverse metric follows $\|h^{-1}\|_2 \leq \frac{1}{\lambda} = \mathcal{O}(1)$\footnote{We assume this for $h$, not $H$, at least in this result only. Recall we have regularized the loss so the metric is full rank. We make this remark to be compatible with our theory in \ref{app:KPL_failure}, \ref{app:calabi_yau_oscillation}.}. By Cauchy-Schwarz,
\begin{align}
\|\dot{\theta}_0\|_h \leq \sqrt{\|h^{-1}\|_2} \|\nabla \mathcal{L}(\theta_0)\|_2 = \mathcal{O}(\frac{1}{\sqrt{m}}) .
\end{align}
Via Fundamental Theorem of Calculus (we have sufficient smoothness), and applying an inequality,
\begin{align}
\label{eqn:nabla_L_FCT}
\|\nabla_h \mathcal{L}(\theta_s)\|_h \leq \|\nabla_h \mathcal{L}(\theta_0)\|_h + \int_0^s \|\nabla_{\omega}^2 \mathcal{L}(\theta_\tau)\|_h \|\dot{\theta}_\tau\|_h d\tau .
\end{align}
Since $\dot{\theta}_s = -h^{-1} \nabla \mathcal{L}(\theta_s)$, we can relate \ref{eqn:nabla_L_FCT} to this and we get
\begin{align}
\|\dot{\theta}_s\|_h \leq \|\dot{\theta}_0\|_h  + \mathcal{O}(\frac{1}{\sqrt{m}}) \int_0^s  \|\dot{\theta}_\tau\|_h  d\tau  ,
\end{align}
again noting the inverse metric is $\mathcal{O}(1)$. This line is nontrivial because \ref{eqn:nabla_L_FCT} is with a different norm than the spectral norm, which our original result used; however, this is salvageable and correct because the spectral norm and the $h$-norm can be related via a maximum eigenvalue. $\lambda_{\text{max}}$ is $\mathcal{O}(1)$, and so is the condition number. Therefore $\|\nabla^2 \mathcal{L}\|_h \leq \mathcal{O}(1) \|\nabla^2 \mathcal{L}\|_2 = \mathcal{O}(1) \cdot \mathcal{O}\left(\frac{1}{\sqrt{m}}\right) = \mathcal{O}\left(\frac{1}{\sqrt{m}}\right)$. Moreover, we have noted the Hessian results of \ref{app:2,0_hess_bounds}, \ref{app:dolb_hessian_bounds}. Via Grönwall's inequality, $\|\dot{\theta}_s\|_h \leq \|\dot{\theta}_0\|_h \exp\left(\mathcal{O}(\frac{1}{\sqrt{m}}) s\right)$. Therefore, we conclude
\begin{align}
\int_0^t  \|\dot{\theta}_s\|_hds \leq \int_0^t  \mathcal{O}(\frac{1}{\sqrt{m}}) ds= \mathcal{O}(\frac{1}{\sqrt{m}}) ,
\end{align}
since $\|\dot{\theta}_0\|_h  = \mathcal{O}(\frac{1}{\sqrt{m}})$.

\noindent $\square$

\vspace{2mm}

\begin{figure}[htbp]
  \centering
  \includegraphics[width=0.65\textwidth]{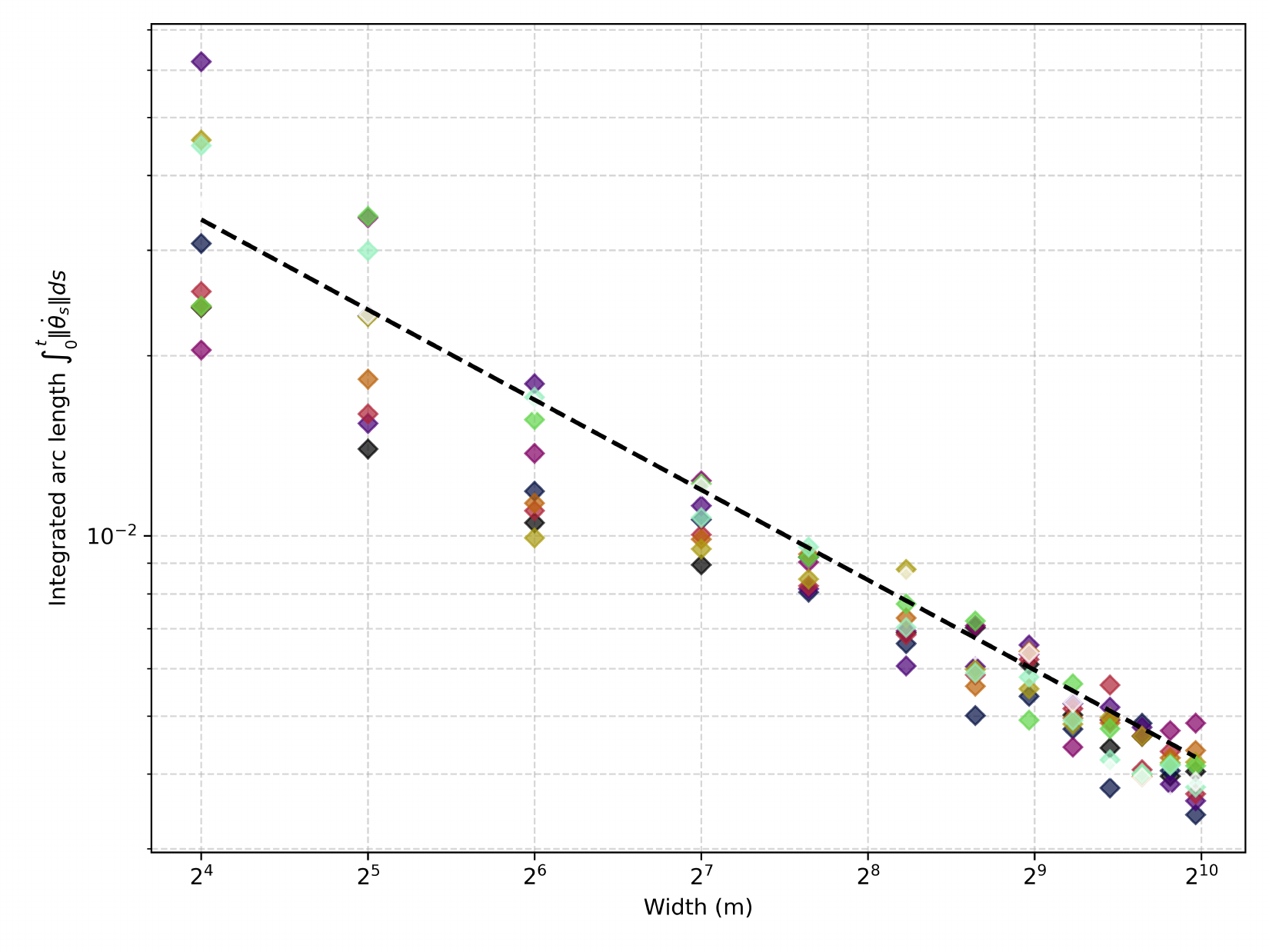}
  \caption{We plot the asymptotics of $\int \| \dot{\theta}\|_h ds$ near initialization across $10 \cdot (\text{number of widths)}$ distinct calculations. We use batch size 64, learning rate 0.01, and 15 steps. The line is the $\frac{1}{\sqrt{m}}$ asymptotic line. The neural network is consistent with \ref{sec:network_setup}.}
  \label{fig:integrated_arc_length_scaling}

  \vspace{10mm}

  \centering
  \includegraphics[width=0.65\textwidth]{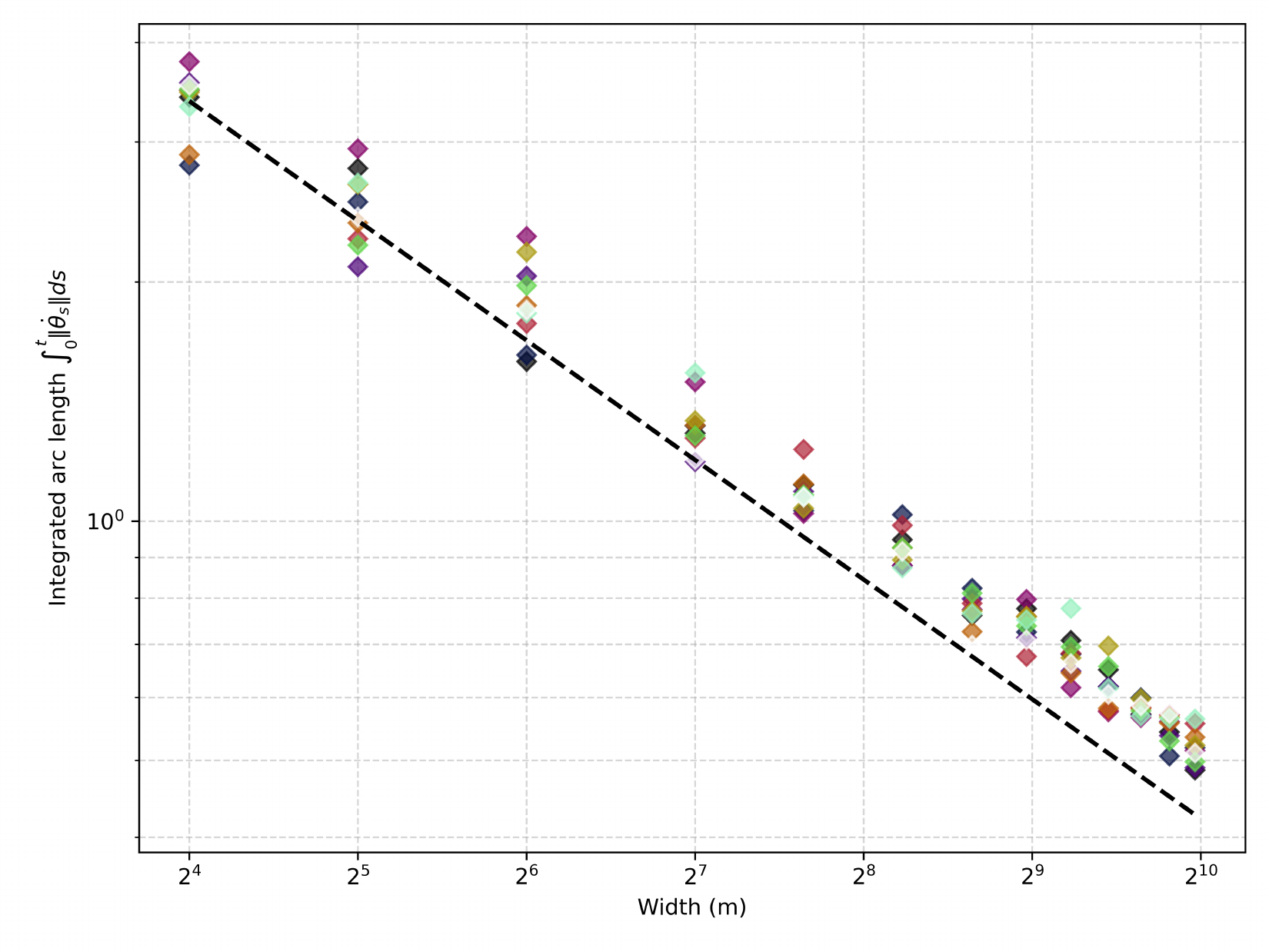}
  \caption{We plot the asymptotics of $\int \| \dot{\theta}\|_h ds$ near initialization across $10 \cdot (\text{number of widths)}$ distinct calculations. We use batch size 64, learning rate 0.5, and 50 steps, meaning we deviate from initialization much further than \ref{fig:integrated_arc_length_scaling}. The line is consistent with Figure \ref{fig:integrated_arc_length_scaling}, and the network consistent with \ref{sec:network_setup}.}
  \label{fig:integrated_arc_length_scaling_highstepslr}
\end{figure}

\vspace{2mm}

\noindent \textit{Proof of Lemma 10.}
Under natural gradient descent, $\dot{\theta}^i = -h^{i\overline{j}} \partial_{\overline{j}} \mathcal{L}$. We define the gradient norm $V(t) = \|\nabla_h \mathcal{L}\|_h^2$. Differentiating yields
\begin{align}
\label{eqn:ddt_v}
\frac{d}{dt} v(t) = - 2 \text{Re}(\nabla_h \mathcal{L}^T \nabla^{2,0}_{\omega} \mathcal{L} \nabla_h \mathcal{L}) - 2 \nabla_h \mathcal{L}^\dagger \nabla^{1,1}_{\omega} \mathcal{L} \nabla_h \mathcal{L} .
\end{align}
We define uniform bounds over $\Omega$ at time $t$
\begin{align}
H(t) = \sup_{\theta \in \Omega} \|\nabla^{2,0}_{\omega} \mathcal{L}\|_h \quad \text{and} \quad \mu(t) = \inf_{\theta \in \Omega} \lambda_{\text{min}}(\nabla^{1,1}_{\omega} \mathcal{L}) .
\end{align}
Applying Cauchy-Schwarz to the $(2,0)$, we obtain the inequality using \ref{eqn:ddt_v}
\begin{align}
\frac{d}{dt} V(t) \leq 2 H(t) V(t) - 2 \mu(t) V(t) .
\end{align}
By Grönwall's inequality, we bound the trajectory
\begin{align}
V(t) \leq V(0) \exp\left( 2 \int_0^t H(s) ds - 2 \int_0^t \mu(s) ds \right) .
\end{align}
We are given the initialization bounds at $t=0$ using the results of \ref{app:dolb_hessian_bounds}
\begin{align}
H(0) = \mathcal{O}(\frac{1}{\sqrt{m}}) \quad \text{and} \quad \mu(0) = \mathcal{O}(\frac{1}{\sqrt{m}}) .
\end{align}
We assume the Hessians are locally Lipschitz with respect to the Kähler metric connection with constant $L$. We can note via Fundamental Theorem of Calculus (take absolute continuity)
\begin{align}
\|\nabla^{2,0}_{\omega} \mathcal{L}(\theta_t)\|_h - \|\nabla^{2,0}_{\omega} \mathcal{L}(\theta_t)|_{t=0} \|_h = \int_0^t \frac{d}{ds} \|\nabla^{2,0}_{\omega} \mathcal{L}(\theta_s)\|_h ds 
\end{align}
and moreover via chain rule
\begin{align}
\frac{d}{ds} \|\nabla^{2,0}_{\omega} \mathcal{L}(\theta_t)\|_h \leq \|\nabla_h (\nabla^{2,0}_{\omega} \mathcal{L}(\theta_s))\|_h \|\dot{\theta}_s\|_h \leq L \|\dot{\theta}_s\|_h  ,
\end{align}
so under a Lipschitz condition, we see
\begin{align}
H(t) \leq H(0) + L \int_0^t \|\dot{\theta}_s\|_h ds .
\end{align}
Similarly,
\begin{align}
\mu(t) \geq \mu(0) - L \int_0^t \|\dot{\theta}_s\|_h ds .
\end{align}
Since $\int_0^t \|\dot{\theta}_s\|_h ds = \mathcal{O}(\frac{1}{\sqrt{m}})$, we can substitute this into our Lipschitz bounds, so
\begin{align}
H(t) = \mathcal{O}(\frac{1}{\sqrt{m}}) + L \cdot \mathcal{O}(\frac{1}{\sqrt{m}}) = \mathcal{O}(\frac{1}{\sqrt{m}}) ,
\end{align}
and moreover,
\begin{align}
\mu(t) = \mathcal{O}(\frac{1}{\sqrt{m}}) - L \cdot \mathcal{O}(\frac{1}{\sqrt{m}}) = \mathcal{O}(\frac{1}{\sqrt{m}}) .
\end{align}
We return to the Grönwall bound established previously. Using what we found
\begin{align}
V(t) & \leq V(0) \exp\left( 2 \int_0^t \mathcal{O}(\frac{1}{\sqrt{m}}) ds - 2 \int_0^t \mathcal{O}(\frac{1}{\sqrt{m}}) ds \right)
\\
& \leq V(0) \exp\left( 2t \cdot \mathcal{O}(\frac{1}{\sqrt{m}}) \right) .
\end{align}
The constants do not cancel, which holds almost surely.

\vspace{2mm}

\noindent To guarantee convergence for finite $m$, we bound the integral of $\mu(s)$ via a lower bound. This shows $\mu$ and its integral are sufficiently large and therefore the exponential with this negative integral in its exponent is dissipative, giving a convergence result. Let us establish an eigenvalue dissipation rate geometrically. Let $M_{i\overline{j}} = \nabla_{i\overline{j}} \mathcal{L}$ denote the $(1,1)$ Hessian, and let $N_{ij} = \nabla_{ij} \mathcal{L}$ denote the $(2,0)$ Hessian. Let us consider the natural gradient flow $\dot{\theta}^i = -h^{i\overline{j}} \partial_{\overline{j}} \mathcal{L} = -\nabla^i \mathcal{L}$ as in \ref{sec:geometry}. Since the parameter is a function of time, we can perform the chain rule
\begin{align}
\partial_t \mathcal{L} = \frac{d}{dt} \mathcal{L}(\theta(t), \overline{\theta}(t)) = \sum_p \frac{\partial \mathcal{L}}{\partial \theta^p} \dot{\theta}^p + \sum_q \frac{\partial \mathcal{L}}{\partial \overline{\theta}^q} \dot{\overline{\theta}}^q .
\end{align}
We compute the material derivative
\begin{align}
\frac{D}{dt} M_{i\overline{j}} = \dot{\theta}^p \nabla_p M_{i\overline{j}} + \dot{\overline{\theta}}^q \nabla_{\overline{q}} M_{i\overline{j}} .
\end{align}
Substituting our gradient flow,
\begin{align}
\frac{D}{dt} M_{i\overline{j}} = - \nabla^p \mathcal{L} \nabla_p (\nabla_{i\overline{j}} \mathcal{L}) - \nabla^{\overline{q}} \mathcal{L} \nabla_{\overline{q}} (\nabla_{i\overline{j}} \mathcal{L}) .
\end{align}
Since the metric is Kähler, $\nabla_{i\overline{j}p} \mathcal{L} = \nabla_{p i\overline{j}} \mathcal{L}$, using notation $\nabla_{i\overline{j}p} \mathcal{L} = \nabla_i \nabla_{\overline{j}} \nabla_p \mathcal{L}$. For the antiholomorphic commutativity, we can note using Riemannian curvature
\begin{align}
\nabla_{\overline{q} i\overline{j}} \mathcal{L} = \nabla_{i\overline{j}\overline{q}} \mathcal{L} - R_{i\overline{j}p\overline{q}} \nabla^p \mathcal{L} .
\end{align}
Substituting back in, the material derivative obeys
\begin{align}
\label{eqn:mat_derivative_M}
\frac{D}{dt} M_{i\overline{j}} & = - \nabla^p \mathcal{L} \nabla_{i\overline{j}p} \mathcal{L} - \nabla^{\overline{q}} \mathcal{L} (\nabla_{i\overline{j}\overline{q}} \mathcal{L} - R_{i\overline{j}p\overline{q}} \nabla^p \mathcal{L}) 
\\
&  = - (\nabla^p \mathcal{L} \nabla_{i\overline{j}p} \mathcal{L} + \nabla^{\overline{q}} \mathcal{L} \nabla_{i\overline{j}\overline{q}} \mathcal{L}) + R_{i\overline{j}p\overline{q}} \nabla^p \mathcal{L} \nabla^{\overline{q}} \mathcal{L} .
\end{align}
Now, we analyze the spatial Hessian of the squared gradient norm, $F = \|\nabla \mathcal{L}\|_h^2 = h^{p\overline{q}} \nabla_p \mathcal{L} \nabla_{\overline{q}} \mathcal{L}$. Applying the covariant derivative $\nabla_{i\overline{j}}$ and the product rule generates four terms
\begin{align}
\nabla_{i\overline{j}} (\|\nabla \mathcal{L}\|_h^2) & = h^{p\overline{q}} (\nabla_{i\overline{j}p} \mathcal{L}) \nabla_{\overline{q}} \mathcal{L} + h^{p\overline{q}} \nabla_p \mathcal{L} (\nabla_{i\overline{j}\overline{q}} \mathcal{L}) + h^{p\overline{q}} \nabla_{ip} \mathcal{L} \nabla_{\overline{j}\overline{q}} \mathcal{L} + h^{p\overline{q}} \nabla_{\overline{j}p} \mathcal{L} \nabla_{i\overline{q}} \mathcal{L} 
\\
& = \nabla^p \mathcal{L} \nabla_{i\overline{j}p} \mathcal{L} + \nabla^{\overline{q}} \mathcal{L} \nabla_{i\overline{j}\overline{q}} \mathcal{L} + N_{ip} N^p_{\overline{j}} + M_{i\overline{k}} M^{\overline{k}}_{\overline{j}} ,
\end{align}
which we have rewrote using our definitions. Rearranging,
\begin{align}
- (\nabla^p \mathcal{L} \nabla_{i\overline{j}p} \mathcal{L} + \nabla^{\overline{q}} \mathcal{L} \nabla_{i\overline{j}\overline{q}} \mathcal{L}) = - \nabla_{i\overline{j}} (\|\nabla \mathcal{L}\|_h^2) + M_{i\overline{k}} M^{\overline{k}}_{\overline{j}} + N_{ip} N^p_{\overline{j}} .
\end{align}
Substituting back into \ref{eqn:mat_derivative_M},
\begin{align}
\frac{D}{dt} M_{i\overline{j}} = M_{i\overline{k}} M^{\overline{k}}_{\overline{j}} + N_{ip} N^p_{\overline{j}} + R_{i\overline{j}p\overline{q}} \nabla^p \mathcal{L} \nabla^{\overline{q}} \mathcal{L} - \nabla_{i\overline{j}} (\|\nabla \mathcal{L}\|_h^2) .
\end{align}
For normalized eigenvector $X$, we have the minimum eigenvalue equation of the Hessian (in a geometric sense, so scaled by the metric) can be calculated via the standard eigenvalue equation
\begin{align}
M_{i\overline{j}} X^i = \mu(t) h_{i\overline{j}} X^i .
\end{align}
We can contract with $X^{\overline{j}}$ and under a normalized, unit eigenvector property $h_{i\overline{j}} X^i X^{\overline{j}} = 1$, we get the eigenvalue follows under these two equations
\begin{align}
\mu(t) = M_{i\overline{j}} X^i X^{\overline{j}} .
\end{align}
Differentiating,
\begin{align}
\frac{d}{dt} \mu(t) = (\frac{D}{dt} M_{i\overline{j}}) X^i X^{\overline{j}} .
\end{align}
Because $X$ is an eigenvector,
\begin{align}
M_{i\overline{k}} M^{\overline{k}}_{\overline{j}} X^i X^{\overline{j}} = (M_{i\overline{k}} X^i) (\overline{M_{j\overline{l}} X^j}) = (\mu(t) X_{\overline{k}}) (\mu(t) X^{\overline{k}}) = \mu(t)^2  .
\end{align}
The above commutes since the contraction makes it scalar. Moreover, we can note
\begin{align}
 N_{ip} N^p_{\overline{j}} X^i X^{\overline{j}} = \|N(X, \cdot)\|_h^2 .
\end{align}
Assuming the holomorphic bisectional curvature is positive and bounded below by a constant $\kappa > 0$, we have $R_{i\overline{j}p\overline{q}} X^i X^{\overline{j}} \nabla^p \mathcal{L} \nabla^{\overline{q}} \mathcal{L} \geq \kappa V(t)$, where $V(t) = \|\nabla \mathcal{L}\|_h^2$. We can rewrite
\begin{align}
\frac{d}{dt} \mu(t) \geq \mu(t)^2 + \|N(X, \cdot)\|_h^2 + \kappa V(t) - X^i X^{\overline{j}} \nabla_{i\overline{j}} (\|\nabla \mathcal{L}\|_h^2)   .
\end{align}
Contracting with simplified \ref{eqn:mat_derivative_M}, and putting everything together
\begin{align}
\frac{d}{dt} \mu(t) & = M_{i\overline{k}} M^{\overline{k}}_{\overline{j}} X^i X^{\overline{j}} + N_{ip} N^p_{\overline{j}} X^i X^{\overline{j}}+ R_{i\overline{j}p\overline{q}} X^i X^{\overline{j}} \nabla^p \mathcal{L} \nabla^{\overline{q}} \mathcal{L} - X^i X^{\overline{j}} \nabla_{i\overline{j}} (\|\nabla \mathcal{L}\|_h^2) 
\\
& \geq \mu(t)^2 + \|N(X, \cdot)\|_h^2 + R_{i\overline{j}p\overline{q}} X^i X^{\overline{j}} \nabla^p \mathcal{L} \nabla^{\overline{q}} \mathcal{L} - X^i X^{\overline{j}} \nabla_{i\overline{j}} (\|\nabla \mathcal{L}\|_h^2)
\\
& \geq \mu(t)^2 + \|N(X, \cdot)\|_h^2 + \kappa V(t) - X^i X^{\overline{j}} \nabla_{i\overline{j}} (\|\nabla \mathcal{L}\|_h^2)  .
\end{align}

\noindent $ \square $

\vspace{2mm}

\noindent \textit{Remark.} Let us return to \ref{eqn:ddt_v}. We will rewrite it using physics notation since this is compatible and standard with Rayleigh quotients. Denote $|\Psi(t)\rangle := |\nabla_h L(t)\rangle$. We get
\begin{align}
\frac{d}{dt}\langle \Psi | \Psi \rangle_h = -2\text{Re}\langle \overline{\Psi} | \nabla_\omega^{2,0} L | \Psi \rangle_h - 2\langle \Psi | \nabla_\omega^{1,1} L | \Psi \rangle_h .
\end{align}
Factoring out $v(t) = \langle \Psi | \Psi \rangle_h$,
\begin{align}
\frac{d}{dt}v(t) & = - \left[ 2\frac{\text{Re}\langle \overline{\Psi} | \nabla_\omega^{2,0} L | \Psi \rangle_h}{\langle \Psi | \Psi \rangle_h} + 2\frac{\langle \Psi | \nabla_\omega^{1,1} L | \Psi \rangle_h}{\langle \Psi | \Psi \rangle_h} \right] v(t) 
\\
& = - \frac{1}{\sqrt{m}} \left[ \underbrace{ 2\frac{\text{Re}\langle \overline{\Psi} | \widetilde{H}^{2,0} | \Psi \rangle_h}{\langle \Psi | \Psi \rangle_h} + 2\frac{\langle \Psi | \widetilde{H}^{1,1} | \Psi \rangle_h}{\langle \Psi | \Psi \rangle_h} }_{\displaystyle := \langle \mathcal{T}_{\text{def}}(t) \rangle_\Psi } \right] v(t).
\end{align}
We can notice in the interior there are two Rayleigh quotients. Define two scaled Hessians such that $\nabla_\omega^{2,0} L = \frac{1}{\sqrt{m}} \widetilde{H}^{2,0}$, $\nabla_\omega^{1,1} L = \frac{1}{\sqrt{m}} \widetilde{H}^{1,1}$, and define $\langle \mathcal{T}_{\text{def}}(t) \rangle_\Psi$ in the above. Therefore, we can note a reformulation with \ref{eqn:rayleigh}
\begin{align}
\frac{d}{dt}v(t) + \frac{1}{\sqrt{m}} \langle \mathcal{T}_{\text{def}}(t) \rangle_\Psi v(t) = 0 .
\end{align}
Via power series
\begin{align}
\left( \frac{d}{dt} + \frac{1}{\sqrt{m}} \langle \mathcal{T}_{\text{def}}(t) \rangle_\Psi \right) \left( v_0(t) + \frac{1}{\sqrt{m}} v_1(t) + \frac{1}{m} v_2(t) + \dots \right) = 0 .
\end{align}
Permit $m$ to be variable. In reality, $m$ is fixed, but say in an asymptotic limit $m\rightarrow \infty$, $m$ is nonconstant. We will show the constants are zero. Denote $\epsilon_i = 1/\sqrt{m_i}$ for short. We can write this as a system
\begin{align}
\begin{cases}  \mathcal{E}_0(t) + \epsilon_1 \mathcal{E}_1(t) + \epsilon_1^2 \mathcal{E}_2(t) + \dots + \epsilon_1^{N-1} \mathcal{E}_{N-1}(t) = 0 \\ \mathcal{E}_0(t) + \epsilon_2 \mathcal{E}_1(t) + \epsilon_2^2 \mathcal{E}_2(t) + \dots + \epsilon_2^{N-1} \mathcal{E}_{N-1}(t) = 0 \\ \quad \vdots \\ \mathcal{E}_0(t) + \epsilon_N \mathcal{E}_1(t) + \epsilon_N^2 \mathcal{E}_2(t) + \dots + \epsilon_N^{N-1} \mathcal{E}_{N-1}(t) = 0 .\end{cases} 
\end{align}
For particular choice,
\begin{align}
\begin{bmatrix} 1 & \epsilon_1 & \epsilon_1^2 & \dots & \epsilon_1^{N-1} \\ 1 & \epsilon_2 & \epsilon_2^2 & \dots & \epsilon_2^{N-1} \\ \vdots & \vdots & \vdots & \ddots & \vdots \\ 1 & \epsilon_N & \epsilon_N^2 & \dots & \epsilon_N^{N-1} \end{bmatrix} \begin{bmatrix} \mathcal{E}_0(t) \\ \mathcal{E}_1(t) \\ \vdots \\ \mathcal{E}_{N-1}(t) \end{bmatrix} = \begin{bmatrix} 0 \\ 0 \\ \vdots \\ 0  \end{bmatrix} ,
\end{align}
which is Vandermonde and invertible, and writing $V \mathcal{E}(t)=0$, this has determinant
\begin{align}
\det(V) = \prod_{1 \leq i < j \leq N} (\epsilon_j - \epsilon_i) .
\end{align}
Use shorthand $\mathcal{D}_0 := \frac{d}{dt}$ denote the time derivative linear evolution operator infinite-width, and we have
\begin{align}
\mathcal{D}_0 v_1(t) = - \langle \mathcal{T}_{\text{def}}(t) \rangle_\Psi v_0(t),
\end{align}
or equivalently,
\begin{align}
v_k(t) = -\int_0^t \langle \mathcal{T}_{\text{def}}(\tau) \rangle_\Psi v_{k-1}(\tau) d\tau .
\end{align}
What this primarily shows is that parameter evolution retains a "memory" of previous states, and that subsequent states are largely determined by their previous states, geometrically speaking. In general, a Rayleigh quotient takes the form
\begin{align}
\text{Rayleigh}(M, x) = \frac{x^\dagger M x}{x^\dagger x} ,
\end{align}
which is a way to stretch the space in the direction of $x$. Moreover, $\lambda_{\text{min}} \leq \frac{x^\dagger M x}{x^\dagger x} \leq \lambda_{\text{max}}$ for eigenvalues of $M$.

\subsection{Gradient flux searches}
\label{app:gradient_flux_searches}

In this section, we attempt to characterize the total amount of flexibility across gradients in an $\epsilon$-ball around initialization. The motivation for this is we will examine the flux across varying radii via
\begin{align}
\label{eqn:oint_flux_noW}
\int_0^\epsilon \left( \oint_{\partial B_r(\theta_0)} d^c \mathcal{L} \wedge \omega^{K-1} \right) dr ,
\end{align}
and attempt to quantify this flux value. The above characterizes how the curvature of the manifold affects the net flow of gradients around initialization. Generally, a larger negative value is more desirable. We highlight:
\begin{enumerate}[label=\Roman*., nosep]
\item Positive flux: net outward gradient flow; corresponds to net inward descent. Note that gradient $+\nabla_h \mathcal{L}$ points inward towards increasing loss.
\item Negative flux: net inward gradient flow; corresponds to net outward descent. Note that gradient descent $- \nabla_h \mathcal{L}$ points outward towards decreasing loss.
\end{enumerate}
We crucially remark our sign convention in \ref{eqn:oint_flux_noW} uses a positive sign convention, which affects the above. This is under the convention the reference normal vector is outward-facing. This implies there is greater variability for trajectory paths, and greater ability to escape the $\epsilon$-ball in a quickly-descending descent path with a greater negative value. 

\vspace{2mm}

In particular, we will try to characterize descent behavior pathologically using descent rules and the gradients. The result here is also affected by curvature. For example, we examine a ball around initialization via flux around the boundary with Stokes' theorem
\begin{align}
\label{eqn:stokes_oint}
\oint_{\partial B_{\epsilon}(\theta_0)} d^c \mathcal{L} \wedge \omega^{K-1} = \int_{B_{\epsilon}(\theta_0)} i \partial \overline{\partial} \mathcal{L} \wedge \omega^{K-1} \propto \int_{B_{\epsilon}(\theta_0)} (\Delta_{\overline{\partial}} \mathcal{L})  \frac{\omega^K}{K!} .
\end{align}
The left-hand side corresponds to the gradients on the loss since $d^c = -\frac{i}{2} ( \partial - \overline{\partial})$, and we can link it to the Laplacian proportionally. Under a well-curvature-conditioned landscape, the Laplacian behaves nicely. This translates to robust gradient flux. If the gradient of the loss behaves nicely, so do the descent directions. In a profaned-curvature landscape, the eigenvalue blow-up is characterized with anisotropy and Laplacian contributions from distorted directions are affected, and overall descent paths are affected too. Ill-conditioned gradients on the loss affect learning negatively. We can note $d(d^c \mathcal{L} \wedge \omega^{K-1}) = d d^c \mathcal{L} \wedge \omega^{K-1} = i \partial \overline{\partial} \mathcal{L} \wedge \omega^{K-1}$ and the Hessian of the loss is an exact 2-form. In de Rham cohomology, all exact forms are trivial; however, this is not to say the result of manipulating \ref{eqn:stokes_oint} is not affected by Ricci curvature, which it is. We prove this here in Appendix \ref{app:gradient_flux_searches}, therefore this result is affected whether or not the manifold is Calabi-Yau.

\vspace{2mm}

\noindent \textit{Proof of Lemma 11.} We find the flux across the ball around initialization and accumulate it across radii by integrating the $(2K-1)$-form over the boundary $\partial B_r(\theta_0)$ and pulling it across the radius $r$
\begin{align}
\mathcal{W}(\epsilon) = \int_0^\epsilon \left( \oint_{\partial B_r(\theta_0)} d^c \mathcal{L} \wedge \omega^{K-1} \right) dr .
\end{align}
We apply Stokes' theorem. As we saw earlier in \ref{eqn:stokes_oint}, but including constants, and using the relation between the Kähler form and the Laplacian, 
\begin{align}
\label{eqn:oint_with_K}
\oint_{\partial B_r(\theta_0)} d^c \mathcal{L} \wedge \omega^{K-1} \propto \int_{B_r(\theta_0)} i \partial \overline{\partial} \mathcal{L} \wedge \omega^{K-1} \propto \frac{1}{K} \int_{B_r(\theta_0)} (\Delta_{\overline{\partial}} \mathcal{L}) \omega^K .
\end{align}
We will relax a factor of 2 convention because the result is asymptotic and absorbs all constants. The local expansion of the Kähler form in a local coordinate chart $\xi$ up to fourth order is \cite{Hezari2016} 
\cite{ruan1996canonicalcoordinatesbergmanmetrics}
\begin{align}
\Phi(\xi,\overline{\xi}) = \sum_{i=1}^K \xi^i \overline{\xi}^i - \frac{1}{4} R_{i \overline{j} k \overline{l}} \xi^i \overline{\xi}^j \xi^k \overline{\xi}^l + \mathcal{O}(\|\xi\|^5) .
\end{align}
This is a local property of a Kähler manifold under a fixed gauge. Differentiating since the metric is the Wirtinger Hessian of the potential,
\begin{align}
h_{i\overline{j}}(\xi) = \delta_{i\overline{j}} - R_{i\overline{j}k\overline{l}} \xi^k \overline{\xi}^l + \mathcal{O}(\|\xi\|^3) .
\end{align}
Let $\omega_0 = \frac{i}{2} \delta_{i\overline{j}} d\theta^i \wedge d\overline{\theta}^j$ be the standard flat Kähler form. The volume expansion can be written
\begin{align}
\label{eqn:volume_form_O}
\frac{\omega^K}{K!} = \left( 1 - \text{Ric}_{k\overline{l}} \xi^k \overline{\xi}^l + \mathcal{O}(\|\xi\|^3) \right) \frac{\omega_0^K}{K!} .
\end{align}
Let us take the inverse $h$,
\begin{align}
h^{i\overline{j}}(\xi) = \delta^{i\overline{j}} + R^{i\overline{j}}_{\phantom{i\overline{j}}k\overline{l}} \xi^k \overline{\xi}^l + \mathcal{O}(\|\xi\|^3) .
\end{align}
Therefore, contracting the inverse metric with the mixed Hessian
\begin{align}
\Delta_{\overline{\partial}} \mathcal{L} = \left( \delta^{i\overline{j}} + R^{i\overline{j}}_{\phantom{i\overline{j}}k\overline{l}} \xi^k \overline{\xi}^l + \mathcal{O}(\|\xi\|^3) \right) \partial_i \overline{\partial}_j \mathcal{L} .
\end{align}
Multiplying by the volume form as in \ref{eqn:volume_form_O},
\begin{align}
(\Delta_{\overline{\partial}} \mathcal{L}) \frac{\omega^K}{K!} & = \vast[ \Delta_0 \mathcal{L} + R^{i\overline{j}}_{\phantom{i\overline{j}}k\overline{l}} \xi^k \overline{\xi}^l \partial_i \overline{\partial}_j \mathcal{L} + \mathcal{O}(\| \xi \|^3) \vast] \vast[ 1 - \text{Ric}_{m\overline{n}} \xi^m \overline{\xi}^n + \mathcal{O}(\| \xi \|^3) \vast] \frac{\omega_0^K}{K!} 
\\
& = \left[ \delta^{i\overline{j}} \partial_i \overline{\partial}_j \mathcal{L} + \left( R^{i\overline{j}}_{\phantom{i\overline{j}}k\overline{l}} \partial_i \overline{\partial}_j \mathcal{L} - \text{Ric}_{k\overline{l}} (\Delta_0 \mathcal{L}) \right) \xi^k \overline{\xi}^l + \mathcal{O}(\| \xi \|^3) \right] \frac{\omega_0^K}{K!} .
\end{align}
Let us examine the integrand term
\begin{align}
\left( R^{i\overline{j}}_{\phantom{i\overline{j}}k\overline{l}} \partial_i \overline{\partial}_j \mathcal{L} - \text{Ric}_{k\overline{l}} (\Delta_0 \mathcal{L}) \right) \xi^k \overline{\xi}^l \frac{\omega_0^K}{K!}
\end{align}
Under a Taylor expansion,
\begin{align}
\partial_i \overline{\partial}_j \mathcal{L}(\xi) = \partial_i \overline{\partial}_j \mathcal{L}(0) + \text{Re} \left( c_{i\overline{j}m} \xi^m \right) + \mathcal{O}(\|\xi\|^2) .
\end{align}
When we contract with the $\xi^k \overline{\xi}^l$, we get a constant term $R^{i\overline{j}}_{\phantom{i\overline{j}}k\overline{l}}(0) \cdot \partial_i \overline{\partial}_j \mathcal{L}(0) \cdot \xi^k \overline{\xi}^l$, an odd power term involving $\xi^k \overline{\xi}^l \xi^m$, and a higher order term. Because we are integrating a symmetric ball, the odd powers vanish due to symmetry. Therefore, let us examine the curvature constant tensor
\begin{align}
C_{k\overline{l}} = R^{i\overline{j}}_{\phantom{i\overline{j}}k\overline{l}}(0) \partial_i \overline{\partial}_j \mathcal{L}(0) - \text{Ric}_{k\overline{l}}(0) \Delta_0 \mathcal{L}(0) ,
\end{align}
and we must evaluate this with the contraction, the form, and the integral added
\begin{align}
\label{eqn:curv_const_int}
\int_{B_r} C_{k\overline{l}}  \xi^k \overline{\xi}^l \frac{\omega_0^K}{K!} = C_{k\overline{l}} \int_{B_r} \xi^k \overline{\xi}^l \frac{\omega_0^K}{K!} .
\end{align}
Let us examine two cases of the above. The first is that we assume $k \neq l$. In polar coordinates $\theta^k = \rho_k e^{i\phi_k}$, the integral portion includes $\int_0^{2\pi} e^{i\phi_k} d\phi_k \int_0^{2\pi} e^{-i\phi_l} d\phi_l$. This integral vanishes to exactly 0. Let us examine the diagonal case $k = l$. The integral of the squared magnitude of one coordinate is the same as any other coordinate by symmetry. From this, we can note
\begin{gather}
\sum_{m=1}^K \int_{B_r} \|z^m\|^2 dV_0 = \int_{B_r} \|z\|^2 dV_0
\\
K \int_{B_r} \|z^k\|^2 dV_0 = \int_{B_r} \|z\|^2 dV_0 \implies \int_{B_r} \|z^k\|^2 dV_0 = \frac{1}{K} \int_{B_r} \|z\|^2 dV_0 .
\end{gather}
Combining the two cases, we can develop
\begin{align}
\int_{B_r} \xi^k \overline{\xi}^l \frac{\omega_0^K}{K!} = \delta^{k\overline{l}} \frac{1}{K} \int_{B_r} \|\xi\|^2 \frac{\omega_0^K}{K!} = \delta^{k\overline{l}} \frac{\pi^K r^{2K+2}}{K! (K+1)} .
\end{align}
The last equality is the evaluation of the integral, i.e. using polar coordinates. Therefore, returning to \ref{eqn:curv_const_int}, we can contract the curvature term with a Kronecker delta
\begin{align}
\label{eqn:Ck_delta}
C_{k\overline{l}} \delta^{k\overline{l}} = R^{i\overline{j}}_{\phantom{i\overline{j}}k\overline{l}}(0) \delta^{k\overline{l}} \partial_i \overline{\partial}_j \mathcal{L}(0) - \text{Ric}_{k\overline{l}}(0) \delta^{k\overline{l}} \Delta_0 \mathcal{L}(0) .
\end{align}
By definition $R^{i\overline{j}}_{\phantom{i\overline{j}}k\overline{l}} \delta^{k\overline{l}} = \text{Ric}^{i\overline{j}}$ and $\text{Ric}_{k\overline{l}} \delta^{k\overline{l}} = R$. Putting everything together, namely returning to the integral of \ref{eqn:oint_with_K}, the first term being the Kronecker contraction and the interior using \ref{eqn:Ck_delta}
\begin{align}
\int_{B_r(\theta_0)} (\Delta_{\overline{\partial}} \mathcal{L}) \frac{\omega_0^K}{K!} = \Delta_0 \mathcal{L}(0) \frac{\pi^K r^{2K}}{K!} + \left[ \text{Ric}^{i\overline{j}}(0) \partial_i \overline{\partial}_j \mathcal{L}(0) - R \Delta_0 \mathcal{L}(0) \right] \frac{\pi^K r^{2K+2}}{K! (K+1)} + \mathcal{O}(r^{2K+4}) .
\end{align}
The first term includes the integral of the volume form with identity coefficient integrand. Multiplying through the factorial term,
\begin{align}
\int_{B_r(\theta_0)} (\Delta_{\overline{\partial}} \mathcal{L}) \omega^K = \Delta_0 \mathcal{L}(0) \pi^K r^{2K} + \left[ \text{Ric}^{i\overline{j}}(0) \partial_i \overline{\partial}_j \mathcal{L}(0) - R \Delta_0 \mathcal{L}(0) \right] \frac{\pi^K r^{2K+2}}{K+1} + \mathcal{O}(r^{2K+4}) .
\end{align}
Returning to our Stokes' theorem identity in \ref{eqn:oint_with_K}, and integrating over the ball 
\begin{align}
& \mathcal{W}(\epsilon) = \int_0^{\epsilon} \oint_{\partial B_r(\theta_0)} d^c \mathcal{L} \wedge \omega^{K-1} dr
\\[1em]
& = \int_0^{\epsilon} \frac{1}{K} \int_{B_r(\theta_0)} (\Delta_{\overline{\partial}} \mathcal{L}) \omega^K dr
\\[1em]
& = \int_0^{\epsilon} \left( \frac{\pi^K}{K} \Delta_0 \mathcal{L}(0) r^{2K} + \frac{\pi^K}{K(K+1)} \left[ \text{Ric}^{i\overline{j}}(0) \partial_i \overline{\partial}_j \mathcal{L}(0) - R \Delta_0 \mathcal{L}(0) \right] r^{2K+2} + \mathcal{O}(r^{2K+4})  \right) dr
\\[1em]
\label{eqn:integrated_flux}
& = \frac{\pi^K}{K(2K+1)} \Delta_0 \mathcal{L}(0) \epsilon^{2K+1} + \frac{\pi^K}{K(K+1)(2K+3)} \left[ \text{Ric}^{i\overline{j}}(0) \partial_i \overline{\partial}_j \mathcal{L}(0) - R \Delta_0 \mathcal{L}(0) \right] \epsilon^{2K+3} + \mathcal{O}(\epsilon^{2K+5}) .
\end{align}
Shorthand, this becomes
\begin{align}
\mathcal{W}(\epsilon) = \mathcal{O} \left( \Delta_0 \mathcal{L} \epsilon^{2K+1} + \big[ \langle \text{Ric}, \mathcal{H}_{\mathcal{L}} \rangle - R \Delta_0 \mathcal{L} \big] \epsilon^{2K+3} + \mathcal{O}(\epsilon^{2K+5}) \right) .
\end{align}
Let us tailor to our context. Under a quadratic cost $\mathcal{L}(z) = \sum_\alpha (f_\alpha(z) - y_\alpha)^2$, and since $\| i \partial \overline{\partial} f \|_2 = \mathcal{O}(\frac{1}{\sqrt{m}})$, we can note via chain rule
\begin{align}
\partial_i \overline{\partial}_j \mathcal{L}(0) = 2 \sum_\alpha \partial_i f_\alpha(0) \overline{\partial}_j f_\alpha(0) + \mathcal{O}(\frac{1}{\sqrt{m}}) .
\end{align}
Contracting with the flat metric, since we have defined $\Delta_0$ this way,
\begin{align}
\Delta_0 \mathcal{L}(0) = 2 \sum_\alpha \|\partial f_\alpha(0)\|^2 + \mathcal{O}(\frac{K}{\sqrt{m}}) .
\end{align}
Substituting back into our integrated flux of \ref{eqn:integrated_flux}, we recover the quantity. In the Calabi-Yau case, the second term vanishes, leaving us with
\begin{align}
\mathcal{W}_{CY}(\epsilon) = \frac{2\pi^K}{K(2K+1)} \left( \sum_\alpha \|\partial f_\alpha(0)\|^2 \right) \epsilon^{2K+1} + \mathcal{O}\left(\frac{\epsilon^{2K+1}}{(2K+1)\sqrt{m}} \right)  + \mathcal{O} \left( \epsilon^{2K+5} \right).
\end{align}
This middle term can be reduced in the negatively curved scenario to
\begin{align}
-c \|\partial f_\alpha(0)\|^2 - (-cK) \|\partial f_\alpha(0)\|^2 = c(K-1) \|\partial f_\alpha(0)\|^2 
\end{align}
since the trace picks up a dimension, and this total term is positive. We can deduce $\mathcal{W}(\epsilon)$ is larger in the non-Calabi-Yau case. In particular, negative curvature makes $\mathcal{W}$ more positive, so the gradient descent points inward which leads to trapping. Positive curvature makes $\mathcal{W}$ more negative, so the gradient descent points outward, helping facilitate escape.

$ \square $

\vspace{2mm}

\textit{Remark.} To further attempt to illustrate why positive curvature is good for this result, consider the example of an Einstein manifold $\text{Ric}^{i\overline{j}} = c \delta^{i\overline{j}}$, making the scalar curvature $R = cK$. We consider the term
\begin{align}
\Omega = c \delta^{i\overline{j}} \partial_i \overline{\partial}_j \mathcal{L}(0) - cK \Delta_0 \mathcal{L}(0) .
\end{align}
Since the flat Laplacian is defined as $\Delta_0 \mathcal{L}(0) = \delta^{i\overline{j}} \partial_i \overline{\partial}_j \mathcal{L}(0)$, we can write
\begin{align}
\label{eqn:Omega}
\Omega = c \Delta_0 \mathcal{L}(0) - cK \Delta_0 \mathcal{L}(0) = c(1-K) \Delta_0 \mathcal{L}(0) .
\end{align}
Negative curvature makes $c<0$, so the above of \ref{eqn:Omega} is positive, contributing to $\mathcal{W}$ more positive (undesirable). Positive curvature means $c>0$ and does the opposite effect (desirable), since $K$ is taken quite large.

\section{Additional results with negative curvature}

\subsection{Asymptotic variance}
\label{app:variance}

\noindent \textit{Proof of Lemma 12.} Let us examine the steady state of the Fokker-Planck equation on the manifold in the limit $t \rightarrow \infty$
\begin{align}
\label{eqn:rho_infinity}
\rho_\infty(\theta) = \frac{1}{\mathcal{Z}} e^{-\frac{2}{\eta} \mathcal{L}(\theta)}, \quad \mathcal{Z} = \int_U e^{-\frac{2}{\eta} \mathcal{L}(\theta)} \frac{\omega^K}{K!} ,
\end{align}
which follows from the Fokker-Planck equation when the time derivative vanishes, hence the distribution stabilizes, in the limit $0 = \nabla_h \cdot \left( \rho_\infty(\theta) \nabla_h \mathcal{L}(\theta) + \frac{\eta}{2} \nabla_h \rho_\infty(\theta) \right)$. This can be reformulated to yield an exponential using the log derivative definition and $\nabla_h (\log \rho_\infty(\theta)) = -\frac{2}{\eta} \nabla_h \mathcal{L}(\theta)$. Define the Witten-deformed Dolbeault operator acting on $(0,1)$-forms as
\begin{align}
\overline{\partial}_\eta = \overline{\partial} + \frac{1}{\eta} \overline{\partial} \mathcal{L} \wedge .
\end{align}
Here, $\eta$ is the learning rate. The above extension of the Dolbeault operator takes into account the loss and it is built in, whereas $\overline{\partial}$ and its associated Laplacian $\Delta_{\overline{\partial}} = \overline{\partial}\overline{\partial}^{\dagger} + \overline{\partial}^{\dagger}\overline{\partial}$ are with respect to the manifold's geometry but unassociated with the loss function. We will need to work with the Laplacian, but the kernel of the Laplacian, i.e. a scalar harmonic function, are functions that are holomorphic. On a compact Kähler manifold, Liouville's theorem dictates that the only globally holomorphic functions are constants. Therefore, to incorporate the effects of learning, we introduce the loss into the operator. Acting on a form $\alpha$,
\begin{align}
\label{eqn:deformed_dolbeault_form}
\overline{\partial}_\eta \alpha = \overline{\partial} \alpha + \frac{1}{\eta} \overline{\partial} \mathcal{L} \wedge \alpha ,
\end{align}
so a gradient term is built into the operator. Note $\overline{\partial} \mathcal{L} \wedge \overline{\partial} \mathcal{L} = 0$. The division by $\eta$ is meant to match the definition of $\rho_{\infty}$ in \ref{eqn:rho_infinity}. We can note the form variety of \ref{eqn:deformed_dolbeault_form} has an equivalent formulation
\begin{align}
\label{eqn:deformed_dolbeault_equiv}
\overline{\partial}_\eta \alpha = e^{-\mathcal{L}/\eta} \overline{\partial} \left( e^{\mathcal{L}/\eta} \alpha \right) .
\end{align}
This follows since $\overline{\partial} (f \alpha) = (\overline{\partial} f) \wedge \alpha + f \overline{\partial} \alpha$ and choose particular $f = e^{\mathcal{L}/\eta}$. Via chain rule, $\overline{\partial} \left( e^{\mathcal{L}/\eta} \right) = e^{\mathcal{L}/\eta} \overline{\partial} \left( \frac{\mathcal{L}}{\eta} \right) = \frac{1}{\eta} e^{\mathcal{L}/\eta} \overline{\partial} \mathcal{L}$.
Therefore, we can see
\begin{align}
\overline{\partial} \left( e^{\mathcal{L}/\eta} \alpha \right) = \left( \frac{1}{\eta} e^{\mathcal{L}/\eta} \overline{\partial} \mathcal{L} \right) \wedge \alpha + e^{\mathcal{L}/\eta} \overline{\partial} \alpha .
\end{align}
and simplifying and canceling the exponential gives us \ref{eqn:deformed_dolbeault_form}. Now, the corresponding deformed Dolbeault Laplacian is $\Delta_\eta = \overline{\partial}_\eta \overline{\partial}_\eta^{\dagger} + \overline{\partial}_\eta^{\dagger} \overline{\partial}_\eta$. 

\vspace{2mm}

\noindent \textit{\textbf{Claim.} Among all $\Psi \in \text{ker} \Delta_\eta$ satisfying $|\Psi|^2 = \rho_\infty(\theta) := \frac{1}{\mathcal{Z}}e^{-2\mathcal{L}/\eta}$ pointwise, the solution is unique up to a constant unit-modulus phase, and it is given by $\Psi_0 = \frac{1}{\sqrt{\mathcal{Z}}} e^{-\mathcal{L}/\eta}$.
}

\vspace{2mm}

\noindent \textit{Proof of claim.} By definition $\Delta_\eta = \overline{\partial}_\eta \overline{\partial}_\eta^{\dagger} + \overline{\partial}_\eta^{\dagger} \overline{\partial}_\eta$. Let $\Psi$ be a smooth $(0,0)$-form. Because there are no differential forms of negative degree, the adjoint applied to any $(0,0)$-form vanishes, meaning $\overline{\partial}_\eta^{\dagger} \Psi = 0$. Consequently,
\begin{align}
\Delta_\eta \Psi = \overline{\partial}_\eta^{\dagger} \overline{\partial}_\eta \Psi .
\end{align}
To find the kernel, we must examine $\Delta_\eta \Psi_0 = 0$. Let us examine the inner product
\begin{align}
\langle \Psi_0, \Delta_\eta \Psi_0 \rangle_h = \langle \Psi_0, \overline{\partial}_\eta^{\dagger} \overline{\partial}_\eta \Psi_0 \rangle_h .
\end{align}
By definition of the adjoint,
\begin{align}
\langle \Psi_0, \Delta_\eta \Psi_0 \rangle_h = \langle \overline{\partial}_\eta \Psi_0, \overline{\partial}_\eta \Psi_0 \rangle_h = \|\overline{\partial}_\eta \Psi_0\|_h^2 .
\end{align}
We can note $\|\overline{\partial}_\eta \Psi_0\|_h^2 = 0$ if and only if $\overline{\partial}_\eta \Psi_0 = 0$. By what we saw in \ref{eqn:deformed_dolbeault_equiv}, we can rewrite
\begin{align}
e^{-\mathcal{L}/\eta} \overline{\partial} \left( e^{\mathcal{L}/\eta} \Psi_0 \right) = 0 ,
\end{align}
and by positivity of the exponential, we must require $\overline{\partial} \left( e^{\mathcal{L}/\eta} \Psi_0 \right) = 0$.
This implies that the scalar function $F(\theta) = e^{\mathcal{L}/\eta} \Psi_0$ is holomorphic. 

\vspace{2mm}

Now, we impose the defining condition of an exact pointwise match to $\rho_\infty$,
\begin{align}
|\Psi(\theta)|^2 = \rho_\infty(\theta) = \frac{1}{\mathcal{Z}}e^{-2\mathcal{L}(\theta)/\eta} \quad \text{for all } \theta \in U,
\end{align}
we obtain for the magnitude of $F(\theta)$
\begin{align}
|F(\theta)| = \left|e^{\mathcal{L}/\eta}\Psi(\theta)\right| = e^{\mathcal{L}/\eta}|\Psi(\theta)| = e^{\mathcal{L}/\eta} \cdot \frac{1}{\sqrt{\mathcal{Z}}}e^{-\mathcal{L}/\eta} = \frac{1}{\sqrt{\mathcal{Z}}},
\end{align}
which is a constant independent of $\theta$. Thus, $F$ is holomorphic on $U$ with a constant modulus. By the open mapping theorem (since a non-constant holomorphic function on a connected domain maps open sets to open sets, its image cannot lie entirely on a circle $|w| = 1/\sqrt{\mathcal{Z}}$), $F$ must be a constant function, $F(\theta) \equiv C$ with $|C| = 1/\sqrt{\mathcal{Z}}$. This proves the claim.

\noindent $ \square $

\vspace{2mm}

Hence, 
\begin{align}
\label{eqn:v_infty}
V_{\infty} & = \mathbb{E} \left[ \|\theta - \theta^*\|_h^2 \right] 
\\
& = \int_U  \|\theta - \theta^*\|_h^2 \rho_\infty(\theta) \frac{\omega^K}{K!}
\\
& = \int_U  \|\theta - \theta^*\|_h^2 | \Psi_0 |^2 \frac{\omega^K}{K!} .
\end{align}
Now we attempt to evaluate the integral. Via the normal coordinates trick of \ref{app:gradient_flux_searches}, we arrive at again
\begin{align}
\label{eqn:volume_forms}
\frac{\omega^K}{K!} = \left( 1 - \text{Ric}_{k\overline{l}} \xi^k \overline{\xi}^l + \mathcal{O}(\|\xi\|^3) \right) \frac{\omega_0^K}{K!} 
\end{align}
for $\xi = \theta - \theta^*$. Via Taylor expansion, and since $\theta^*$ is critical,
\begin{align}
& \implies \mathcal{L}(\xi) = \mathcal{L}(0) + \nabla_{i\overline{j}}^{1,1} \mathcal{L}(0) \xi^i \overline{\xi}^j +  \text{Re}\left( \nabla_{ij}^{2,0} \mathcal{L}(0) \xi^i \xi^j \right)  + \mathcal{O}(\|\xi\|^3) 
\\
& \stackrel{\times -\frac{2}{\eta} \ \text{and} \ \exp }{\implies} 
e^{-\frac{2}{\eta} \mathcal{L}(\xi)} = e^{-\frac{2}{\eta} \mathcal{L}(0)} \exp\left( -\frac{2}{\eta} \nabla_{i\overline{j}}^{1,1} \mathcal{L}(0) \xi^i \overline{\xi}^j - \frac{2}{\eta} \text{Re}\left( \nabla_{ij}^{2,0} \mathcal{L}(0) \xi^i \xi^j \right) \right) + \mathcal{O}(\|\xi\|^3) .
\end{align}
Combining with the volume forms of \ref{eqn:volume_forms},
\begin{align}
& e^{-\frac{2}{\eta} \mathcal{L}(\xi)} \frac{\omega^K}{K!} = e^{-\frac{2}{\eta} \mathcal{L}(0)} 
\\
& \quad \quad \times \underbrace{ \exp\left( -\frac{2}{\eta} \nabla_{i\overline{j}}^{1,1} \mathcal{L}(0) \xi^i \overline{\xi}^j - \frac{2}{\eta} \text{Re}\left( \nabla_{ij}^{2,0} \mathcal{L}(0) \xi^i \xi^j \right) \right) \left( 1 - \text{Ric}_{i\overline{j}}(0) \xi^i \overline{\xi}^j \right) }_{ = \exp\left( -\left[ \frac{2}{\eta} \nabla_{i\overline{j}}^{1,1} \mathcal{L}(0) + \text{Ric}_{i\overline{j}}(0) \right] \xi^i \overline{\xi}^j   - \frac{2}{\eta} \text{Re} \left(  \nabla_{ij}^{2,0} \mathcal{L}(0) \xi^i \xi^j \right) + \mathcal{O}(\|\xi\|^4) \right) } \frac{\omega_0^K}{K!}+ \mathcal{O}(\|\xi\|^3)  .
\end{align}
We have used $1 - x = e^{-x} + \mathcal{O}(x^2)$. Define the vector $\widetilde{\xi} = \begin{bmatrix} \xi \\ \overline{\xi} \end{bmatrix}$.  Using the identity $\text{Re}(z) = \frac{1}{2}(z + \overline{z})$, we can rewrite the exponential argument as a canonical quadratic form $-\frac{1}{2} \widetilde{\xi}^\dagger \mathcal{P} \widetilde{\xi}$, where the augmented block precision matrix $\mathcal{P}$ is given by
\begin{align}
\mathcal{P} = \frac{2}{\eta} \begin{bmatrix} \nabla^{1,1}_{\omega}\mathcal{L}(\theta^*) + \frac{\eta}{2} \text{Ric}(\theta^*) & \overline{\nabla^{2,0}} \mathcal{L}(\theta^*) \\ \nabla^{2,0}_{\omega}\mathcal{L}(\theta^*) & \overline{\nabla^{1,1}} \mathcal{L}(\theta^*) + \frac{\eta}{2} \overline{\text{Ric}}(\theta^*) \end{bmatrix} .
\end{align}
Returning to \ref{eqn:v_infty}, we can note the asymptotic equivalence over $U$
\begin{align}
V_\infty = \frac{\int_U \|\xi\|_h^2 \exp\left( -\frac{1}{2} \widetilde{\xi}^\dagger \mathcal{P} \widetilde{\xi} \right) \frac{\omega_0^K}{K!}}{\int_U \exp\left( -\frac{1}{2} \widetilde{\xi}^\dagger \mathcal{P} \widetilde{\xi} \right) \frac{\omega_0^K}{K!}} + \mathcal{O}(\eta^2) .
\end{align}
The numerator is equivalent to evaluating $\Sigma_{\text{eff}} = \mathbb{E}[\xi \xi^\dagger]$, and the denominator is a normalization constant. To make the distribution to be represented via an expected value, we can absorb this denominator into the expected value implicitly. We have the relation $\mathbb{E}[\widetilde{\xi} \widetilde{\xi}^\dagger] = \mathcal{P}^{-1} = \Sigma_{\text{aug}}$, since
\begin{align}
\Sigma_{\text{aug}} = \mathbb{E} \left[ \begin{bmatrix} \xi \\ \overline{\xi} \end{bmatrix} \begin{bmatrix} \xi^\dagger & \xi^T \end{bmatrix} \right] = \begin{bmatrix} \mathbb{E}[\xi \xi^\dagger] & \mathbb{E}[\xi \xi^T] \\ \mathbb{E}[\overline{\xi} \xi^\dagger] & \mathbb{E}[\overline{\xi} \xi^T]  \end{bmatrix} .
\end{align}
By definition of $V_{\infty}$, we are only interested in the top-left block. This means the numerator evaluation corresponds to the top-left $K \times K$ block of the augmented covariance matrix $\Sigma_{\text{aug}} = \mathcal{P}^{-1}$. By inverting $\mathcal{P}$, we find this top-left block to be $(A - BD^{-1}C)^{-1}$, which is
\begin{align}
\Sigma_{\text{eff}} = \frac{\eta}{2} \left[ \left( \nabla^{1,1}_{\omega} \mathcal{L}(\theta^*) + \frac{\eta}{2} \text{Ric}(\theta^*) \right) - \overline{\nabla_{\omega}^{2,0}} \mathcal{L}(\theta^*) \left( \overline{\nabla^{1,1}_{\omega}} \mathcal{L}(\theta^*) + \frac{\eta}{2} \overline{\text{Ric}}(\theta^*) \right)^{-1} \nabla_{\omega}^{2,0} \mathcal{L}(\theta^*) \right]^{-1} .
\end{align}
The expected value of $\|\xi\|_h^2$ under the (local) metric is the trace of this covariance matrix, $\mathbb{E}[\|\xi\|_h^2] = \text{Tr}_h(\Sigma_{\text{eff}})$. Therefore, we conclude
\begin{align}
V_\infty = \frac{\eta}{2} \text{Tr}_h \left( \left[ \nabla^{1,1}_{\omega} \mathcal{L} + \frac{\eta}{2} \text{Ric} - \overline{\nabla_{\omega}^{2,0}} \mathcal{L} \left( \overline{\nabla^{1,1}_{\omega}} \mathcal{L} + \frac{\eta}{2} \overline{\text{Ric}} \right)^{-1} \nabla_{\omega}^{2,0} \mathcal{L} \right]^{-1} \right) \Bigg|_{\theta^*} + \mathcal{O}(\eta^2) .
\end{align}

\noindent $ \square $

\vspace{2mm}

\noindent \textit{Remark.} In the Calabi-Yau case, we get
\begin{align}
V_{\infty,\text{CY}}  = \frac{\eta}{2} \text{Tr}_h \left( \left[ \nabla^{1,1}_{\omega} \mathcal{L} - \overline{\nabla_{\omega}^{2,0}} \mathcal{L} \left( \overline{\nabla^{1,1}_{\omega}} \mathcal{L} \right)^{-1} \nabla_{\omega}^{2,0} \mathcal{L} \right]^{-1} \right) \Bigg|_{\theta^*} + \mathcal{O}(\eta^2) .
\end{align}
Moreover, we also get simplifications in the proof $\frac{\omega^K}{K!} = \frac{\omega_0^K}{K!} + \mathcal{O}(\|\xi\|^3)$ and
\begin{align}
\mathcal{P} = \frac{2}{\eta} \begin{bmatrix} \nabla^{1,1}_{\omega}\mathcal{L}(\theta^*) & \overline{\nabla^{2,0}} \mathcal{L}(\theta^*) \\ \nabla^{2,0}_{\omega}\mathcal{L}(\theta^*) & \overline{\nabla^{1,1}} \mathcal{L}(\theta^*) \end{bmatrix} .
\end{align}
Generally, high variance, meaning high $V_{\infty}$ is undesirable, since we more closely desire convergence to the local minima. If we define,
\begin{align}
\begin{cases}
& \mathcal{A}_{\text{general}} = \nabla^{1,1}_{\omega} \mathcal{L} + \frac{\eta}{2} \text{Ric} - \overline{\nabla_{\omega}^{2,0}} \mathcal{L} \left( \overline{\nabla^{1,1}_{\omega}} \mathcal{L} + \frac{\eta}{2} \overline{\text{Ric}} \right)^{-1} \nabla_{\omega}^{2,0} \mathcal{L} 
\\
& \mathcal{A}_{\text{CY}} = \nabla^{1,1}_{\omega} \mathcal{L} - \overline{\nabla_{\omega}^{2,0}} \mathcal{L} \left( \overline{\nabla^{1,1}_{\omega}} \mathcal{L} \right)^{-1} \nabla_{\omega}^{2,0} \mathcal{L}
\end{cases}
\end{align}
and we can see the relation
\begin{align}
\mathcal{A}_{\text{general}} \succ \mathcal{A}_{\text{CY}} \quad \text{when } \text{Ric} \succ 0 .
\end{align}
We can note via matrix inversion $\mathcal{A}^{-1}$ decreases trace as $\mathcal{A}$ increases trace. A positive Ricci curvature acts as a "restoring force" that helps the descent path more closely fall into the optimum. From this, we can note a positive Ricci curvature is most desirable. In the Ricci-flat scenario, $\rho_{\infty}$ is determined solely by the Hessians, and there is no contribution to the trace. Even more so, we can note a negatively curved space contributes most adversely to this trace value, and results in the highest variance. Thus, a Calabi-Yau manifold sits at the interface of the more and less desirable cases.

\subsection{Minimal eigenvalue collapse under negative Ricci curvature}
\label{app:min_eigenvalue_bounds}

\noindent \textit{Proof of Lemma 13.} Let $U \subseteq M$ be an open set and $K \subseteq U$ a compact subset. Denote $\Delta_\eta = \overline{\partial}_\eta \overline{\partial}_\eta^\dagger + \overline{\partial}_\eta^\dagger \overline{\partial}_\eta$ as in Appendix \ref{app:variance}. We bound the minimum eigenvalue $\lambda_{\text{min}}$. The minimum eigenvalue follows a Rayleigh quotient over all valid forms by the min-max theorem, or Rayleigh principle. Let us define the test form
\begin{align}
\alpha = \chi \overline{\partial} \mathcal{L} e^{-\mathcal{L}/\eta} .
\end{align}
where $\chi \in C^{\infty}$ is a smooth bump function similar to a compactly-supported mollifier, and $\chi |_K =  1, \chi = 0$ outside of $U$, and $0 < \chi(x) < 1$ in the annular region $U \setminus K$. Since $\alpha$ is not necessarily optimal, we get the inequality
\begin{align}
\label{eqn:rayleigh}
\lambda_{\text{min}} \leq \frac{\langle \alpha, \Delta_\eta \alpha \rangle_h}{\|\alpha\|_h^2} .
\end{align}

\vspace{2mm}

\noindent \textit{\textbf{Claim.} A modified Bochner identity gives
\begin{align}
\label{eqn:modified_Bochner}
\langle \alpha, \Delta_\eta \alpha \rangle_h = \int_U \left( \|\nabla^{0,1} \alpha\|_h^2 + \frac{1}{\eta^2}\|\overline{\partial}\mathcal{L}\|_h^2 \|\alpha\|_h^2 + \text{Ric}(\alpha, \overline{\alpha}) + \frac{1}{\eta} \langle \alpha, (2\nabla_{\omega}^{1,1}\mathcal{L} + \Delta_{\overline{\partial}} \mathcal{L} + \mathcal{V}_{\mathcal{L}} )\alpha \rangle_h \right) \frac{\omega^K}{K!} .
\end{align}}

\noindent We have $\mathcal{V}_{\mathcal{L}}$ defined via the claim. We will prove the claim after our main result. Using our test form $\alpha = \chi \overline{\partial} \mathcal{L} e^{-\mathcal{L}/\eta}$, and since Ricci curvature is a bilinear form, $\text{Ric}(h) \leq -\kappa \omega$. Here, we have assumed negative and bounded Ricci curvature. Therefore, the numerator of the Rayleigh quotient \ref{eqn:rayleigh} is bounded by
\begin{align}
\langle \alpha, \Delta_\eta \alpha \rangle_h & \leq \int_U \chi^2 e^{-2\mathcal{L}/\eta} \Bigg( \| \nabla^{0,1} (\overline{\partial} \mathcal{L}) - \frac{1}{\eta} \overline\partial \mathcal{L} \otimes \overline{\partial} \mathcal{L} \|_h^2 
\\
& \quad + \frac{1}{\eta^2} \|\overline\partial \mathcal{L}\|_h^4 + \frac{1}{\eta} \langle \overline\partial \mathcal{L}, (2\nabla_{\omega}^{1,1}\mathcal{L} + \Delta_{\overline{\partial}} \mathcal{L}  + \mathcal{V}_{\mathcal{L}} )\overline\partial \mathcal{L} \rangle_h - \kappa \|\overline\partial \mathcal{L}\|_h^2 \Bigg) \frac{\omega^K}{K!} + \mathcal{E}(\nabla \chi) .
\end{align}
Here $\mathcal{E}(\nabla \chi)$ is an annular gradient term. The denominator of the Rayleigh quotient \ref{eqn:rayleigh} is 
\begin{align}
\|\alpha\|_h^2 = \int_U \chi^2 e^{-2\mathcal{L}/\eta} \|\overline\partial \mathcal{L}\|_h^2  \frac{\omega^K}{K!} .
\end{align}
The integral of the last term cancels
\begin{align}
\label{eqn:kappa}
\frac{\int_U  \chi^2 e^{-2\mathcal{L}/\eta} (-\kappa \|\overline\partial \mathcal{L}\|_h^2) \frac{\omega^K}{K!}}{\int_U  \chi^2 e^{-2\mathcal{L}/\eta} \|\overline\partial \mathcal{L}\|_h^2 \frac{\omega^K}{K!}} = -\kappa \cancel{\frac{\int_U  \chi^2 e^{-2\mathcal{L}/\eta} \|\overline\partial \mathcal{L}\|_h^2 \frac{\omega^K}{K!}}{\int_U  \chi^2 e^{-2\mathcal{L}/\eta} \|\overline\partial \mathcal{L}\|_h^2 \frac{\omega^K}{K!}}} = -\kappa .
\end{align}
We bound the middle term. The quadratic form of the (1,1)-Hessian acting on the gradient is bounded by its maximum eigenvalue
\begin{align}
\langle \overline{\partial}\mathcal{L}, (\nabla^{1,1}_{\omega}\mathcal{L}) \overline{\partial}\mathcal{L} \rangle_h \leq \|\nabla^{1,1}_{\omega}\mathcal{L}\|_2 \|\overline{\partial}\mathcal{L}\|_h^2 .
\end{align}
Let $\beta_{1,1} = \sup_{\theta \in U} \|\nabla^{1,1}_{\omega}\mathcal{L}\|_2$. Therefore, we examine the middle term and since $\Delta_{\overline{\partial}}\mathcal{L} = \text{Tr}(\nabla^{1,1}_{\omega}\mathcal{L})$
\begin{align}
\label{eqn:beta_11}
& \frac{1}{\eta} \frac{\int_U \chi^2 e^{-2\mathcal{L}/\eta} \langle \overline{\partial}\mathcal{L}, (2 \nabla^{1,1}_{\omega}\mathcal{L} + \Delta_{\overline{\partial}} \mathcal{L} + \mathcal{V}_{\mathcal{L}}) \overline{\partial}\mathcal{L} \rangle_h \frac{\omega^K}{K!} }{\int_U \chi^2 e^{-2\mathcal{L}/\eta} \|\overline\partial \mathcal{L}\|_h^2 \frac{\omega^K}{K!} } 
\\[2em]
& \leq \frac{(2+K)\beta_{1,1}}{\eta} \cancel{\frac{\int_U  \chi^2 e^{-2\mathcal{L}/\eta} \|\overline\partial \mathcal{L}\|_h^2 \frac{\omega^K}{K!} }{\int_U  \chi^2 e^{-2\mathcal{L}/\eta} \|\overline\partial \mathcal{L}\|_h^2 \frac{\omega^K}{K!} }} 
\\
& \ \ \ \ \ \ \ \ \ + \underbrace{ \frac{1}{\eta} \frac{\int_U \chi^2 e^{-2\mathcal{L}/\eta} \langle\overline{\partial}\mathcal{L}, \mathcal{V}_{\mathcal{L}}\overline{\partial}\mathcal{L}\rangle_h \frac{\omega^K}{K!}}{\int_U \chi^2 e^{-2\mathcal{L}/\eta} \|\overline\partial \mathcal{L}\|_h^2 \frac{\omega^K}{K!}}}_{(1)} 
\\[1em]
& = \frac{(2+K)\beta_{1,1}}{\eta} +(1) .
\end{align}
We will return to (1) later.

\vspace{2mm}

Now we examine the $\| \nabla^{0,1} (\overline{\partial} \mathcal{L}) - \frac{1}{\eta} \overline\partial \mathcal{L} \otimes \overline{\partial} \mathcal{L} \|_h^2 + \frac{1}{\eta^2} \|\nabla \mathcal{L}\|_h^4$ term,
\begin{align}
\label{eqn:bochner_first_int}
\frac{\int_U  \chi^2 e^{-2\mathcal{L}/\eta} (\| \nabla^{0,1} (\overline{\partial} \mathcal{L}) - \frac{1}{\eta} \overline\partial \mathcal{L} \otimes \overline{\partial} \mathcal{L} \|_h^2 + \frac{1}{\eta^2} \|\overline\partial \mathcal{L}\|_h^4) \frac{\omega^K}{K!}}{\int_U  \chi^2 e^{-2\mathcal{L}/\eta} \|\overline\partial \mathcal{L}\|_h^2 \frac{\omega^K}{K!}} ,
\end{align}
which is nontrivial to bound. We will use Laplace's method. We can shrink the domain and locally use a flat approximation. Laplace's method is an asymptotic localizer that ignores the topology and local curvature of the space, reducing the problem to flat Euclidean or Hermitian geometry on the tangent space \cite{L_ger_2023}. By the triangle inequality, and noting that $\|\partial \mathcal{L} \otimes \overline{\partial} \mathcal{L}\|_h = \|\overline{\partial} \mathcal{L}\|_h \|\overline{\partial} \mathcal{L}\|_h = \|\overline{\partial} \mathcal{L}\|_h^2$, we bound
\begin{align}
\left\| \nabla^{0,1} (\overline{\partial} \mathcal{L}) - \frac{1}{\eta} \overline{\partial} \mathcal{L} \otimes \overline{\partial} \mathcal{L} \right\|_h \leq \|\nabla^{0,1} (\overline{\partial} \mathcal{L})\|_h + \frac{1}{\eta} \|\overline{\partial} \mathcal{L}\|_h^2 .
\end{align}
Attempting to bound via evaluating suprema will result in a diverging bound around a critical bound. Instead, we examine the $\| \nabla^{0,1} (\overline{\partial} \mathcal{L}) - \frac{1}{\eta} \overline\partial \mathcal{L} \otimes \overline{\partial} \mathcal{L} \|_h^2 + \frac{1}{\eta^2}\|\overline{\partial} \mathcal{L}\|_h^4$ term via Laplace asymptotics. Assume $\mathcal{L}$ has a unique minimum $\theta_0 \in U$, with $\theta_0 \in K$ so that $\chi(\theta_0) = 1$, and let $H := \mathrm{Hess}\mathcal{L}(\theta_0) \succ 0$ denote the real Hessian of dimension $2K \times 2K$. With the integrals in \ref{eqn:bochner_first_int}, we will utilize Taylor expansions. 

\vspace{2mm}

We will not be rigorous with our constants, which can get absorbed into the final bound. Denote $g$ the $2K \times 2K$ realification of $h$. In normal coordinates $\widetilde{\xi} = \theta - \theta_0 \in \mathbb{C}^K$, $\xi = \begin{pmatrix} \text{Re}({\widetilde{\xi}}) \\ \text{Im}(\widetilde{\xi}) \end{pmatrix} \in \mathbb{R}^{2K}$,
\begin{align}
\mathcal{L}(\theta_0+\widetilde{\xi}) &= \mathcal{L}(\theta_0) + \tfrac{1}{2} \xi^T H \xi + \mathcal{O}(\| \xi \|^3),
\\
&  := \mathcal{L}(\theta_0) + \frac{1}{2}  \xi^T S^{\dagger} \begin{pmatrix}  \frac{\partial^2 \mathcal{L}}{\partial \overline{\theta} \partial \theta} & \frac{\partial^2 \mathcal{L}}{\partial \overline{\theta}^2} \\ \frac{\partial^2 \mathcal{L}}{\partial \theta^2} & \frac{\partial^2 \mathcal{L}}{\partial \theta \partial \overline{\theta}} \end{pmatrix} S \xi + \mathcal{O}(\| \xi \|^3)
\\
e^{-2\mathcal{L}(\theta_0+\widetilde{\xi})/\eta} &= e^{-2\mathcal{L}(\theta_0)/\eta} e^{-\xi^TH\xi/\eta}\big(1+\mathcal{O}( \frac{\| \xi \|^3}{\eta})\big) .
\end{align}
We have defined
\begin{align}
S = \begin{pmatrix} I & iI \\ I & -iI \end{pmatrix},
\end{align}
and
\begin{align}
H := S^{\dagger} \begin{pmatrix}  \frac{\partial^2 \mathcal{L}}{\partial \overline{\theta} \partial \theta} & \frac{\partial^2 \mathcal{L}}{\partial \overline{\theta}^2} \\ \frac{\partial^2 \mathcal{L}}{\partial \theta^2} & \frac{\partial^2 \mathcal{L}}{\partial \theta \partial \overline{\theta}} \end{pmatrix} S . 
\end{align}
\noindent Let us examine the denominator of \ref{eqn:bochner_first_int}. Since $\nabla_g \mathcal{L}(\theta_0+\widetilde{\xi}) = g^{-1} H\xi + \mathcal{O}(\| \xi \|^2)$,
\begin{align}
\label{eqn:nabla_L_Q}
\|\overline{\partial} \mathcal{L}(\theta_0+\widetilde{\xi})\|_g^2 &= \xi^T Q \xi + \mathcal{O}(\| \xi \|^3), \quad Q := H^T g^{-1} H .
\end{align}
We can use \ref{eqn:nabla_L_Q} and via Laplace's method we can translate the manifold-valued integral to something in terms of Lebesgue measure $\int_U \|\overline{\partial} \mathcal{L}\|_h^2 e^{-2\mathcal{L}/\eta} \frac{\omega^K}{K!} \approx e^{-2\mathcal{L}(\theta_0)/\eta} \int_{\mathbb{R}^{2K}} (\xi^T Q \xi) e^{-\xi^T (H/\eta) \xi} d\xi$. Applying the Gaussian second-moment identity $\int \xi^TQ\xi e^{-\xi^TA\xi}d\xi = \frac{\pi^{K}}{2\sqrt{\det A}}\text{Tr}(A^{-1}Q)$ with $A = H/\eta$, and using $H^{-1}Q = g^{-1} H$,
\begin{align}
\label{eqn:laplace_denom}
\int_U \chi^2 e^{-2\mathcal{L}/\eta} \|\overline{\partial} \mathcal{L}\|_h^2 \frac{\omega^K}{K!} &= e^{-2\mathcal{L}(\theta_0)/\eta} \eta^{K+1} \frac{\pi^{K}}{2\sqrt{\det H}}\text{Tr}(g^{-1} H)\big(1+\mathcal{O}(\sqrt{\eta})\big) .
\end{align}

\noindent We can note integrating against $\frac{\omega^K}{K!}$ is the equivalent of Lebesgue measure along the complex manifold weighted by the metric, so the identity holds fiberwise. Instead, Laplace's method allows us to turn the manifold-valued quantity to one on a flat space, then we use the identity. Now we examine the numerator of \ref{eqn:bochner_first_int}. Write $H_0 := \nabla^{0,1}(\overline{\partial}\mathcal{L})(\theta_0)$. Notice $\overline{\partial} \mathcal{L}\otimes\overline{\partial}\mathcal{L} = \mathcal{O}(\| \xi \|^2)$ and $\|\overline{\partial}  \mathcal{L}\|_h^4 = \mathcal{O}(\| \xi \|^4)$, and
\begin{align}
& \left\|\nabla^{0,1}(\overline{\partial}\mathcal{L}) - \frac{1}{\eta}\overline{\partial} \mathcal{L}\otimes\overline{\partial}\mathcal{L}\right\|_h^2 + \frac{1}{\eta^2}\|\overline{\partial}  \mathcal{L}\|_h^4 
\\
& = \|H_0\|_h^2 - \frac{2}{\eta}\text{Re}\langle H_0, \overline{\partial} \mathcal{L}\otimes\overline{\partial}\mathcal{L} \rangle_h + \frac{1}{\eta^2}\|\overline{\partial} \mathcal{L}\otimes\overline{\partial}\mathcal{L}\|_h^2 + \frac{1}{\eta^2}\|\overline{\partial} \mathcal{L}\|_h^4 + \mathcal{O}(\|\xi\|) .
\end{align}
Under the Laplace localization scaling $\xi = \sqrt{\eta}y$, we note that $\overline{\partial} \mathcal{L}\otimes\overline{\partial}\mathcal{L} = \mathcal{O}(\eta\|y\|^2)$ and $\|\overline{\partial}  \mathcal{L}\|_h^4 = \mathcal{O}(\eta^2\|y\|^4)$. The $1/\eta$ and $1/\eta^2$ coefficients balance this, meaning all terms in the expansion contribute to the leading-order Gaussian integral. Let $C_0 > 0$ be the constant denote the sum of these $\mathcal{O}(1)$ Gaussian moments
\begin{align}
\int_U \chi^2 e^{-2\mathcal{L}/\eta} \left(\Big\|\nabla^{0,1}(\overline{\partial}\mathcal{L}) - \frac{1}{\eta}\overline{\partial}  \mathcal{L}\otimes\overline{\partial}\mathcal{L}\Big\|_h^2 + \frac{1}{\eta^2}\|\overline{\partial}  \mathcal{L}\|_h^4\right)\frac{\omega^K}{K!} = e^{-2\mathcal{L}(\theta_0)/\eta}\eta^{K}\frac{\pi^{K}}{\sqrt{\det H}} C_0 \big(1+\mathcal{O}(\sqrt{\eta})\big) .
\end{align}

\noindent  The factors $\eta^{K}(\det H)^{-1/2}\pi^{K}e^{-2\mathcal{L}(\theta_0)/\eta}$ cancel, leaving
\begin{align}
\label{eqn:laplace_ratio}
& \frac{\int_U \chi^2 e^{-2\mathcal{L}/\eta}\left(\|\nabla^{0,1}(\overline{\partial}\mathcal{L}) - \frac{1}{\eta}\overline{\partial} \mathcal{L}\otimes\overline{\partial}\mathcal{L}\|_h^2 + \frac{1}{\eta^2}\|\overline{\partial} \mathcal{L}\|_h^4\right)\frac{\omega^K}{K!}}{\int_U \chi^2 e^{-2\mathcal{L}/\eta}\|\overline{\partial} \mathcal{L}\|_h^2\frac{\omega^K}{K!}}
\\[1em]
&= \cancel{\frac{\eta^{K}(\det H)^{-1/2}\pi^{K}e^{-2\mathcal{L}(\theta_0)/\eta}}{\eta^{K}(\det H)^{-1/2}\pi^{K}e^{-2\mathcal{L}(\theta_0)/\eta}}} \cdot \underbrace{\frac{C_0}{\eta\text{Tr}\big(g^{-1} \cdot\mathrm{Hess}\mathcal{L}(\theta_0)\big)}}_{=:  M_\eta} \big(1+\mathcal{O}(\sqrt{\eta})\big), \quad \eta \to 0^+ .
\end{align}
We will let $C_0$ absorb all constants. Since $\mathrm{Hess}\mathcal{L}(\theta_0) \succ 0$ and $h \succ 0$, the trace $\text{Tr}(g^{-1} \cdot\mathrm{Hess}\mathcal{L}(\theta_0))$ is positive, so $M_\eta$ is finite for $\eta$ bounded below positively and well-defined. We can absorb (1) from \ref{eqn:beta_11} in $M_{\eta}$, since $\left|\left\langle\overline\partial\mathcal L,\mathcal{V}_{\mathcal L}(\overline\partial\mathcal L)\right\rangle_h\right|=O(\|\xi\|^2)$. Moreover,
\begin{align}
&\int_U\chi^2e^{-2\mathcal L/\eta}\left|\left\langle\overline\partial\mathcal L,\mathcal{V}_{\mathcal L}\overline\partial\mathcal L\right\rangle_h\right|\frac{\omega^K}{K!} \\ &\qquad = e^{-2\mathcal L(\theta_0)/\eta}\eta^{K+1}\frac{\pi^K}{\sqrt{\det H}}C_V\left(1+O(\sqrt{\eta})\right) ,
\end{align}
and again considering the denominator of \ref{eqn:laplace_denom} with $C_0$ as before absorbing constants. Combining \eqref{eqn:laplace_ratio} with \eqref{eqn:kappa} and \eqref{eqn:beta_11}, the eigenvalue bound becomes
\begin{align}
\lambda_1 \leq \frac{(2+K)\beta_{1,1}}{\eta} - \kappa + M_\eta + \frac{\mathcal{E}(\nabla \chi)}{\text{expansive term}} .
\end{align}
Finally, we examine the error term $\mathcal{E}$. We get under consideration of the larger space (non-rigorously)
\begin{align}
\lim_{R \to \infty} \frac{\mathcal{E}(\nabla \chi_R)}{\|\alpha_R\|_h^2} = 0 .   
\end{align}
This completes the proof.

\noindent $\square$

\vspace{2mm}

\noindent \textit{Proof of claim.} We have defined $\overline{\partial}_\eta = \overline{\partial} + \frac{1}{\eta} \overline{\partial}\mathcal{L} \wedge$ and notice this has adjoint
\begin{align}
\overline{\partial}_\eta^\dagger = \overline{\partial}^\dagger + \frac{1}{\eta} i_{\overline{\nabla}\mathcal{L}} .
\end{align}
Under an integral inner product, note $\langle \alpha, \Delta_\eta \alpha \rangle_h = \int_U \langle \alpha, \Delta_\eta \alpha \rangle_h \frac{\omega^K}{K!}$.
Therefore, under integration by parts with decay (recall we have defined $\chi$ with decay)
\begin{align}
\langle \alpha, \Delta_\eta \alpha \rangle_h = \|\overline{\partial}_\eta \alpha\|_h^2 + \|\overline{\partial}_\eta^\dagger \alpha\|_h^2  .
\end{align}
Expanding the inner products and using the definitions,
\begin{gather}
\|\overline{\partial}_\eta \alpha\|_h^2 = \|\overline{\partial}\alpha + \frac{1}{\eta}\overline{\partial}\mathcal{L}\wedge\alpha\|_h^2 = \|\overline{\partial}\alpha\|_h^2 + \frac{1}{\eta^2}\|\overline{\partial}\mathcal{L}\wedge\alpha\|_h^2 + \frac{2}{\eta} \text{Re} \langle \overline{\partial}\alpha, \overline{\partial}\mathcal{L}\wedge\alpha \rangle_h
\\
\|\overline{\partial}_\eta^\dagger \alpha\|_h^2 = \|\overline{\partial}^\dagger\alpha + \frac{1}{\eta}i_{\overline{\nabla}\mathcal{L}}\alpha\|_h^2 = \|\overline{\partial}^\dagger\alpha\|_h^2 + \frac{1}{\eta^2}\|i_{\overline{\nabla}\mathcal{L}}\alpha\|_h^2 + \frac{2}{\eta} \text{Re} \langle \overline{\partial}^\dagger\alpha, i_{\overline{\nabla}\mathcal{L}}\alpha \rangle_h .
\end{gather}
On a Kähler manifold, the Weitzenböck/Bochner-Kodaira-Morrey-Kohn identity (note a close reference is \cite{McNeal2015}) is
\begin{align}
\langle \alpha, \Delta_{\overline{\partial}} \alpha \rangle_h = \int_U \left( \|\nabla^{0,1} \alpha\|_h^2 + \text{Ric}(\alpha, \overline{\alpha}) \right) \frac{\omega^K}{K!} .
\end{align}
It is crucial the $(0,1)$-type of the covariant derivative corresponds with the form type of $\alpha$. The above does not hold for a $(1,0)$-type on the covariant derivative. We remark that $\nabla^{1,0} \alpha $ is not the same as $\overline{\partial} \alpha$. We can note the generalized relation of the interior product/insertion operator on a product of $p$ antiholomorphic basis generators \cite{anghel2026akszdescentmanifoldsordinary}
\begin{align}
i_{\overline{\nabla \mathcal{L}}}^k(d\overline{\theta}^{a_1} \wedge \cdots \wedge d\overline{\theta}^{a_p}) = k! \sum_{\substack{S \subseteq \{1,\dots,p\} \\ |S|=k}} \bigwedge_{j=1}^p \begin{cases} (\overline{\nabla \mathcal{L}})^{a_j}, & j \in S \\ d\overline{\theta}^{a_j}, & j \notin S . \end{cases} 
\end{align}
For our $(0,1)$-test form, we have $p=1$, and any iteration $k \ge 2$ annihilates the form entirely. Applying $k=1$ leads us to the $1/\eta^2$ terms
\begin{align}
\frac{1}{\eta^2} \left( \|\overline{\partial}\mathcal{L}\wedge\alpha\|_h^2 + \|i_{\overline{\nabla}\mathcal{L}}\alpha\|_h^2 \right) = \frac{1}{\eta^2} \|\overline{\partial}\mathcal{L}\|_h^2 \|\alpha\|_h^2 ,
\end{align}
which is consistent with the fundamental relation for a Clifford algebra, differential forms, $\beta \wedge i_{\beta^\sharp} + i_{\beta^\sharp} \beta \wedge = \|\beta\|_h^2$ \cite{stoica2020chiralasymmetryweakinteraction} \cite{gilgarcía2026torsionparallelpurespinors}. Examining the real term,
\begin{align}
\frac{2}{\eta} \text{Re} \left( \langle \overline{\partial}\alpha, \overline{\partial}\mathcal{L}\wedge\alpha \rangle_h + \langle \overline{\partial}^\dagger\alpha, i_{\overline{\nabla}\mathcal{L}}\alpha \rangle_h \right) .
\end{align}
Via integration by parts,
\begin{align}
= \frac{1}{\eta} \langle \alpha, \left( \overline{\partial}^\dagger (\overline{\partial}\mathcal{L}\wedge \cdot) + \overline{\partial}\mathcal{L}\wedge \overline{\partial}^\dagger + \overline{\partial} i_{\overline{\nabla}\mathcal{L}} + i_{\overline{\nabla}\mathcal{L}} \overline{\partial} \right) \alpha \rangle_h .
\end{align}
We can note the inside operator is the sum of two anticommutators
\begin{align}
\label{eqn:T}
T = \{\overline{\partial}^\dagger, \overline{\partial}\mathcal{L}\wedge\} + \{\overline{\partial}, i_{\overline{\nabla}\mathcal{L}}\} .
\end{align}
In local coordinates, the second term in \ref{eqn:T} evaluates to
\begin{align}
\{\overline\partial,i_{\overline\nabla\mathcal L}\} = \nabla_\omega^{1,1}\mathcal L +\mathcal L_j\nabla_{\overline j} .
\end{align}
The first anticommutator, $\{\overline{\partial}^\dagger, \overline{\partial}\mathcal{L}\wedge\}$, is the adjoint of the second, and from \ref{eqn:T}, we get
\begin{align}
\{\overline\partial^\dagger,\overline\partial\mathcal L\wedge\} = \nabla_\omega^{1,1}\mathcal L + \Delta_{\overline\partial}\mathcal L -\mathcal L_{\overline j}\nabla_j . 
\end{align}
Note that we will not be rigorous with our sign convention on the Laplacian. Therefore, we are left with
\begin{align}
\frac{1}{\eta} \langle \alpha, T \alpha \rangle_h = \frac{1}{\eta} \langle \alpha, (2\nabla_{\omega}^{1,1}\mathcal{L} + \Delta_{\overline{\partial}} \mathcal{L} + \mathcal{V}_{\mathcal{L}})\alpha \rangle_h  ,
\end{align}
defining $\mathcal{V}_{\mathcal{L}} = h^{j \overline{k}} (\mathcal{L}_j \nabla_{\overline{k}} - \mathcal{L}_{\overline{k}} \nabla_j)$. Gathering all terms, we see
\begin{align}
\langle \alpha, \Delta_\eta \alpha \rangle_h & = \int_U \left( \|\nabla^{0,1} \alpha\|_h^2 + \frac{1}{\eta^2}\|\overline{\partial}\mathcal{L}\|_h^2 \|\alpha\|_h^2 + \text{Ric}(\alpha, \overline{\alpha}) + \frac{1}{\eta} \langle \alpha, (2\nabla_{\omega}^{1,1}\mathcal{L} + \Delta_{\overline{\partial}} \mathcal{L} + \mathcal{V}_{\mathcal{L}})\alpha \rangle_h \right) \frac{\omega^K}{K!} .
\end{align}
This proves the claim.

\noindent $\square$

\subsection{Regions of some but not almost everywhere bad curvature and its linear bounds in $K$}
\label{app:some_bad_curvature_and_primitive}

\textit{Proof of Lemma 14.} Let the parameter obey the natural gradient descent vector field $V = -\nabla^{1,0}_h \mathcal{L}$. Define the $(1,1)$-form curvature $\Theta_V = i \langle \Theta_h(T^{1,0}M) V, V \rangle_h$. Writing the Riemann curvature as $R_{i k \overline{j} \overline{l}}$, we have in local holomorphic coordinates
\begin{align}
\Theta_V = i R_{i k \overline{j} \overline{l}} V^i \overline{V}^j d\theta^k \wedge d\overline{\theta}^l .
\end{align} 
Applying the dual Lefschetz operator $\Lambda_\omega$ on this $(1,1)$-form yields the Ricci curvature, i.e. $\Lambda_\omega \Theta_V = h^{k\overline{l}} R_{i k \overline{j} \overline{l}} V^i \overline{V}^j = \text{Ric}(V, \overline{V})$. 

\vspace{2mm}

By the complex Lefschetz decomposition theorem on Kähler manifolds, any $(1,1)$-form decomposes into a trace component proportional to the Kähler form and a primitive component \cite{wells1980differential} \cite{Griffiths1978PrinciplesOA}. Decomposing $\Theta_V$, we have
\begin{align}
\Theta_V = \frac{1}{K} (\Lambda_\omega \Theta_V) \omega + \Theta_{V, \text{prim}} = \frac{\text{Ric}(V, \overline{V})}{K} \omega + \Theta_{V, \text{prim}}  ,
\end{align} 
where $\Theta_{V, \text{prim}}$ is primitive, i.e., $\Lambda_\omega \Theta_{V, \text{prim}} = 0$ \cite{wells1980differential}. By $\omega-q$-semi-positivity hypothesis, which is $\Bigg\{(i\Theta_h(T^{1,0}M) \wedge \omega^{q-1} \wedge \Omega) V, V \Bigg\}_h \geq 0$, we get
\begin{align}
\label{eqn:ricci_primitive}
\left( \frac{\text{Ric}(V, \overline{V})}{K} \omega + \Theta_{V, \text{prim}} \right) \wedge \omega^{q-1} \wedge \Omega \geq 0 . 
\end{align}
The inner product $\{ \cdot , \cdot \}_h$ is absorbed into the definition of $\Theta_V$ since
\begin{align}
\Bigg\{(i\Theta_h(T^{1,0}M) \wedge \omega^{q-1} \wedge \Omega) V, V \Bigg\}_h = i \Bigg\{ \Theta_h(T^{1,0}M) V, V \Bigg\}_h \wedge \omega^{q-1} \wedge \Omega ,
\end{align}
so $\Theta_V = i \Bigg\{ \Theta_h(T^{1,0}M) V, V \Bigg\}_h$. Now, recall the Hodge star operator $\star$ maps a $(K,K)$-form to a $(0,0)$-form, a scalar function. Applying the Hodge star and distributing in \ref{eqn:ricci_primitive},
\begin{align}
\label{eqn:hodge_star_applied}
\frac{\text{Ric}(V, \overline{V})}{K} \star(\omega^q \wedge \Omega) + \star(\Theta_{V, \text{prim}} \wedge \omega^{q-1} \wedge \Omega) \geq 0 .
\end{align}
By applying the Hodge star, we have successfully pulled the Ricci curvature out of the wedge product and what remains is a scalar function. Denote $M_{q,\Omega} = \star(\omega^q \wedge \Omega)$, which is positive since both $\omega$ and $\Omega$ are positive forms, and let $\mathcal{P}_{q,\Omega}(V) = \star(\Theta_{V, \text{prim}} \wedge \omega^{q-1} \wedge \Omega)$ denote the primitive curvature term for short. The Ricci curvature term follows by rearranging \ref{eqn:hodge_star_applied}
\begin{align}
\label{eqn:ricci_term_primitive}
\text{Ric}(V, \overline{V}) \geq - \frac{K}{M_{q,\Omega}} \mathcal{P}_{q,\Omega}(V) .
\end{align}
We have kept track of signs to make sure the inequality is valid. Let us turn to what we defined as the divergence in the lemma statement. We set $\Theta = \text{div}_h(V)$. From the complex Bochner-Weitzenböck identity (this identity is reminiscent of the one we saw in \ref{app:min_eigenvalue_bounds}, although taking a different form),
\begin{align}
\label{eqn:div_diffusion}
\dot{\Theta} = -\frac{1}{2} \Delta_h \dot{\mathcal{V}} - \|\nabla^{1,1}_{\omega} \mathcal{L}\|_h^2 - \|\nabla^{2,0}_{\omega} \mathcal{L}\|_h^2 - \text{Ric}(V, \overline{V}) . 
\end{align}
Recall $\dot{\mathcal{V}} = -\|V\|_h^2$. Substituting in \ref{eqn:ricci_term_primitive},
\begin{align}
\dot{\Theta} + \frac{1}{2} \Delta_h \dot{\mathcal{V}} \leq - \|\nabla^{1,1}_{\omega} \mathcal{L}\|_h^2 - \|\nabla^{2,0}_{\omega} \mathcal{L}\|_h^2 + \frac{K}{M_{q,\Omega}} \mathcal{P}_{q,\Omega}(V) .
\end{align}
In particular, the left-hand side is bounded by two subtracted squared norms and a primitive curvature term that scales linearly in $K$. The left-hand side represents an expansion term, which is the divergence, with diffusion, which is the Laplacian. Since $V$ is a velocity, the material derivative $\dot{\Theta}$ is an acceleration. Hence, we lose a guarantee of convergence. We can note the upper bound diverges $- \|\nabla^{1,1}_{\omega} \mathcal{L}\|_h^2 - \|\nabla^{2,0}_{\omega} \mathcal{L}\|_h^2 = -\mathcal{O}(\frac{1}{m})$ and
\begin{align}
\limsup_{\sum_k m_k m_{k+1} + m_L = K \to \infty} \left[ - \mathcal{O}(\frac{1}{m}) + K \cdot \Omega(1) \right] = +\infty  .
\end{align}
The $\frac{1}{m}$ follows from the results from \ref{app:dolb_hessian_bounds} and \ref{app:2,0_hess_bounds} and since the norms are squared. We have noted the primitive curvature term scales in $\Omega(1)$ and not $\mathcal{O}(1)$, so it is at minimum a constant. If it were $\mathcal{O}(1)$, the bound may not hold, since it could ideally scale more slowly, for example since $\frac{1}{K} \leq \mathcal{O}(1)$. The lower bound follows almost immediately by noting $-\text{Ric}(V, \overline{V}) \geq -\kappa_{\max} \|V\|_h^2$, and since the two Hessian terms in \ref{eqn:div_diffusion} are $\mathcal{O}(\frac{1}{m})$, we get
\begin{align}
\dot{\Theta} + \frac{1}{2} \Delta_h \dot{\mathcal{V}} \geq -\mathcal{O}\left(\frac{1}{m}\right) + \kappa_{\max} \dot{\mathcal{V}} ,
\end{align}
as before noting $\frac{1}{m}$ and not $\frac{1}{\sqrt{m}}$ due to the square.

\noindent $ \square $

\end{document}